\documentclass[11pt]{article}
\usepackage{fullpage}
\usepackage[T1]{fontenc}
\usepackage{lmodern}
\usepackage{amsmath,amssymb,amsfonts,amsthm}
\usepackage[authoryear]{natbib}
\usepackage{graphicx}
\usepackage[dvipsnames]{xcolor}
\usepackage{array,multirow}
\usepackage{booktabs}
\providecommand{\botrule}{\bottomrule}
\usepackage{adjustbox}
\usepackage{etoolbox}
\NewEnvironmentCopy{arxivtabular}{tabular}
\newcommand{\arxivtablelayout}{%
  \RenewDocumentEnvironment{tabular}{O{c} m +b}{%
    \begin{adjustbox}{max width=\linewidth}%
      \begin{arxivtabular}[##1]{##2}##3\end{arxivtabular}%
    \end{adjustbox}%
  }{}%
}
\AtBeginEnvironment{table}{\arxivtablelayout}
\usepackage{algorithm,algpseudocode}
\usepackage{enumitem}
\usepackage{appendix}
\usepackage{tikz}
\usepackage{microtype}
\makeatletter
\let\th@thmstyleone\th@plain
\let\th@thmstyletwo\th@remark
\let\th@thmstylethree\th@definition
\makeatother

\makeatletter
\renewcommand{\@maketitle}{%
  \newpage\null\vskip 0.25em
  \begingroup
  \centering
  {\fontsize{18}{22}\selectfont\bfseries\@title\par}%
  \vskip 1.1em
  {\begin{tabular}{@{}c@{}}\@author\end{tabular}\par}%
  \endgroup
  \vskip 1.0em
}
\makeatother

\graphicspath{{figures/}}

\usepackage{booktabs}
\usepackage{setspace}
\usepackage{bm}
\definecolor{Blue}{HTML}{174E66}
\definecolor{Green}{HTML}{65B66E}
\definecolor{Orange}{HTML}{D99A42}
\definecolor{Red}{HTML}{C65353}
\definecolor{Grey}{HTML}{999999}

\newcommand{\pilotaxes}[1]{%
  \draw[Blue,line width=.65pt] (0,0) rectangle (2,2);
  \foreach \v/\tick in {0/0,.5/{0.5},1/1,1.5/{1.5},2/2}{
    \draw[Blue] (\v,0) -- ++(0,-.025);
    \node[anchor=north,inner sep=1.4pt] at (\v,0) {\tick};
    \draw[Blue] (0,\v) -- ++(-.025,0);
    \ifdim\v pt>0pt \node[anchor=east,inner sep=1.4pt] at (0,\v) {\tick};\fi
  }
  \node[anchor=south,font=\sffamily\bfseries\fontsize{8}{9}\selectfont] at (1,2.045) {#1};
  \node at (1,-.30) {batch size exponent $\kappa_{b}$};}

\DeclareMathOperator*{\argmin}{arg\,min}

\newcommand{\E}{\mathbb{E}}
\newcommand{\tr}{\operatorname{tr}}
\newcommand{\Var}{\operatorname{Var}}
\newcommand{\Cov}{\operatorname{Cov}}
\providecommand{\bottomrule}{\botrule}

\theoremstyle{thmstyleone}%
\newtheorem{theorem}{Theorem}

\newtheorem{assumption}{Assumption}
\newtheorem{lemma}{Lemma}

\theoremstyle{thmstyletwo}%

\theoremstyle{thmstylethree}%

\usepackage[colorlinks=true, linkcolor=blue, citecolor=blue, urlcolor=blue]{hyperref}
\usepackage{etoc}

\hypersetup{pdftitle={U-Statistics-Powered LLM Reasoning},pdfauthor={Shijin Gong, Kai Ye, Victor Chernozhukov, Xinyu Zhang, Chengchun Shi}}
\title{U-Statistics-Powered LLM Reasoning}
\author{
\normalsize Shijin Gong$^{1,\ast}$\quad Kai Ye$^{2,\ast}$\quad Victor Chernozhukov$^{3,\dagger}$\quad Xinyu Zhang$^{4,1}$\quad Chengchun Shi$^{2,\dagger}$ \\[0.65em]
\small $^{1}$\,University of Science and Technology of China \\
\small $^{2}$\,London School of Economics and Political Science \\
\small $^{3}$\,Massachusetts Institute of Technology \\
\small $^{4}$\,Chinese Academy of Sciences
}
\date{}

\begin{document}
\etocdepthtag.toc{main}
\maketitle
\begingroup
\renewcommand{\thefootnote}{\fnsymbol{footnote}}
\footnotetext[1]{These authors contributed equally.\quad}
\footnotetext[2]{Co-corresponding authors: Victor Chernozhukov (chernozhukov@gmail.com) \& Chengchun Shi (c.shi7@lse.ac.uk)}
\endgroup

\begin{abstract}
Since its introduction by DeepSeek, group relative policy optimization (GRPO) has been widely adopted for training large reasoning models and agents, spurring a proliferation of its variants. Yet the group size, a crucial hyperparameter in these GRPO-type algorithms, remains largely chosen by trial and error.  Exploiting the U-statistic structure of GRPO's policy gradient, we recast the selection of the group size as a classical statistical problem of variance ratio estimation. This problem is particularly challenging when the denominator of the ratio is small, rendering naive plug-in estimators unstable. To address this challenge, we propose a pessimistic ratio estimator that accounts for the denominator's estimation uncertainty. We characterise the statistical properties of this estimator across three sample-size regimes. Empirically, using  synthetic tasks from Reasoning Gym and real-world benchmarks covering both mathematical reasoning and code generation, we show that our adaptive group size selection reduces the mean squared error of the resulting policy gradient and improves policy optimisation over fixed-group-size baselines, with modest additional computational cost.
\end{abstract}


\pagestyle{plain}
\allowdisplaybreaks[2]

\section{Introduction}
\label{sec:intro}
Reasoning has emerged as one of the most powerful techniques in the evolution of large language models (LLMs). By guiding models to think step by step, reasoning enables LLMs to solve mathematical, scientific, and coding problems at, or even beyond, expert-human level \citep{wei2022chain}. As a prominent example, Google's Gemini Deep Think achieved gold-medal-level performance at the 2025 International Mathematical Olympiad  \citep{castelvecchi2025deepmind}. 
More recently, Anthropic's Claude, in collaboration with mathematician Levent Alp{\"o}ge, has made a breakthrough on the longstanding Riemann hypothesis, producing a verified proof of the conjecture \citep{leanmathlib2026riemann}.

Group relative policy optimization (GRPO), introduced in DeepSeekMath \citep{shao2024deepseekmath}, has become a cornerstone algorithm for training LLMs to reason. It gained widespread popularity following the release of DeepSeek-R1 in early 2025, which achieved performance comparable to or even better than OpenAI's o1 on several reasoning benchmarks \citep{guo2025deepseek}. The algorithm was subsequently adopted and extended in numerous open-source large reasoning models \citep{liu2025fin}. 
These efforts, collectively, resulted in a rapidly growing literature on GRPO-type algorithms, including Dr. GRPO \citep{liu2025understanding}, GSPO \citep{zheng2025group} and GPG \citep{chu2025gpg} as popular variants. These algorithms have been further extended beyond LLM reasoning to agentic settings \citep{feng2025group}.  

At their core, GRPO-type algorithms are stochastic gradient ascent (SGA) algorithms. At each iteration, they sample a minibatch of prompts and a group of responses for each prompt in the minibatch, score these responses, and compute a stochastic gradient based on each response's performance relative to its group average. The computational cost is largely dominated by the generation of responses, which can be extremely expensive for long reasoning trajectories. Fixing the number of sampled responses per iteration, a \textit{larger} group size allows more responses per prompt but reduces the number of prompts sampled, whereas a \textit{smaller} group size increases prompt diversity at the expense of fewer responses per prompt. This creates a trade-off in selecting the group size. Despite the practical success of GRPO-type algorithms, there remains a lack of principled guidance on choosing this crucial hyperparameter.

This paper fills in this gap. Exploiting the U-statistic structure of GRPO's stochastic gradient, the optimal group size that minimises the mean squared error (MSE) of the gradient depends on the ratio of two variance components arising from the Hoeffding decomposition \citep{ustat2026demystifying}. Building on this observation, we study how to estimate this variance ratio, recasting group size selection in LLM reasoning as a classical statistical estimation problem.

\begin{figure}
\centering
\includegraphics[width=0.9\linewidth]{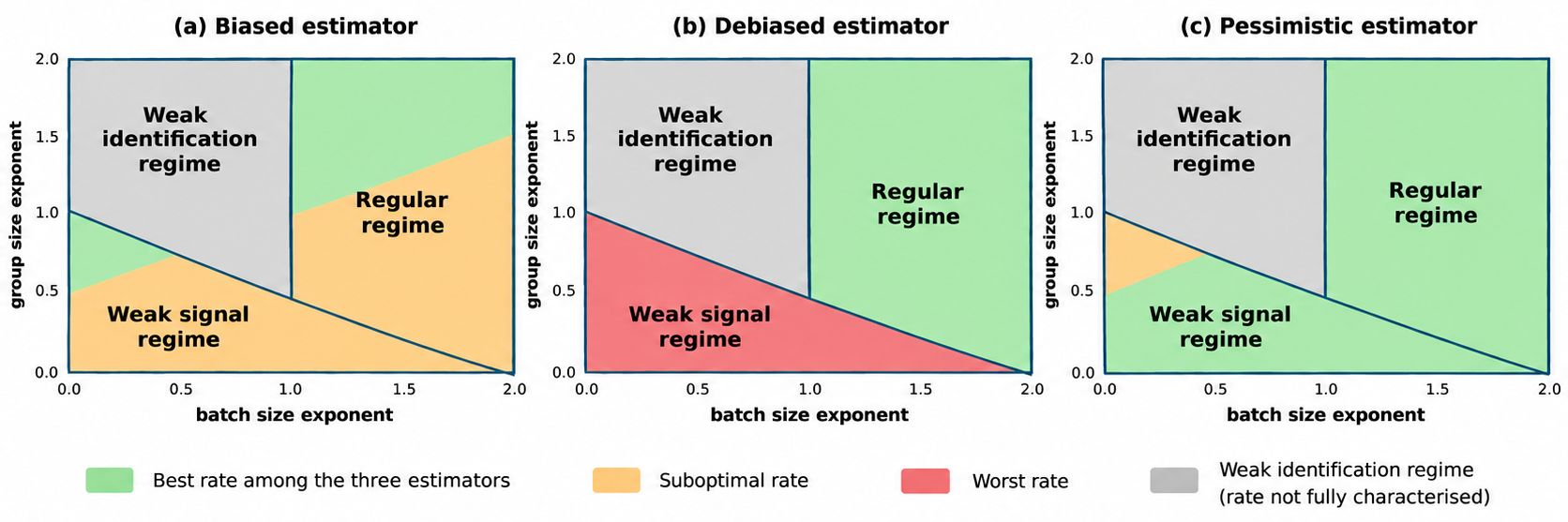}
\caption{Polynomial rates of expected excess gradient MSE for the three group size estimators. %
Blue lines delimit the regimes defined in Section~\ref{sec:tradeoffgroupest}. In the interiors, {\color{Green}green} indicates the best rate (allowing ties), {\color{Orange}orange} a slower rate, and {\color{Red}red} the worst rate. {\color{Grey}Grey} marks weak identification, where the available bounds do not give a general ranking among the three estimators.}\label{fig:estcomparison} %
\end{figure}

A natural approach is to estimate the two variance components (the numerator and denominator) and plug their estimates into the ratio. We investigate two such estimators: (i) a biased estimator that ignores the finite-group bias arising from the limited number of responses per prompt, and (ii) a debiased estimator that  corrects this bias. Although bias correction is a classical technique for improving statistical estimation \citep{firth1993bias}, we uncover a surprising phenomenon: debiasing does \textit{not} necessarily improve ratio estimation. When the denominator is small, removing its positive bias can substantially amplify estimation errors, leading to highly unstable ratio estimates.

To address this instability, we develop a third estimator: a pessimistic estimator that penalises each candidate group size according to the standard error of its estimated gradient MSE, to discourage large ratio arising from the estimated denominator close to zero. %

Figure~\ref{fig:estcomparison} compares the convergence rates of the three estimators. Its horizontal and vertical axes show how quickly the number of prompts and the number of responses per prompt grow as the total sample size. Their precise definitions are given in Section~\ref{sec:tradeoffgroupest}. Larger values on these axes indicate faster growth rates. Based on these growth rates, Figure~\ref{fig:estcomparison} partitions the sample-size space into three regimes: (a) a regular regime where the estimation error of the denominator is small compared to its ground truth; (b) a weak signal regime where the estimation error dominates its ground truth; and (c) a weak identification regime where the convergence rates of the estimators are not fully determined. 

The theoretical advantages of pessimistic estimation are evident in the figure, which shows that the pessimistic estimator achieves the best convergence rate over most of the sample-size space. 
In the regular regime, where the sample sizes are sufficiently large for the debiased estimator to accurately estimate the denominator and attain the fastest polynomial convergence rate, the pessimistic estimator achieves the same rate. 
Under weak signal, where the sample sizes are insufficient to accurately estimate the denominator, debiasing without accounting for estimation uncertainty incurs the worst rate, whereas pessimism retains the best rate over most of this region. The biased estimator, meanwhile, suffers from a suboptimal convergence rate across substantial portions of both regimes due to its finite-group bias. Under weak identification, no general ranking among the three estimators can be established, as the convergence rates of the two plug-in estimators remain undetermined.

Empirically, we conduct extensive experiments on synthetic tasks from Reasoning Gym \citep{stojanovski2025reasoninggym} and real-world benchmarks covering mathematical reasoning and code generation to validate our theoretical findings. In particular, the proposed pessimistic estimator generally yields more accurate policy gradient estimates (Figure \ref{fig:mse-mechanisms}). More importantly, these improvements translate into better policy optimisation: reasoning models trained with the proposed adaptive group size selection algorithm consistently outperform those trained with fixed group sizes (Figure \ref{fig:policy-current}, Tables \ref{tab:qwen3-current} \& \ref{tab:coding-current}), while incurring only a modest increase in training time.

\section{Background and Problem Setup}\label{sec:grpo}
\subsection{RL for LLM Reasoning}
GRPO is a reinforcement learning \citep[RL,][]{sutton2018reinforcement} algorithm for post-training LLMs to enhance their reasoning capabilities. To better understand GRPO, we first review how RL can be applied to LLM reasoning.

Suppose we have a base LLM pretrained on large text corpora. Given a \textit{prompt}, i.e., a mathematical question, denoted by $X$, the model generates a \textit{completion} $Y$, consisting of a reasoning trajectory describing how it arrives at the answer and the final answer itself. Both $X$ and $Y$ are represented as finite sequences of tokens drawn from a discrete vocabulary. We denote by $\pi_\theta(\bullet| X)$ the conditional probability mass function of the completion given the prompt, where $\theta$ denotes the model parameters. 

Our goal is to fine-tune $\theta$ to enhance the model's reasoning capabilities in a specific domain, such as mathematics. To this end, we need a deterministic reward function $r(X,Y)$ that assigns a \textit{reward} $Z=r(X,Y)$ to each prompt-completion pair. For instance, in mathematical reasoning, the reward may be a binary function indicating whether the final answer is correct. We aim to identify an optimal parameter $\theta^\star$ that maximizes the expected reward:
\begin{equation}\label{eqn:reward}
    \theta^* \in \arg\max_\theta
\mathbb{E}_{X\sim\mathcal D,Y\sim\pi_\theta(\bullet\mid X)}
[r(X,Y)],
\end{equation}
where $\mathcal D$ denotes the empirical distribution over a set of training prompts (e.g., mathematical questions), and the expectation is taken over both the prompt and the completion generated by the model.

The optimisation in \eqref{eqn:reward} can be naturally formulated as a policy optimisation problem and thus solved via RL. Specifically, we view the prompt $X$ as the context, the completion $Y$ as the action, and $\pi_\theta$ as the policy, following the RL terminology. Maximizing the expected reward thus amounts to identifying the optimal policy parameter $\theta^*$.  

In the context of LLMs, the action space, i.e., the space of all completions, is enormous. To handle this large action space, policy gradient algorithms are commonly employed in RL \citep[Chapter 13]{sutton2018reinforcement}. Their underlying principle closely follows SGA. Specifically, to maximize the expected reward in \eqref{eqn:reward}, we seek to estimate its gradient with respect to $\theta$, denoted by ${\bm g}(\theta)$, referred to as the policy gradient in RL. We then use its estimator ${\widehat{\bm g}}(\theta)$ to update the model parameters via gradient ascent, $\theta \leftarrow \theta + \eta {\widehat{\bm g}}(\theta)$, where $\eta$ denotes the learning rate. 

It remains to estimate ${\bm g}(\theta)$. By adapting the policy gradient theorem  \citep{sutton1999policy} to our setting, it is easy to show that
\begin{equation}\label{eqn:policygradient}
    {\bm g}(\theta)=\mathbb{E}_{X\sim \mathcal{D},Y\sim \pi_{\theta}(\bullet|X)} [\nabla_{\theta} \log \pi_{\theta}(Y|X) r(X,Y)].
\end{equation}
Here, $\nabla_{\theta} \log \pi_{\theta}(Y|X)$ corresponds to the score function of $\pi_{\theta}$, and is referred to as the \textit{policy score} in RL. Equation \eqref{eqn:policygradient} suggests a Monte Carlo procedure for estimating the policy gradient, which gives rise to the classical REINFORCE algorithm \citep{williams1992simple}. Specifically, at each iteration, we first sample a minibatch of prompts $X_1,\ldots,X_{{B}}$ from $\mathcal D$. For each prompt $X_i$, we then generate a completion $Y_i$ from the current policy $\pi_\theta(\bullet|X_i)$ and evaluate its reward $Z_i=r(X_i,Y_i)$, yielding
\begin{equation}\label{eqn:reinforcegradient}
{\widehat{\bm g}}_{\mathrm{REINFORCE}}(\theta)
=\frac{1}{{{B}}}\sum_{i=1}^{{B}}
\nabla_\theta\log\pi_\theta(Y_i|X_i)Z_i.
\end{equation}

By construction, \eqref{eqn:reinforcegradient} is unbiased for ${\bm g}(\theta)$. However, a well-known limitation of REINFORCE is that its gradient estimator can suffer from high variance. A classical RL approach to variance reduction is to introduce a baseline function $b(X)$ that depends on the prompt but not on the sampled completion and to replace $Z_i$ in \eqref{eqn:reinforcegradient} with the residual $Z_i-b(X_i)$ when multiplying the policy score $\nabla_\theta\log\pi_\theta(Y_i|X_i)$. Its rationale lies in that the score function has conditional mean zero given $X_i$, so subtracting such a baseline preserves the unbiasedness of the estimator. Meanwhile, a properly chosen baseline can substantially reduce its variance. In particular, under suitable conditions, the optimal baseline that minimises the variance of the resulting gradient coincides with the conditional expected reward $\mathbb E(Z|X)$ \citep{greensmith2004variance}, referred to as the value function $V^{\pi_\theta}(X)$ in RL.

Motivated by this observation, modern policy gradient algorithms such as proximal policy optimization (PPO) employ an additional neural network to estimate the value function and uses the estimated values to construct the policy gradient estimates \citep{schulman2017proximal,ouyang2022training}. However, training and maintaining this additional network can incur substantial computational costs in LLM reasoning tasks \citep{shao2024deepseekmath}.

\subsection{GRPO-Type Algorithms}
We next review GRPO, which addresses the aforementioned computational bottleneck by eliminating the additional value network. Its main idea is simple: instead of generating a single completion for each prompt $X_i$, GRPO samples a group of conditionally i.i.d. completions $\{Y_{i,j}\}_{j=1}^{{G}}$ and evaluates their rewards $\{Z_{i,j}\}_{j=1}^{{G}}$. It then employs the group average ${{G}}^{-1}\sum_{j=1}^{{G}} Z_{i,j}$ to approximate the value function, thus avoiding the separate value network and considerably facilitating the computation.

\begin{figure}[t]
    \centering
    \includegraphics[width=0.65\linewidth]{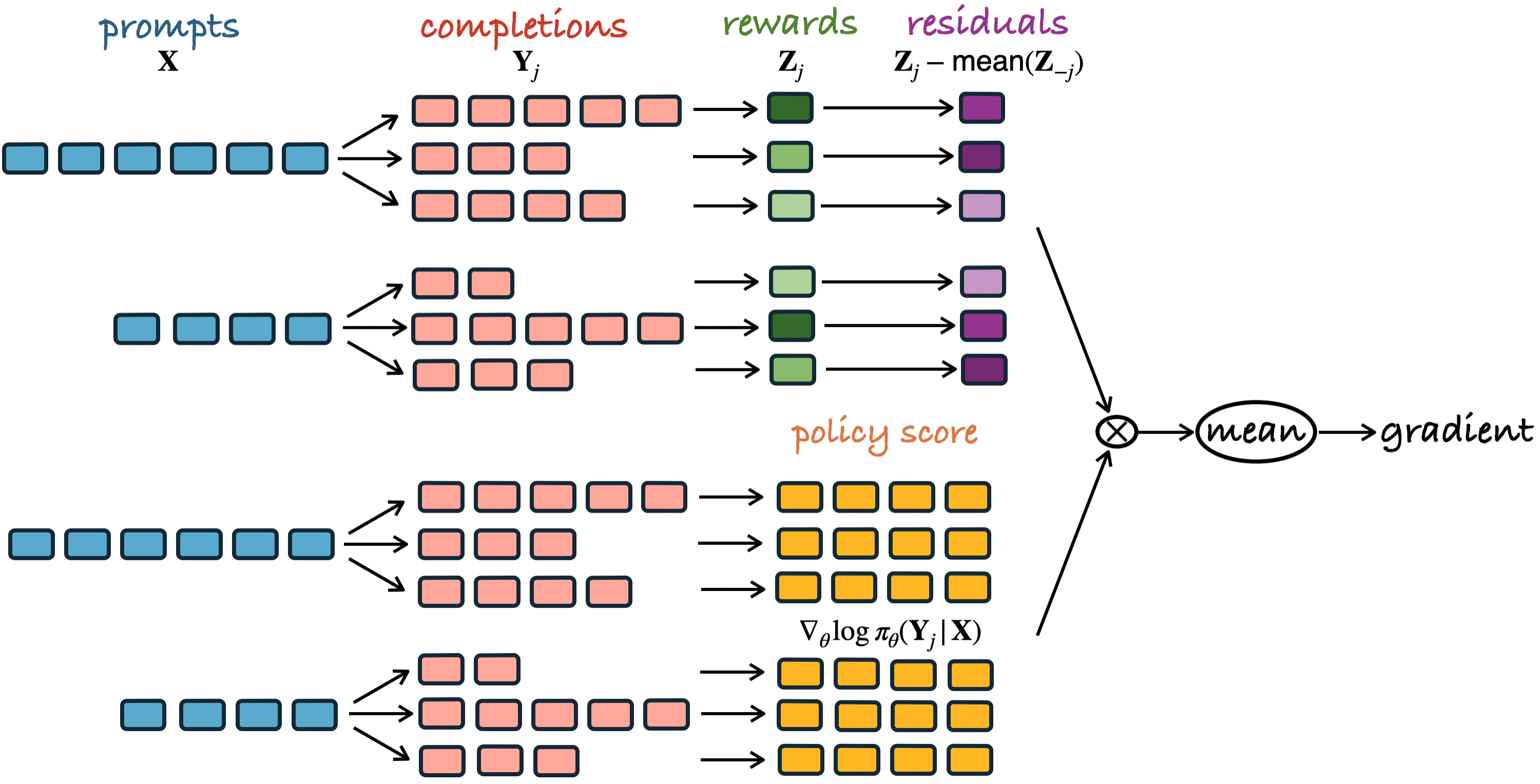}
    \caption{Illustration of how a GRPO-type algorithm estimates the policy gradient. Each rectangle represents either a token or a numerical value (e.g., a reward or residual). A sequence of rectangles forms a prompt, a completion, or a numerical vector (e.g., a policy score).}\label{fig:grpoillustration}
\end{figure}

For illustration purposes, we focus on a simplified version of GRPO in the main paper and discuss its extensions to more general formulations in Section \ref{sec:discussion}. Similar simplifications have been adopted in later variants, including Dr. GRPO and GPG. 
Figure \ref{fig:grpoillustration} visualises the pipeline through which the algorithm estimates the policy gradient. In particular, we follow the leave-one-out approach \citep{ahmadian2024back} that use the average reward of the other ${{G}}-1$ completions $\bar{Z}_{i,-j}=({{G}}-1)^{-1}\sum_{k\neq j}Z_{i,k}$ to approximate the value function for each completion, leading to the following policy gradient estimator,
\begin{equation}\label{eqn:pgest}
    {\widehat{\bm g}}_{{{B}},{{G}}}(\theta)=\frac{1}{{{B}}}\sum_{i=1}^{{B}} {\widehat{\bm g}}_{{G}}(X_i;\theta)=\frac{1}{{{B}}}\sum_{i=1}^{{B}} \Big[\frac{1}{{{G}}}\sum_{j=1}^{{G}} \nabla_{\theta} \log \pi_{\theta}(Y_{i,j}|X_i)(Z_{i,j}-\bar{Z}_{i,-j})\Big].
\end{equation}
The leave-one-out procedure ensures the unbiasedness of the resulting policy gradient estimator. In contrast, if the completion's own reward is included in the group average, the resulting estimator is biased, with its expectation equal to $({{G}}-1)/{{G}}$ times the true policy gradient.

\begin{figure}[t]
\centering
\includegraphics[width=0.85\textwidth]{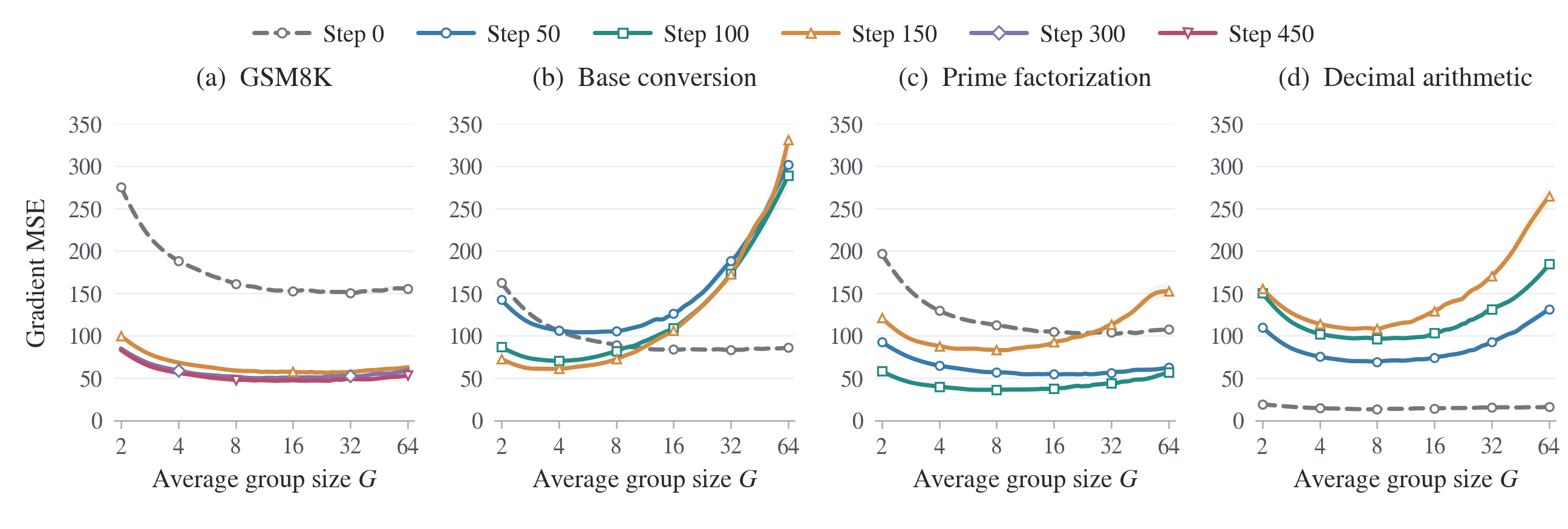}
\caption{MSE of the policy gradient as a function of group size ${{G}}$, with the total number of sampled completions ${{B}}{{G}}$ held fixed. Panels (a), (b), (c), and (d) correspond to the GSM8K dataset and three synthetic reasoning tasks from Reasoning Gym: base conversion, prime factorisation, and decimal arithmetic, respectively. Results are shown at steps 0, 150, 300, and 450 for GSM8K, and at steps 0, 50, 100, and 150 for the synthetic reasoning tasks.}
\label{fig:mse-geometry}
\end{figure}

By replacing a separately trained value network with group-based sampling, GRPO-type algorithms improve computational efficiency, but introduce a crucial hyperparameter: the group size ${{G}}$. 
To illustrate the importance of this hyperparameter, Figure \ref{fig:mse-geometry} plots the MSE of the policy gradient estimates against different group sizes at four checkpoints for each of four datasets, while fixing the total sampling budget ${{B}}{{G}}$. 
It can be seen that, across all four datasets, the gradient MSE can be highly sensitive to the choice of ${{G}}$ at certain training checkpoints. However, large gradient MSE can translate into unstable policy updates, leading to policy collapse in severe cases \citep{hu2025reinforce++,gong2026basis,gong2026kernelized}. This motivates us to adaptively select the group size to minimise the gradient MSE under a fixed sampling budget. In the next subsection, we recast this problem as a statistical estimation problem of the variance ratio.

\subsection{From Group Size Selection to Variance Ratio Estimation}
Interestingly, the MSE of GRPO's policy gradient estimator can be characterised through its connection to a classical concept in statistics: U-statistics. To see this, consider the gradient ${\widehat{\bm g}}_{{G}}(X_i;\theta)$ for an individual prompt $X_i$ defined by the expression inside the square brackets in \eqref{eqn:pgest}. A key observation is that,  ${\widehat{\bm g}}_{{G}}(X_i;\theta)$ involves two layers of averaging, an outer average over the ${{G}}$ sampled completions and an inner average over the leave-one-out group mean $\bar Z_{i,-j}=({{G}}-1)^{-1}\sum_{k\neq j}Z_{i,k}$, making it a second-order U-statistic \citep{ustat2026demystifying}.

Applying the Hoeffding decomposition \citep{hoeffding1948}, we can express ${\widehat{\bm g}}_{{G}}(X_i;\theta)$ as the sum of three orthogonal components: its conditional expectation given the prompt (denoted by ${\bm g}(X_i;\theta)$), a first-order projection, and a degenerate second-order term. Averaging these components over the sampled prompts $\{X_i\}_i$ and taking the trace of their covariance matrices yields a three-term decomposition of the MSE of ${\widehat{\bm g}}_{{{B}},{{G}}}(\theta)$, shown in the following lemma. 
\begin{lemma}[The GRPO scaling law]\label{prop:score-closed-forms}
Suppose $\{X_i\}_{i=1}^{{B}}$ are i.i.d. sampled with replacement from $\mathcal{D}$. Then
\begin{equation}\label{eq:scalinglaw}
\operatorname{MSE}\{{\widehat{\bm g}}_{{{B}},{{G}}}(\theta)\}:=\E\|{\widehat{\bm g}}_{{{B}},{{G}}}(\theta)-{\bm g}(\theta)\|^2=\frac{c_1}{{{B}}}+\frac{c_2}{{{B}}{{G}}}+\frac{c_3}{{{B}}{{G}}({{G}}-1)},
\end{equation}
where ${\bm g}(\theta)$ denotes the true gradient defined in \eqref{eqn:policygradient}, and the three variance components are given by 
\begin{align*}
c_1&=\E\|{\bm g}(X;\theta)-{\bm g}(\theta)\|^2,\\
c_2&=\E\|\nabla_{\theta} \log \pi_{\theta}(Y|X) \{Z-V^{\pi_\theta}(X)\}-{\bm g}(X;\theta)\|^2,\\
c_3&=\E\Big[\E \big\{ \|\nabla_{\theta} \log \pi_{\theta}(Y|X)\|^2| X\big\} \Var(Z| X)\Big]+\E\|{\bm g}(X;\theta)\|^2,
\end{align*}
where $(X,Y,Z)$ denotes a prompt-completion-reward triplet. 
Moreover, under a fixed total sampling budget ${{N}}={{B}}{{G}}$, there exists an optimal group size ${{G}}$ that minimises \eqref{eq:scalinglaw}, given by
\begin{equation}\label{eq:gstar}
{{G}}^*=1+\sqrt{c_3/c_1}.
\end{equation}
\end{lemma}
Lemma \ref{prop:score-closed-forms} is a slight refinement of Proposition 3 and Theorem 7 of \citet{ustat2026demystifying}. Its first part provides a closed-form expression for the MSE of ${\widehat{\bm g}}_{{{B}},{{G}}}(\theta)$. As commented earlier, the three terms on the  right-hand-side (RHS) of \eqref{eq:scalinglaw} correspond to the traces of the covariance matrices of the zeroth-, first-, and second-order terms arising from the Hoeffding decomposition, aggregated over the sampled prompts in the minibatch. 
The variance component $c_1$ captures between-prompt variability, whereas $c_2$ and $c_3$ capture within-prompt variability. All three components are independent of the two sample-size parameters, ${{B}}$ and ${{G}}$. 

Under a fixed sampling budget ${{N}}={{B}}{{G}}$, the three terms on the RHS of \eqref{eq:scalinglaw} behave differently as ${{G}}$ varies: (0) the zeroth order term can be rewritten as $c_1{{G}}/{{N}}$ and increases linearly with ${{G}}$, as a larger group size reduces the batch size ${{B}}$ and increases the prompt-level variability; (1) the first-order term becomes $c_2/{{N}}$ and remains constant as a function of ${{G}}$; (2) the second-order term becomes $c_3/\{{{N}}({{G}}-1)\}$ and decreases with ${{G}}$. Consequently, as illustrated in Figure \ref{fig:mse-geometry}, there exists an optimal group size, neither too small nor too large, that balances the zeroth- and second-order terms and admits the simple closed-form expression in \eqref{eq:gstar}. 

Equation \eqref{eq:gstar} recasts the selection of the optimal group size as a statistical estimation problem of the variance ratio $c_3/c_1$, which we study in detail in the next section. We refer to \eqref{eq:gstar} as the GRPO scaling law, as it characterises the optimal choice of this crucial hyperparameter. 

To conclude this section, we remark that Lemma \ref{prop:score-closed-forms} assumes that prompts are sampled independently with replacement from $\mathcal{D}$, whereas in practice they are sampled without replacement. The latter introduces a finite population correction factor $1-{{B}}/(m-1)$ to the zeroth order term $c_1/{{B}}$, where $m$ denotes the total number of prompts in the training dataset. In typical LLM training, ${{B}}$ ranges from a few tens to a few hundred, whereas $m$ ranges from several thousand to tens of thousands. Consequently, this correction factor is close to one, making the sample-with-replacement assumption a reasonable approximation.

\section{Biased and Debiased Estimators: Their Trade-Offs}\label{sec:tradeoff}
\subsection{Adaptive Group Size Selection}
Figure \ref{fig:mse-geometry} illustrates that the optimal group size can vary across datasets and training iterations. The scaling law in \eqref{eq:gstar} further implies that this optimal value is determined by the ratio $c_3/c_1$. This motivates us to adaptively estimate these variance components for group size selection throughout training.  

\begin{algorithm}[t]
\caption{GRPO-type algorithms with adaptive group sizes.}
\label{alg:adaptive-group}
\begingroup\setstretch{1}
\begin{algorithmic}[1]
\Require Initial policy parameter $\theta_0$; sampling budget ${{N}}$; number of iterations $T$; learning rate $\eta$; pilot batch/group sizes ${b}$/${g}$; update set $\mathcal T$.
\For{$t=0,\ldots,T-1$}
  \If{$t\in\mathcal T$}
    \State Independently sample ${b}$ prompts with replacement from $\mathcal D$ and ${g}$ completions per prompt from $\pi_{\theta_t}$.
    \State Compute the rewards for each prompt-completion pair.  
    \State \label{line:allocation-step}Choose ${{G}}_t$ by the biased rule \eqref{eqn:biasest}, the debiased rule \eqref{eqn:debiasest}, or a pessimistic rule \eqref{eq:main-floored-pess} (Section~\ref{sec:pessimistic}).
  \Else
    \State Retain ${{G}}_t\gets {{G}}_{t-1}$.
  \EndIf
  \State Set ${{B}}_t\gets {{N}}/{{G}}_t$. 
  \State Sample ${{B}}_t$ prompts from $\mathcal{D}$ and ${{G}}_t$ completions per prompt, and evaluate the rewards for each prompt-completion pair. 
  \State Compute the RLOO gradient ${\widehat{\bm g}}_{{{B}}_t,{{G}}_t}(\theta_t)$ in \eqref{eqn:pgest} using these prompt-completion-reward triplets. 
  \State Update the policy parameter  $\theta_{t+1}\gets\theta_t+\eta {\widehat{\bm g}}_{{{B}}_t,{{G}}_t}(\theta_t)$.
\EndFor
\Ensure Final policy $\pi_{\theta_T}$.
\end{algorithmic}
\endgroup
\end{algorithm}

We summarise our procedure in Algorithm \ref{alg:adaptive-group}. At a given training iteration, the sampling of prompts and their completions may be performed twice, for different purposes. The first sampling step (Line 3) serves as a pilot study to estimate the variance ratio and determine the group size, whereas the second (Line 10) is the standard sampling step in GRPO-type algorithms, using the group size determined by the pilot study to estimate the policy gradient according to \eqref{eqn:pgest}. Throughout, we use lowercase $b$, $g$, and $n=bg$ to denote the batch size, group size, and total sample size in the pilot study, respectively, and the corresponding uppercase letters $B$, $G$, and $N=BG$ for policy gradient estimation.

Ideally, the pilot sampling step would be performed at every training iteration to adapt the group size to change in its optimal value throughout training. 
However, doing so would incur substantial computational costs. To this end, we perform the estimation only at certain training iterations, denoted by the update set \(\mathcal T\), and retain the previously selected group size at other iterations; see Lines 2 -- 7 of Algorithm \ref{alg:adaptive-group}.

In our implementation, we estimate the variance ratio at $t=0$ and $t=25$, and subsequently every 50 iterations, starting at $t=50$. The additional update at $t=25$ is motivated by our observation that the initial iteration tends to select a group size that is too large for later iterations. Re-estimating the variance ratio at $t=25$ allows us to adjust the group size earlier, so that the gradient estimation can be improved over the following 25 iterations.

We also remark that the prompts and completions in the pilot data could, in principle, be reused for gradient estimation in the second step. However, such reuse would introduce dependence between the selected group size and the estimated gradient. To avoid this complication, we use independent samples for the two steps.

It remains to specify the estimation procedure for the group size ${{G}}_t$ in Line 5. In the rest of this section, we investigate two plug-in estimators, the biased and debiased estimators: both use the ${b}$ sampled prompts and their associated ${g}$ completion-reward pairs from Lines 3 -- 5 of Algorithm \ref{alg:adaptive-group} to estimate $c_1$ and $c_3$, and then plug their estimates into \eqref{eq:gstar} for group size selection.  %

Interestingly, our analysis reveals a counterintuitive phenomenon: debiasing, a classical statistical technique, does not necessarily improve estimation accuracy of the ratio and can even worsen it substantially. To conclude this section, we summarise the trade-offs between the biased and debiased estimators, with the technical details provided in Sections \ref{sec:tradeoffvarest} and \ref{sec:tradeoffgroupest}. On the \textit{positive} side, debiasing offers three advantages: 
\begin{enumerate}
    \item[(i)] It removes the biases in estimating $c_1$ and $c_3$ (Theorem \ref{thm:varianceest}(a));
    \item[(ii)] The debiased estimates are consistent as ${b}\to\infty$, even when ${g}$ remains fixed, whereas consistency of the biased estimates generally requires both ${b}$ and ${g}$ to diverge (Theorem \ref{thm:varianceest}(b));
    \item[(iii)] The advantages in (i) and (ii) for estimating the two variance components also translate into more accurate group size selection: in the regular regime, where the sample sizes are sufficiently large for the estimation error in $c_1$ to be negligible compared to $c_1$ itself, the policy gradient estimator in \eqref{eqn:pgest} using the debiased group size estimator, achieves an excess MSE (the increase in its MSE compared to using the optimal group size) that is asymptotically no larger or much smaller than that using the biased estimator (Theorem \ref{thm:ratioest}).
\end{enumerate}
On the \textit{negative} side, debiasing introduces two drawbacks:
\begin{enumerate}
    \item[(i)] It removes the positive bias that stabilises the denominator $c_1$ in variance ratio estimation. While the biased estimator of $c_1$ is bounded below by a positive constant multiple of its bias with high probability, its debiased version can be non-positive with non-vanishing probability 
    for sufficiently small $c_1$ (Theorem \ref{thm:varianceest}(c)), 
    rendering its group size estimator based on \eqref{eq:gstar} ill-defined.
    \item[(ii)] This instability can translate into substantially worse group size selection: in the weak signal regime where $c_1$ is much smaller compared to its estimation error, the debiased estimator may select the maximum allowable group size, causing its gradient estimator's excess MSE to attain the largest possible asymptotic order and even diverge to infinity, whereas the biased estimator's excess MSE can converge to zero (Theorem \ref{thm:ratioest}).
\end{enumerate}
The weak signal regime is likely to arise in practice. To illustrate this, Table \ref{tab:varest} reports the estimated variance ratios $c_3/c_1$ and $c_3/c_2$, computed using large samples from three synthetic tasks in Reasoning Gym and two real-world benchmarks, GSM8K and DAPO-Math-17k. The ratio of our interest, $c_3/c_1$, is generally large, particularly for the two real-world datasets. By contrast, $c_3/c_2$ remains close to one across all five tasks, indicating that $c_2$ and $c_3$ are comparable and both are substantially larger than $c_1$. Consequently, sufficiently accurate estimation of $c_1$, with estimation error negligible compared to $c_1$ itself, may require considerably large pilot sample sizes ${b}$ and ${g}$ in Line 3 of Algorithm \ref{alg:adaptive-group}. However, increasing these sample sizes incurs additional computational costs, making the weak signal regime practically relevant.

\begin{table}[t]
\centering
\caption{Estimates of the variance ratios $c_3/c_1$ and $c_3/c_2$ at training step 150. We use large datasets with ${b}={g}=3200$ for estimation, except for DAPO-Math-17k, where both are reduced to 640. For computational efficiency, the variance components are computed using only the parameters in the last four of 28 Transformer blocks of Qwen2.5-1.5B-Instruct for Reasoning Gym and GSM8K, and Qwen2.5-Math-7B for DAPO-Math-17k.}\label{tab:varest}
\setlength{\tabcolsep}{5pt}
\fontsize{8pt}{10pt}\selectfont
\begin{tabular}{@{}lccccc@{}}
\toprule
Ratio & \shortstack{Base\\ conversion}
& \shortstack{Prime\\ factorisation}
& \shortstack{Decimal\\ arithmetic}
& GSM8K & DAPO-Math-17k \\
\midrule
$c_3/c_1$ & $6.97$ & $40.87$ & $28.20$ & $290.93$ & $507.31$ \\
$c_3/c_2$ & $0.93$ & $0.91$ & $1.01$ & $0.92$ & $0.93$ \\
\bottomrule
\end{tabular}
\end{table}

\subsection{Trade-Offs of Debiasing in Variance Estimation}\label{sec:tradeoffvarest}
We first introduce the biased estimators for $c_1$ and $c_3$. 
Recall that $c_1$ can be expressed as the trace of the covariance matrix $\Cov\{{\bm g}(X;\theta)\}$. If ${\bm g}(X_i;\theta)$ were available for each sampled prompt $X_i$, we could estimate $c_1$ by applying the classical sample variance formula to $\{{\bm g}(X_i;\theta)\}_{i=1}^{b}$. However, ${\bm g}(X_i;\theta)$ is a conditional expectation over the completion distribution and is generally intractable to evaluate exactly. A natural approach is to replace ${\bm g}(X_i;\theta)$ with its finite-group estimator ${\widehat{\bm g}}_{g}(X_i;\theta)$ (see \eqref{eqn:pgest}), computed using ${g}$ sampled completions for each prompt $X_i$, yielding
\begin{equation*}
\widehat{c}_1
=\frac{1}{{b}-1}\sum_{i=1}^{b} \operatorname{tr}\Big[\{{\widehat{\bm g}}_{g}(X_i;\theta)-{\widehat{\bm g}}_{{b},{g}}(\theta)\}\{{\widehat{\bm g}}_{g}(X_i;\theta)-{\widehat{\bm g}}_{{b},{g}}(\theta)\}^\top\Big]
=\frac{1}{{b}-1}\sum_{i=1}^{b} \|{\widehat{\bm g}}_{g}(X_i;\theta)-{\widehat{\bm g}}_{{b},{g}}(\theta)\|^2.
\end{equation*}
However, this approximation introduces a bias, as ${\widehat{\bm g}}_{g}(X_i;\theta)$ contains additional variability arising from the finite number of completions per prompt. Indeed, as we establish in Theorem \ref{thm:varianceest} later, its bias can be quantified as
\begin{equation}\label{eqn:c1bias}
\mathbb{E}(\widehat{c}_1)-c_1=\frac{c_2}{{g}}+\frac{c_3}{{g}({g}-1)}.
\end{equation}
Thus, the bias is positive and of order $O({g}^{-1})$, reflecting the effect of using ${g}$ many completions per prompt.

The estimation of $c_3$ follows a similar principle, where we replace the population moments with their finite sample estimates, yielding
\begin{equation*}
    \widehat{c}_3=\frac{1}{{b}}\sum_{i=1}^{b} \left[\Big\{\frac{1}{{g}}\sum_{j=1}^{g}\|\nabla_{\theta} \log \pi_{\theta}(Y_{i,j}|X_i)\|^2\Big\}\Big\{\frac{1}{{g}-1}\sum_{j=1}^{g} (Z_{i,j}-\bar{Z}_i)^2 \Big\}+\|{\widehat{\bm g}}_{g}(X_i;\theta)\|^2\right].
\end{equation*}
Again, the use of finite group estimator $\|{\widehat{\bm g}}_{g}(X_i;\theta)\|^2$ for approximating $\|{\bm g}(X_i;\theta)\|^2$ introduces bias. As we establish in Theorem \ref{thm:varianceest}, this bias can be explicitly calculated as
\begin{equation}\label{eqn:c3bias}
    \mathbb{E} (\widehat{c}_3)-c_3=-\frac{{g}-2}{{g}^2-{g}}c_3+\frac{2(c_0+c_1+c_2)}{{g}},
\end{equation}
where we use $c_0$ to denote $\|{\bm g}(\theta)\|^2$.

We next introduce the debiased estimator. Debiasing is a classical statistical technique that aims to improve estimation accuracy by removing systematic bias from an estimator \citep{firth1993bias}. To apply this idea, note that, according to \eqref{eqn:c1bias} and \eqref{eqn:c3bias}, the biases of $\widehat{c}_1$ and $\widehat{c}_3$ can be expressed as linear combinations of $c_0,c_1,c_2$ and $c_3$. We further estimate $c_0$ and $c_2$ using simple plug-in estimators,
\begin{equation*}
    \widehat{c}_0=\|{\widehat{\bm g}}_{{b},{g}}(\theta)\|^2\,\,\hbox{and}\,\,\widehat{c}_2=\frac{1}{{b}({g}-1)}\sum_{i=1}^{b} \sum_{j=1}^{g} \|\nabla_{\theta}\log \pi_{\theta}(Y_{i,j}|X_i)(Z_{i,j}-\bar{Z}_{i,-j})-{\widehat{\bm g}}_{g}(X_i;\theta)\|^2,
\end{equation*}
whose biases can also be expressed as linear combinations of these four variance components. Writing $\widehat{\bm{c}}=(\widehat{c}_0,\widehat{c}_1,\widehat{c}_2,\widehat{c}_3)^\top$ and $\bm{c}=(c_0,c_1,c_2,c_3)^\top$, we obtain $\mathbb{E}(\widehat{\bm{c}})=\bm{A} \bm{c}$ 
where $\bm{A}$ is a deterministic invertible matrix whose explicit form is derived in \eqref{eq:bias-matrix} in Section~\ref{app:varianceest-proof}. In particular, its second and fourth rows follow from \eqref{eqn:c1bias} and \eqref{eqn:c3bias}, respectively. Inverting this linear relationship yields the debiased estimator 
\begin{equation}\label{eqn:debias}
    \widetilde{\bm{c}}=\bm{A}^{-1} \widehat{\bm{c}}.
\end{equation}
whose estimates for $c_1$ and $c_3$ are denoted by $\widetilde{c}_1$ and $\widetilde{c}_3$, respectively. 

Finally, we formalise the trade-offs between the biased and debiased estimators in terms of variance and ratio estimation. Before proceeding, we impose an upper bound condition and a lower bound condition.   

\begin{assumption}[Upper bound conditions]\label{assump:upper}
    The absolute reward $|r(X,Y)|$, the parameter dimension and the $\ell_2$-norm of the policy score $\|\nabla_{\theta}\log\pi_{\theta}(Y|X)\|$ are uniformly bounded from above by a finite positive constant.
\end{assumption}

\begin{assumption}[Lower bound conditions]\label{assump:lower}
    $c_2$, $c_3$, ${b}\Var(\widehat{c}_3)$ and ${b}\Var(\widehat c_1)/(c_1+{g}^{-2})$ are bounded away from zero.
\end{assumption}

The bounded reward condition in Assumption \ref{assump:upper} is mild, as rewards are typically binary, indicating whether a prompt is answered correctly, or take values in $\{0,1,2\}$ when an additional format reward is included to assess whether the completion follows a prescribed format, as in DeepSeek. The boundedness assumptions on the parameter dimension and the policy score are more restrictive. Nevertheless, as commented in Table \ref{tab:varest}, our implementation does not compute policy scores with respect to the full parameter space, but instead restricts them to a substantially smaller subspace involving only the last few Transformer blocks, making this assumption more reasonable. 

In principle, our theory could be extended to accommodate a growing parameter dimension. Such an extension would introduce dimension-dependent factors into the estimation errors of the variances. However, since our primary interest lies in the ratio $c_3/c_1$, any common multiplicative dimension-dependent factors in the numerator and denominator would cancel. Consequently, allowing the dimension to grow may add technical complexity without providing further insight into group size selection. We therefore focus on the fixed-dimensional setting for simplicity.

The lower bound conditions on $c_2$ and $c_3$ in Assumption \ref{assump:lower} are reasonable in light of Table  \ref{tab:varest}, which shows that both components are substantially larger than $c_1$. The lower bound conditions on the variances of $\widehat c_1$ and $\widehat c_3$ are more technical. Under Assumption \ref{assump:upper}, we can show that $\Var(\widehat c_1)=O{(c_1+{g}^{-2})/{b}}$ and $\Var(\widehat c_3)=O({b}^{-1})$. These lower bound conditions are imposed to ensure that these upper bounds are sharp up to constant factors. 

\begin{theorem}[Trade-off of debiasing in variance estimation]\label{thm:varianceest}
Suppose Assumptions \ref{assump:upper} and \ref{assump:lower} hold. Then
\begin{enumerate}
    \item[(a)] The biases of the biased variance estimators $\widehat{c}_1$ and $\widehat{c}_3$ are given by \eqref{eqn:c1bias} and \eqref{eqn:c3bias}, respectively. To the contrary, the debiased estimators $\widetilde{c}_1$ and $\widetilde{c}_3$ given by \eqref{eqn:debias} are unbiased.
    \item[(b)] $\widehat{c}_1$ and $\widehat{c}_3$ are consistent when both ${b}$ and ${g}$ approach infinity. To the contrary, consistencies of $\widetilde{c}_1$ and $\widetilde{c}_3$ require only ${b}$ to approach infinity.
    \item[(c)] The biased estimator $\widehat c_1$ is non-negative, almost surely. Additionally, there exist some constants $c$,  $\bar{c}>0$ such that 
    \begin{equation*}
        \mathbb{P}\big(\widehat{c}_1<\frac{\bar{c}}{{g}}\big)=O\big(\exp(-c{b})\big). 
    \end{equation*}
    By contrast, for the debiased estimator $\widetilde{c}_1$, there exists some constant $\nu>0$ such that
    \begin{equation}\label{eqn:debiaselowerbound}
        \mathbb{P}(\widetilde{c}_1\leq 0)\geq \Phi\big(-\frac{c_1{g}\sqrt{{b}}}{\nu}\big)-O\big(\frac{1}{\sqrt[3]{{b}}}+\frac{1}{{g}}\big),
    \end{equation}
    where $\Phi$ denotes the cumulative distribution function of a standard normal random variable. 
\end{enumerate}
\end{theorem}
Theorem \ref{thm:varianceest} characterises the trade-offs between the biased and debiased estimators of the variance components $c_1$ and $c_3$. Specifically, Parts (a) and (b) establish the benefits of debiasing in terms of unbiasedness and consistency under weaker conditions, without requiring ${g}$ to diverge. Part (c), on the other hand, reveals its potential drawback: the debiased denominator estimate can be non-positive. To better understand \eqref{eqn:debiaselowerbound}, consider fixing ${b}$ and ${g}$ and letting $c_1\to0$. The RHS of \eqref{eqn:debiaselowerbound} then approaches $1/2$, up to an error of order $O({b}^{-1/3}+{g}^{-1})$. Thus, provided ${b}$ and ${g}$ are not overly small, the probability that $\widetilde{c}_1$ is non-positive remains bounded away from zero. By contrast, thanks to its positive bias of order ${g}^{-1}$ (see \eqref{eqn:c1bias}), the biased estimator $\widehat{c}_1$ is bounded below by a positive multiple of ${g}^{-1}$ with high probability, with the probability of violating this bound decaying exponentially in ${b}$.

\subsection{Trade-Offs of Debiasing in Group Size Selection}\label{sec:tradeoffgroupest}
In this section, we demonstrate how the trade-offs in variance estimation translate into group size selection. We begin by introducing the biased and debiased group size estimators. 

According to Theorem \ref{thm:varianceest}(c), the biased estimator $\widehat c_1$ is non-negative almost surely and strictly positive with high probability, making the ratio estimator $\widehat{c}_3/\widehat{c}_1$ well defined. 
Plugging $\widehat c_1$ and $\widehat c_3$ into \eqref{eq:gstar} yields its group size estimator
\begin{equation}\label{eqn:biasest}
{\widehat {G}}_{\rm bias}=\min\big(1+\sqrt{\widehat c_3/\widehat c_1},{{G}}_{\max}\big),
\end{equation}
where ${{G}}_{\max}$ denotes the maximum allowable group size. We impose this upper bound to prevent excessively large variance ratio estimate $\widehat c_3/\widehat c_1$.

To the contrary, since the debiased estimator $\widetilde c_1$ can be non-positive with non-vanishing probability, directly applying the scaling law \eqref{eq:gstar} may yield an ill-defined group size estimate. To address this, we define 
\begin{equation}\label{eqn:debiasest}
{\widehat{{G}}}_{\rm debias}=
\begin{cases}
{{G}}_{\max}, & \widetilde{c}_1\leq 0,\\
\min\Big\{1+\sqrt{\max(\widetilde{c}_3/{\widetilde{c}_1},1)},{{G}}_{\max}\Big\}, & \widetilde{c}_1>0.
\end{cases}
\end{equation}
First, when $\widetilde c_1\leq 0$, the variance ratio is invalid, and we set the selected group size to ${{G}}_{\max}$. Second, when $\widetilde c_1>0$, we truncate the estimated ratio from below at one to ensure that the selected group size is at least two, so that the leave-one-out reward $\bar{Z}_{i,-j}$ can be computed for each group. We also impose the upper bound ${{G}}_{\max}$ to prevent excessively large group sizes. Third, the lower truncation is unnecessary for the biased estimator in \eqref{eqn:biasest}, since $\widehat c_3\geq\widehat c_1$ always holds (Section~\ref{app:raw-risk-proof}), ensuring that the estimated ratio is at least one.

To compare the two group size estimators in \eqref{eqn:biasest} and \eqref{eqn:debiasest}, we evaluate their performance using the normalised excess gradient MSE. Specifically, for an estimated group size ${\widehat {G}}$, we define
\begin{equation}\label{eqn:RG}
    \mathcal{R}({\widehat{{G}}})={{N}}\mathbb{E}\Big[\text{MSE}\{{\widehat{\bm g}}_{{{N}}/{\widehat{{G}}},{\widehat{{G}}}}(\theta)\}-\text{MSE}\{{\widehat{\bm g}}_{{{N}}/{{G}}^*,{{G}}^*}(\theta)\}\Big],
\end{equation}
where we recall that ${{N}}$ denotes the total sampling budget, ${{G}}^*$ denotes the optimal group size determined by \eqref{eq:gstar}, and the expectation is taken with respect to the variability in ${\widehat{{G}}}$. 

By definition, the excess MSE measures the difference between the MSE of the gradient estimator using the selected ${\widehat {G}}$ and that using the oracle group size ${{G}}^*$. The normalising factor ${{N}}$ is motivated by the fact that the oracle MSE is of order ${{N}}^{-1}$ under Assumptions \ref{assump:upper} and \ref{assump:lower}. Consequently, \eqref{eqn:RG} quantifies the relative increase in the gradient MSE compared with the oracle. 

By Lemma \ref{prop:score-closed-forms}, it can be shown that 
\begin{equation}\label{eqn:RG2}
    \mathcal{R}({\widehat{{G}}})=\mathbb{E}\Big\{\frac{c_1({\widehat{{G}}}-{{G}}^*)^2}{{\widehat{{G}}}-1}\Big\}.
\end{equation}
Consequently, the excess MSE is directly related to the MSE of the estimated ${{G}}^*$, i.e., $\mathbb{E} ({\widehat{{G}}}-{{G}}^*)^2$. Meanwhile, the denominator ${\widehat {G}}-1$ in \eqref{eqn:RG2} introduces an inherent asymmetry between over- and underestimation: overestimating the optimal group size incurs a smaller excess MSE than underestimating it.

\begin{figure}[t]
\centering
\includegraphics[width=\linewidth]{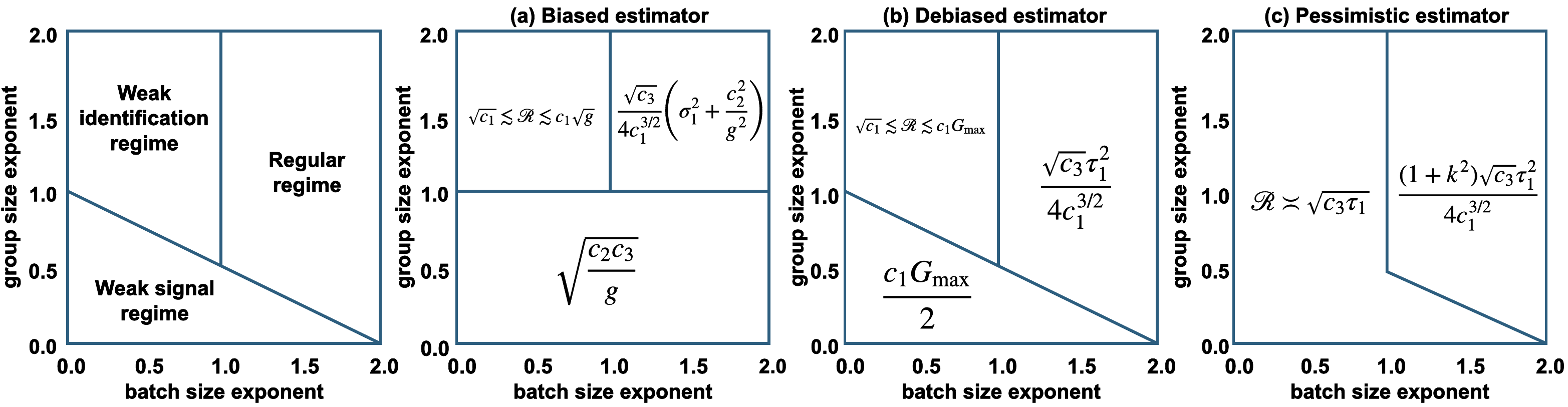}
\caption{Expected excess gradient MSE $\mathcal R$ across the three regimes. The last three panels summarise the asymptotic behaviour of $\mathcal R$ for each estimator. Expressions without symbols represent the leading terms of asymptotic equivalents, whereas $\asymp$ denotes equivalence up to multiplicative constants, and $\lesssim$ denotes asymptotic lower or upper bounds. Here, $\sigma_1^2$ and $\tau_1^2$ denote the variances of the biased estimator $\widehat{c}_1$ and the debiased estimator $\widetilde{c}_1$, respectively, both of order $(c_1+{g}^{-2})/{b}$. See Section~\ref{app:pessimistic-risk-proof} for precise mathematical statements.}
\label{fig:estasymptotic}
\end{figure}

We next describe the asymptotic regime considered in our analysis. To simplify the theoretical comparisons among different estimators, we let the total pilot sample size ${{n}}={b}{g}$ approach infinity and characterise the leading terms of their excess MSE, or establish lower and upper bounds when their asymptotic orders cannot be determined. Note that ${{n}}$ differs from ${{N}}$: the former denotes the sampling budget used in the pilot study for group size selection, whereas the latter denotes the sampling budget used to construct the policy gradient estimator in \eqref{eqn:pgest} with the selected group size.

More specifically, we consider a joint asymptotic regime in which the pilot sample sizes ${b}$ and ${g}$, together with the maximum allowable group size ${{G}}_{\max}$, diverge as ${{n}}\to\infty$, while the denominator $c_1$ converges to zero, reflecting its small magnitude compared to $c_3$. To formalise this asymptotic regime, we further assume that ${b}$, ${g}$, ${{G}}_{\max}$ and $c_1$ scale polynomially with ${{n}}$, as specified in Assumption \ref{assump:asym} below. 

\begin{assumption}[Asymptotic regime]\label{assump:asym}
    The quantities $c_1$, ${b}$, ${g}$ and ${{G}}_{\max}$ are proportional to ${{n}}^{-\kappa_1}$, ${{n}}^{\kappa_2}$, ${{n}}^{\kappa_3}$ and ${{n}}^{\kappa_4}$, respectively, for some positive constants $\kappa_1,\kappa_2,\kappa_3$ and $\kappa_4$. We further assume $\kappa_4>\kappa_1$.
\end{assumption}
Under this condition, we define two sample size exponents: the batch size exponent $\kappa_{b}=\kappa_2/\kappa_1$ and the group size exponent $\kappa_{g}=\kappa_3/\kappa_1$. These two exponents correspond to the horizontal and vertical axes in Figure \ref{fig:estasymptotic}, and measure how fast ${b}$ and ${g}$ grow compared to $c_1^{-1}$. In particular, $\kappa_{b}>1$ or $\kappa_{g}>1$ indicates that ${b}$ or ${g}$, respectively, grows faster than $c_1^{-1}$. These exponents play crucial roles in determining the asymptotic regimes. The leftmost panel of Figure~\ref{fig:estasymptotic} visualises the three regimes. 
\begin{enumerate}
    \item \textbf{Regular regime}: This regime corresponds to the classical asymptotic setting where estimation errors are sufficiently small for standard asymptotic approximations to apply. In our problem, the potentially small denominator $c_1$ requires its estimation error to be negligible compared to $c_1$ itself. Specifically, it can be shown that $\tau_1^2=\Var(\widetilde{c}_1)$ is of the order $(c_1+{g}^{-2})/{b}$ (Lemma~\ref{lem:coefficient-variance}). Thus, the regular regime requires $(c_1+{g}^{-2})/{b}\ll c_1^2$, or equivalently, both ${b}c_1\to\infty$ and $c_1{g}\sqrt{{b}}\to\infty$. Under Assumption \ref{assump:asym}, these conditions translate into $\kappa_{b}>1$ and $\kappa_{g}+\kappa_{b}/2>1$, which occupies the upper-right region of the sample-size space.
    
    \item \textbf{Weak signal regime}: This regime is motivated by Theorem \ref{thm:varianceest}(c), which shows that when $c_1{g}\sqrt{{b}}$ is sufficiently small, the debiased estimate $\widetilde{c}_1$ can be non-positive with non-vanishing probability. The name ``weak signal'' reflects the fact that the signal contained in the denominator $c_1$ is so weak that even after being scaled by ${g}\sqrt{{b}}$, it remains too small to ensure reliable estimation. Under Condition \ref{assump:asym}, this regime corresponds to the subspace $\kappa_{g}+\kappa_{b}/2< 1$, which occupies the bottom-left corner of the sample-size space.
    
    \item \textbf{Weak identification regime}: The remaining upper left region, defined by $\kappa_{b}\leq 1$ and $\kappa_{g}+\kappa_{b}/2>1$, forms the weak identification regime. Its name reflects the fact that, as we establish in Theorem \ref{thm:ratioest} below, the available bounds for the two plug-in estimators need not determine how their excess MSEs compare in this regime.
\end{enumerate}

We next characterise the excess MSEs of the biased and debiased estimators. 
\begin{theorem}[Excess MSEs of biased and debiased estimators]\label{thm:ratioest}
Suppose Assumptions \ref{assump:upper} -- \ref{assump:asym} hold. Then the excess MSEs of the biased and debiased group size estimators satisfy the asymptotic expressions and bounds presented in the middle two panels of Figure \ref{fig:estasymptotic}, respectively. 
\end{theorem}

Theorem \ref{thm:ratioest} reveals the trade-offs of debiasing in group size selection. 
In particular:
\begin{enumerate}
    \item In the \textbf{regular regime}, when ${g}\gg c_1^{-1}$ (i.e., $\kappa_{g}>1$), the biased estimator incurs an additional excess MSE of $\sqrt{c_3}c_2^2/(4c_1^{3/2}{g}^2)$ compared to the debiased estimator, due to its finite-group bias. This additional term dominates when ${g}\ll\sqrt{{b}/c_1}$, yielding $\mathcal{R}({\widehat{{G}}}_{\text{bias}})\gg\mathcal{R}({\widehat{{G}}}_{\text{debias}})$, as illustrated by the orange region in Figure \ref{fig:estcomparison}. The same conclusion holds for smaller ${g}$ in the remaining part of the regular regime. These results suggest that when the denominator $c_1$ can be estimated sufficiently accurately, debiasing improves variance ratio estimation and, consequently, group size selection.
    \item In the \textbf{weak signal regime}, the debiased estimator incurs an excess MSE of order $c_1{{G}}_{\max}$, which diverges at the largest possible asymptotic order under Assumption \ref{assump:asym}. By contrast, the biased estimator's excess MSE converges to zero as ${{n}}\to\infty$. Interestingly, this suggests that the positive bias of $\widehat c_1$, while undesirable for variance estimation, acts as a regularisation that stabilises the estimated ratio. Removing this bias can therefore substantially worsen group size selection. 
    \item In the \textbf{weak identification} regime, our theoretical bounds do not establish whether either estimator is superior. Although both have excess MSE lower bounds of order $\sqrt{c_1}$, the debiased estimator has a potentially much larger upper bound of order $c_1{{G}}_{\max}$.
\end{enumerate}
In summary, both the biased and debiased estimators have their own limitations. The biased estimator suffers from finite-group bias, which can be particularly pronounced when ${g}$ is small. Although the debiased estimator eliminates this bias, it can become highly unstable when the denominator $c_1$ is small, as estimation errors in the denominator are amplified in the resulting ratio. These limitations motivate our proposed pessimistic estimator, which builds on the debiased estimator while explicitly accounting for estimation uncertainty to mitigate the instability caused by small denominators. We introduce this estimator in the next section.

\section{The Pessimistic Estimator}\label{sec:pessimistic} 
The failure of debiasing highlights that minimising the estimated gradient MSE can favour group sizes whose losses happen to be underestimated, leading to substantially high true MSEs. We therefore stay pessimistic by minimising an upper confidence bound on the gradient MSE. Similar adjustments arise in work on intersection bounds, data integration and offline reinforcement learning  \citep{chernozhukov2013intersection,swaminathan2015batch,swaminathan2015self,liu2020provably,jin2021pessimism,rashidinejad2021bridging,yang2023pessimistic,wu2025pessimistic}, among others.

To elaborate the pessimistic principle, we draw an analogy between our group size selection and the classical multi-armed bandit problem \citep{lai1985asymptotically}. We regard each candidate group size ${G}$ as an arm and the MSE of the corresponding policy gradient estimator as its cost. %
According to Lemma \ref{prop:score-closed-forms}, under a fixed sampling budget ${{N}}$, this cost is given by $c_1{G}/{{N}}+c_2/{{N}}+c_3/{{{N}}({G}-1)}$. Since the second term $c_2/{{N}}$ is independent of ${G}$, minimising this cost is equivalent to minimising
\[
    \ell({G})=c_1{G}+\frac{c_3}{{G}-1}.
\]
Our objective is therefore to identify the optimal arm that minimises $\ell({G})$.

From this perspective, the debiased estimator follows a non-pessimistic, greedy action selection approach: it first estimates the loss function $\ell(G)$ and then selects the arm that minimises the estimated loss $\widehat{\ell}(G)=\widetilde{c}_1 G+\widetilde{c}_3/(G-1)$. However, this approach ignores the estimation uncertainty. When $c_1$ is small, even a slight underestimation of $c_1$ can substantially underestimate the first loss component $c_1G$ for large $G$, leading to the selection of an excessively large group size and, consequently, a substantial loss.

To address this limitation, we introduce a pessimistic rule, which enforces the known sign of $c_1$ and adds a penalty for uncertainty in the estimated loss. To see how the penalty affects group size selection, we first treat the covariance matrix of $(\widetilde c_1,\widetilde c_3)$ as known. Both variance components $c_1$ and $c_3$ are still estimated from the pilot. The standard deviation $\sigma({G})$ of $\widehat\ell({G})$ satisfies
\begin{equation}\label{eq:loss-variance}
\sigma^2({G})={G}^2\Var(\widetilde c_1)
 +\frac{2{G}}{{G}-1}\Cov(\widetilde c_1,\widetilde c_3)
 +\frac{\Var(\widetilde c_3)}{({G}-1)^2}.
\end{equation}
For a fixed penalty multiplier $k>0$, write $(u)_+=\max\{u,0\}$, we select the group size by minimising the following pessimistically adjusted loss:
\begin{equation}\label{eqn:pessiest}
{\widehat {G}}_{\rm pes}=\argmin_{2\le {G}\le {{G}}_{\max}} Q_k({G}),\qquad
Q_k({G}):={G}(\widetilde c_1)_++\frac{\widetilde c_3}{{G}-1}+k\sigma({G}).
\end{equation}

To see how pessimism discourages excessively large group sizes, consider a conservative upper bound on the standard deviation, obtained by the Cauchy--Schwarz inequality:
\[
\bar\sigma(G)=G\operatorname{sd}(\widetilde c_1)+\frac{\operatorname{sd}(\widetilde c_3)}{G-1}\geq \sigma(G).
\]
Replacing $\sigma(G)$ in \eqref{eqn:pessiest} with $\bar\sigma(G)$ yields a closed-form solution:
\begin{eqnarray}\label{eqn:pessi2}
\widehat G_{\rm pes}
=1+\sqrt{\frac{\widetilde c_3+k\operatorname{sd}(\widetilde c_3)}
{(\widetilde c_1)_++k\operatorname{sd}(\widetilde c_1)}},
\end{eqnarray}
when the lower- and upper-bound constraints on $G$ are ignored. Compared with the debiased estimator in \eqref{eqn:debiasest}, \eqref{eqn:pessi2} incorporates estimation uncertainty into both the numerator and denominator. Together with clipping, this adjustment ensures that the denominator is strictly positive, which guards against underestimating $c_1$ and excessively large group sizes.

Importantly, this additional protection does not sacrifice the polynomial convergence rate in the regular regime, %
as demonstrated in the following theorem. %

\begin{theorem}[Excess MSE of the pessimistic estimator]\label{thm:pessi}
Under Assumptions~\ref{assump:upper}--\ref{assump:asym}, for each fixed $k>0$, the expected excess MSE of the pessimistic estimator in \eqref{eqn:pessiest} satisfies the asymptotic relations shown in the rightmost panel of Figure~\ref{fig:estasymptotic}.
\end{theorem}

The comparison with the plug-in estimators in the middle panels of Figure~\ref{fig:estasymptotic} depends on how precisely the pilot estimates $c_1$.
\begin{enumerate}
    \item In the \textbf{regular regime}, pessimism has the same rate as debiasing and increases the leading constant by a factor of $1+k^2$. The factor measures the cost of adding $k$ standard errors when the pilot already estimates $c_1$ precisely. Our numerical results in Section~\ref{sec:experiments} show that this constant-factor increase has little practical impact, as both estimators achieve similarly small MSEs for large pilot sample sizes.
    \item In the \textbf{weak signal regime}, the pessimistic estimator has vanishing excess MSE, while the debiased estimator incurs excess MSE asymptotic to $c_1{{G}}_{\max}/2$. To compare pessimism with the biased estimator, note that $\tau_1^2=\Var(\widetilde{c}_1)\asymp(c_1+{g}^{-2})/{b}$ and that the biased estimator's excess MSE is of order ${g}^{-1/2}$. The ratio of their risks, pessimistic to biased, has order $\{(1+{g}^2c_1)/{b}\}^{1/4}$. Pessimism therefore has the faster rate when ${g}^2c_1=o({b})$, which holds over a substantial portion of the regime shown in Figure~\ref{fig:estcomparison}.
    \item In the \textbf{weak identification regime}, the pessimistic estimator's excess MSE has order $\sqrt{c_3{{\tau_1}}}$ and tends to zero. This rate is asymptotically much smaller than the upper bound of order $c_1G_{\max}$ for the debiased estimator.
\end{enumerate}
In our implementation, the covariance matrix of $(\widetilde c_1,\widetilde c_3)^\top$ in \eqref{eq:loss-variance} is unknown and is estimated from the ${b}$ sampled prompts using the grouped jackknife estimator (See Section~\ref{app:jackknife-cov} for details), yielding the estimated standard error $\widehat\sigma({G})$. For large $G$, this standard error is dominated by $G\widehat\tau_1$, where $\widehat\tau_1$ denotes the estimated standard error of the denominator $\widetilde c_1$.  
In small samples, or when the ${b}$ sampled prompts miss rare prompt types that drive between-prompt variation, $\widehat\tau_1$ can occasionally be close to zero when $\widetilde c_1\le0$, weakening the penalty on large ${G}$. To guard against this finite-sample degeneracy, we can impose a floor $\epsilon_0\ge0$ on $\widehat\tau_1$, leading to
\begin{equation}\label{eq:main-floored-pess}
\widehat Q_k({G})={G}(\widetilde c_1)_++\frac{\widetilde c_3}{{G}-1}
+k\bigl\{\widehat\sigma({G})+{G}(\epsilon_0-\widehat\tau_1)_+\bigr\}.
\end{equation}
Setting $\epsilon_0=0$ recovers the clipped pessimistic rule in \eqref{eqn:pessiest}, with $\sigma(G)$ replaced by its jackknife estimate, whereas $\epsilon_0>0$ yields the floored pessimistic rule. For large $G$, even when $\widehat\tau_1$ and hence $\widehat\sigma(G)$ approach zero, the floor ensures that the uncertainty penalty is approximately $kG\epsilon_0$, which thus discourages large group sizes.

In Section~\ref{app:covariance-pessimistic-risk}, we analyse \eqref{eq:main-floored-pess} and establish that it preserves the asymptotic guarantees of Theorem~\ref{thm:pessi}. We further show that calibrating $k$ using standard normal or Student-$t$ quantiles, as done in our implementation (Section \ref{sec:experiments}), allows $\widehat Q_k(G)$ to be interpreted as an asymptotically valid upper confidence bound for the true loss $\ell(G)$ at a prescribed nominal confidence level of $1-\alpha$.

\section{Experiments}\label{sec:experiments}
In this section, we conduct extensive experiments to evaluate the proposed pessimistic group size selection from two perspectives:
\begin{enumerate}
    \item \textbf{Gradient estimation} (Section~\ref{sec:offline-diagnosis}): We compare the MSE of the policy gradient estimator in \eqref{eqn:pgest} under group sizes selected by the biased, debiased and pessimistic estimators (Figure~\ref{fig:mse-mechanisms}). The results are consistent with our theoretical findings in Figure~\ref{fig:estcomparison}, with the pessimistic estimators achieving the lowest average excess MSE in \textbf{27} out of \textbf{30} experimental settings across different sample sizes.
    \item \textbf{Policy optimisation} (Section~\ref{sec:policy-optimization}): The improvements in gradient estimation accuracy translate into better downstream policy optimisation. GRPO-type algorithms with our proposed adaptive group sizes generally outperform those with fixed group sizes across synthetic reasoning tasks from Reasoning Gym as well as mathematical reasoning and code generation benchmarks (Figure~\ref{fig:policy-current}, Tables~\ref{tab:qwen3-current} \& \ref{tab:coding-current}), while incurring only a $7.03\%$--$10.33\%$ increase in training time (Table~\ref{tab:policy-estimation-overhead}). 
\end{enumerate}

We next present the two sets of experiments, with additional details deferred to Section~\ref{app:current-experiments} to save space.

\subsection{Gradient Estimation}\label{sec:offline-diagnosis}
We first compare the MSEs of policy gradient estimators under group sizes selected by the biased, debiased, clipped and floored pessimistic estimators. Here, both pessimistic rules minimise \eqref{eq:main-floored-pess} using the grouped jackknife standard error $\widehat\sigma({G})$ and the $90\%$ standard normal quantile for $k$, with $\epsilon_0=0$ (clipped) or $\epsilon_0=1000/(g\sqrt b)$ (floored) to assess the effect of the standard-error floor. We consider 6 synthetic reasoning tasks from Reasoning Gym (base conversion, prime factorisation, decimal arithmetic, equation solving, fraction arithmetic, and knights-and-knaves puzzles) together with the real-world mathematical reasoning benchmark GSM8K. For each dataset, we first post-train a \texttt{Qwen2.5-1.5B-Instruct} model using standard GRPO. We then select model checkpoints covering the early, middle, and late stages of training, including those shown in Figure~\ref{fig:mse-geometry}, and evaluate the MSE of the policy gradient at the model parameters corresponding to these checkpoints.

\begin{figure}[!tbp]
\centering
\includegraphics[width=\textwidth]{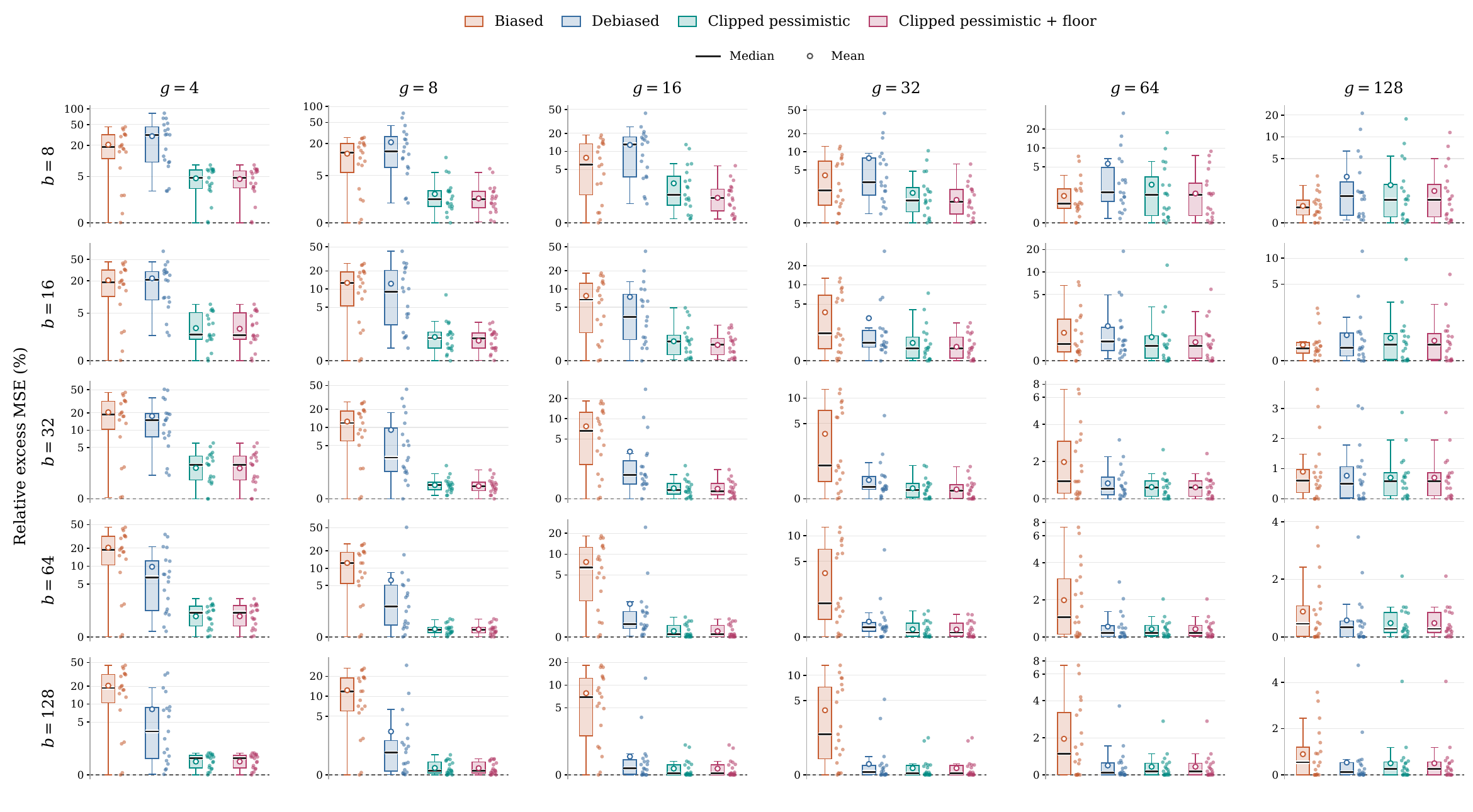}
\caption{Boxplots of relative excess MSEs across different dataset-checkpoint combinations. The black lines and white circles visualise the medians and means, respectively, across these combinations.}\label{fig:mse-mechanisms}
\end{figure}

For each checkpoint, we approximate the ground truth policy gradient using Monte Carlo estimation with 1024 prompts and 256 completions per prompt. We then vary the pilot sample sizes $b\in\{8,16,32,64,128\}$ and $g\in\{4,8,16,32,64,128\}$ to explore different sample-size regimes. For each pair $({b},{g})$, we generate the corresponding prompt-completion-reward data and use the four estimators to select the group size ${\widehat {G}}$. We then fix the sampling budget for gradient estimation at ${{N}}=512$ and compute the policy gradient estimator ${\widehat{\bm g}}_{{{N}}/{\widehat {G}},{\widehat {G}}}$ in \eqref{eqn:pgest} using independently sampled data. Since these gradient estimators are unbiased, their MSEs equal the traces of their covariance matrices. We therefore generate 256 independent gradient estimates for each selected ${\widehat {G}}$ and estimate the corresponding MSE using the trace of their sample covariance matrix. Next, we repeat the entire procedure 25 times, each with an independently selected ${\widehat {G}}$, and average the resulting MSE estimates to account for randomness in group size selection. We further compute the relative excess MSE, defined as the difference between the estimated MSE and an approximation to the oracle MSE under the optimal group size, divided by the latter. Finally, for each sample size configuration $({b},{g})$, we average the relative excess MSE across different dataset-checkpoint combinations and report the results in Figure  \ref{fig:mse-mechanisms}.

We make three observations: (i) When both ${b}$ and ${g}$ are small, the debiased estimator incurs the largest relative excess MSE.
This observation is consistent with our theoretical findings in the weak signal regime. (ii) When the sample sizes become large, the debiased and pessimistic estimators achieve comparably small relative excess MSEs, whereas the biased estimator still suffers without a sufficiently large ${g}$. This agrees with our theoretical findings in the regular regime. (iii) Overall, both pessimistic estimators demonstrate competitive performance across different combinations of sample sizes. The clipped pessimistic estimator achieves lower average relative excess MSE than both biased and debiased estimators in \textbf{27} out of \textbf{30} settings. Adding the standard-error floor further reduces the average relative excess MSE in \textbf{23} out of \textbf{30} settings.

\subsection{Policy optimisation}
\label{sec:policy-optimization}
We next demonstrate that the pessimistic estimator's advantage in reducing the gradient MSE translates into better downstream policy optimisation. In addition to post-training \texttt{Qwen2.5-1.5B-Instruct} on Reasoning Gym and GSM8K, we further consider the mathematical reasoning dataset DAPO \citep{yu2025dapo} and code generation tasks from AceCode \citep{zeng2025acecoder}. To evaluate the performance of our proposal across different models, we use the larger \texttt{Qwen3-8B} model for DAPO and \texttt{DeepSeek-Coder-1.3B-Base} for AceCode. 

For each training dataset, we post-train the corresponding model using the GRPO-type algorithm with our proposed adaptive group size selection method (Algorithm~\ref{alg:adaptive-group}). Specifically, we adopt the clipped pessimistic estimator in \eqref{eq:main-floored-pess} with $\epsilon_0=0$ and $k$ being the 90\% standard normal quantile. We compare its performance against that of models trained with fixed group sizes ${{G}}\in \{2,4,8,16,32,64\}$ throughout training, and evaluate each model's accuracy on the corresponding test datasets, measured by the percentage of test prompts answered correctly. We summarise the results as follows:

\begin{figure}[t]
\centering
\includegraphics[width=0.95\textwidth]{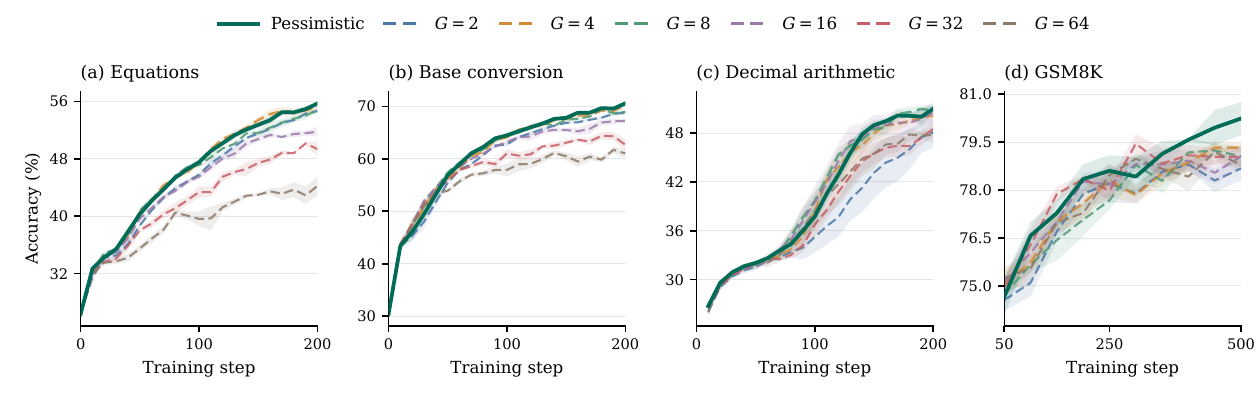}
\caption{Test accuracy curves of models trained on Reasoning Gym and GSM8K, averaged over five random seeds, with shaded areas representing standard errors. Solid lines indicate models trained using the proposed pessimistic group size selection method, whereas dashed lines represent fixed-group-size baselines.}\label{fig:policy-current}
\end{figure}
\begin{enumerate}
    \item Figure \ref{fig:policy-current} presents the mean test accuracies throughout training for models trained on Reasoning Gym and GSM8K, averaged over five random seeds, where the shaded areas represent the corresponding standard deviations. The final test accuracies at the last training iteration are reported in Table {\ref{tab:policy-current}} (Section {\ref{app:policy-details}}). Here, the test datasets are generated from Reasoning Gym or the test split of GSM8K, respectively. The proposed method (denoted by ``Pessimistic'') generally achieves the highest test accuracy throughout training. Its accuracy curves consistently lie above those of most fixed-group-size baselines, with substantial performance gaps against several baselines in the middle and later stages of training. 
    \item Table~\ref{tab:qwen3-current} reports the test accuracies of the post-trained \texttt{Qwen3-8B} models across ten mathematics benchmarks, ranging from school-level to university-level and olympiad problems. These test benchmarks are separate datasets from the DAPO training dataset. The proposed adaptive algorithm generally achieves the best performance: it attains the highest accuracy on six of the ten benchmarks and the highest average accuracy across all ten.
    \item Table {\ref{tab:coding-current}} (Section {\ref{app:coding-details}}) reports greedy accuracy, the average pass rate over eight sampled completions (Average@8), and the proportion of problems solved by at least one completion (Pass@8) of the fine-tuned \texttt{DeepSeek-Coder-1.3B-Base} models on HumanEval+, MBPP+ \citep{liu2023evalplus} and the test split of AceCode. The proposed adaptive algorithm again achieves the best overall performance, attaining the highest Average@8 on all three benchmarks, as well as the highest greedy accuracy on MBPP+ and AceCode and Pass@8 on MBPP+.
\end{enumerate}
In summary, these results demonstrate the effectiveness of the proposed adaptive group size selection in improving downstream LLM post-training performance across a broad range of synthetic and real-world benchmarks. %

\begin{table}[t]
\centering
\small
\renewcommand{\arraystretch}{1.04}
\setlength{\tabcolsep}{4pt}
\caption{Test accuracies (\%) of \texttt{Qwen3-8B} models trained on mathematical reasoning problems from the DAPO collection, evaluated on ten mathematics benchmarks and their average (Avg.). The best and second-best accuracies in each column are highlighted in bold and underlined, respectively.}
\label{tab:qwen3-current}
\begin{tabular}{@{}l*{3}{r}@{\hspace{12pt}}*{7}{r}@{\hspace{7pt}}r@{}}
\toprule
& \multicolumn{3}{c}{Greedy accuracy} & \multicolumn{7}{c}{Average@16} & \\
\cmidrule(lr{6pt}){2-4}\cmidrule(l{6pt}r){5-11}
Method & \shortstack{MATH\\500} & Minerva & Olymp. & \shortstack{AIME\\2024} & \shortstack{AIME\\2025} & \shortstack{AMC\\2023} & \shortstack{HMMT\\2024} & \shortstack{HMMT\\2025} & \shortstack{BRUMO\\2025} & \shortstack{CMIMC\\2025} & \textbf{Avg.}\\
\midrule
${{G}}=4$ & 88.40 & \textbf{48.53} & 57.04 & 33.75 & 25.83 & 77.97 & 12.50 & 11.25 & 36.25 & 16.56 & 40.81\\
${{G}}=8$ & \underline{90.80} & 46.32 & 60.15 & 38.33 & \underline{31.04} & \underline{80.78} & \underline{18.33} & 10.83 & \underline{43.96} & 16.25 & 43.68\\
${{G}}=16$ & 89.80 & \underline{47.79} & \underline{61.04} & \textbf{40.21} & 30.00 & \underline{80.78} & 16.67 & 12.92 & \textbf{44.17} & \underline{18.59} & 44.20\\
${{G}}=32$ & 90.20 & 43.38 & 59.26 & 32.08 & 28.12 & 77.66 & 13.75 & 10.83 & 41.46 & 17.19 & 41.39\\
${{G}}=64$ & \underline{90.80} & 47.43 & \underline{61.04} & \underline{40.00} & \textbf{31.25} & 80.31 & 17.08 & \underline{14.17} & \textbf{44.17} & 18.44 & \underline{44.47}\\
\midrule
\textbf{Pessimistic} & \textbf{92.20} & 45.59 & \textbf{62.96} & 38.75 & 30.00 & \textbf{84.38} & \textbf{19.58} & \textbf{14.79} & 43.33 & \textbf{19.84} & \textbf{45.14}\\
\botrule
\end{tabular}
\end{table}

\section{Discussion}\label{sec:discussion}
How to allocate limited sampling budgets to obtain accurate policy gradient estimates and improve downstream policy optimisation remains an open question in training modern large reasoning models. This paper brings classical statistical tools, including U-statistics, variance  estimation and uncertainty quantification, to address such a problem in GRPO-type training. We develop a comprehensive theory to characterise the estimation errors of different estimators across a range of asymptotic regimes (Figure \ref{fig:estasymptotic}) and establish favourable convergence guarantees for the proposed pessimistic estimators (Figure \ref{fig:estcomparison}). Several directions warrant further investigation:
\begin{enumerate}
    \item We use a pilot study with $b$ prompts and $g$ completions per prompt to estimate the optimal training group size $G$. A related question is how to optimally choose $g$ itself to enhance the estimation of $G$ under a fixed pilot sampling budget $n=bg$. This gives rise to a classical experimental design problem \citep{Fisher1935}. Our theoretical results suggest that a promising approach is to minimise the variance $\tau_1^2=\Var(\widetilde c_1)$, which governs the leading excess gradient MSE of the pessimistic estimator (Figure \ref{fig:estasymptotic}). In particular, since $\tau_1^2\asymp(c_1+g^{-2})/b=(c_1g+g^{-1})/n$, balancing the two terms yields an optimal pilot group size of order $g\asymp c_1^{-1/2}$. Developing a data adaptive procedure for selecting this optimal $g$ is a particularly exciting direction for future research.
    \item Our analysis focuses on a simplified version of GRPO-type algorithms, omitting reward-variance normalisation, response-length normalisation, PPO-type minibatch updates, and Kullback--Leibler (KL) regularisation. We also consider classical stochastic gradient ascent rather than more widely used adaptive optimisers Adam \citep{kingma2014adam}. Although some of these components have been removed in recent variants, such as Dr. GRPO and GPG, extending our theoretical framework to accommodate more sophisticated optimisation procedures remains a  practically relevant direction.
    \item Our analysis focuses on \textit{within-iteration} sampling allocation, choosing the batch and group sizes under a fixed sampling budget per iteration. However, under a fixed total training budget, an additional trade-off arises in allocating the budget \textit{across iterations}: allocating more samples to each iteration yields more accurate policy gradient estimates, but reduces the total number of training iterations. Jointly optimising the sampling budget per iteration and the number of training iterations, together with group size selection, is therefore an interesting direction for future research. 
    \item Our approach assigns the same number of completions to every sampled prompt. However, different prompts can vary substantially in difficulty and thus reward variability. For instance, very easy or very difficult prompts may provide limited information from additional completions, whereas prompts with more variable rewards may benefit from larger groups \citep{sun2026curverl}. This motivates us to select prompt-dependent group sizes.
\end{enumerate}

\bibliography{reference}

@inproceedings{ahmadian2024back,
  title={Back to basics: Revisiting {REINFORCE}-style optimization for learning from human feedback in {LLMs}},
  author={Ahmadian, Arash and Cremer, Chris and Gall{\'e}, Matthias and Fadaee, Marzieh and Kreutzer, Julia and Pietquin, Olivier and {\"U}st{\"u}n, Ahmet and Hooker, Sara},
  booktitle={Proceedings of the 62nd Annual Meeting of the Association for Computational Linguistics (Volume 1: Long Papers)},
  pages={12248--12267},
  year={2024}
}

@misc{leanmathlib2026riemann,
  author = {{Lean FRO and Mathlib Initiative}},
  title  = {{Lean} verification of the {Riemann} Hypothesis},
  year   = {2026},
  month  = nov,
  note   = {Joint statement, November 21, 2026},
  url    = {https://leanprover-community.github.io/blog/posts/riemann-hypothesis/}
}

@book{Fisher1935,
  author    = {Fisher, R. A.},
  title     = {The Design of Experiments},
  publisher = {Oliver and Boyd},
  address   = {Edinburgh},
  year      = {1935}
}

@inproceedings{wu2025pessimistic,
  title={Pessimistic Data Integration for Policy Evaluation},
  author={Wu, Xiangkun and Li, Ting and Aminian, Gholamali
          and Behnamnia, Armin and Rabiee, Hamid R.
          and Shi, Chengchun},
  booktitle={Advances in Neural Information Processing Systems},
  volume={38},
  pages={118854--118903},
  year={2025}
}

@inproceedings{liu2020provably,
  title={Provably Good Batch Off-Policy Reinforcement Learning Without Great Exploration},
  author={Liu, Yao and Swaminathan, Adith and Agarwal, Alekh and Brunskill, Emma},
  booktitle={Advances in Neural Information Processing Systems},
  volume={33},
  year={2020}
}

@inproceedings{yang2023pessimistic,
  title={Pessimistic model selection for offline deep reinforcement learning},
  author={Yang, Chao-Han Huck and Qi, Zhengling and Cui, Yifan and Chen, Pin-Yu},
  booktitle={Uncertainty in Artificial Intelligence},
  pages={2379--2389},
  year={2023},
  organization={PMLR}
}

@inproceedings{swaminathan2015self,
  title={The Self-Normalized Estimator for Counterfactual Learning},
  author={Swaminathan, Adith and Joachims, Thorsten},
  booktitle={Advances in Neural Information Processing Systems},
  volume={28},
  year={2015}
}

@article{kingma2014adam,
  title={Adam: A method for stochastic optimization},
  author={Kingma, Diederik P and Ba, Jimmy},
  journal={arXiv preprint arXiv:1412.6980},
  year={2014}
}

@article{lai1985asymptotically,
  title={Asymptotically efficient adaptive allocation rules},
  author={Lai, Tze Leung and Robbins, Herbert},
  journal={Advances in applied mathematics},
  volume={6},
  number={1},
  pages={4--22},
  year={1985},
  publisher={Academic Press, Inc. Orlando, FL, USA}
}

@article{firth1993bias,
  title={Bias reduction of maximum likelihood estimates},
  author={Firth, David},
  journal={Biometrika},
  pages={27--38},
  year={1993},
  publisher={JSTOR}
}

@article{liu2025fin,
  title={{Fin-R1}: A large language model for financial reasoning through reinforcement learning},
  author={Liu, Zhaowei and Guo, Xin and Yang, Zhi and Lou, Fangqi and Zeng, Lingfeng and Niu, Jinyi and Li, Mengping and Qi, Qi and Liu, Zhiqiang and Han, Yiyang and others},
  journal={arXiv preprint arXiv:2503.16252},
  year={2025}
}

@article{hoeffding1948,
  title = {A Class of Statistics with Asymptotically Normal Distribution},
  author = {Hoeffding, Wassily},
  journal = {The Annals of Mathematical Statistics},
  volume = {19},
  number = {3},
  pages = {293--325},
  year = {1948},
  month = {September},
  publisher = {Institute of Mathematical Statistics},
  doi = {10.1214/aoms/1177730196},
}

@article{greensmith2004variance,
  title = {Variance Reduction Techniques for Gradient Estimates in Reinforcement Learning},
  author = {Greensmith, Evan and Bartlett, Peter L. and Baxter, Jonathan},
  journal = {Journal of Machine Learning Research},
  volume = {5},
  number = {Nov},
  pages = {1471--1530},
  year = {2004},
}

@article{ustat2026demystifying,
  title={Demystifying group relative policy optimization: Its policy gradient is a {U}-statistic},
  author={Zhou, Hongyi and Ye, Kai and Xu, Erhan and Zhu, Jin and Yang, Ying and Gong, Shijin and Shi, Chengchun},
  journal={arXiv preprint arXiv:2603.01162},
  year={2026}
}

@article{zheng2025group,
  title={Group sequence policy optimization},
  author={Zheng, Chujie and Liu, Shixuan and Li, Mingze and Chen, Xiong-Hui and Yu, Bowen and Gao, Chang and Dang, Kai and Liu, Yuqiong and Men, Rui and Yang, An and others},
  journal={arXiv preprint arXiv:2507.18071},
  year={2025}
}

@article{guo2025deepseek,
  title = {{DeepSeek-R1} Incentivizes Reasoning in {LLMs} through Reinforcement Learning},
  author = {Guo, Daya and Yang, Dejian and Zhang, Haowei and Song, Junxiao and Wang, Peiyi and Zhu, Qihao and Xu, Runxin and Zhang, Ruoyu and Ma, Shirong and Bi, Xiao and others},
  journal = {Nature},
  volume = {645},
  number = {8081},
  pages = {633--638},
  year = {2025},
  publisher = {Nature Portfolio},
  doi = {10.1038/s41586-025-09422-z},
}

@article{liu2025understanding,
  title={Understanding {R1-Zero}-like training: A critical perspective},
  author={Liu, Zichen and Chen, Changyu and Li, Wenjun and Qi, Penghui and Pang, Tianyu and Du, Chao and Lee, Wee Sun and Lin, Min},
  journal={arXiv preprint arXiv:2503.20783},
  year={2025}
}

@inproceedings{ouyang2022training,
  author = {Ouyang, Long and Wu, Jeffrey and Jiang, Xu and Almeida, Diogo and Wainwright, Carroll and Mishkin, Pamela and Zhang, Chong and Agarwal, Sandhini and Slama, Katarina and Ray, Alex and Schulman, John and Hilton, Jacob and Kelton, Fraser and Miller, Luke and Simens, Maddie and Askell, Amanda and Welinder, Peter and Christiano, Paul F. and Leike, Jan and Lowe, Ryan},
  booktitle = {Advances in Neural Information Processing Systems},
  editor = {Koyejo, S. and Mohamed, S. and Agarwal, A. and Belgrave, D. and Cho, K. and Oh, A.},
  pages = {27730--27744},
  publisher = {Curran Associates, Inc.},
  title = {Training Language Models to Follow Instructions with Human Feedback},
  volume = {35},
  year = {2022}
}

@article{schulman2017proximal,
  title={Proximal policy optimization algorithms},
  author={Schulman, John and Wolski, Filip and Dhariwal, Prafulla and Radford, Alec and Klimov, Oleg},
  journal={arXiv preprint arXiv:1707.06347},
  year={2017}
}

@article{shao2024deepseekmath,
  title={{DeepSeekMath}: Pushing the limits of mathematical reasoning in open language models},
  author={Shao, Zhihong and Wang, Peiyi and Zhu, Qihao and Xu, Runxin and Song, Junxiao and Bi, Xiao and Zhang, Haowei and Zhang, Mingchuan and Li, YK and Wu, Yang and others},
  journal={arXiv preprint arXiv:2402.03300},
  year={2024}
}

@inproceedings{wei2022chain,
  title = {Chain-of-Thought Prompting Elicits Reasoning in Large Language Models},
  author = {Jason Wei and Xuezhi Wang and Dale Schuurmans and Maarten Bosma and Brian Ichter and Fei Xia and Ed Chi and Quoc Le and Denny Zhou},
  booktitle = {Advances in Neural Information Processing Systems},
  volume = {35},
  year = {2022},
  eprint = {2201.11903},
  archivePrefix = {arXiv},
  primaryClass = {cs.CL},
}

@article{williams1992simple,
  author = {Williams, Ronald J.},
  title = {Simple Statistical Gradient-Following Algorithms for Connectionist Reinforcement Learning},
  journal = {Machine Learning},
  year = {1992},
  month = {May},
  volume = {8},
  number = {3},
  pages = {229--256},
  issn = {1573-0565},
  doi = {10.1007/BF00992696},
}

@inproceedings{yu2025dapo,
  title={{DAPO}: An open-source {LLM} reinforcement learning system at scale},
  author={Yu, Qiying and Zhang, Zheng and Zhu, Ruofei and Yuan, Yufeng and Zuo, Xiaochen and Yue, Yu and Dai, Weinan and Fan, Tiantian and Liu, Gaohong and Liu, Juncai and others},
  booktitle={Advances in Neural Information Processing Systems},
  volume={38},
  pages={113222--113244},
  year={2025}
}

@inproceedings{sutton1999policy,
  title={Policy Gradient Methods for Reinforcement Learning
         with Function Approximation},
  author={Sutton, Richard S. and McAllester, David A.
          and Singh, Satinder P. and Mansour, Yishay},
  booktitle={Advances in Neural Information Processing Systems},
  volume={12},
  pages={1057--1063},
  year={1999}
}

@inproceedings{feng2025group,
  title = {Group-in-Group Policy Optimization for {LLM} Agent Training},
  author = {Feng, Lang and Xue, Zhenghai and Liu, Tingcong and An, Bo},
  booktitle = {Advances in Neural Information Processing Systems},
  volume = {38},
  year = {2025}
}

@article{chu2025gpg,
  title={{GPG}: A simple and strong reinforcement learning baseline for model reasoning},
  author={Chu, Xiangxiang and Huang, Hailang and Zhang, Xiao and Wei, Fei and Wang, Yong},
  journal={arXiv preprint arXiv:2504.02546},
  year={2025}
}

@article{gong2026kernelized,
  title={Kernelized advantage estimation: From nonparametric statistics to {LLM} reasoning},
  author={Gong, Shijin and Ye, Kai and Zhu, Jin and Zhang, Xinyu and Zhou, Hongyi and Shi, Chengchun},
  journal={arXiv preprint arXiv:2604.28005},
  year={2026}
}

@article{swaminathan2015batch,
  title = {Batch Learning from Logged Bandit Feedback through Counterfactual Risk Minimization},
  author = {Swaminathan, Adith and Joachims, Thorsten},
  journal = {Journal of Machine Learning Research},
  volume = {16},
  number = {52},
  pages = {1731--1755},
  year = {2015},
}

@inproceedings{jin2021pessimism,
  title = {Is Pessimism Provably Efficient for Offline {RL}?},
  author = {Jin, Ying and Yang, Zhuoran and Wang, Zhaoran},
  booktitle = {Proceedings of the 38th International Conference on Machine Learning},
  series = {Proceedings of Machine Learning Research},
  volume = {139},
  pages = {5084--5096},
  publisher = {PMLR},
  year = {2021},
}

@inproceedings{rashidinejad2021bridging,
  title = {Bridging Offline Reinforcement Learning and Imitation Learning: A Tale of Pessimism},
  author = {Rashidinejad, Paria and Zhu, Banghua and Ma, Cong and Jiao, Jiantao and Russell, Stuart},
  booktitle = {Advances in Neural Information Processing Systems},
  volume = {34},
  pages = {11702--11716},
  year = {2021},
}

@article{castelvecchi2025deepmind,
  title   = {{DeepMind} and {OpenAI} models solve maths problems at level of top students},
  author  = {Castelvecchi, Davide},
  journal = {Nature},
  volume  = {644},
  pages   = {20},
  year    = {2025},
  doi     = {10.1038/d41586-025-02343-x}
}

@book{sutton2018reinforcement,
  title = {Reinforcement Learning: An Introduction},
  author = {Sutton, Richard S. and Barto, Andrew G.},
  edition = {2nd},
  publisher = {MIT Press},
  address = {Cambridge, MA},
  year = {2018}
}

@inproceedings{zeng2025acecoder,
  title = {{AceCoder}: Acing Coder {RL} via Automated Test-Case Synthesis},
  author = {Zeng, Huaye and Jiang, Dongfu and Wang, Haozhe and Nie, Ping and Chen, Xiaotong and Chen, Wenhu},
  booktitle = {Proceedings of the 63rd Annual Meeting of the Association for Computational Linguistics (Volume 1: Long Papers)},
  pages = {12023--12040},
  year = {2025},
  publisher = {Association for Computational Linguistics},
  doi = {10.18653/v1/2025.acl-long.587},
}

@article{guo2024deepseekcoder,
  title={{DeepSeek-Coder}: when the large language model meets programming--the rise of code intelligence},
  author={Guo, Daya and Zhu, Qihao and Yang, Dejian and Xie, Zhenda and Dong, Kai and Zhang, Wentao and Chen, Guanting and Bi, Xiao and Wu, Yifan and Li, YK and others},
  journal={arXiv preprint arXiv:2401.14196},
  year={2024}
}

@inproceedings{liu2023evalplus,
  title = {Is Your Code Generated by {ChatGPT} Really Correct? Rigorous Evaluation of Large Language Models for Code Generation},
  author = {Liu, Jiawei and Xia, Chunqiu Steven and Wang, Yuyao and Zhang, Lingming},
  booktitle = {Advances in Neural Information Processing Systems},
  volume = {36},
  year = {2023},
}

@inproceedings{stojanovski2025reasoninggym,
  title={{Reasoning Gym}: Reasoning environments for reinforcement learning with verifiable rewards},
  author={Stojanovski, Zafir and Stanley, Oliver and Sharratt, Joe and Jones, Richard and Adefioye, Abdulhakeem and Kaddour, Jean and K{\"o}pf, Andreas},
  booktitle={Advances in Neural Information Processing Systems},
  volume={38},
  year={2025}
}

@article{cobbe2021training,
  title={Training verifiers to solve math word problems},
  author={Cobbe, Karl and Kosaraju, Vineet and Bavarian, Mohammad and Chen, Mark and Jun, Heewoo and Kaiser, Lukasz and Plappert, Matthias and Tworek, Jerry and Hilton, Jacob and Nakano, Reiichiro and others},
  journal={arXiv preprint arXiv:2110.14168},
  year={2021}
}

@inproceedings{lightman2024verify,
  title = {Let's Verify Step by Step},
  author = {Lightman, Hunter and Kosaraju, Vineet and Burda, Yura and Edwards, Harri and Baker, Bowen and Lee, Teddy and Leike, Jan and Schulman, John and Sutskever, Ilya and Cobbe, Karl},
  booktitle = {International Conference on Learning Representations},
  year = {2024},
}

@inproceedings{lewkowycz2022solving,
  title = {Solving Quantitative Reasoning Problems with Language Models},
  author = {Lewkowycz, Aitor and Andreassen, Anders and Dohan, David and Dyer, Ethan and Michalewski, Henryk and Ramasesh, Vinay and Slone, Ambrose and Anil, Cem and Schlag, Imanol and Gutman-Solo, Theo and Wu, Yuhuai and Neyshabur, Behnam and Gur-Ari, Guy and Misra, Vedant},
  booktitle = {Advances in Neural Information Processing Systems},
  volume = {35},
  year = {2022},
}

@inproceedings{he2024olympiadbench,
  title = {{OlympiadBench}: A Challenging Benchmark for Promoting {AGI} with Olympiad-Level Bilingual Multimodal Scientific Problems},
  author = {He, Chaoqun and Luo, Renjie and Bai, Yuzhuo and Hu, Shengding and Thai, Zhen and Shen, Junhao and Hu, Jinyi and Han, Xu and Huang, Yujie and Zhang, Yuxiang and Liu, Jie and Qi, Lei and Liu, Zhiyuan and Sun, Maosong},
  booktitle = {Proceedings of the 62nd Annual Meeting of the Association for Computational Linguistics (Volume 1: Long Papers)},
  year = {2024},
  publisher = {Association for Computational Linguistics},
  pages = {3828--3850},
  doi = {10.18653/v1/2024.acl-long.211},
}

@article{rosenthal1970subspaces,
  author = {Rosenthal, Haskell P.},
  title = {On the subspaces of {$L^p$} ({$p>2$}) spanned by sequences of independent random variables},
  journal = {Israel Journal of Mathematics},
  volume = {8},
  pages = {273--303},
  year = {1970},
  doi = {10.1007/BF02771562},
}

@book{dubhashi2009concentration,
  author = {Dubhashi, Devdatt P. and Panconesi, Alessandro},
  title = {Concentration of Measure for the Analysis of Randomized Algorithms},
  publisher = {Cambridge University Press},
  year = {2009},
}

@inproceedings{balunovic2025matharena,
  title = {{MathArena}: Evaluating {LLMs} on Uncontaminated Math Competitions},
  author = {Balunovi{\'c}, Mislav and Dekoninck, Jasper and Petrov, Ivo and Jovanovi{\'c}, Nikola and Vechev, Martin},
  booktitle = {Advances in Neural Information Processing Systems},
  volume = {38},
  year = {2025},
}

@article{chernozhukov2013intersection,
  author = {Chernozhukov, Victor and Lee, Sokbae and Rosen, Adam M.},
  title = {Intersection Bounds: Estimation and Inference},
  journal = {Econometrica},
  year = {2013},
  volume = {81},
  number = {2},
  pages = {667--737},
  doi = {10.3982/ECTA8718}
}

@inproceedings{gong2026basis,
  title={{BASIS}: Batchwise Advantage Estimation from Single-Rollout Information Sharing for {LLM} Reasoning},
  author={Gong, Shijin and Xu, Erhan and Ye, Kai and Livieri, Giulia and Quinzan, Francesco and Shi, Chengchun},
  booktitle={Findings of the Association for Computational Linguistics: EMNLP 2026},
  pages={6208--6226},
  year={2026}
}

@article{sun2026curverl,
  title={{CurveRL}: Principled Distribution-Aware Context Reweighting for {LLM} Reasoning},
  author={Sun, Ke and Zhao, Yizhou and Xin, Jiayi and Long, Qi and Su, Weijie},
  journal={arXiv preprint arXiv:2605.24331},
  year={2026}
}

@article{hu2025reinforce++,
  title={{REINFORCE++}: Stabilizing critic-free policy optimization with global advantage normalization},
  author={Hu, Jian and Liu, Jason Klein and Xu, Haotian and Shen, Wei},
  journal={arXiv preprint arXiv:2501.03262},
  year={2025}
}

\newpage

\clearpage
\hypersetup{plainpages=false}

\begin{appendices}
\numberwithin{equation}{section}
\renewcommand{\theHequation}{appendix.\Alph{section}.\arabic{equation}}

\begin{center}
{\LARGE\bfseries Appendix for U-Statistics-Powered LLM Reasoning\par}
\end{center}

\etocdepthtag.toc{appendix}
\begingroup
\etocsettagdepth{main}{none}
\etocsettagdepth{appendix}{subsubsection}
\etocsettocstyle{\section*{Contents}}{}
\normalsize
\makeatletter
\def\@pnumwidth{2.5em}
\def\@tocrmarg{3.5em}
\def\@dotsep{4.5}
\renewcommand*\l@section[2]{\addvspace{5pt}{\bfseries\@dottedtocline{1}{0em}{2em}{#1}{#2}}}
\renewcommand*\l@subsection{\@dottedtocline{2}{1.6em}{2.9em}}
\renewcommand*\l@subsubsection{\@dottedtocline{3}{4.5em}{3.5em}}
\makeatother
\tableofcontents
\endgroup
\clearpage

\section{Proofs of the main theorems}
\label{sec:theory-proofs}

\theoremstyle{thmstyleone}
\newtheorem{appendixtheorem}[theorem]{Theorem}
\newtheorem{appendixproposition}[theorem]{Proposition}
\newtheorem{appendixlemma}{Lemma}
\renewcommand{\theappendixlemma}{A.\arabic{appendixlemma}}
\newtheorem*{restatedtheoremone}{Theorem~\ref{thm:varianceest}}
\newtheorem*{restatedtheoremtwo}{Theorem~\ref{thm:ratioest}}
\newtheorem*{restatedtheoremthree}{Theorem~\ref{thm:pessi}}

Sections~\ref{app:gradient-mse-proof}--\ref{app:pessimistic-risk-proof} present the detailed statements and proofs of Lemma~\ref{prop:score-closed-forms} and Theorems~\ref{thm:varianceest}--\ref{thm:pessi}. Sections~\ref{app:covariance-pessimistic-risk} and~\ref{app:pilot-confidence} further extend the theoretical analysis of the pessimistic estimator in \eqref{eq:main-floored-pess} to grouped jackknife standard errors with a floor $\epsilon_0$, integer group sizes, and quantile calibration of the penalty multiplier $k$.

Throughout this section, $c,C_0,K,K_0,K_1$ denote positive constants independent of ${b}$, ${g}$ and ${{n}}$ and may vary from line to line. Constants in conditional bounds are also independent of the conditioning variables.

\subsection{Proof of Lemma~\ref{prop:score-closed-forms}}
\label{app:gradient-mse-proof}

\begin{proof}[Proof of Lemma~\ref{prop:score-closed-forms}.]
For a fixed parameter $\theta$ and prompt $x$, let
\[
Y_1,\ldots,Y_{{G}}\mid X=x\overset{\mathrm{i.i.d.}}{\sim}\pi_\theta(\cdot\mid x),
\qquad Z_j=r(x,Y_j),\quad j=1,\ldots,{{G}},
\]
and define the score function as
$S_\theta(x,Y)=\nabla_\theta\log\pi_\theta(Y\mid x)$.
Define the conditional variance components by
\begin{align}
c_2(x)&=\E\Bigl[\bigl\|S_\theta(x,Y)\{Z-V^{\pi_\theta}(x)\}-{\bm g}(x;\theta)\bigr\|^2\mid X=x\Bigr],\notag\\
c_3(x)&=\E[\|S_\theta(x,Y)\|^2\mid X=x]\Var(Z\mid X=x)+\|{\bm g}(x;\theta)\|^2.
\label{eq:varianceest-conditional-c3}
\end{align}
Then $\E[c_j(X)]=c_j$ for $j=2,3$.

We first express ${\widehat{\bm g}}_{{G}}(x;\theta)$ as a second-order U-statistic. By definition, we have
\begin{equation*}
{\widehat{\bm g}}_{{G}}(x;\theta)=\frac1{{G}}\sum_{j=1}^{{G}} S_\theta(x,Y_j)(Z_j-\bar Z_{-j})
=\frac1{{{G}}({{G}}-1)}\sum_{j\ne k}S_\theta(x,Y_j)(Z_j-Z_k).
\end{equation*}
Applying a standard symmetrization argument, it follows that
\begin{align}
{\widehat{\bm g}}_{{G}}(x;\theta)
&=\frac1{2{{G}}({{G}}-1)}\sum_{j\ne k}S_\theta(x,Y_j)(Z_j-Z_k)
+\frac1{2{{G}}({{G}}-1)}\sum_{j\ne k}S_\theta(x,Y_k)(Z_k-Z_j)\notag\\
&=\frac1{2{{G}}({{G}}-1)}\sum_{j\ne k}
\Bigl[S_\theta(x,Y_j)Z_j-S_\theta(x,Y_j)Z_k-S_\theta(x,Y_k)Z_j+S_\theta(x,Y_k)Z_k\Bigr]\notag\\
&=\frac1{2{{G}}({{G}}-1)}\sum_{j\ne k}\bigl\{S_\theta(x,Y_j)-S_\theta(x,Y_k)\bigr\}(Z_j-Z_k)\notag\\
&=\binom {{G}}2^{-1}\sum_{1\le j<k\le {{G}}}h\big((Y_j,Z_j),(Y_k,Z_k);x\big),
\label{eq:lemma-gradient-ustat}
\end{align}
where the symmetric kernel
\[
h\big((y,z),(y',z');x\big)=\frac12\bigl\{S_\theta(x,y)-S_\theta(x,y')\bigr\}(z-z').
\]

Recall that, for i.i.d. random elements $T_1,\ldots,T_{m}$ and a symmetric kernel $h$ with $\E\|h(T_1,T_2)\|^2<\infty$, the second-order U-statistic $U_{m}=\binom {m}2^{-1}\sum_{j<k}h(T_j,T_k)$ satisfies the following Hoeffding decomposition and MSE identity \citep{hoeffding1948,ustat2026demystifying}:
\begin{equation}
U_{m}-h_0
=\frac1{m}\sum_{j=1}^{m}\zeta_1(T_j)
+\binom {m}2^{-1}\sum_{j<k}\zeta_2(T_j,T_k),
\label{eq:lemma-ustat-decomposition}
\end{equation}
\begin{equation}
\E\|U_{m}-h_0\|^2
=\frac1{m}\E\|\zeta_1(T_1)\|^2
+\frac2{{m}({m}-1)}\E\|\zeta_2(T_1,T_2)\|^2,
\label{eq:lemma-ustat-mse}
\end{equation}
where
\begin{align*}
h_0&=\E[h(T_1,T_2)],\qquad h_1(t)=\E[h(t,T_2)],\\
\zeta_1(t)&=2\{h_1(t)-h_0\},\qquad \zeta_2(t,t')=h(t,t')-h_1(t)-h_1(t')+h_0.
\end{align*}

For the U-statistic ${\widehat{\bm g}}_{{G}}(x;\theta)$ in \eqref{eq:lemma-gradient-ustat}, since $\E[S_\theta(x,Y)\mid X=x]=0$ and $\E[S_\theta(x,Y)Z\mid X=x]={\bm g}(x;\theta)$, we have
\begin{align*}
h_0(x)&=\E\Bigl[h\big((Y_1,Z_1),(Y_2,Z_2);x\big)\mid X=x\Bigr]\\
&=\E[S_\theta(x,Y)Z\mid X=x]-\E[S_\theta(x,Y)\mid X=x]\E[Z\mid X=x]={\bm g}(x;\theta),\\
h_1(y,z;x)&=\E\Bigl[h\big((y,z),(Y,Z);x\big)\mid X=x\Bigr]\\
&=\frac12\big[S_\theta(x,y)\{z-V^{\pi_\theta}(x)\}-z\E[S_\theta(x,Y)\mid X=x]+\E[S_\theta(x,Y)Z\mid X=x]\big]\\
&=\frac12\big[S_\theta(x,y)\{z-V^{\pi_\theta}(x)\}+{\bm g}(x;\theta)\big].
\end{align*}
Applying \eqref{eq:lemma-ustat-mse} with ${m}={{G}}$ under this conditional distribution, we obtain
\begin{equation}
\E\big[\|{\widehat{\bm g}}_{{G}}(x;\theta)-{\bm g}(x;\theta)\|^2\mid X=x\big]
=\frac1{{G}}\E[\|\zeta_{1,1}(x)\|^2\mid X=x]
+\frac2{{{G}}({{G}}-1)}\E[\|\zeta_{2,12}(x)\|^2\mid X=x],
\label{eq:lemma-gradient-conditional-mse}
\end{equation}
where
\begin{align}
\zeta_{1,j}(x)
&=2\{h_1(Y_j,Z_j;x)-h_0(x)\}
=S_\theta(x,Y_j)\{Z_j-V^{\pi_\theta}(x)\}-{\bm g}(x;\theta),\notag\\
\zeta_{2,jk}(x)
&=h\big((Y_j,Z_j),(Y_k,Z_k);x\big)-h_1(Y_j,Z_j;x)-h_1(Y_k,Z_k;x)+h_0(x)\notag\\
&=-\frac12\big[S_\theta(x,Y_j)\{Z_k-V^{\pi_\theta}(x)\}
+S_\theta(x,Y_k)\{Z_j-V^{\pi_\theta}(x)\}\big].
\label{eq:lemma-gradient-projections}
\end{align}
By \eqref{eq:varianceest-conditional-c3} and \eqref{eq:lemma-gradient-projections}, we have
\begin{align}
\E[\|\zeta_{1,1}(x)\|^2\mid X=x]
&=c_2(x),
\label{eq:lemma-gradient-first-moment}\\
2\E[\|\zeta_{2,12}(x)\|^2\mid X=x]
&=\frac12\E\big[\|S_\theta(x,Y_1)\|^2\{Z_2-V^{\pi_\theta}(x)\}^2\mid X=x\big]\notag\\
&\quad+\frac12\E\big[\|S_\theta(x,Y_2)\|^2\{Z_1-V^{\pi_\theta}(x)\}^2\mid X=x\big]\notag\\
&\quad+\E\big[\{S_\theta(x,Y_1)\}^\top S_\theta(x,Y_2)
\cdot\{Z_1-V^{\pi_\theta}(x)\}\{Z_2-V^{\pi_\theta}(x)\}\mid X=x\big]\notag\\
&=\E[\|S_\theta(x,Y)\|^2\mid X=x]\Var(Z\mid X=x)+\|{\bm g}(x;\theta)\|^2=c_3(x),
\label{eq:lemma-gradient-second-moment}
\end{align}
where the second equality in \eqref{eq:lemma-gradient-second-moment} follows from $(Y_1,Z_1)\perp(Y_2,Z_2)\mid X=x$ and $\E[S_\theta(x,Y)\{Z-V^{\pi_\theta}(x)\}\mid X=x]={\bm g}(x;\theta)$. This implies, by \eqref{eq:lemma-gradient-conditional-mse}, \eqref{eq:lemma-gradient-first-moment} and \eqref{eq:lemma-gradient-second-moment},
\begin{equation}
\E[\|{\widehat{\bm g}}_{{G}}(x;\theta)-{\bm g}(x;\theta)\|^2\mid X=x]
=\frac{c_2(x)}{{G}}+\frac{c_3(x)}{{{G}}({{G}}-1)}.
\label{eq:lemma-gradient-conditional-closed-form}
\end{equation}

Next, we derive the MSE of ${\widehat{\bm g}}_{{{B}},{{G}}}(\theta)$. By definition, we have 
\begin{align*}
\operatorname{MSE}\{{\widehat{\bm g}}_{{{B}},{{G}}}(\theta)\}
&=\E\|{\widehat{\bm g}}_{{{B}},{{G}}}(\theta)-{\bm g}(\theta)\|^2\\
&=\E\Bigl\|\frac1{{B}}\sum_{i=1}^{{B}}[{\widehat{\bm g}}_{{G}}(X_i;\theta)-{\bm g}(X_i;\theta)]
+\frac1{{B}}\sum_{i=1}^{{B}}[{\bm g}(X_i;\theta)-{\bm g}(\theta)]\Bigr\|^2\\
&=\E\Bigl\|\frac1{{B}}\sum_{i=1}^{{B}}[{\widehat{\bm g}}_{{G}}(X_i;\theta)-{\bm g}(X_i;\theta)]\Bigr\|^2\\
&\quad+\E\Bigl\|\frac1{{B}}\sum_{i=1}^{{B}}[{\bm g}(X_i;\theta)-{\bm g}(\theta)]\Bigr\|^2\\
&=\frac1{{B}}\E\|{\widehat{\bm g}}_{{G}}(X;\theta)-{\bm g}(X;\theta)\|^2
+\frac1{{B}}\E\|{\bm g}(X;\theta)-{\bm g}(\theta)\|^2.
\end{align*}
where the third equality follows from $\E[{\widehat{\bm g}}_{{G}}(X_i;\theta)\mid X_i]={\bm g}(X_i;\theta)$, and the last equality follows from the independence of the sampled groups.
By \eqref{eq:lemma-gradient-conditional-closed-form} and $\E[c_j(X)]=c_j$ for $j=2,3$, we obtain
\[
\E\|{\widehat{\bm g}}_{{{B}},{{G}}}(\theta)-{\bm g}(\theta)\|^2
=\frac{c_1}{{B}}+\frac{c_2}{{{B}}{{G}}}+\frac{c_3}{{{B}}{{G}}({{G}}-1)},
\]
which proves \eqref{eq:scalinglaw}.

Under a fixed sampling budget ${{N}}={{B}}{{G}}$, the RHS of \eqref{eq:scalinglaw} becomes
\begin{equation*}
\frac{c_1{{G}}}{{{N}}}+\frac{c_2}{{{N}}}+\frac{c_3}{{{N}}({{G}}-1)}
=\frac{c_1+c_2}{{{N}}}+\frac1{{N}}\Bigl\{c_1({{G}}-1)+\frac{c_3}{{{G}}-1}\Bigr\}
\ge\frac{c_1+c_2+2\sqrt{c_1c_3}}{{{N}}},
\end{equation*}
where the inequality follows from the Cauchy--Schwarz inequality, and equality holds if and only if ${{G}}=1+\sqrt{c_3/c_1}$. This proves \eqref{eq:gstar}. Thus, we complete the proof of Lemma~\ref{prop:score-closed-forms}.
\end{proof}

\medskip
\subsection{Proof of Theorem~\ref{thm:varianceest}}
\label{app:varianceest-proof}

{For convenience, we first restate Theorem~\ref{thm:varianceest} below.}

\begin{restatedtheoremone}
Suppose Assumptions \ref{assump:upper} and \ref{assump:lower} hold. Then
\begin{enumerate}
    \item[(a)] The biases of the biased variance estimators $\widehat{c}_1$ and $\widehat{c}_3$ are given by \eqref{eqn:c1bias} and \eqref{eqn:c3bias}, respectively. To the contrary, the debiased estimator $\widetilde{c}_1$ and $\widetilde{c}_3$ given by \eqref{eqn:debias} are unbiased.
    \item[(b)] $\widehat{c}_1$ and $\widehat{c}_3$ are consistent when both ${b}$ and ${g}$ approach infinity. To the contrary, consistencies of $\widetilde{c}_1$ and $\widetilde{c}_3$ require only ${b}$ to approach infinity.
    \item[(c)] The biased estimator $\widehat c_1$ is non-negative, almost surely. Additionally, there exist some constants $c$, $\bar{c}>0$ such that
    \begin{equation*}
        \mathbb{P}\big(\widehat{c}_1<\frac{\bar{c}}{{g}}\big)=O\big(\exp(-c{b})\big).
    \end{equation*}
    By contrast, for the debiased estimator $\widetilde{c}_1$, there exists some constant $\nu>0$ such that
    \begin{equation*}
        \mathbb{P}(\widetilde{c}_1\leq 0)\geq \Phi\big(-\frac{c_1{g}\sqrt{{b}}}{\nu}\big)-O\big(\frac{1}{\sqrt[3]{{b}}}+\frac{1}{{g}}\big),
    \end{equation*}
    where $\Phi$ denotes the cumulative distribution function of a standard normal random variable.
\end{enumerate}
\end{restatedtheoremone}

{To prove Theorem~\ref{thm:varianceest}, we first present three general probability inequalities (Lemmas~\ref{lem:conditional-vector-rosenthal}--\ref{lem:conditional-paley-zygmund}), and then present the proofs of parts~(a),~(b) and~(c) in Sections~\ref{app:varianceest-a},~\ref{app:varianceest-b} and~\ref{app:varianceest-c}, respectively, where Section~\ref{app:varianceest-c} also establishes two auxiliary lemmas on conditional moment bounds and the conditional decomposition of $\widetilde c_1$ (Lemmas~\ref{lem:within-prompt-moment-bounds} and~\ref{lem:conditional-decomposition-variance}).}

\begin{appendixlemma}[Conditional Rosenthal inequality for random vectors]
\label{lem:conditional-vector-rosenthal}
Let $T_1,\ldots,T_{g}\in\mathbb R^d$ be conditionally independent given $X$, with $\E(T_j\mid X)=0$ and $\|T_j\|\le L$ almost surely for $j=1,\ldots,{g}$, where $d\le d_0$. Then for any $p\ge2$, there exists a constant $a_p>0$, depending only on $p$ and $d_0$, such that almost surely,
\begin{align}
\E\Bigl[\bigl\|\frac1{g}\sum_{j=1}^{g} T_j\bigr\|^p\,\Bigm|\,X\Bigr]
&\le\frac{a_p}{{g}^p}\Bigl[\sum_{j=1}^{g}\E(\|T_j\|^p\mid X)
+\Bigl\{\sum_{j=1}^{g}\E(\|T_j\|^2\mid X)\Bigr\}^{p/2}\Bigr]\notag\\
&\le K_p{g}^{-p/2},
\label{eq:conditional-independent-mean-moments}
\end{align}
where $K_p=2a_pL^p$.
\end{appendixlemma}

\begin{proof}[Proof of Lemma~\ref{lem:conditional-vector-rosenthal}.]
Let $T_j=(T_{j1},\ldots,T_{jd})^\top$ for $j=1,\ldots,{g}$. Conditional on $X$, for each coordinate $\ell=1,\ldots,d$, the random variables $T_{1\ell},\ldots,T_{{g}\ell}$ are independent and have mean zero. By Rosenthal's inequality \citep{rosenthal1970subspaces}, for any $p\ge2$, there exists a constant $C_p>0$ depending only on $p$ such that, almost surely,
\begin{equation}
\E\Bigl[\bigl|\sum_{j=1}^{g} T_{j\ell}\bigr|^p\,\Bigm|\,X\Bigr]
\le C_p\Bigl[\sum_{j=1}^{g}\E(|T_{j\ell}|^p\mid X)
+\Bigl\{\sum_{j=1}^{g}\E(T_{j\ell}^2\mid X)\Bigr\}^{p/2}\Bigr].
\label{eq:rosenthal-coordinate}
\end{equation}

By H\"older's inequality, we have $\|v\|^p\le d^{p/2-1}\sum_{\ell=1}^d|v_\ell|^p$ for any $v=(v_1,\ldots,v_d)^\top\in\mathbb R^d$ and $p\ge2$. Combining this inequality with \eqref{eq:rosenthal-coordinate}, we obtain
\begin{align*}
\E\Bigl[\bigl\|\sum_{j=1}^{g} T_j\bigr\|^p\,\Bigm|\,X\Bigr]
&\le d^{p/2-1}\sum_{\ell=1}^d\E\Bigl[\bigl|\sum_{j=1}^{g} T_{j\ell}\bigr|^p\,\Bigm|\,X\Bigr]\\
&\le C_pd^{p/2-1}\Bigl[\sum_{j=1}^{g}\sum_{\ell=1}^d\E(|T_{j\ell}|^p\mid X)
+\sum_{\ell=1}^d\Bigl\{\sum_{j=1}^{g}\E(T_{j\ell}^2\mid X)\Bigr\}^{p/2}\Bigr]\\
&\le a_p\Bigl[\sum_{j=1}^{g}\E(\|T_j\|^p\mid X)
+\Bigl\{\sum_{j=1}^{g}\E(\|T_j\|^2\mid X)\Bigr\}^{p/2}\Bigr],
\end{align*}
where $a_p=C_pd_0^{p/2-1}$, and the last inequality follows from $d\le d_0$, $\sum_{\ell=1}^d|v_\ell|^p\le\|v\|^p$ and $\sum_{\ell=1}^d b_\ell^{p/2}\le(\sum_{\ell=1}^d b_\ell)^{p/2}$ for $b_\ell\ge0$. Dividing both sides by ${g}^p$ yields the first inequality in \eqref{eq:conditional-independent-mean-moments}. Finally, since $\|T_j\|\le L$ almost surely, we have
\[
\E\Bigl[\bigl\|\frac1{g}\sum_{j=1}^{g} T_j\bigr\|^p\,\Bigm|\,X\Bigr]
\le a_pL^p\bigl({g}^{1-p}+{g}^{-p/2}\bigr)
\le2a_pL^p{g}^{-p/2},
\]
which proves the second inequality in \eqref{eq:conditional-independent-mean-moments}. Thus, we complete the proof of Lemma~\ref{lem:conditional-vector-rosenthal}.
\end{proof}

\medskip
\begin{appendixlemma}[Multiplicative Chernoff bound]
\label{lem:bernoulli-chernoff}
Let $I_1,\ldots,I_m$ be independent Bernoulli random variables with $\Pr(I_i=1)=p_i$ for $i=1,\ldots,m$. Define $S_m=\sum_{i=1}^m I_i$ and $\mu=\E S_m=\sum_{i=1}^m p_i$. Then, for any $0<\delta<1$,
\[
\Pr\{S_m\le(1-\delta)\mu\}\le\exp\Bigl(-\frac{\delta^2\mu}{2}\Bigr).
\]
\end{appendixlemma}

\begin{proof}[Proof of Lemma~\ref{lem:bernoulli-chernoff}.]
The result follows from Theorem~1.1 of \citet{dubhashi2009concentration}.
\end{proof}

\medskip
\begin{appendixlemma}[Conditional Paley--Zygmund inequality]
\label{lem:conditional-paley-zygmund}
Let $T$ be a nonnegative random variable with $\E T^2<\infty$, and let $X$ be a random element. Then, for any $0<\delta<1$, almost surely on $\{\E(T\mid X)>0\}$,
\[
\Pr\{T\ge\delta\E(T\mid X)\mid X\}
\ge(1-\delta)^2\frac{\{\E(T\mid X)\}^2}{\E(T^2\mid X)}.
\]
\end{appendixlemma}

\begin{proof}[Proof of Lemma~\ref{lem:conditional-paley-zygmund}.]
By the conditional Cauchy--Schwarz inequality, we have
\begin{align*}
\E(T\mid X)
&\le\delta\E(T\mid X)+\E[T\mathbf 1\{T\ge\delta\E(T\mid X)\}\mid X]\\
&\le\delta\E(T\mid X)+\{\E(T^2\mid X)\}^{1/2}
\Pr\{T\ge\delta\E(T\mid X)\mid X\}^{1/2}.
\end{align*}
Rearranging terms and squaring both sides yields the desired inequality.
\end{proof}

\subsubsection{Theorem~\ref{thm:varianceest}(a): Bias and bias correction}
\label{app:varianceest-a}

\begin{proof}[Proof of Theorem~\ref{thm:varianceest}(a).]
Define $\psi_j(x)=S_\theta(x,Y_j)(Z_j-\bar Z_{-j})$, and we have
\[
{\widehat{\bm g}}_{g}(x;\theta)=\frac1{g}\sum_{j=1}^{g}\psi_j(x),\qquad
{\widehat{\bm g}}_{{b},{g}}(\theta)=\frac1{b}\sum_{i=1}^{b}{\widehat{\bm g}}_{g}(X_i;\theta).
\]
We first calculate the expectations of $\widehat c_j$ for $j=1,2,3$, respectively.

\medskip
\noindent\textbf{(i) Expectation of $\widehat c_1$.} By definition,
\begin{equation}
\widehat c_1=\frac1{{b}-1}\sum_{i=1}^{b}\|{\widehat{\bm g}}_{g}(X_i;\theta)\|^2-\frac {b}{{b}-1}\|{\widehat{\bm g}}_{{b},{g}}(\theta)\|^2.
\label{eq:varianceest-c1-definition}
\end{equation}
By \eqref{eq:lemma-gradient-conditional-closed-form} and $\E[{\widehat{\bm g}}_{g}(x;\theta)\mid X=x]={\bm g}(x;\theta)$, we have
\begin{align}\label{eq:varianceest-prompt-second-moment-x}
\E[\|{\widehat{\bm g}}_{g}(x;\theta)\|^2\mid X=x]
=\|{\bm g}(x;\theta)\|^2+\frac{c_2(x)}{g}+\frac{c_3(x)}{{g}({g}-1)},
\end{align}
which, together with $\E[c_j(X)]=c_j$ for $j=2,3$, implies
\begin{align}
\E\|{\widehat{\bm g}}_{g}(X;\theta)\|^2
&=\E\|{\bm g}(X;\theta)\|^2+\frac{c_2}{g}+\frac{c_3}{{g}({g}-1)}\notag\\
&=\E\|{\bm g}(X;\theta)-{\bm g}(\theta)\|^2+\|{\bm g}(\theta)\|^2+\frac{c_2}{g}+\frac{c_3}{{g}({g}-1)}\notag\\
&=c_0+c_1+\frac{c_2}{g}+\frac{c_3}{{g}({g}-1)}.
\label{eq:varianceest-prompt-second-moment}
\end{align}

Next, we compute $\E\widehat c_0=\E\|{\widehat{\bm g}}_{{b},{g}}(\theta)\|^2$. We have
\begin{align}
\E\|{\widehat{\bm g}}_{{b},{g}}(\theta)\|^2
&=\frac1{{b}^2}\sum_{i=1}^{b}\E\|{\widehat{\bm g}}_{g}(X_i;\theta)\|^2
+\frac1{{b}^2}\sum_{i\ne i'}\E\big[{\widehat{\bm g}}_{g}(X_i;\theta)^\top{\widehat{\bm g}}_{g}(X_{i'};\theta)\big]\notag\\
&=\frac1{b}\E\|{\widehat{\bm g}}_{g}(X;\theta)\|^2+\frac{{b}-1}{b}\|{\bm g}(\theta)\|^2\notag\\
&=c_0+\frac1{b}\Bigl\{c_1+\frac{c_2}{g}+\frac{c_3}{{g}({g}-1)}\Bigr\}.
\label{eq:varianceest-batch-second-moment}
\end{align}
where the second equality follows from $\E{\widehat{\bm g}}_{g}(X_i;\theta)={\bm g}(\theta)$, and the last equality follows from \eqref{eq:varianceest-prompt-second-moment}.
Combining \eqref{eq:varianceest-c1-definition}, \eqref{eq:varianceest-prompt-second-moment} and \eqref{eq:varianceest-batch-second-moment}, we obtain
\[
\E\widehat c_1=c_1+\frac{c_2}{g}+\frac{c_3}{{g}({g}-1)}.
\]
Thus $\widehat c_1$ has a positive finite-group bias, as stated in \eqref{eqn:c1bias}.

\medskip
\noindent\textbf{(ii) Expectation of $\widehat c_2$.} By definition,
\begin{align*}
\widehat c_2
&=\frac1{{b}({g}-1)}\sum_{i=1}^{b}\sum_{j=1}^{g}\bigl\|S_\theta(X_i,Y_{i,j})(Z_{i,j}-\bar Z_{i,-j})-{\widehat{\bm g}}_{g}(X_i;\theta)\bigr\|^2\\
&=\frac1{{b}({g}-1)}\sum_{i=1}^{b}\sum_{j=1}^{g}\|\psi_j(X_i)-{\widehat{\bm g}}_{g}(X_i;\theta)\|^2\\
&=\frac1{{b}({g}-1)}\sum_{i=1}^{b}\Bigl\{\sum_{j=1}^{g}\|\psi_j(X_i)\|^2-{g}\|{\widehat{\bm g}}_{g}(X_i;\theta)\|^2\Bigr\},
\end{align*}
where the last equality follows from ${g}^{-1}\sum_{j=1}^{g}\psi_j(X_i)={\widehat{\bm g}}_{g}(X_i;\theta)$. Since the second moment of ${\widehat{\bm g}}_{g}(X;\theta)$ is given in \eqref{eq:varianceest-prompt-second-moment}, it remains to compute $\E\|\psi_j(X)\|^2$.
For a fixed prompt $x$, we have
\begin{align}
\E[\|\psi_j(x)\|^2\mid X=x]
&=\E\bigl[\|S_\theta(x,Y_j)\|^2(Z_j-\bar Z_{-j})^2\mid X=x\bigr]\notag\\
&=\E\bigl[\|S_\theta(x,Y_j)\|^2\{Z_j-V^{\pi_\theta}(x)+V^{\pi_\theta}(x)-\bar Z_{-j}\}^2\mid X=x\bigr]\notag\\
&=\E\bigl[\|S_\theta(x,Y_j)\|^2\{Z_j-V^{\pi_\theta}(x)\}^2\mid X=x\bigr]\notag\\
&\quad+2\E\bigl[\|S_\theta(x,Y_j)\|^2\{Z_j-V^{\pi_\theta}(x)\}\mid X=x\bigr]\E[V^{\pi_\theta}(x)-\bar Z_{-j}\mid X=x]\notag\\
&\quad+\E[\|S_\theta(x,Y_j)\|^2\mid X=x]\E[\{V^{\pi_\theta}(x)-\bar Z_{-j}\}^2\mid X=x]\notag\\
&=\E\bigl[\|S_\theta(x,Y)\|^2\{Z-V^{\pi_\theta}(x)\}^2\mid X=x\bigr]\notag\\
&\quad+\frac1{{g}-1}\E[\|S_\theta(x,Y)\|^2\mid X=x]\Var(Z\mid X=x)\notag\\
&=c_2(x)+\|{\bm g}(x;\theta)\|^2+\frac{c_3(x)-\|{\bm g}(x;\theta)\|^2}{{g}-1}\notag\\
&=c_2(x)+\frac{{g}-2}{{g}-1}\|{\bm g}(x;\theta)\|^2+\frac{c_3(x)}{{g}-1},
\label{eq:varianceest-psi-second-moment-x}
\end{align}
where the third equality uses $\bar Z_{-j}\mathrel{\perp\!\!\!\perp}(Y_j,Z_j)\mid X=x$, the fourth equality follows from $\Var(\bar Z_{-j}\mid X=x)=\Var(Z\mid X=x)/({g}-1)$, and the fifth equality follows from \eqref{eq:varianceest-conditional-c3}.
This leads to
\[
\E\|\psi_j(X)\|^2=c_2+\frac{{g}-2}{{g}-1}\E\|{\bm g}(X;\theta)\|^2+\frac{c_3}{{g}-1}.
\]
By \eqref{eq:varianceest-prompt-second-moment}, we obtain
\begin{align*}
\E\widehat c_2
&=\frac {g}{{g}-1}\bigl\{\E\|\psi_1(X)\|^2-\E\|{\widehat{\bm g}}_{g}(X;\theta)\|^2\bigr\}\\
&=c_2-\frac{{g}}{({g}-1)^2}\E\|{\bm g}(X;\theta)\|^2+\frac{c_3}{{g}-1}.
\end{align*}

\medskip
\noindent\textbf{(iii) Expectation of $\widehat c_3$.} By definition,
\begin{equation}
\widehat c_3=\frac1{b}\sum_{i=1}^{b}\Bigl[
\Bigl\{\frac1{g}\sum_{j=1}^{g}\|S_\theta(X_i,Y_{i,j})\|^2\Bigr\}
\Bigl\{\frac1{{g}-1}\sum_{j=1}^{g}(Z_{i,j}-\bar Z_i)^2\Bigr\}
+\|{\widehat{\bm g}}_{g}(X_i;\theta)\|^2\Bigr].
\label{eq:varianceest-c3-definition}
\end{equation}
For a fixed prompt $x$, the sample variance satisfies
\begin{align}
\frac1{{g}-1}\sum_{j=1}^{g}(Z_j-\bar Z)^2
&=\frac1{{g}-1}\Bigl[\sum_{j=1}^{g}\{Z_j-V^{\pi_\theta}(x)\}^2
-\frac1{g}\Bigl\{\sum_{j=1}^{g}(Z_j-V^{\pi_\theta}(x))\Bigr\}^2\Bigr]\notag\\
&=\frac1{g}\sum_{j=1}^{g}\{Z_j-V^{\pi_\theta}(x)\}^2-\frac1{{g}({g}-1)}\sum_{j\ne k}\{Z_j-V^{\pi_\theta}(x)\}\{Z_k-V^{\pi_\theta}(x)\}.
\label{eq:varianceest-sample-variance-expansion}
\end{align}
By \eqref{eq:varianceest-sample-variance-expansion}, we have
\begin{equation}
\begin{aligned}
&\E\Bigl[\Bigl\{\frac1{g}\sum_{\ell=1}^{g}\|S_\theta(x,Y_\ell)\|^2\Bigr\}\Bigl\{\frac1{{g}-1}\sum_{j=1}^{g}(Z_j-\bar Z)^2\Bigr\}\,\Bigm|\,X=x\Bigr]\\
&\quad=\frac1{{g}^2}\sum_{\ell=1}^{g}\sum_{j=1}^{g}\E[\|S_\theta(x,Y_\ell)\|^2\{Z_j-V^{\pi_\theta}(x)\}^2\mid X=x]\\
&\qquad-\frac1{{g}^2({g}-1)}\sum_{\ell=1}^{g}\sum_{j\ne k}\E[\|S_\theta(x,Y_\ell)\|^2\{Z_j-V^{\pi_\theta}(x)\}\{Z_k-V^{\pi_\theta}(x)\}\mid X=x]\\
&\quad=\frac1{{g}^2}\sum_{\ell=1}^{g}\sum_{j=1}^{g}\E[\|S_\theta(x,Y_\ell)\|^2\{Z_j-V^{\pi_\theta}(x)\}^2\mid X=x],
\end{aligned}
\label{eq:varianceest-c3-product-reduction}
\end{equation}
where the second equality follows from the vanishing of the cross-terms, as for each $j\ne k$,
\begin{equation*}
\begin{aligned}
&\E[\|S_\theta(x,Y_\ell)\|^2\{Z_j-V^{\pi_\theta}(x)\}\{Z_k-V^{\pi_\theta}(x)\}\mid X=x]\\
&\quad=\begin{cases}
\E[\|S_\theta(x,Y_j)\|^2\{Z_j-V^{\pi_\theta}(x)\}\mid X=x]\,
\E[Z_k-V^{\pi_\theta}(x)\mid X=x]=0, & \ell=j,\\[4pt]
\E[\|S_\theta(x,Y_k)\|^2\{Z_k-V^{\pi_\theta}(x)\}\mid X=x]\,
\E[Z_j-V^{\pi_\theta}(x)\mid X=x]=0, & \ell=k,\\[4pt]
\E[\|S_\theta(x,Y_\ell)\|^2\mid X=x]\,
\E[Z_j-V^{\pi_\theta}(x)\mid X=x]\,
\E[Z_k-V^{\pi_\theta}(x)\mid X=x]=0, & \ell\notin\{j,k\}.
\end{cases}
\end{aligned}
\end{equation*}
Then, we have
\begin{equation}
\begin{aligned}
&\frac1{{g}^2}\sum_{\ell=1}^{g}\sum_{j=1}^{g}\E[\|S_\theta(x,Y_\ell)\|^2\{Z_j-V^{\pi_\theta}(x)\}^2\mid X=x]\\
&\quad=\frac1{{g}^2}\sum_{j=1}^{g}\E[\|S_\theta(x,Y_j)\|^2\{Z_j-V^{\pi_\theta}(x)\}^2\mid X=x]\\
&\qquad+\frac1{{g}^2}\sum_{\ell\ne j}\E[\|S_\theta(x,Y_\ell)\|^2\{Z_j-V^{\pi_\theta}(x)\}^2\mid X=x]\\
&\quad=\frac{{g}}{{g}^2}\E[\|S_\theta(x,Y)\|^2\{Z-V^{\pi_\theta}(x)\}^2\mid X=x]\\
&\qquad+\frac{{g}({g}-1)}{{g}^2}\E[\|S_\theta(x,Y)\|^2\mid X=x]\E[\{Z-V^{\pi_\theta}(x)\}^2\mid X=x]\\
&\quad=\frac1{g}\E\bigl[\|S_\theta(x,Y)\|^2\{Z-V^{\pi_\theta}(x)\}^2\mid X=x\bigr]\\
&\qquad+\frac{{g}-1}{g}\E[\|S_\theta(x,Y)\|^2\mid X=x]\Var(Z\mid X=x)\\
&\quad=\frac1{g}\{c_2(x)+\|{\bm g}(x;\theta)\|^2\}
+\frac{{g}-1}{g}\{c_3(x)-\|{\bm g}(x;\theta)\|^2\},
\end{aligned}
\label{eq:varianceest-c3-diagonal-sum}
\end{equation}
where the second equality follows from $(Y_\ell,Z_\ell)\mathrel{\perp\!\!\!\perp}(Y_j,Z_j)\mid X=x$ for $\ell\ne j$.

Combining \eqref{eq:varianceest-prompt-second-moment}, \eqref{eq:varianceest-c3-definition}, \eqref{eq:varianceest-c3-product-reduction} and \eqref{eq:varianceest-c3-diagonal-sum}, and using $\E[c_j(X)]=c_j$ for $j=2,3$, we obtain
\begin{align*}
\E\widehat c_3
&=\frac1{g}\{c_2+\E\|{\bm g}(X;\theta)\|^2\}
+\frac{{g}-1}{g}\{c_3-\E\|{\bm g}(X;\theta)\|^2\}+\E\|{\widehat{\bm g}}_{g}(X;\theta)\|^2\\
&=\frac{2c_2}{g}+\frac{{g}^2-2{g}+2}{{g}({g}-1)}c_3+\frac2{g}\E\|{\bm g}(X;\theta)\|^2,
\end{align*}
which proves \eqref{eqn:c3bias}.

\medskip
\noindent\textbf{(iv) Bias correction.} In summary, the following identities hold for the estimators:
\begin{align}
\E\widehat c_0&=c_0+\frac{c_1}{b}+\frac{c_2}{{b}{g}}+\frac{c_3}{{b}{g}({g}-1)},\notag\\
\E\widehat c_1&=c_1+\frac{c_2}{g}+\frac{c_3}{{g}({g}-1)},\notag\\
\E\widehat c_2&=c_2+\frac{c_3}{{g}-1}-\frac{{g}(c_0+c_1)}{({g}-1)^2},\notag\\
\E\widehat c_3&=\frac{{g}^2-2{g}+2}{{g}({g}-1)}c_3+\frac{2(c_0+c_1+c_2)}{g}.
\label{eq:varianceest-moment-map}
\end{align}
Collecting the coefficients in \eqref{eq:varianceest-moment-map} yields $\E\widehat{\bm c}=\bm A\bm c$, where
\begin{equation}
\bm A=\begin{pmatrix}
1&{b}^{-1}&({b}{g})^{-1}&[{b}{g}({g}-1)]^{-1}\\
0&1&{g}^{-1}&[{g}({g}-1)]^{-1}\\
-\frac{{g}}{({g}-1)^2}&-\frac{{g}}{({g}-1)^2}&1&\frac1{{g}-1}\\
\frac2{{g}}&\frac2{{g}}&\frac2{{g}}&\frac{{g}^2-2{g}+2}{{g}({g}-1)}
\end{pmatrix}.
\label{eq:bias-matrix}
\end{equation}
Since
\[
\det(\bm A)=\frac{({g}-2)\{({g}-1)^2+1\}}{({g}-1)^3}>0\qquad({g}\ge3),
\]
we can define $\widetilde{\bm c}=(\widetilde c_0,\widetilde c_1,\widetilde c_2,\widetilde c_3)^\top=\bm A^{-1}\widehat{\bm c}$, which satisfies
\begin{equation*}
\E\widetilde{\bm c}=\bm A^{-1}\E\widehat{\bm c}=\bm c.
\end{equation*}
Expanding $\bm A^{-1}\widehat{\bm c}$ yields
\begin{align}
\widetilde c_0&=\widehat c_0-\frac{\widehat c_1}{{b}},\notag\\
\widetilde c_1&=
\frac{({g}-1)^2({g}\widehat c_1-\widehat c_2)-{g}(\widehat c_0-\widehat c_1/{b})}
{{g}({g}^2-2{g}+2)},\notag\\
\widetilde c_2&=
\frac{{g}^3-2{g}+2}{{g}({g}-2)({g}^2-2{g}+2)}
\bigl\{\widehat c_0+(1-{b}^{-1})\widehat c_1\bigr\}\notag\\
&\quad+\frac{({g}-1)^2({g}^3-2{g}^2+2{g}-2)}{{g}^2({g}-2)({g}^2-2{g}+2)}\widehat c_2
-\frac{\widehat c_3}{{g}-2},\notag\\
\widetilde c_3&=\frac{{g}-1}{{g}-2}\widehat c_3
-\frac{2({g}-1)({g}^2-{g}+1)}{{g}({g}-2)({g}^2-2{g}+2)}
\bigl\{\widehat c_0+(1-{b}^{-1})\widehat c_1\bigr\}\notag\\
&\quad-\frac{2({g}-1)^4}{{g}^2({g}-2)({g}^2-2{g}+2)}\widehat c_2.
\label{eq:app-score-unbiased}
\end{align}
In particular, $\widetilde c_1$ and $\widetilde c_3$ are exactly unbiased, without requiring ${g}\to\infty$.
\end{proof}

\subsubsection{Theorem~\ref{thm:varianceest}(b): Consistency}
\label{app:varianceest-b}

\begin{proof}[Proof of Theorem~\ref{thm:varianceest}(b).]
We first prove consistency of the biased estimators as ${b},{g}\to\infty$. By Markov's inequality, for every $\varepsilon>0$ and $j=1,3$, we have
\begin{align}
\Pr\bigl(|\widehat c_j-c_j|>\varepsilon\bigr)
&\le\frac1{\varepsilon^2}\E|\widehat c_j-c_j|^2\notag\\
&=\frac1{\varepsilon^2}\bigl\{\Var(\widehat c_j)+(\E\widehat c_j-c_j)^2\bigr\}\notag\\
&\le\frac1{\varepsilon^2}\Bigl\{\Var(\widehat c_j)+\frac{C_0}{{g}^2}\Bigr\},
\label{eq:varianceest-biased-mse-bound}
\end{align}
where the last inequality follows from \eqref{eq:varianceest-moment-map} and the bounds $0\le c_k\le C_0$, $k=0,1,2,3$, under Assumption~\ref{assump:upper}. 

We next establish an upper bound for $\Var(\widehat c_j)$, $j=0,1,2,3$.
It follows from the definitions of $\widehat c_2$ and $\widehat c_3$ that
\begin{align}
\Var(\widehat c_2)
&=\frac1{b}\Var\Bigl(\frac1{{g}-1}\sum_{j=1}^{g}
\|\psi_j(X_1)-{\widehat{\bm g}}_{g}(X_1;\theta)\|^2\Bigr)\le\frac{C_0}{b},\notag\\
\Var(\widehat c_3)
&=\frac1{b}\Var\Bigl(
\bigl\{\frac1{g}\sum_{j=1}^{g}\|S_\theta(X_1,Y_{1,j})\|^2\bigr\}
\bigl\{\frac1{{g}-1}\sum_{j=1}^{g}(Z_{1,j}-\bar Z_1)^2\bigr\}
+\|{\widehat{\bm g}}_{g}(X_1;\theta)\|^2\Bigr)\le\frac{C_0}{b},
\label{eq:varianceest-c23-variance-bound}
\end{align}
where the inequalities follow from Assumption~\ref{assump:upper}. For $\widehat c_0$, we have
\begin{align}
\Var(\widehat c_0)
&\le \E\bigl(\|{\widehat{\bm g}}_{{b},{g}}(\theta)\|^2-\|{\bm g}(\theta)\|^2\bigr)^2\notag\\
&\le C_0\E\|{\widehat{\bm g}}_{{b},{g}}(\theta)-{\bm g}(\theta)\|^2
=\frac{C_0}{b}\E\|{\widehat{\bm g}}_{g}(X_1;\theta)-{\bm g}(\theta)\|^2
\le\frac{C_0}{b},
\label{eq:varianceest-c0-variance-bound}
\end{align}
where the first inequality follows from $\Var(\widehat c_0)\le\E(\widehat c_0-c_0)^2$, and the second follows from $|\|u\|^2-\|v\|^2|\le(\|u\|+\|v\|)\|u-v\|$. Then, for $\widehat c_1$,
\begin{align}
{\Var(\widehat c_1)}
&=\frac{{b}^2}{({b}-1)^2}
\Var\Bigl(\frac1{b}\sum_{i=1}^{b}\|{\widehat{\bm g}}_{g}(X_i;\theta)\|^2-\widehat c_0\Bigr)\notag\\
&\le\frac{2{b}^2}{({b}-1)^2}
\Bigl\{\frac1{b}\Var\bigl(\|{\widehat{\bm g}}_{g}(X_1;\theta)\|^2\bigr)+\Var(\widehat c_0)\Bigr\}
\le\frac{C_0}{b},
\label{eq:varianceest-c1-variance-bound}
\end{align}
where the first inequality uses $\Var(U-V)\le2\Var(U)+2\Var(V)$. Combining \eqref{eq:varianceest-c23-variance-bound}--\eqref{eq:varianceest-c1-variance-bound}, we have
\begin{equation}
\E\|\widehat{\bm c}-\E\widehat{\bm c}\|^2
=\sum_{j=0}^3\Var(\widehat c_j)\le\frac{C_0}{b}.
\label{eq:varianceest-raw-variance-bound}
\end{equation}

Substituting \eqref{eq:varianceest-raw-variance-bound} into \eqref{eq:varianceest-biased-mse-bound}, we obtain
\[
\Pr\bigl(|\widehat c_j-c_j|>\varepsilon\bigr)
\le\frac1{\varepsilon^2}\E|\widehat c_j-c_j|^2
\le\frac{C_0}{\varepsilon^2}\Bigl(\frac1{b}+\frac1{{g}^2}\Bigr)\longrightarrow0,
\qquad j=1,3,
\]
as ${b},{g}\to\infty$. Hence $\widehat c_j\xrightarrow{p}c_j$ for $j=1,3$.

In contrast, consider ${b}\to\infty$ with ${g}\le\overline {g}<\infty$ and $c_2\ge\underline c_2>0$. For $\delta=\underline c_2/\overline {g}>0$, by \eqref{eq:varianceest-moment-map} and $c_3\ge0$, we have
\[
\E\widehat c_1-c_1
=\frac{c_2}{g}+\frac{c_3}{{g}({g}-1)}\ge\delta.
\]
Consequently, as ${b}\to\infty$,
\begin{align*}
\Pr\bigl(|\widehat c_1-c_1|>\delta/2\bigr)
&\ge\Pr\bigl(|\E\widehat c_1-c_1|-|\widehat c_1-\E\widehat c_1|>\delta/2\bigr)\\
&\ge\Pr\bigl(|\widehat c_1-\E\widehat c_1|<\delta/2\bigr)\\
&\ge1-\frac{4{\Var(\widehat c_1)}}{\delta^2}
\ge1-\frac{4C_0}{{b}\delta^2}\longrightarrow1
\end{align*}
where the third inequality follows from Chebyshev's inequality. Thus $\widehat c_1$ is inconsistent.

Finally, we prove consistency of the debiased estimators as ${b}\to\infty$ for any ${g}\ge3$. Since $\widetilde{\bm c}-\bm c=\bm A^{-1}(\widehat{\bm c}-\E\widehat{\bm c})$, by Markov's inequality, we have
\begin{align}
\Pr\bigl(\|\widetilde{\bm c}-\bm c\|>\varepsilon\bigr)
&\le\frac1{\varepsilon^2}\E\|\widetilde{\bm c}-\bm c\|^2\notag\\
&=\frac1{\varepsilon^2}\E\|\bm A^{-1}(\widehat{\bm c}-\E\widehat{\bm c})\|^2\notag\\
&\le\frac{\|\bm A^{-1}\|_2^2}{\varepsilon^2}
\E\|\widehat{\bm c}-\E\widehat{\bm c}\|^2\notag\\
&\le\frac{C_0}{\varepsilon^2}\E\|\widehat{\bm c}-\E\widehat{\bm c}\|^2
\le\frac{C_0}{{b}\varepsilon^2}\longrightarrow0,
\label{eq:varianceest-debiased-mse-bound}
\end{align}
as ${b}\to\infty$, where $\|\cdot\|_2$ denotes the matrix norm induced by the Euclidean norm. Here, the second inequality follows from $\|\bm A^{-1}\|_2^2\le C_0$ by \eqref{eq:app-score-unbiased}, and the last inequality follows from \eqref{eq:varianceest-raw-variance-bound}. Thus $\widetilde{\bm c}\xrightarrow{p}\bm c$, for any fixed ${g}\ge3$. 

\end{proof}

\subsubsection{Theorem~\ref{thm:varianceest}(c): Probability bounds}
\label{app:varianceest-c}

\begin{appendixlemma}[Conditional moment bounds]
\label{lem:within-prompt-moment-bounds}
Suppose Assumption~\ref{assump:upper} holds, and let ${g}\ge3$. Define
\[
\varepsilon_i={\widehat{\bm g}}_{g}(X_i;\theta)-{\bm g}(X_i;\theta).
\]
For the projections in \eqref{eq:lemma-gradient-projections} and each $q=1,2,3$, we have
\begin{align*}
\E\Bigl[\bigl\|\frac1{g}\sum_{j=1}^{g}\zeta_{1,j}(X_i)\bigr\|^{2q}\,\Bigm|\,X_i\Bigr]
&=O({g}^{-q}),\\
\E\Bigl[\bigl\|\binom {g}2^{-1}\sum_{j<k}\zeta_{2,jk}(X_i)\bigr\|^{2q}\,\Bigm|\,X_i\Bigr]
&=O({g}^{-2q}),\\
\E(\|\varepsilon_i\|^{2q}\mid X_i)
&=O({g}^{-q}),
\end{align*}
where the bounds hold uniformly in ${b}$, ${g}$ and $X_i$.
\end{appendixlemma}

\begin{proof}[Proof of Lemma~\ref{lem:within-prompt-moment-bounds}.]
Applying the Hoeffding decomposition \eqref{eq:lemma-ustat-decomposition} to the U-statistic ${\widehat{\bm g}}_{g}(X_i;\theta)$, we obtain
\begin{equation}
\varepsilon_i
=\frac1{g}\sum_{j=1}^{g}\zeta_{1,j}(X_i)
+\binom {g}2^{-1}\sum_{j<k}\zeta_{2,jk}(X_i),
\label{eq:within-prompt-hoeffding}
\end{equation}
where $\zeta_{1,j}(X_i)$ and $\zeta_{2,jk}(X_i)$ are the projections defined in \eqref{eq:lemma-gradient-projections} with $x=X_i$.
Under Assumption~\ref{assump:upper}, the parameter dimension satisfies $d\le d_0$ for some constant $d_0\ge1$, and $\|\zeta_{1,j}(X_i)\|$, $\|S_\theta(X_i,Y_{i,j})\|$ and $|Z_{i,j}-V^{\pi_\theta}(X_i)|$ are uniformly bounded. Applying Lemma~\ref{lem:conditional-vector-rosenthal} with $X=X_i$, $T_j=\zeta_{1,j}(X_i)$ and $p=2q$, we obtain
\begin{equation}
\E\Bigl[\bigl\|\frac1{g}\sum_{j=1}^{g}\zeta_{1,j}(X_i)\bigr\|^{2q}\,\Bigm|\,X_i\Bigr]
=O({g}^{-q}),\qquad q=1,2,3.
\label{eq:within-prompt-linear-moments}
\end{equation}

For the second term in \eqref{eq:within-prompt-hoeffding}, by \eqref{eq:lemma-gradient-projections}, we have
\begin{align}
\binom {g}2^{-1}\sum_{j<k}\zeta_{2,jk}(X_i)
&=\frac1{{g}({g}-1)}\Bigl[\sum_{j=1}^{g} S_\theta(X_i,Y_{i,j})\{Z_{i,j}-V^{\pi_\theta}(X_i)\}\notag\\
&\qquad-\Bigl\{\sum_{j=1}^{g} S_\theta(X_i,Y_{i,j})\Bigr\}
\Bigl\{\sum_{k=1}^{g}[Z_{i,k}-V^{\pi_\theta}(X_i)]\Bigr\}\Bigr],
\label{eq:within-prompt-product-decomposition}
\end{align}
where the equality follows by separating the diagonal terms from the product of the two sums. Since $\E[S_\theta(X_i,Y_{i,j})\mid X_i]=0$ and $\E[Z_{i,j}-V^{\pi_\theta}(X_i)\mid X_i]=0$, by the Cauchy--Schwarz inequality and Lemma~\ref{lem:conditional-vector-rosenthal} with $p=4q$, we have
\begin{align}
&\E\Bigl[\bigl\|\sum_{j=1}^{g} S_\theta(X_i,Y_{i,j})\bigr\|^{2q}
\bigl|\sum_{k=1}^{g}[Z_{i,k}-V^{\pi_\theta}(X_i)]\bigr|^{2q}\,\Bigm|\,X_i\Bigr]\notag\\
&\quad\le
\Bigl\{\E\Bigl[\bigl\|\sum_{j=1}^{g} S_\theta(X_i,Y_{i,j})\bigr\|^{4q}\,\Bigm|\,X_i\Bigr]\Bigr\}^{1/2}
\Bigl\{\E\Bigl[\bigl|\sum_{k=1}^{g}[Z_{i,k}-V^{\pi_\theta}(X_i)]\bigr|^{4q}\,\Bigm|\,X_i\Bigr]\Bigr\}^{1/2}=O({g}^{2q}).
\label{eq:within-prompt-product-moment-bound}
\end{align}
Moreover, we have
\begin{equation}
\Bigl\|\sum_{j=1}^{g} S_\theta(X_i,Y_{i,j})\{Z_{i,j}-V^{\pi_\theta}(X_i)\}\Bigr\|
\le\sum_{j=1}^{g}\|S_\theta(X_i,Y_{i,j})\|\,|Z_{i,j}-V^{\pi_\theta}(X_i)|
=O({g}).
\label{eq:within-prompt-diagonal-bound}
\end{equation}
Applying $\|u-v\|^{2q}\le2^{2q-1}(\|u\|^{2q}+\|v\|^{2q})$ to \eqref{eq:within-prompt-product-decomposition}, together with \eqref{eq:within-prompt-product-moment-bound} and \eqref{eq:within-prompt-diagonal-bound}, we obtain
\begin{align}
\E\Bigl[\bigl\|\binom {g}2^{-1}\sum_{j<k}\zeta_{2,jk}(X_i)\bigr\|^{2q}\,\Bigm|\,X_i\Bigr]
&=\frac{O({g}^{2q})+O({g}^{2q})}{\{{g}({g}-1)\}^{2q}}=O({g}^{-2q}).
\label{eq:within-prompt-moments}
\end{align}
Combining \eqref{eq:within-prompt-hoeffding} with \eqref{eq:within-prompt-linear-moments} and \eqref{eq:within-prompt-moments}, we obtain
\begin{equation*}
\E(\|\varepsilon_i\|^{2q}\mid X_i)
=O({g}^{-q}+{g}^{-2q})
=O({g}^{-q}),\qquad q=1,2,3.
\end{equation*}
This completes the proof of Lemma~\ref{lem:within-prompt-moment-bounds}.
\end{proof}

\begin{appendixlemma}[Conditional decomposition and variance bounds]
\label{lem:conditional-decomposition-variance}
Let $X_{1:{b}}=(X_1,\ldots,X_{b})$, $Y_{i,1:{g}}=(Y_{i,1},\ldots,Y_{i,{g}})$ and ${\bar{\bm g}}_{b}(\theta)={b}^{-1}\sum_{i=1}^{b} {\bm g}(X_i;\theta)$, and define
\[
\begin{aligned}
c_{1,{b}}=\frac1{{b}-1}\sum_{i=1}^{b}\|{\bm g}(X_i;\theta)&-{\bar{\bm g}}_{b}(\theta)\|^2,\quad
\xi_i={b}\{\E(\widetilde c_1\mid X_{1:{b}},Y_{i,1:{g}})-c_{1,{b}}\},\\
&\sigma_{b}^2=\frac{{g}^2}{{b}}\sum_{i=1}^{b}\Var(\xi_i\mid X_{1:{b}}).
\end{aligned}
\]
Then, under Assumptions~\ref{assump:upper} and~\ref{assump:lower}, the following conclusions hold for ${b}\ge2$ and ${g}\ge3$, and the bounds hold uniformly in ${b}$ and ${g}$.
\begin{enumerate}
\item[(a)] We have $\E(\widetilde c_1\mid X_{1:{b}})=c_{1,{b}}$ and $\E c_{1,{b}}=c_1$. Moreover, conditional on $X_{1:{b}}$, the variables $\xi_1,\ldots,\xi_{b}$ are independent with mean zero, and
\begin{align*}
&\widetilde c_1-c_{1,{b}}
=\frac1{b}\sum_{i=1}^{b}\xi_i-\frac1{{b}({b}-1)}\sum_{i\ne\ell}\varepsilon_i^\top\varepsilon_\ell,\\
&\E\Bigl[\Bigl\{{g}\sqrt {b}(\widetilde c_1-c_{1,{b}})-\frac {g}{\sqrt {b}}\sum_{i=1}^{b}\xi_i\Bigr\}^2\Bigm|X_{1:{b}}\Bigr]
=O({b}^{-1}).
\end{align*}
\item[(b)] We have $\sigma_{b}^2=O(1+{g}c_{1,{b}})$, and there exist fixed constants $\nu,c>0$ and ${g}_0\ge3$ such that
\[
\Pr(\sigma_{b}<\nu)=O(e^{-c{b}}+c_1),\qquad {g}\ge {g}_0.
\]
\end{enumerate}
\end{appendixlemma}

\begin{proof}[Proof of Lemma~\ref{lem:conditional-decomposition-variance}.]
\noindent\textbf{(a) Conditional decomposition.}
Substituting the definitions of $\widehat c_0,\widehat c_1$ and $\widehat c_2$ into \eqref{eq:app-score-unbiased}, we obtain
\begin{align}
\widetilde c_1
&=\widehat c_1+\frac{{g}-2}{{b}({g}^2-2{g}+2)}\sum_{i=1}^{b}\|{\widehat{\bm g}}_{g}(X_i;\theta)\|^2
-\frac{{g}-1}{{b}{g}({g}^2-2{g}+2)}\sum_{i=1}^{b}\sum_{j=1}^{g}\|\psi_j(X_i)\|^2\notag\\
&=\frac{{g}({g}-1)}{{b}({g}^2-2{g}+2)}\sum_{i=1}^{b}
\Bigl\{\|{\widehat{\bm g}}_{g}(X_i;\theta)\|^2-\frac1{{g}^2}\sum_{j=1}^{g}\|\psi_j(X_i)\|^2\Bigr\}-\frac1{{b}({b}-1)}\sum_{i\ne\ell}{\widehat{\bm g}}_{g}(X_i;\theta)^\top{\widehat{\bm g}}_{g}(X_\ell;\theta).
\label{eq:conditional-group-representation}
\end{align}
Define
\[
Q_i=\|\varepsilon_i\|^2-\frac1{{g}^2}\sum_{j=1}^{g}\|\psi_j(X_i)-{\bm g}(X_i;\theta)\|^2.
\]
Since ${\widehat{\bm g}}_{g}(X_i;\theta)={\bm g}(X_i;\theta)+\varepsilon_i={g}^{-1}\sum_{j=1}^{g}\psi_j(X_i)$, we have
\begin{align}
&\|{\widehat{\bm g}}_{g}(X_i;\theta)\|^2-\frac1{{g}^2}\sum_{j=1}^{g}\|\psi_j(X_i)\|^2=\|{\bm g}(X_i;\theta)\|^2+2{\bm g}(X_i;\theta)^\top\varepsilon_i+\|\varepsilon_i\|^2\notag\\
&\qquad-\frac1{{g}^2}\sum_{j=1}^{g}\Bigl[\|{\bm g}(X_i;\theta)\|^2+2{\bm g}(X_i;\theta)^\top\{\psi_j(X_i)-{\bm g}(X_i;\theta)\}+\|\psi_j(X_i)-{\bm g}(X_i;\theta)\|^2\Bigr]\notag\\
&\quad=\|{\bm g}(X_i;\theta)\|^2+2{\bm g}(X_i;\theta)^\top\varepsilon_i+\|\varepsilon_i\|^2
-\frac1{g}\|{\bm g}(X_i;\theta)\|^2\notag\\
&\qquad-\frac2{g}{\bm g}(X_i;\theta)^\top\varepsilon_i
-\frac1{{g}^2}\sum_{j=1}^{g}\|\psi_j(X_i)-{\bm g}(X_i;\theta)\|^2\notag\\
&\quad=\frac{{g}-1}{g}\|{\bm g}(X_i;\theta)\|^2+\frac{2({g}-1)}{g}{\bm g}(X_i;\theta)^\top\varepsilon_i+Q_i.
\label{eq:conditional-within-group-expansion}
\end{align}
where the second equality follows from $\sum_{j=1}^{g}\{\psi_j(X_i)-{\bm g}(X_i;\theta)\}={g}\varepsilon_i$.
By \eqref{eq:varianceest-prompt-second-moment-x}, \eqref{eq:varianceest-psi-second-moment-x} and $\E\{\psi_j(X_i)\mid X_i\}={\bm g}(X_i;\theta)$, we obtain
\begin{align}
\E(Q_i\mid X_i)
&=\frac{c_2(X_i)}{g}+\frac{c_3(X_i)}{{g}({g}-1)}
-\frac1{g}\Bigl\{c_2(X_i)+\frac{c_3(X_i)-\|{\bm g}(X_i;\theta)\|^2}{{g}-1}\Bigr\}=\frac{\|{\bm g}(X_i;\theta)\|^2}{{g}({g}-1)}.
\label{eq:conditional-quadratic-mean}
\end{align}
Taking the conditional expectation given $X_{1:{b}}$ in \eqref{eq:conditional-group-representation} and applying \eqref{eq:conditional-within-group-expansion} and \eqref{eq:conditional-quadratic-mean}, we obtain
\begin{align}
\E(\widetilde c_1\mid X_{1:{b}})
&=\frac1{b}\sum_{i=1}^{b}\|{\bm g}(X_i;\theta)\|^2
-\frac1{{b}({b}-1)}\sum_{i\ne\ell}{\bm g}(X_i;\theta)^\top {\bm g}(X_\ell;\theta)\notag\\
&=\frac1{{b}-1}\sum_{i=1}^{b}\|{\bm g}(X_i;\theta)-{\bar{\bm g}}_{b}(\theta)\|^2=c_{1,{b}},\notag\\
\E c_{1,{b}}&=\E\|{\bm g}(X;\theta)\|^2-\|{\bm g}(\theta)\|^2=c_1.
\label{eq:conditional-centering}
\end{align}

By definition,
\begin{equation}
\frac{\xi_i}{{b}}=\E(\widetilde c_1\mid X_{1:{b}},Y_{i,1:{g}})-c_{1,{b}}.
\label{eq:conditional-first-projection}
\end{equation}
Conditional on $X_{1:{b}}$, each $\xi_i$ depends only on $Y_{i,1:{g}}$, so $\xi_1,\ldots,\xi_{b}$ are independent, and $\E(\xi_i\mid X_{1:{b}})=0$ by \eqref{eq:conditional-centering}. Substituting \eqref{eq:conditional-group-representation} into \eqref{eq:conditional-first-projection} and using \eqref{eq:conditional-within-group-expansion} and \eqref{eq:conditional-quadratic-mean}, we obtain
\begin{align}
\xi_i
&=\frac{2({g}-1)^2}{{g}^2-2{g}+2}{\bm g}(X_i;\theta)^\top\varepsilon_i
-\frac2{{b}-1}\sum_{\ell\ne i}{\bm g}(X_\ell;\theta)^\top\varepsilon_i
+\frac{{g}({g}-1)}{{g}^2-2{g}+2}\{Q_i-\E(Q_i\mid X_i)\}\notag\\
&=\frac{2{b}}{{b}-1}\{{\bm g}(X_i;\theta)-{\bar{\bm g}}_{b}(\theta)\}^\top\varepsilon_i
-\frac2{{g}^2-2{g}+2}{\bm g}(X_i;\theta)^\top\varepsilon_i+\frac{{g}({g}-1)}{{g}^2-2{g}+2}\{Q_i-\E(Q_i\mid X_i)\},
\label{eq:conditional-first-projection-explicit}
\end{align}
where the first equality follows from $\E[{\widehat{\bm g}}_{g}(X_\ell;\theta)\mid X_{1:{b}},Y_{i,1:{g}}]={\bm g}(X_\ell;\theta)$ for $\ell\ne i$, and the second equality follows from $\sum_{\ell\ne i}{\bm g}(X_\ell;\theta)={b}{\bar{\bm g}}_{b}(\theta)-{\bm g}(X_i;\theta)$.
Moreover, for $i\ne\ell$, we have
\[
{\widehat{\bm g}}_{g}(X_i;\theta)^\top{\widehat{\bm g}}_{g}(X_\ell;\theta)-{\bm g}(X_i;\theta)^\top {\bm g}(X_\ell;\theta)={\bm g}(X_i;\theta)^\top\varepsilon_\ell+\varepsilon_i^\top {\bm g}(X_\ell;\theta)+\varepsilon_i^\top\varepsilon_\ell,
\]
which, together with \eqref{eq:conditional-group-representation}--\eqref{eq:conditional-centering} and \eqref{eq:conditional-first-projection-explicit}, leads to
\begin{equation}
\widetilde c_1-c_{1,{b}}
=\frac1{b}\sum_{i=1}^{b}\xi_i-\frac1{{b}({b}-1)}\sum_{i\ne\ell}\varepsilon_i^\top\varepsilon_\ell.
\label{eq:conditional-exact-decomposition}
\end{equation}

By \eqref{eq:conditional-exact-decomposition}, we have
\[
{g}\sqrt {b}(\widetilde c_1-c_{1,{b}})-\frac {g}{\sqrt {b}}\sum_{i=1}^{b}\xi_i
=-\frac{{g}}{\sqrt {b}({b}-1)}\sum_{i\ne\ell}\varepsilon_i^\top\varepsilon_\ell,
\]
and hence
\begin{align*}
&\E\Bigl[\Bigl\{{g}\sqrt {b}(\widetilde c_1-c_{1,{b}})-\frac {g}{\sqrt {b}}\sum_{i=1}^{b}\xi_i\Bigr\}^2\Bigm|X_{1:{b}}\Bigr]\\
&\quad=\frac{{g}^2}{{b}({b}-1)^2}\E\Bigl[\Bigl(\sum_{i\ne\ell}\varepsilon_i^\top\varepsilon_\ell\Bigr)^2\Bigm|X_{1:{b}}\Bigr]\\
&\quad=\frac{{g}^2}{{b}({b}-1)^2}\sum_{i\ne\ell}\sum_{r\ne s}
\E[(\varepsilon_i^\top\varepsilon_\ell)(\varepsilon_r^\top\varepsilon_s)\mid X_{1:{b}}]\\
&\quad=\frac{2{g}^2}{{b}({b}-1)^2}\sum_{i\ne\ell}\E[(\varepsilon_i^\top\varepsilon_\ell)^2\mid X_{1:{b}}]\\
&\quad\le\frac{2{g}^2}{{b}({b}-1)^2}\sum_{i\ne\ell}
\E[\|\varepsilon_i\|^2\|\varepsilon_\ell\|^2\mid X_{1:{b}}]\\
&\quad=\frac{2{g}^2}{{b}({b}-1)^2}\sum_{i\ne\ell}\E(\|\varepsilon_i\|^2\mid X_i)\E(\|\varepsilon_\ell\|^2\mid X_\ell)\\
&\quad=\frac{2{g}^2}{{b}({b}-1)^2}\cdot {b}({b}-1)\cdot O({g}^{-2})
=O({b}^{-1}),
\end{align*}
where the third equality holds because $\{\varepsilon_i\}_{i=1}^{b}$ are conditionally independent with mean zero given $X_{1:{b}}$, so that $\E[(\varepsilon_i^\top\varepsilon_\ell)(\varepsilon_r^\top\varepsilon_s)\mid X_{1:{b}}]=0$ unless $\{r,s\}=\{i,\ell\}$. Here, the first inequality follows from the Cauchy--Schwarz inequality, the fourth equality again uses conditional independence, and the final bound follows from Lemma~\ref{lem:within-prompt-moment-bounds} with $q=1$. This proves part~(a).

\smallskip
\noindent\textbf{(b) Conditional variance bounds.}
Recall that the conditional variance of $({g}/\sqrt {b})\sum_{i=1}^{b}\xi_i$ is
\begin{equation}
\sigma_{b}^2=\frac{{g}^2}{{b}}\sum_{i=1}^{b}\Var(\xi_i\mid X_{1:{b}}).
\label{eq:conditional-variance-definition}
\end{equation}
To establish the bounds for $\sigma_{b}$, we approximate $\varepsilon_i$ and $Q_i-\E(Q_i\mid X_i)$ in \eqref{eq:conditional-first-projection-explicit} by conditionally uncorrelated linear and quadratic terms, respectively.

By \eqref{eq:within-prompt-hoeffding}, we have
\begin{equation}
\varepsilon_i=\bar\zeta_1(X_i)+\bar\zeta_2(X_i),
\label{eq:conditional-error-projections}
\end{equation}
where we define
\[
\bar\zeta_1(X_i)=\frac1{g}\sum_{j=1}^{g}\zeta_{1,j}(
X_i),\qquad
\bar\zeta_2(X_i)=\binom {g}2^{-1}\sum_{j<k}\zeta_{2,jk}(X_i).
\]
Then, we define the quadratic approximation to $Q_i$ as
\begin{equation}
Q_i^0=\|\bar\zeta_1(X_i)\|^2-\frac1{{g}^2}\sum_{j=1}^{g}\|\zeta_{1,j}(X_i)\|^2
=\frac1{{g}^2}\sum_{j\ne k}\zeta_{1,j}(X_i)^\top\zeta_{1,k}(X_i),
\label{eq:quadratic-approx}
\end{equation}
and the conditional covariance matrix as $\Sigma_i=\Cov\{\zeta_{1,j}(X_i)\mid X_i\}$. Since $\{\zeta_{1,j}(X_i)\}_{j=1}^{g}$ are conditionally i.i.d. given $X_i$ and satisfy $\E\{\zeta_{1,j}(X_i)\mid X_i\}=0$, it follows from \eqref{eq:quadratic-approx} that
\begin{align}
\E(Q_i^0\mid X_i)&=0,\qquad \E[\bar\zeta_1(X_i)Q_i^0\mid X_i]=0,\qquad \Var(Q_i^0\mid X_i)=\frac{2({g}-1)}{{g}^3}\tr(\Sigma_i^2).
\label{eq:conditional-quadratic-leading-moments}
\end{align}
Thus $Q_i^0$ is uncorrelated with every linear form in $\bar\zeta_1(X_i)$ conditional on $X_i$. To use this variance calculation for $\xi_i$, we first bound the approximation error. Substituting \eqref{eq:conditional-error-projections} into the definition of $Q_i$ and $Q_i^0$, we obtain
\begin{align}
Q_i-Q_i^0
&=\{2\bar\zeta_1(X_i)+\bar\zeta_2(X_i)\}^\top\bar\zeta_2(X_i)-\frac1{{g}^2}\sum_{j=1}^{g}
\bigl\{\|\psi_j(X_i)-{\bm g}(X_i;\theta)\|^2-\|\zeta_{1,j}(X_i)\|^2\bigr\}.
\label{eq:conditional-quadratic-expansion}
\end{align}
For the second term on the RHS of \eqref{eq:conditional-quadratic-expansion}, we have
\begin{align}
&\E\Bigl[\bigl\{\|\psi_j(X_i)-{\bm g}(X_i;\theta)\|^2-\|\zeta_{1,j}(X_i)\|^2\bigr\}^2\mid X_i\Bigr]\notag\\
&\quad\le C_0\E\bigl[\|\psi_j(X_i)-{\bm g}(X_i;\theta)-\zeta_{1,j}(X_i)\|^2\mid X_i\bigr]\notag\\
&\quad=C_0\E\bigl[\|S_\theta(X_i,Y_{i,j})\|^2\{\bar Z_{i,-j}-V^{\pi_\theta}(X_i)\}^2\mid X_i\bigr]\notag\\
&\quad=\frac{C_0}{{g}-1}\E\bigl[\|S_\theta(X_i,Y_{i,j})\|^2\mid X_i\bigr]\Var(Z_{i,j}\mid X_i)
=O({g}^{-1}),
\label{eq:conditional-diagonal-error}
\end{align}
where $C_0>0$ is a constant independent of ${b}$ and ${g}$. Here, the inequality follows from $(\|u\|^2-\|v\|^2)^2\le(\|u\|+\|v\|)^2\|u-v\|^2$ and Assumption~\ref{assump:upper}, which also implies the last equality. Then, by \eqref{eq:conditional-quadratic-expansion}, we obtain
\begin{align}
\E[(Q_i-Q_i^0)^2\mid X_i]
&\le2\E\Bigl[\bigl\{\{2\bar\zeta_1(X_i)+\bar\zeta_2(X_i)\}^\top\bar\zeta_2(X_i)\bigr\}^2\mid X_i\Bigr]\notag\\
&\quad+\frac2{{g}^4}\E\biggl[\Bigl\{\sum_{j=1}^{g}\bigl(\|\psi_j(X_i)-{\bm g}(X_i;\theta)\|^2-\|\zeta_{1,j}(X_i)\|^2\bigr)\Bigr\}^2\Bigm|X_i\biggr]\notag\\
&\le2\E\bigl[\|2\bar\zeta_1(X_i)+\bar\zeta_2(X_i)\|^2\|\bar\zeta_2(X_i)\|^2\mid X_i\bigr]\notag\\
&\quad+\frac2{{g}^3}\sum_{j=1}^{g}
\E\Bigl[\bigl\{\|\psi_j(X_i)-{\bm g}(X_i;\theta)\|^2-\|\zeta_{1,j}(X_i)\|^2\bigr\}^2\mid X_i\Bigr]\notag\\
&\le2\Bigl\{\E[\|2\bar\zeta_1(X_i)+\bar\zeta_2(X_i)\|^4\mid X_i]\cdot
\E[\|\bar\zeta_2(X_i)\|^4\mid X_i]\Bigr\}^{1/2}+O({g}^{-3})\notag\\
&\le2\Bigl\{\bigl(128\E[\|\bar\zeta_1(X_i)\|^4\mid X_i]
+8\E[\|\bar\zeta_2(X_i)\|^4\mid X_i]\bigr)\cdot\E[\|\bar\zeta_2(X_i)\|^4\mid X_i]\Bigr\}^{1/2}+O({g}^{-3})\notag\\
&=O\bigl\{\bigl(({g}^{-2}+{g}^{-4}){g}^{-4}\bigr)^{1/2}\bigr\}+O({g}^{-3})
=O({g}^{-3}).
\label{eq:conditional-quadratic-error}
\end{align}
Here, the first inequality follows from $(a-b)^2\le2a^2+2b^2$, and the second inequality applies the Cauchy--Schwarz inequality to both the inner product and the finite sum. Then, the third inequality combines the conditional Cauchy--Schwarz inequality with \eqref{eq:conditional-diagonal-error}, which bounds the second term by ${g}^{-3}\cdot {g}\cdot O({g}^{-1})=O({g}^{-3})$, the fourth follows from the triangle inequality and $(a+b)^4\le8(a^4+b^4)$, and the last equality follows from Lemma~\ref{lem:within-prompt-moment-bounds} with $q=2$.

Subsequently, we approximate $\xi_i$ in \eqref{eq:conditional-first-projection-explicit} by
\begin{equation}
\xi_i^0=\frac{2{b}}{{b}-1}\{{\bm g}(X_i;\theta)-{\bar{\bm g}}_{b}(\theta)\}^\top\bar\zeta_1(X_i)
+\frac{{g}({g}-1)}{{g}^2-2{g}+2}Q_i^0,
\label{eq:conditional-leading-summand}
\end{equation}
so that
\begin{align}
{g}(\xi_i-\xi_i^0)
&=\frac{2{b}{g}}{{b}-1}\{{\bm g}(X_i;\theta)-{\bar{\bm g}}_{b}(\theta)\}^\top\bar\zeta_2(X_i)\notag\\
&\quad-\frac{2{g}}{{g}^2-2{g}+2}{\bm g}(X_i;\theta)^\top\varepsilon_i
+\frac{{g}^2({g}-1)}{{g}^2-2{g}+2}\{Q_i-Q_i^0-\E(Q_i\mid X_i)\}.
\label{eq:conditional-summand-error-expansion}
\end{align}
For the last term on the RHS of \eqref{eq:conditional-summand-error-expansion}, we have
\begin{align}
\E[\{Q_i-Q_i^0-\E(Q_i\mid X_i)\}^2\mid X_i]
&=\Var(Q_i-Q_i^0\mid X_i)\notag\\
&\le\E[(Q_i-Q_i^0)^2\mid X_i]=O({g}^{-3}),
\label{eq:conditional-centered-quadratic-error}
\end{align}
where the first equality uses $\E(Q_i^0\mid X_i)=0$ from \eqref{eq:conditional-quadratic-leading-moments}, and the final bound follows from \eqref{eq:conditional-quadratic-error}. Applying these bounds to \eqref{eq:conditional-summand-error-expansion}, we obtain
\begin{align}
\frac{{g}^2}{{b}}\sum_{i=1}^{b}\E[(\xi_i-\xi_i^0)^2\mid X_{1:{b}}]&\le\frac3{b}\Bigl(\frac{2{b}{g}}{{b}-1}\Bigr)^2\sum_{i=1}^{b}
\|{\bm g}(X_i;\theta)-{\bar{\bm g}}_{b}(\theta)\|^2\E[\|\bar\zeta_2(X_i)\|^2\mid X_i]\notag\\
&\quad\quad+\frac3{b}\Bigl(\frac{2{g}}{{g}^2-2{g}+2}\Bigr)^2\sum_{i=1}^{b}
\|{\bm g}(X_i;\theta)\|^2\E[\|\varepsilon_i\|^2\mid X_i]\notag\\
&\quad\quad+\frac3{b}\Bigl(\frac{{g}^2({g}-1)}{{g}^2-2{g}+2}\Bigr)^2\sum_{i=1}^{b}
\E[\{Q_i-Q_i^0-\E(Q_i\mid X_i)\}^2\mid X_i]\notag\\
&=O\Bigl(\frac1{b}\sum_{i=1}^{b}\|{\bm g}(X_i;\theta)-{\bar{\bm g}}_{b}(\theta)\|^2\Bigr)+O({g}^{-3})+O({g}^{-1})\notag\\
&=O(c_{1,{b}}+{g}^{-1}),
\label{eq:conditional-summand-approximation}
\end{align}
where the inequality follows from $(a+b+c)^2\le3(a^2+b^2+c^2)$ and the Cauchy--Schwarz inequality. Here, for the first equality, we bound the first two terms using Lemma~\ref{lem:within-prompt-moment-bounds} with $q=1$, together with the boundedness of ${\bm g}(X_i;\theta)$ under Assumption~\ref{assump:upper} for the second term, and the third term using \eqref{eq:conditional-centered-quadratic-error}. 

Additionally, the two terms in \eqref{eq:conditional-leading-summand} are conditionally uncorrelated by \eqref{eq:conditional-quadratic-leading-moments}. Thus,
\begin{align}
&\frac{{g}^2}{{b}}\sum_{i=1}^{b}\Var(\xi_i^0\mid X_{1:{b}})\notag\\
&\quad=\frac{{g}^2}{{b}}\sum_{i=1}^{b}\Bigl[
\Bigl(\frac{2{b}}{{b}-1}\Bigr)^2\{{\bm g}(X_i;\theta)-{\bar{\bm g}}_{b}(\theta)\}^{\top}
\frac{\Sigma_i}{{g}}\{{\bm g}(X_i;\theta)-{\bar{\bm g}}_{b}(\theta)\}\notag\\
&\qquad\qquad\qquad+\Bigl(\frac{{g}({g}-1)}{{g}^2-2{g}+2}\Bigr)^2\frac{2({g}-1)}{{g}^3}\tr(\Sigma_i^2)\Bigr]\notag\\
&\quad=\frac{2{g}({g}-1)^3}{{b}({g}^2-2{g}+2)^2}\sum_{i=1}^{b}\tr(\Sigma_i^2)
+\frac{4{b}{g}}{({b}-1)^2}\sum_{i=1}^{b}
\{{\bm g}(X_i;\theta)-{\bar{\bm g}}_{b}(\theta)\}^{\top}\Sigma_i\{{\bm g}(X_i;\theta)-{\bar{\bm g}}_{b}(\theta)\}\notag\\
&\quad=\frac2{b}\sum_{i=1}^{b}\tr(\Sigma_i^2)
+\frac{4{b}{g}}{({b}-1)^2}\sum_{i=1}^{b}
\{{\bm g}(X_i;\theta)-{\bar{\bm g}}_{b}(\theta)\}^{\top}\Sigma_i\{{\bm g}(X_i;\theta)-{\bar{\bm g}}_{b}(\theta)\}
+O({g}^{-1}),
\label{eq:conditional-leading-variance}
\end{align}
where the first equality follows from $\Cov\{\bar\zeta_1(X_i)\mid X_{1:{b}}\}={g}^{-2}\sum_{j=1}^{g}\Sigma_i=\Sigma_i/{g}$ and \eqref{eq:conditional-quadratic-leading-moments}, and the last equality follows from
\[
\frac{2{g}({g}-1)^3}{({g}^2-2{g}+2)^2}-2=O({g}^{-1})
\]
and the boundedness of $\Sigma_i$ under Assumption~\ref{assump:upper}. Since $\Sigma_i$ is positive semidefinite, we have
\begin{equation}
\frac{{g}^2}{{b}}\sum_{i=1}^{b}\Var(\xi_i^0\mid X_{1:{b}})
\ge\frac2{b}\sum_{i=1}^{b}\tr(\Sigma_i^2)-\frac{C_\Sigma}{{g}},
\label{eq:conditional-leading-variance-lower-bound}
\end{equation}
where $C_\Sigma>0$ is a constant independent of ${b}$, ${g}$ and $X_{1:{b}}$. Here, the inequality follows from \eqref{eq:conditional-leading-variance} by discarding the nonnegative quadratic form and bounding the remainder below by $-C_\Sigma/{g}$. 

Both $\xi_i$ and $\xi_i^0$ have conditional mean zero given $X_{1:{b}}$. Consequently,
\begin{align}
\sigma_{b}
&=\Bigl\{\frac{{g}^2}{{b}}\sum_{i=1}^{b}\E(\xi_i^2\mid X_{1:{b}})\Bigr\}^{1/2}\notag\\
&\ge\Bigl\{\frac{{g}^2}{{b}}\sum_{i=1}^{b}\E[(\xi_i^0)^2\mid X_{1:{b}}]\Bigr\}^{1/2}
-\Bigl\{\frac{{g}^2}{{b}}\sum_{i=1}^{b}\E[(\xi_i-\xi_i^0)^2\mid X_{1:{b}}]\Bigr\}^{1/2}\notag\\
&=\Bigl\{\frac{{g}^2}{{b}}\sum_{i=1}^{b}\Var(\xi_i^0\mid X_{1:{b}})\Bigr\}^{1/2}
-O(\sqrt{c_{1,{b}}}+{g}^{-1/2})\notag\\
&\ge\Bigl\{\frac{{g}^2}{{b}}\sum_{i=1}^{b}\Var(\xi_i^0\mid X_{1:{b}})+\frac{C_\Sigma}{{g}}\Bigr\}^{1/2}
-\sqrt{C_\Sigma}\,{g}^{-1/2}-O(\sqrt{c_{1,{b}}}+{g}^{-1/2})\notag\\
&\ge\Bigl\{\frac2{b}\sum_{i=1}^{b}\tr(\Sigma_i^2)\Bigr\}^{1/2}
-K(\sqrt{c_{1,{b}}}+{g}^{-1/2}),
\label{eq:conditional-variance-lower-bound}
\end{align}
where $K>0$ is a constant independent of ${b}$, ${g}$ and $X_{1:{b}}$. The first inequality is the reverse triangle inequality for the conditional $L^2$ norm, and the second equality follows from \eqref{eq:conditional-summand-approximation}. The second last inequality follows from $\sqrt{a+b}\le\sqrt a+\sqrt b$ for $a,b\ge0$, and the last follows from \eqref{eq:conditional-leading-variance-lower-bound}.

By \eqref{eq:varianceest-conditional-c3} and \eqref{eq:lemma-gradient-projections}, we have
\[
\tr(\Sigma_i)=\E[\|\zeta_{1,j}(X_i)\|^2\mid X_i]=c_2(X_i),
\qquad \E\tr(\Sigma_i)=c_2.
\]
It follows that
\begin{align}
\E\tr(\Sigma_i^2)
\ge\frac1d\{\E\tr(\Sigma_i)\}^2=\frac{c_2^2}{d}
\ge\frac{\underline c_2^{\,2}}{d_0}=:m>0,
\label{eq:exp-sigma-bound}
\end{align}
where $d$ is the dimension of $\theta$, and the first inequality holds for any symmetric random matrix. The last inequality uses $d\le d_0$ from Assumption~\ref{assump:upper} and $c_2\ge\underline c_2>0$ from Assumption~\ref{assump:lower}.

For any $\nu>0$, by \eqref{eq:conditional-variance-lower-bound}, we have
\begin{align}
\Pr(\sigma_{b}<\nu)
&\le\Pr\Bigl\{\frac2{b}\sum_{i=1}^{b}\tr(\Sigma_i^2)<4\nu^2\Bigr\}+\Pr\{K(\sqrt{c_{1,{b}}}+{g}^{-1/2})>\nu\}.
\label{eq:conditional-normal-event-split}
\end{align}
For the first term on the RHS of \eqref{eq:conditional-normal-event-split}, let $\nu>0$ be fixed such that $2\nu^2\le m/2$. By Assumption~\ref{assump:upper}, there exists a constant $L>0$ such that $\|\zeta_{1,j}(X_i)\|\le L$ almost surely. Hence,
\begin{equation}
0\le\tr(\Sigma_i^2)\le\{\tr(\Sigma_i)\}^2
=\bigl\{\E[\|\zeta_{1,j}(X_i)\|^2\mid X_i]\bigr\}^2\le L^4,
\label{eq:conditional-covariance-upper-bound}
\end{equation}
where the first inequality follows from the positive semidefiniteness of $\Sigma_i$. Hence,
\begin{align}
\Pr\Bigl\{\frac2{b}\sum_{i=1}^{b}\tr(\Sigma_i^2)<4\nu^2\Bigr\}&\le\Pr\Bigl\{\frac1{b}\sum_{i=1}^{b}\tr(\Sigma_i^2)<m/2\Bigr\}\notag\\
&\le\Pr\Bigl\{\frac1{b}\sum_{i=1}^{b}[\tr(\Sigma_i^2)-\E\tr(\Sigma_i^2)]<-m/2\Bigr\}\le e^{-c{b}},
\label{eq:conditional-trace-lower-tail}
\end{align}
where the second inequality follows from \eqref{eq:exp-sigma-bound}, and the last inequality follows from applying Bernstein's inequality to the independent variables $\tr(\Sigma_i^2)$, bounded by $L^4$, for some fixed $c>0$.

For the second term on the RHS of \eqref{eq:conditional-normal-event-split}, choose a fixed ${g}_0\ge3$ such that $K{g}_0^{-1/2}\le\nu/2$. Then, for ${g}\ge {g}_0$, we obtain
\begin{align}
\Pr\{K(\sqrt{c_{1,{b}}}+{g}^{-1/2})>\nu\}
&\le\Pr(K\sqrt{c_{1,{b}}}>\nu/2)\notag\\
&=\Pr\Bigl(c_{1,{b}}>\frac{\nu^2}{4K^2}\Bigr)\le\frac{4K^2}{\nu^2}\E c_{1,{b}}
=\frac{4K^2c_1}{\nu^2}=O(c_1),
\label{eq:conditional-prompt-variation-tail}
\end{align}
where the second inequality follows from Markov's inequality, and the second equality follows from \eqref{eq:conditional-centering}. Combining \eqref{eq:conditional-normal-event-split}, \eqref{eq:conditional-trace-lower-tail} and \eqref{eq:conditional-prompt-variation-tail}, we obtain
\begin{equation*}
\Pr(\sigma_{b}<\nu)=O(e^{-c{b}}+c_1),\qquad {g}\ge {g}_0.
\end{equation*}

Finally, by \eqref{eq:conditional-variance-definition}, \eqref{eq:conditional-leading-summand} and $\E(\xi_i\mid X_{1:{b}})=\E(\xi_i^0\mid X_{1:{b}})=0$, we have
\begin{align*}
\sigma_{b}^2
&=\frac{{g}^2}{{b}}\sum_{i=1}^{b}\E[\{(\xi_i-\xi_i^0)+\xi_i^0\}^2\mid X_{1:{b}}]\\
&\le\frac{2{g}^2}{{b}}\sum_{i=1}^{b}\E[(\xi_i-\xi_i^0)^2\mid X_{1:{b}}]
+\frac{2{g}^2}{{b}}\sum_{i=1}^{b}\Var(\xi_i^0\mid X_{1:{b}})\\
&=O(c_{1,{b}}+{g}^{-1})+O(1+{g}c_{1,{b}})=O(1+{g}c_{1,{b}}),
\end{align*}
where the $O(c_{1,{b}}+{g}^{-1})$ bound follows from \eqref{eq:conditional-summand-approximation} and the $O(1+{g}c_{1,{b}})$ bound follows from using \eqref{eq:conditional-leading-variance} with \eqref{eq:conditional-covariance-upper-bound} and
\[
\frac{{b}{g}}{({b}-1)^2}\sum_{i=1}^{b}\|{\bm g}(X_i;\theta)-{\bar{\bm g}}_{b}(\theta)\|^2
=\frac{{b}{g}}{{b}-1}c_{1,{b}}=O({g}c_{1,{b}}).
\]
This completes the proof of Lemma~\ref{lem:conditional-decomposition-variance}.
\end{proof}

\begin{proof}[Proof of Theorem~\ref{thm:varianceest}(c).]
First, $\widehat c_1$ is non-negative almost surely by definition.

\medskip
\noindent\textbf{(i) Lower tail of $\widehat c_1$.} We first show that, for some constants $v,p'>0$ independent of ${b}$ and ${g}$,
\[
\Pr\bigl\{\|{\widehat{\bm g}}_{g}(X_1;\theta)-{\widehat{\bm g}}_{g}(X_2;\theta)\|^2\ge v/(2{g})\bigr\}\ge p'.
\]
We begin by bounding the conditional fourth moment of ${\widehat{\bm g}}_{g}(X_1;\theta)-{\widehat{\bm g}}_{g}(X_2;\theta)$. Specifically, we have
\begin{align}
\E\bigl[\|{\widehat{\bm g}}_{g}(X_1;\theta)-&{\widehat{\bm g}}_{g}(X_2;\theta)\|^4\mid X_1,X_2\bigr]=\E\bigl[\|{\bm g}(X_1;\theta)-{\bm g}(X_2;\theta)
+\varepsilon_1-\varepsilon_2\|^4\mid X_1,X_2\bigr]\notag\\
&\le27\|{\bm g}(X_1;\theta)-{\bm g}(X_2;\theta)\|^4
+27\sum_{i=1}^2\E(\|\varepsilon_i\|^4\mid X_i)\notag\\
&{=27\|{\bm g}(X_1;\theta)-{\bm g}(X_2;\theta)\|^4+O({g}^{-2})},
\label{eq:paired-gradient-fourth-moment}
\end{align}
{where the inequality uses $\|u+v+w\|^4\le27(\|u\|^4+\|v\|^4+\|w\|^4)$, and the last equality follows from Lemma~\ref{lem:within-prompt-moment-bounds} with $q=2$.}

Since $c_2(X)$ is uniformly bounded and $\E c_2(X)=c_2\ge\underline c_2>0$, we can choose constants $v,p>0$, independent of ${b}$ and ${g}$, such that $\Pr\{c_2(X)\ge v\}\ge p$. On the event $\{c_2(X_1)\ge v\}$, by \eqref{eq:varianceest-prompt-second-moment-x}, we have
\begin{align}
&\E\bigl[\|{\widehat{\bm g}}_{g}(X_1;\theta)-{\widehat{\bm g}}_{g}(X_2;\theta)\|^2\mid X_1,X_2\bigr]\notag\\
&\quad=\|{\bm g}(X_1;\theta)-{\bm g}(X_2;\theta)\|^2
+\frac{c_2(X_1)+c_2(X_2)}{g}
+\frac{c_3(X_1)+c_3(X_2)}{{g}({g}-1)}\notag\\
&\quad\ge\|{\bm g}(X_1;\theta)-{\bm g}(X_2;\theta)\|^2+\frac v{g}.
\label{eq:paired-gradient-second-moment}
\end{align}
Combining \eqref{eq:paired-gradient-fourth-moment} and \eqref{eq:paired-gradient-second-moment}, we obtain, on $\{c_2(X_1)\ge v\}$,
\begin{align}
&\E\bigl[\|{\widehat{\bm g}}_{g}(X_1;\theta)-{\widehat{\bm g}}_{g}(X_2;\theta)\|^4\mid X_1,X_2\bigr]\le{C_0}\bigl\{\|{\bm g}(X_1;\theta)-{\bm g}(X_2;\theta)\|^4+v^2{g}^{-2}\bigr\}\notag\\
&\qquad\qquad\le{C_0}\Bigl\{\|{\bm g}(X_1;\theta)-{\bm g}(X_2;\theta)\|^2+\frac v{g}\Bigr\}^2\notag\\
&\qquad\qquad\le{C_0}
\Bigl\{\E\bigl[\|{\widehat{\bm g}}_{g}(X_1;\theta)-{\widehat{\bm g}}_{g}(X_2;\theta)\|^2\mid X_1,X_2\bigr]\Bigr\}^2,
\label{eq:paired-gradient-moment-comparison}
\end{align}
{where $C_0>0$ is a constant independent of ${b}$, ${g}$, $X_1$ and $X_2$.}
Applying Lemma~\ref{lem:conditional-paley-zygmund} with $\delta=1/2$ on $\{c_2(X_1)\ge v\}$, together with \eqref{eq:paired-gradient-second-moment} and \eqref{eq:paired-gradient-moment-comparison}, we further obtain
\begin{align*}
&\Pr\bigl\{\|{\widehat{\bm g}}_{g}(X_1;\theta)-{\widehat{\bm g}}_{g}(X_2;\theta)\|^2\ge v/(2{g})\mid X_1,X_2\bigr\}\\
&\quad\ge\Pr\Bigl\{\|{\widehat{\bm g}}_{g}(X_1;\theta)-{\widehat{\bm g}}_{g}(X_2;\theta)\|^2\ge\frac12\E[\|{\widehat{\bm g}}_{g}(X_1;\theta)-{\widehat{\bm g}}_{g}(X_2;\theta)\|^2\mid X_1,X_2]\,\Bigm|\,X_1,X_2\Bigr\}\\
&\quad\ge\frac{\bigl\{\E[\|{\widehat{\bm g}}_{g}(X_1;\theta)-{\widehat{\bm g}}_{g}(X_2;\theta)\|^2\mid X_1,X_2]\bigr\}^2}
{4\E[\|{\widehat{\bm g}}_{g}(X_1;\theta)-{\widehat{\bm g}}_{g}(X_2;\theta)\|^4\mid X_1,X_2]}\ge{\frac1{4C_0}}.
\end{align*}
By the law of total expectation and $\Pr\{c_2(X_1)\ge v\}\ge p$, we have
\begin{align}
&\Pr\bigl\{\|{\widehat{\bm g}}_{g}(X_1;\theta)-{\widehat{\bm g}}_{g}(X_2;\theta)\|^2\ge v/(2{g})\bigr\}\notag\\
&\quad=\E\Bigl[\Pr\bigl\{\|{\widehat{\bm g}}_{g}(X_1;\theta)-{\widehat{\bm g}}_{g}(X_2;\theta)\|^2\ge v/(2{g})\mid X_1,X_2\bigr\}\Bigr]\notag\\
&\quad\ge\E\Bigl[\mathbf 1\{c_2(X_1)\ge v\}\Pr\bigl\{\|{\widehat{\bm g}}_{g}(X_1;\theta)-{\widehat{\bm g}}_{g}(X_2;\theta)\|^2\ge v/(2{g})\mid X_1,X_2\bigr\}\Bigr]\notag\\
&\quad{\ge\frac{\Pr\{c_2(X_1)\ge v\}}{4C_0}
\ge\frac{p}{4C_0}=:p'>0.}
\label{eq:paired-gradient-small-ball}
\end{align}

To bound $\Pr(\widehat c_1<\bar{c}/{g})$, define
\[
N_{b}=\sum_{i=1}^{\lfloor {b}/2\rfloor}
\mathbf 1\bigl\{\|{\widehat{\bm g}}_{g}(X_{2i-1};\theta)-{\widehat{\bm g}}_{g}(X_{2i};\theta)\|^2\ge v/(2{g})\bigr\}.
\]
Since the $\lfloor {b}/2\rfloor$ pairs are mutually independent, $N_{b}$ is a sum of independent Bernoulli random variables. By \eqref{eq:paired-gradient-small-ball}, we have
\begin{equation}
\E N_{b}\ge p'\bigl\lfloor\frac {b}2\bigr\rfloor\ge\frac{p'{b}}{3}.
\label{eq:paired-count-mean}
\end{equation}
Moreover, grouping the first $2\lfloor {b}/2\rfloor$ summands in $\widehat c_1$ into disjoint pairs, we have
\begin{align}
\widehat c_1
&=\frac1{{b}-1}\sum_{i=1}^{b}\|{\widehat{\bm g}}_{g}(X_i;\theta)-{\widehat{\bm g}}_{{b},{g}}(\theta)\|^2\notag\\
&\ge\frac1{{b}-1}\sum_{i=1}^{\lfloor {b}/2\rfloor}
\Bigl\{\|{\widehat{\bm g}}_{g}(X_{2i-1};\theta)-{\widehat{\bm g}}_{{b},{g}}(\theta)\|^2
+\|{\widehat{\bm g}}_{g}(X_{2i};\theta)-{\widehat{\bm g}}_{{b},{g}}(\theta)\|^2\Bigr\}\notag\\
&\ge\frac1{2({b}-1)}
\sum_{i=1}^{\lfloor {b}/2\rfloor}
\|{\widehat{\bm g}}_{g}(X_{2i-1};\theta)-{\widehat{\bm g}}_{g}(X_{2i};\theta)\|^2\ge\frac{vN_{b}}{4{g}({b}-1)},
\label{eq:paired-count-lower-bound}
\end{align}
where the second inequality follows from $\|u-w\|^2+\|v-w\|^2\ge\|u-v\|^2/2$, and the last follows from $t\ge{v}/({2{g}})\mathbf 1\{t\ge v/(2{g})\}$ for $t\ge0$.

Taking $\bar{c}=vp'/24$, we obtain
\begin{align}
\Pr(\widehat c_1<\bar{c}/{g})
&\le\Pr\Bigl\{N_{b}<\frac{p'{b}}{6}\Bigr\}
\le\Pr\{N_{b}\le\E N_{b}/2\}\notag\\
&\le\exp\{-\E N_{b}/8\}
\le e^{-c{b}},
\label{eq:raw-denominator-lower-tail}
\end{align}
where $c=p'/24$, the first inequality follows from \eqref{eq:paired-count-lower-bound}, the second and last from \eqref{eq:paired-count-mean}, and the third from Lemma~\ref{lem:bernoulli-chernoff} with $\delta=1/2$. This proves the first part of Theorem~\ref{thm:varianceest}(c).

\medskip
\noindent\textbf{(ii) Lower bound for $\Pr(\widetilde c_1\le0)$.} {Recall the definitions of $c_{1,{b}}$, $\xi_i$, $\sigma_{b}$ and $Q_i$ from Lemma~\ref{lem:conditional-decomposition-variance}. We first establish a conditional normal approximation for ${g}\sqrt {b}(\widetilde c_1-c_{1,{b}})$ given $X_{1:{b}}$, using $({g}/\sqrt {b})\sum_i\xi_i$ in Lemma~\ref{lem:conditional-decomposition-variance}(a) as an intermediate approximation.} For any $h>0$, we consider separately the event $\{|{g}\sqrt {b}(\widetilde c_1-c_{1,{b}})-({g}/\sqrt {b})\sum_i\xi_i|\le h\}$ and its complement. On the event $\{\sigma_{b}\ge\nu\}$, comparing the resulting conditional probability bounds with $\Phi(t/\sigma_{b})$, we obtain
\begin{align}
&\sup_t\Bigl|\Pr\{{g}\sqrt {b}(\widetilde c_1-c_{1,{b}})\le t\mid X_{1:{b}}\}-\Phi(t/\sigma_{b})\Bigr|\notag\\
&\quad\le\underbrace{\sup_t\Bigl|\Pr\Bigl(\frac {g}{\sqrt {b}}\sum_i\xi_i\le t\Bigm|X_{1:{b}}\Bigr)-\Phi(t/\sigma_{b})\Bigr|}_{(\mathrm{I})}\notag\\
&\qquad+\underbrace{\sup_t\bigl[\Phi\{(t+h)/\sigma_{b}\}-\Phi\{(t-h)/\sigma_{b}\}\bigr]}_{(\mathrm{II})}\notag\\
&\qquad+\underbrace{\Pr\Bigl\{\Bigl|{g}\sqrt {b}(\widetilde c_1-c_{1,{b}})-\frac {g}{\sqrt {b}}\sum_i\xi_i\Bigr|>h\Bigm|X_{1:{b}}\Bigr\}}_{(\mathrm{III})},
\label{eq:conditional-normal-transfer-bound}
\end{align}
where the inequality follows by adding and subtracting $\Phi\{(t+h)/\sigma_{b}\}$ in the upper bound and $\Phi\{(t-h)/\sigma_{b}\}$ in the lower bound, and using the monotonicity of $\Phi$.

To bound $(\mathrm{I})$ by the Berry--Esseen inequality, we first bound the third absolute moments of ${g}\xi_i$. For the centred quadratic term, we have
\begin{align}
\E[|Q_i-\E(Q_i\mid X_i)|^3\mid X_i]
&\le8\E[|Q_i|^3\mid X_i]\notag\\
&=8\E\biggl[\Bigl|\|\varepsilon_i\|^2-\frac1{{g}^2}\sum_{j=1}^{g}\|\psi_j(X_i)-{\bm g}(X_i;\theta)\|^2\Bigr|^3\Bigm|X_i\biggr]\notag\\
&\le32\E[\|\varepsilon_i\|^6\mid X_i]+32\E\biggl[\Bigl\{\frac1{{g}^2}\sum_{j=1}^{g}\|\psi_j(X_i)-{\bm g}(X_i;\theta)\|^2\Bigr\}^3\Bigm|X_i\biggr]\notag\\
&\le32\E[\|\varepsilon_i\|^6\mid X_i]
+\frac{32}{{g}^4}\sum_{j=1}^{g}\E[\|\psi_j(X_i)-{\bm g}(X_i;\theta)\|^6\mid X_i]=O({g}^{-3}),
\label{eq:conditional-quadratic-third-moment}
\end{align}
where the first inequality follows from $|a-b|^3\le4(|a|^3+|b|^3)$ and Jensen's inequality, the second and third inequalities follow from $|a-b|^3\le4(|a|^3+|b|^3)$ and H\"older's inequality, and the last equality follows from {Lemma~\ref{lem:within-prompt-moment-bounds}} with $q=3$ and Assumption~\ref{assump:upper}.

Also, by the Cauchy--Schwarz inequality and {Lemma~\ref{lem:within-prompt-moment-bounds}} with $q=3$, we have
\begin{equation}
\E[\|\varepsilon_i\|^3\mid X_i]
\le\{\E[\|\varepsilon_i\|^6\mid X_i]\}^{1/2}=O({g}^{-3/2}).
\label{eq:conditional-error-third-moment}
\end{equation}
By \eqref{eq:conditional-first-projection-explicit}, we obtain
\begin{align}
\E[|{g}\xi_i|^3\mid X_{1:{b}}]&\le9\Bigl(\frac{2{b}{g}}{{b}-1}\Bigr)^3
\|{\bm g}(X_i;\theta)-{\bar{\bm g}}_{b}(\theta)\|^3\E[\|\varepsilon_i\|^3\mid X_i]\notag\\
&\quad\quad+9\Bigl(\frac{2{g}}{{g}^2-2{g}+2}\Bigr)^3
\|{\bm g}(X_i;\theta)\|^3\E[\|\varepsilon_i\|^3\mid X_i]\notag\\
&\quad\quad+9\Bigl(\frac{{g}^2({g}-1)}{{g}^2-2{g}+2}\Bigr)^3
\E[|Q_i-\E(Q_i\mid X_i)|^3\mid X_i]\notag\\
&=O\bigl({g}^3\|{\bm g}(X_i;\theta)-{\bar{\bm g}}_{b}(\theta)\|^3{g}^{-3/2}\bigr)
+O({g}^{-3}{g}^{-3/2})+O({g}^3{g}^{-3})\notag\\
&=O\{1+{g}^{3/2}\|{\bm g}(X_i;\theta)-{\bar{\bm g}}_{b}(\theta)\|^3\},
\label{eq:conditional-summand-moments}
\end{align}
where the inequality follows from $|a+b+c|^3\le9(|a|^3+|b|^3+|c|^3)$ and $|u^\top v|\le\|u\|\|v\|$. In the first equality, we bound the first two terms and the third term using \eqref{eq:conditional-error-third-moment} and \eqref{eq:conditional-quadratic-third-moment}, respectively, together with the boundedness of ${\bm g}(X_i;\theta)$ under Assumption~\ref{assump:upper}.

By {Lemma~\ref{lem:conditional-decomposition-variance}(a)}, the conditionally independent, centred variables ${g}\xi_i$ satisfy
\[
\sum_{i=1}^{b}\Var({g}\xi_i\mid X_{1:{b}})={b}\sigma_{b}^2.
\]
On the event $\{\sigma_{b}\ge\nu\}$, applying the Berry--Esseen inequality conditionally on $X_{1:{b}}$, we obtain
\begin{align}
\sup_t\Bigl|\Pr\Bigl(\frac {g}{\sqrt {b}}\sum_{i=1}^{b}\xi_i\le t\,\Bigm|\,X_{1:{b}}\Bigr)-\Phi(t/\sigma_{b})\Bigr|&\le\frac{C_{\rm BE}}{{b}^{3/2}\sigma_{b}^3}\sum_{i=1}^{b}\E[|{g}\xi_i|^3\mid X_{1:{b}}]\notag\\
&=O\Bigl\{{b}^{-1/2}+\frac{{g}^{3/2}}{{b}^{3/2}}\sum_{i=1}^{b}
\|{\bm g}(X_i;\theta)-{\bar{\bm g}}_{b}(\theta)\|^3\Bigr\}\notag\\
&=O\{{b}^{-1/2}+({g}c_{1,{b}})^{3/2}\},
\label{eq:conditional-full-sum-normal}
\end{align}
where $C_{\rm BE}$ is the Berry--Esseen constant. Here, the first equality follows from \eqref{eq:conditional-summand-moments}, and the last equality follows from
\[
\sum_{i=1}^{b}\|{\bm g}(X_i;\theta)-{\bar{\bm g}}_{b}(\theta)\|^3
\le\Bigl\{\sum_{i=1}^{b}\|{\bm g}(X_i;\theta)-{\bar{\bm g}}_{b}(\theta)\|^2\Bigr\}^{3/2}
=\{({b}-1)c_{1,{b}}\}^{3/2}.
\]

For $(\mathrm{II})$, on the event $\{\sigma_{b}\ge\nu\}$, we have
\begin{equation}
\sup_t\bigl[\Phi\{(t+h)/\sigma_{b}\}-\Phi\{(t-h)/\sigma_{b}\}\bigr]
\le\frac{2h}{\sigma_{b}}\sup_u\Phi'(u)
\le\frac{2h}{\sqrt{2\pi}\nu}.
\label{eq:conditional-normal-shift}
\end{equation}

{For $(\mathrm{III})$, by Chebyshev's inequality and Lemma~\ref{lem:conditional-decomposition-variance}(a), for any $h>0$, we have}
\begin{align}
&\Pr\Bigl\{\Bigl|{g}\sqrt {b}(\widetilde c_1-c_{1,{b}})-\frac {g}{\sqrt {b}}\sum_i\xi_i\Bigr|>h\,\Bigm|\,X_{1:{b}}\Bigr\}\notag\\
&\quad\le\frac1{h^2}\E\Bigl[\Bigl\{{g}\sqrt {b}(\widetilde c_1-c_{1,{b}})-\frac {g}{\sqrt {b}}\sum_i\xi_i\Bigr\}^2\Bigm|X_{1:{b}}\Bigr]
=O\{({b}h^2)^{-1}\}.
\label{eq:conditional-remainder-tail}
\end{align}
On the event $\{\sigma_{b}\ge\nu\}$, substituting the bounds for $(\mathrm{I})$, $(\mathrm{II})$ and $(\mathrm{III})$ in \eqref{eq:conditional-full-sum-normal}, \eqref{eq:conditional-normal-shift} and \eqref{eq:conditional-remainder-tail}, respectively, into \eqref{eq:conditional-normal-transfer-bound}, we obtain
\begin{align*}
&\sup_t\Bigl|\Pr\{{g}\sqrt {b}(\widetilde c_1-c_{1,{b}})\le t\mid X_{1:{b}}\}-\Phi(t/\sigma_{b})\Bigr|\\
&\quad\le O\{{b}^{-1/2}+({g}c_{1,{b}})^{3/2}\}+\frac{2h}{\sqrt{2\pi}\nu}+O\{({b}h^2)^{-1}\}\\
&\quad=O\Bigl\{{b}^{-1/2}+({g}c_{1,{b}})^{3/2}+h+\frac1{{b}h^2}\Bigr\}.
\end{align*}
Taking $h={b}^{-1/3}$, on $\{\sigma_{b}\ge\nu\}$, we have
\begin{align}
&\sup_t\Bigl|\Pr\{{g}\sqrt {b}(\widetilde c_1-c_{1,{b}})\le t\mid X_{1:{b}}\}-\Phi(t/\sigma_{b})\Bigr|=O\{{b}^{-1/3}+({g}c_{1,{b}})^{3/2}\}.
\label{eq:conditional-completion-clt}
\end{align}

{We next evaluate \eqref{eq:conditional-completion-clt} at $t=-{g}\sqrt {b}\,c_{1,{b}}$ and take expectations over the prompts. On $\{\sigma_{b}\ge\nu\}$, we have}
\begin{align}
&\Bigl|\Pr(\widetilde c_1\le0\mid X_{1:{b}})-\Phi\Bigl(-\frac{{g}\sqrt {b}c_{1,{b}}}{\sigma_{b}}\Bigr)\Bigr|\le C_{\rm N}\{{b}^{-1/3}+({g}c_{1,{b}})^{3/2}\},
\label{eq:conditional-sign-normal-approximation}
\end{align}
where $C_{\rm N}\ge1$ is a constant independent of ${b}$ and ${g}$. Since $c_{1,{b}}\ge0$, on the event $\{\sigma_{b}\ge\nu\}$, we have
\[
\Phi\Bigl(-\frac{{g}\sqrt {b} c_{1,{b}}}{\nu}\Bigr)
\le\Phi\Bigl(-\frac{{g}\sqrt {b} c_{1,{b}}}{\sigma_{b}}\Bigr)\le\frac12.
\]
It follows from \eqref{eq:conditional-sign-normal-approximation} that, on $\{\sigma_{b}\ge\nu\}$,
\begin{equation}
\Pr(\widetilde c_1\le0\mid X_{1:{b}})
\ge\Phi\Bigl(-\frac{{g}\sqrt {b} c_{1,{b}}}{\nu}\Bigr)
-C_{\rm N}\{{b}^{-1/3}+({g}c_{1,{b}})^{3/2}\}.
\label{eq:conditional-sign-probability}
\end{equation}
By the law of total expectation, for ${g}\ge {g}_0$, we obtain
\begin{align}
\Pr(\widetilde c_1\le0)
&\ge\E\bigl[\mathbf1_{\{\sigma_{b}\ge\nu\}}\Pr(\widetilde c_1\le0\mid X_{1:{b}})\bigr]\notag\\
&\ge\E\Bigl[\mathbf1_{\{\sigma_{b}\ge\nu\}}
\Phi\Bigl(-\frac{{g}\sqrt {b} c_{1,{b}}}{\nu}\Bigr)\Bigr]
-C_{\rm N}\bigl[{b}^{-1/3}+\E\min\{1,({g}c_{1,{b}})^{3/2}\}\bigr]\notag\\
&\ge\E\Bigl[\mathbf1_{\{\sigma_{b}\ge\nu\}}
\Phi\Bigl(-\frac{{g}\sqrt {b} c_{1,{b}}}{\nu}\Bigr)\Bigr]
-C_{\rm N}({b}^{-1/3}+{g}c_1)\notag\\
&\ge\E\Phi\Bigl(-\frac{{g}\sqrt {b} c_{1,{b}}}{\nu}\Bigr)
-\Pr(\sigma_{b}<\nu)-C_{\rm N}({b}^{-1/3}+{g}c_1)\notag\\
&=\E\Phi\Bigl(-\frac{{g}\sqrt {b} c_{1,{b}}}{\nu}\Bigr)
-O({b}^{-1/3}+{g}c_1+c_1)\notag\\
&\ge\Phi\Bigl(-\frac{c_1{g}\sqrt {b}}{\nu}\Bigr)
-O({b}^{-1/3}+{g}c_1+c_1),
\label{eq:conditional-lower-integrated}
\end{align}
where the second inequality follows from \eqref{eq:conditional-sign-probability} and the fact that both probabilities lie in $[0,1]$, the third inequality follows from $\min\{1,x^{3/2}\}\le x$ for $x\ge0$, the fourth inequality follows from $\mathbf1_{\{\sigma_{b}\ge\nu\}}\Phi(\cdot)\ge\Phi(\cdot)-\mathbf1_{\{\sigma_{b}<\nu\}}$, and the equality follows from {Lemma~\ref{lem:conditional-decomposition-variance}(b)} and $e^{-c{b}}=O({b}^{-1/3})$. The final inequality follows from Jensen's inequality and {Lemma~\ref{lem:conditional-decomposition-variance}(a)}, since $v\mapsto\Phi(-{g}\sqrt {b}v/\nu)$ is convex for $v\ge0$.

We consider two cases for ${g}\ge {g}_0$. If $c_1{g}\sqrt {b}\le {b}^{1/6}$, then $c_1\le {g}c_1=(c_1{g}\sqrt {b})/\sqrt {b}\le {b}^{-1/3}$, and hence
\begin{equation}
\Pr(\widetilde c_1\le0)
\ge\Phi\Bigl(-\frac{c_1{g}\sqrt {b}}{\nu}\Bigr)-O({b}^{-1/3}).
\label{eq:conditional-lower-small-shift}
\end{equation}
If $c_1{g}\sqrt {b}>{b}^{1/6}$, the standard normal tail bound implies
\[
\Phi\Bigl(-\frac{c_1{g}\sqrt {b}}{\nu}\Bigr)
\le\frac{\nu^2}{c_1^2{g}^2{b}}\le\nu^2{b}^{-1/3},
\]
so that
\begin{equation}
\Pr(\widetilde c_1\le0)\ge0
\ge\Phi\Bigl(-\frac{c_1{g}\sqrt {b}}{\nu}\Bigr)-\nu^2{b}^{-1/3}.
\label{eq:conditional-lower-large-shift}
\end{equation}
For $3\le {g}<{g}_0$, since $\Phi(-c_1{g}\sqrt {b}/\nu)\le1/2\le {g}_0/(2{g})$, we have
\begin{equation}
\Pr(\widetilde c_1\le0)\ge0
\ge\Phi\Bigl(-\frac{c_1{g}\sqrt {b}}{\nu}\Bigr)-\frac{{g}_0}{2{g}}.
\label{eq:conditional-lower-small-group}
\end{equation}
Combining \eqref{eq:conditional-lower-small-shift}--\eqref{eq:conditional-lower-small-group}, we obtain
\begin{equation}
\Pr(\widetilde c_1\le0)
\ge\Phi\Bigl(-\frac{c_1{g}\sqrt {b}}{\nu}\Bigr)
-C_0({b}^{-1/3}+{g}^{-1}),
\label{eq:appendix-half-probability}
\end{equation}
where $C_0>0$ is a constant independent of ${b}$ and ${g}$, chosen sufficiently large to cover the three cases.
This establishes \eqref{eqn:debiaselowerbound} and completes the proof of Theorem~\ref{thm:varianceest}(c).
\end{proof}

\medskip
\subsection{Proof of Theorem~\ref{thm:ratioest}}
\label{app:ratio-tradeoff}

{We first restate Theorem~\ref{thm:ratioest} below, with the asymptotic expressions and bounds in Figure~\ref{fig:estasymptotic} stated explicitly.}

\begin{restatedtheoremtwo}[Detailed statement]
Suppose Assumptions~\ref{assump:upper}--\ref{assump:asym} hold. For the estimators in \eqref{eqn:biasest} and \eqref{eqn:debiasest}, the excess MSE defined in \eqref{eqn:RG} satisfies the following conclusions.
\begin{enumerate}
\renewcommand{\theenumi}{(\alph{enumi})}
\renewcommand{\labelenumi}{\theenumi}
\item\label{part:ratioest-debiased} \textit{Debiased estimator.} We have
\[
\mathcal R({\widehat {G}}_{\rm debias})\sim
\begin{cases}
\dfrac{c_1{{G}}_{\max}}2,
& c_1{g}\sqrt {b}\to0,\\[6pt]
\dfrac{\sqrt{c_3}{{\tau_1}^2}}{4c_1^{3/2}},
& {b}c_1\to\infty,\quad c_1{g}\sqrt {b}\to\infty.
\end{cases}
\]
If ${b}c_1=O(1)$ and $c_1{g}\sqrt {b}\to\infty$, then
\[
\sqrt{c_1}\lesssim\mathcal R({\widehat {G}}_{\rm debias})\lesssim c_1{{G}}_{\max}.
\]
\item\label{part:ratioest-biased} \textit{Biased estimator.} We have
\[
\mathcal R({\widehat {G}}_{\rm bias})\sim
\begin{cases}
\sqrt{c_2c_3/{g}},
& {g}c_1\to0,\\[6pt]
\dfrac{\sqrt{c_3}}{4c_1^{3/2}}
\Bigl\{{{\sigma_1}^2}+\dfrac{c_2^2}{{g}^2}\Bigr\},
& {b}c_1\to\infty,\quad {g}c_1\to\infty.
\end{cases}
\]
If ${b}c_1=O(1)$ and ${g}c_1\to\infty$, then
\[
\sqrt{c_1}\lesssim\mathcal R({\widehat {G}}_{\rm bias})\lesssim c_1\sqrt {g}.
\]
\end{enumerate}
\end{restatedtheoremtwo}

{To prove Theorem~\ref{thm:ratioest}, we first establish three auxiliary lemmas on concentration and moment bounds for the variance estimators (Lemma~\ref{lem:coefficient-concentration}), their variance scales (Lemma~\ref{lem:coefficient-variance}) and the local approximation of the excess MSE (Lemma~\ref{lem:local-excess-risk}), which will also be used in the proof of Theorem~\ref{thm:pessi}. We then present the proofs of parts~\ref{part:ratioest-debiased} and~\ref{part:ratioest-biased} in Sections~\ref{app:debiased-risk-proof} and~\ref{app:raw-risk-proof}, respectively.}

\begin{appendixlemma}[Concentration and moment bounds]
\label{lem:coefficient-concentration}
Under Assumption~\ref{assump:upper}, let ${b}\ge2$ and ${g}\ge3$. The following bounds hold uniformly in ${b}$ and ${g}$.
\begin{enumerate}
\item[(a)] There exists a constant $K>0$, independent of ${b}$ and ${g}$, such that, for every $T\in\{\widehat c_j,\widetilde c_j:j=1,3\}$ and $u\ge1$,
\begin{equation}
\Pr\Bigl\{|T-\E T|>K\Bigl[\sqrt{u\Var(T)}+\frac{u}{{b}}\Bigr]\Bigr\}\le2e^{-u}.
\label{eq:coefficient-bernstein}
\end{equation}
\item[(b)] The estimators satisfy
\begin{align*}
{\Var(\widehat c_1)}+{\Var(\widetilde c_1)}&=O\{(c_1+{g}^{-2})/{b}\},\\
\Var(\widetilde c_1-\widehat c_1)&=O({b}^{-1}{g}^{-2}),\\
\E(\widehat c_3-c_3)^2&=O({b}^{-1}+{g}^{-2}),\\
\E(\widetilde c_3-c_3)^2&=O({b}^{-1}).
\end{align*}
\item[(c)] If Assumption~\ref{assump:lower} also holds, then there exists a constant $a>0$, independent of ${b}$ and ${g}$, such that, for sufficiently large ${b}$ and ${g}$,
\[
\Pr\{\widehat c_3\notin[c_3/2,2c_3]\}\le2e^{-a{b}}.
\]
\end{enumerate}
\end{appendixlemma}

\begin{proof}[Proof of Lemma~\ref{lem:coefficient-concentration}.]
\noindent\textbf{(i) Concentration bound.}
For $T\in\{\widehat c_j,\widetilde c_j:j=1,3\}$, we prove \eqref{eq:coefficient-bernstein} by separating $T-\E T$ into the independent centred sum $\sum_{i=1}^{b}\{\E(T\mid X_i,Y_{i,1:{g}})-\E T\}$ and a cross-prompt term. By direct calculations, we have
\begin{align}
T-\E T
&=\sum_{i=1}^{b}\{\E(T\mid X_i,Y_{i,1:{g}})-\E T\}\notag\\
&\quad+\Bigl[\sum_{i\ne\ell}
\{{\widehat{\bm g}}_{g}(X_i;\theta)-{\bm g}(\theta)\}^{\top}
\{{\widehat{\bm g}}_{g}(X_\ell;\theta)-{\bm g}(\theta)\}\Bigr]\notag\\
&\qquad\times
\begin{cases}
-\{{b}({b}-1)\}^{-1},&T\in\{\widehat c_1,\widetilde c_1\},\\
0,&T\in\{\widehat c_3,\widetilde c_3\}.
\end{cases}
\label{eq:prompt-hoeffding-coefficients}
\end{align}
For $i\ne\ell$ and $1\le r\le {b}$, we have
\begin{align*}
&\E\bigl[\{{\widehat{\bm g}}_{g}(X_i;\theta)-{\bm g}(\theta)\}^{\top}
\{{\widehat{\bm g}}_{g}(X_\ell;\theta)-{\bm g}(\theta)\}\mid X_r,Y_{r,1:{g}}\bigr]\\
&\quad=
\begin{cases}
\{{\widehat{\bm g}}_{g}(X_i;\theta)-{\bm g}(\theta)\}^{\top}
\E\{{\widehat{\bm g}}_{g}(X_\ell;\theta)-{\bm g}(\theta)\},&r=i,\\[3pt]
\bigl[\E\{{\widehat{\bm g}}_{g}(X_i;\theta)-{\bm g}(\theta)\}\bigr]^{\top}
\{{\widehat{\bm g}}_{g}(X_\ell;\theta)-{\bm g}(\theta)\},&r=\ell,\\[3pt]
\bigl[\E\{{\widehat{\bm g}}_{g}(X_i;\theta)-{\bm g}(\theta)\}\bigr]^{\top}
\E\{{\widehat{\bm g}}_{g}(X_\ell;\theta)-{\bm g}(\theta)\},&r\notin\{i,\ell\}.
\end{cases}
\end{align*}
where all three expressions are zero since $\E{\widehat{\bm g}}_{g}(X_i;\theta)=\E{\widehat{\bm g}}_{g}(X_\ell;\theta)={\bm g}(\theta)$. Taking conditional expectations given $(X_r,Y_{r,1:{g}})$, we obtain
\begin{align}
&\E\Bigl[\{\E(T\mid X_r,Y_{r,1:{g}})-\E T\}
\{{\widehat{\bm g}}_{g}(X_i;\theta)-{\bm g}(\theta)\}^{\top}
\{{\widehat{\bm g}}_{g}(X_\ell;\theta)-{\bm g}(\theta)\}\Bigr]\notag\\
&\quad=\E\Bigl[\{\E(T\mid X_r,Y_{r,1:{g}})-\E T\}
\E\bigl[\{{\widehat{\bm g}}_{g}(X_i;\theta)-{\bm g}(\theta)\}^{\top}
\{{\widehat{\bm g}}_{g}(X_\ell;\theta)-{\bm g}(\theta)\}\mid X_r,Y_{r,1:{g}}\bigr]\Bigr]=0.
\label{eq:corr-ET}
\end{align}
Summing over $r$ and $i\ne\ell$ in \eqref{eq:corr-ET}, we obtain that the two terms in \eqref{eq:prompt-hoeffding-coefficients} are uncorrelated. Consequently,
\begin{align}
\sum_{i=1}^{b}\Var\{\E(T\mid X_i,Y_{i,1:{g}})\}
&=\Var\Bigl[\sum_{i=1}^{b}\{\E(T\mid X_i,Y_{i,1:{g}})-\E T\}\Bigr]\notag\\
&=\Var(T)-\Var\Bigl[T-\E T-\sum_{i=1}^{b}\{\E(T\mid X_i,Y_{i,1:{g}})-\E T\}\Bigr]\le\Var(T).
\label{eq:coefficient-projection-variance}
\end{align}
Moreover, by Assumption~\ref{assump:upper}, there exists a constant $K_0>0$ independent of ${b}$ and ${g}$ such that
\[
|\E(T\mid X_i,Y_{i,1:{g}})-\E T|\le \frac{K_0}{{b}},
\]
Therefore, for the first term in \eqref{eq:prompt-hoeffding-coefficients}, we apply Bernstein's inequality, together with \eqref{eq:coefficient-projection-variance}, we obtain
\begin{equation}
\Pr\Bigl\{
\Bigl|\sum_{i=1}^{b}\{\E(T\mid X_i,Y_{i,1:{g}})-\E T\}\Bigr|
>\sqrt{2u\Var(T)}+\frac{2K_0u}{3{b}}\Bigr\}\le2e^{-u}.
\label{eq:coefficient-projection-tail}
\end{equation}

We next establish a tail bound for the second term in \eqref{eq:prompt-hoeffding-coefficients}. By Assumption~\ref{assump:upper}, there exists a constant $K_1>0$, independent of ${b}$ and ${g}$, such that $\|{\widehat{\bm g}}_{g}(X_i;\theta)-{\bm g}(\theta)\|\le K_1$ uniformly. Since ${\widehat{\bm g}}_{{b},{g}}(\theta)-{\bm g}(\theta)={b}^{-1}\sum_{i=1}^{b}\{{\widehat{\bm g}}_{g}(X_i;\theta)-{\bm g}(\theta)\}$ and the coefficient of the cross-prompt sum in \eqref{eq:prompt-hoeffding-coefficients} has absolute value at most $\{{b}({b}-1)\}^{-1}\le 2/{b}^2$ for ${b}\ge2$, we have
\begin{align}
\Bigl|T-\E T-\sum_{i=1}^{b}\{\E(T\mid X_i,Y_{i,1:{g}})-\E T\}\Bigr|
&\le\frac2{{b}^2}\Bigl|\sum_{i\ne\ell}
\{{\widehat{\bm g}}_{g}(X_i;\theta)-{\bm g}(\theta)\}^{\top}
\{{\widehat{\bm g}}_{g}(X_\ell;\theta)-{\bm g}(\theta)\}\Bigr|\notag\\
&=2\Bigl|\|{\widehat{\bm g}}_{{b},{g}}(\theta)-{\bm g}(\theta)\|^2
-\frac1{{b}^2}\sum_{i=1}^{b}\|{\widehat{\bm g}}_{g}(X_i;\theta)-{\bm g}(\theta)\|^2\Bigr|\notag\\
&\le2\|{\widehat{\bm g}}_{{b},{g}}(\theta)-{\bm g}(\theta)\|^2+\frac{2K_1^2}{{b}}.
\label{eq:coefficient-pair-bound}
\end{align}
By \eqref{eq:coefficient-pair-bound}, it suffices to bound the upper tail of $\|{\widehat{\bm g}}_{{b},{g}}(\theta)-{\bm g}(\theta)\|$ via McDiarmid's inequality. First, since the prompt groups are independent with $\E{\widehat{\bm g}}_{g}(X_i;\theta)={\bm g}(\theta)$, the Cauchy--Schwarz inequality yields
\begin{align}
\E\|{\widehat{\bm g}}_{{b},{g}}(\theta)-{\bm g}(\theta)\|
&\le\{\E\|{\widehat{\bm g}}_{{b},{g}}(\theta)-{\bm g}(\theta)\|^2\}^{1/2}\notag\\
&=\Bigl\{\frac1{b}\E\|{\widehat{\bm g}}_{g}(X_1;\theta)-{\bm g}(\theta)\|^2\Bigr\}^{1/2}
\le\frac{K_1}{\sqrt {b}}.
\label{eq:coefficient-gradient-mean}
\end{align}
Next, $\|{\widehat{\bm g}}_{{b},{g}}(\theta)-{\bm g}(\theta)\|$ is a function of the independent prompt groups $\{(X_i,Y_{i,1:{g}})\}_{i=1}^{b}$. Since $\|{\widehat{\bm g}}_{g}(X_i;\theta)-{\bm g}(\theta)\|\le K_1$, replacing any single group $(X_i,Y_{i,1:{g}})$ changes ${\widehat{\bm g}}_{g}(X_i;\theta)$ by at most $2K_1$, so $\|{\widehat{\bm g}}_{{b},{g}}(\theta)-{\bm g}(\theta)\|$ changes by at most $2K_1/{b}$ by the reverse triangle inequality. Therefore, by applying McDiarmid's inequality, we have, for every $t>0$,
\begin{align*}
&\Pr\bigl\{\|{\widehat{\bm g}}_{{b},{g}}(\theta)-{\bm g}(\theta)\|
-\E\|{\widehat{\bm g}}_{{b},{g}}(\theta)-{\bm g}(\theta)\|>t\bigr\}\\
&\quad\le\exp\Bigl\{-\frac{2t^2}{\sum_{i=1}^{b}(2K_1/{b})^2}\Bigr\}
=\exp\Bigl(-\frac{{b}t^2}{2K_1^2}\Bigr).
\end{align*}
Taking $t=K_1\sqrt{2u/{b}}$ for $u>0$ and using \eqref{eq:coefficient-gradient-mean}, we obtain
\begin{equation}
\Pr\Bigl\{\|{\widehat{\bm g}}_{{b},{g}}(\theta)-{\bm g}(\theta)\|
>\frac{K_1(1+\sqrt{2u})}{\sqrt {b}}\Bigr\}\le e^{-u}.
\label{eq:coefficient-gradient-tail}
\end{equation}
Now let $u\ge1$. Since $(1+\sqrt{2u})^2+1\le2(1+2u)+1=3+4u\le7u$, combining \eqref{eq:coefficient-pair-bound} and \eqref{eq:coefficient-gradient-tail} yields
\begin{align}
&\Pr\Bigl\{\Bigl|T-\E T-\sum_{i=1}^{b}\{\E(T\mid X_i,Y_{i,1:{g}})-\E T\}\Bigr|>\frac{14K_1^2u}{{b}}\Bigr\}\notag\\
&\quad\le\Pr\Bigl\{2\|{\widehat{\bm g}}_{{b},{g}}(\theta)-{\bm g}(\theta)\|^2+\frac{2K_1^2}{{b}}
>\frac{2K_1^2}{{b}}\{(1+\sqrt{2u})^2+1\}\Bigr\}\notag\\
&\quad=\Pr\Bigl\{\|{\widehat{\bm g}}_{{b},{g}}(\theta)-{\bm g}(\theta)\|
>\frac{K_1(1+\sqrt{2u})}{\sqrt {b}}\Bigr\}\le e^{-u}.
\label{eq:coefficient-cross-prompt-tail}
\end{align}
Finally, combining \eqref{eq:coefficient-projection-tail} and \eqref{eq:coefficient-cross-prompt-tail} via the triangle inequality and the union bound gives, for every $u\ge1$,
\begin{align*}
&\Pr\Bigl\{|T-\E T|>\sqrt{2u\Var(T)}
+\Bigl(\frac{2K_0}{3}+14K_1^2\Bigr)\frac u{b}\Bigr\}\\
&\quad\le\Pr\Bigl\{\Bigl|\sum_{i=1}^{b}\{\E(T\mid X_i,Y_{i,1:{g}})-\E T\}\Bigr|
>\sqrt{2u\Var(T)}+\frac{2K_0u}{3{b}}\Bigr\}\\
&\qquad+\Pr\Bigl\{\Bigl|T-\E T-\sum_{i=1}^{b}\{\E(T\mid X_i,Y_{i,1:{g}})-\E T\}\Bigr|
>\frac{14K_1^2u}{{b}}\Bigr\}\\
&\quad\le2e^{-u}+e^{-u}=3e^{-u}.
\end{align*}
Evaluating this bound at $u+\log(3/2)$ yields a tail probability bound of $3e^{-u-\log(3/2)}=2e^{-u}$. Since $1\le u+\log(3/2)\le\{1+\log(3/2)\}u$ for all $u\ge1$, \eqref{eq:coefficient-bernstein} follows with a sufficiently large constant $K>0$ independent of ${b}$ and ${g}$.

\smallskip
\noindent\textbf{(ii) Moment bounds.}
We first prove ${\Var(\widehat c_1)}=O\{(c_1+{g}^{-2})/{b}\}$. We use a paired representation in which each pairing contains $\lfloor {b}/2\rfloor$ independent terms, so that its variance reduces to the variance of a single pair divided by $\lfloor {b}/2\rfloor$. By direct calculation and \eqref{eq:varianceest-c1-definition}, we have
\begin{align*}
\sum_{i<\ell}\|{\widehat{\bm g}}_{g}(X_i;\theta)-{\widehat{\bm g}}_{g}(X_\ell;\theta)\|^2
&=({b}-1)\sum_{i=1}^{b}\|{\widehat{\bm g}}_{g}(X_i;\theta)\|^2
-2\sum_{i<\ell}{\widehat{\bm g}}_{g}(X_i;\theta)^\top{\widehat{\bm g}}_{g}(X_\ell;\theta)\\
&={b}\sum_{i=1}^{b}\|{\widehat{\bm g}}_{g}(X_i;\theta)\|^2
-\Bigl\|\sum_{i=1}^{b}{\widehat{\bm g}}_{g}(X_i;\theta)\Bigr\|^2
={b}({b}-1)\widehat c_1.
\end{align*}
Let $\pi$ be a permutation of $\{1,\ldots,{b}\}$, and let $\Pi$ denote the set of all such permutations. Consequently,
\begin{align}
\widehat c_1
&=\binom {b}2^{-1}\sum_{i<\ell}
\frac12\|{\widehat{\bm g}}_{g}(X_i;\theta)-{\widehat{\bm g}}_{g}(X_\ell;\theta)\|^2\notag\\
&=\frac1{{b}!}\sum_{\pi\in\Pi}\frac1{\lfloor {b}/2\rfloor}\sum_{i=1}^{\lfloor {b}/2\rfloor}
\frac12\|{\widehat{\bm g}}_{g}(X_{\pi(2i-1)};\theta)-{\widehat{\bm g}}_{g}(X_{\pi(2i)};\theta)\|^2.
\label{eq:paired-sample-variance}
\end{align}
The second equality follows because each pairwise term appears the same number of times in the sum over $\pi\in\Pi$. By Jensen's inequality,
\begin{align}
{\Var(\widehat c_1)}
&\le\frac1{{b}!}\sum_{\pi\in\Pi}\E\Bigl|\frac1{\lfloor {b}/2\rfloor}\sum_{i=1}^{\lfloor {b}/2\rfloor}
\frac12\|{\widehat{\bm g}}_{g}(X_{\pi(2i-1)};\theta)-{\widehat{\bm g}}_{g}(X_{\pi(2i)};\theta)\|^2-\E\widehat c_1\Bigr|^2\notag\\
&=\E\biggl|\frac1{\lfloor {b}/2\rfloor}\sum_{i=1}^{\lfloor {b}/2\rfloor}
\bigl\{\frac12\|{\widehat{\bm g}}_{g}(X_{2i-1};\theta)-{\widehat{\bm g}}_{g}(X_{2i};\theta)\|^2-\E\widehat c_1\bigr\}\biggr|^2\notag\\
&=\frac{\Var\{\|{\widehat{\bm g}}_{g}(X_1;\theta)-{\widehat{\bm g}}_{g}(X_2;\theta)\|^2/2\}}{\lfloor {b}/2\rfloor},
\label{eq:biased-denominator-moments}
\end{align}
Moreover, we have
\begin{align}
\Var\Bigl\{\frac12\|{\widehat{\bm g}}_{g}(X_1;\theta)-{\widehat{\bm g}}_{g}(X_2;\theta)\|^2\Bigr\}
&\le\frac14\E\|{\widehat{\bm g}}_{g}(X_1;\theta)-{\widehat{\bm g}}_{g}(X_2;\theta)\|^4\notag\\
&\le\frac{27}{4}\E\|{\bm g}(X_1;\theta)-{\bm g}(X_2;\theta)\|^4+O({g}^{-2})
=O(c_1+{g}^{-2}),
\label{eq:biased-pair-moments}
\end{align}
where the second inequality follows by taking expectations in \eqref{eq:paired-gradient-fourth-moment}, and the last equality follows from Assumption~\ref{assump:upper} and $\E\|{\bm g}(X_1;\theta)-{\bm g}(X_2;\theta)\|^2=2c_1$. Substituting \eqref{eq:biased-pair-moments} into \eqref{eq:biased-denominator-moments}, we obtain ${\Var(\widehat c_1)}=O\{(c_1+{g}^{-2})/{b}\}$.

To obtain the corresponding variance bound for $\widetilde c_1$, we next control $\Var(\widetilde c_1-\widehat c_1)$. By definition,
\[
\widetilde c_1-\widehat c_1
=-\frac{\widehat c_1+\widetilde c_0}{{g}^2-2{g}+2}
-\frac{({g}-1)^2}{{g}({g}^2-2{g}+2)}\widehat c_2.
\]
Applying $(a+b+c)^2\le3(a^2+b^2+c^2)$ to the centred terms, we obtain
\begin{align}
\Var(\widetilde c_1-\widehat c_1)
&=\E\biggl[\Bigl\{\frac{\widehat c_1-\E\widehat c_1}{{g}^2-2{g}+2}
+\frac{\widetilde c_0-\E\widetilde c_0}{{g}^2-2{g}+2}+\frac{({g}-1)^2}{{g}({g}^2-2{g}+2)}(\widehat c_2-\E\widehat c_2)\Bigr\}^2\biggr]\notag\\
&\le\frac3{({g}^2-2{g}+2)^2}
\{{\Var(\widehat c_1)}+\Var(\widetilde c_0)\}+3\Bigl\{\frac{({g}-1)^2}{{g}({g}^2-2{g}+2)}\Bigr\}^2\Var(\widehat c_2)\notag\\
&=O({g}^{-4}{b}^{-1})+O({g}^{-2}{b}^{-1})
=O({g}^{-2}{b}^{-1}),
\label{eq:correction-error-moments}
\end{align}
where the second equality follows from \eqref{eq:varianceest-raw-variance-bound} and \eqref{eq:varianceest-debiased-mse-bound}. Combining \eqref{eq:biased-denominator-moments}--\eqref{eq:correction-error-moments}, we have
\begin{align*}
{\Var(\widehat c_1)}+{\Var(\widetilde c_1)}
&\le3{\Var(\widehat c_1)}+2\Var(\widetilde c_1-\widehat c_1)\\
&=O\{(c_1+{g}^{-2})/{b}\}.
\end{align*}
For the numerator estimators, \eqref{eq:varianceest-moment-map}, \eqref{eq:varianceest-raw-variance-bound} and \eqref{eq:varianceest-debiased-mse-bound} imply
\begin{equation*}
\E(\widehat c_3-c_3)^2=O({b}^{-1}+{g}^{-2}),\qquad
\E(\widetilde c_3-c_3)^2=O({b}^{-1}).
\end{equation*}

\smallskip
\noindent\textbf{(iii) Tail bound for $\widehat c_3$.}
To prove part~(c), we apply \eqref{eq:coefficient-bernstein} with $u=a{b}$ for a fixed $a>0$, chosen so that its deviation threshold is smaller than $c_3/4$. By Assumption~\ref{assump:lower}, there exists a fixed constant $\underline c_3>0$ such that $c_3\ge\underline c_3$. The variance bound in \eqref{eq:varianceest-c23-variance-bound} allows us to choose $a>0$ sufficiently small that
\[
K\Bigl[\sqrt{a{b}\Var(\widehat c_3)}+a\Bigr]\le\underline c_3/8<c_3/4.
\]
Moreover, \eqref{eq:varianceest-moment-map} implies $|\E\widehat c_3-c_3|\le c_3/4$ for sufficiently large ${g}$. We obtain
\begin{align*}
\Pr\{\widehat c_3\notin[c_3/2,2c_3]\}
&\le\Pr(|\widehat c_3-\E\widehat c_3|\ge c_3/4)
\le2e^{-a{b}},
\end{align*}
where the last inequality follows from \eqref{eq:coefficient-bernstein} with $u=a{b}\ge1$ for sufficiently large ${b}$. This completes the proof of Lemma~\ref{lem:coefficient-concentration}.
\end{proof}

\medskip
\begin{appendixlemma}[Variance scales]
\label{lem:coefficient-variance}
Under Assumptions~\ref{assump:upper}--\ref{assump:asym},
\begin{equation*}
{\Var(\widehat c_1)}\asymp{\Var(\widetilde c_1)}\asymp\frac{c_1+{g}^{-2}}{b}.
\end{equation*}
\end{appendixlemma}

\begin{proof}[Proof of Lemma~\ref{lem:coefficient-variance}.]
The upper bounds follow from Lemma~\ref{lem:coefficient-concentration}(b), and the lower bound for ${\Var(\widehat c_1)}$ holds by Assumption~\ref{assump:lower}. For the debiased estimator, we first derive the $({b}{g}^2)^{-1}$ lower bound. For $i\ne\ell$ and $1\le r\le {b}$, we have
\[
\E(\xi_r\varepsilon_i^\top\varepsilon_\ell\mid X_{1:{b}})=0,
\]
since $\xi_r$ depends only on the completions for prompt $r$ given $X_{1:{b}}$, and taking the conditional expectation over the completions for $i$ or $\ell$ different from $r$ yields zero. Thus, by {Lemma~\ref{lem:conditional-decomposition-variance}(a)},
\begin{align}
\E[(\widetilde c_1-c_{1,{b}})^2\mid X_{1:{b}}]
&=\frac{\sigma_{b}^2}{{b}{g}^2}
+\frac1{{b}^2({b}-1)^2}\E\Bigl[\Bigl(\sum_{i\ne\ell}\varepsilon_i^\top\varepsilon_\ell\Bigr)^2\Bigm|X_{1:{b}}\Bigr]\notag\\
&=\frac{\sigma_{b}^2}{{b}{g}^2}+O\Bigl(\frac1{{b}^2{g}^2}\Bigr).
\label{eq:conditional-variance-decomposition}
\end{align}
By the law of total variance, {Lemma~\ref{lem:conditional-decomposition-variance}(b)} and \eqref{eq:conditional-variance-decomposition}, we obtain
\begin{align*}
{\Var(\widetilde c_1)}
&\ge\E[\Var(\widetilde c_1\mid X_{1:{b}})]
\ge\frac{\E\sigma_{b}^2}{{b}{g}^2}\\
&\ge\frac{\nu^2}{{b}{g}^2}\Pr(\sigma_{b}\ge\nu)
\sim\frac{\nu^2}{{b}{g}^2},
\end{align*}
where the final relation follows from ${b},{g}\to\infty$ and $c_1\to0$ in Assumption~\ref{assump:asym}.

For the $c_1/{b}$ lower bound, Assumption~\ref{assump:lower} and Lemma~\ref{lem:coefficient-concentration}(b) imply that there exist fixed constants $\underline v,K_0>0$ such that
\[
{b}{\Var(\widehat c_1)}\ge\underline v c_1,\qquad
{b}{g}^2\Var(\widetilde c_1-\widehat c_1)\le K_0.
\]
If ${g}^2c_1\ge4K_0/\underline v$, the reverse triangle inequality implies
\begin{align*}
{\sqrt{\Var(\widetilde c_1)}}
&\ge{\sqrt{\Var(\widehat c_1)}}-\sqrt{\Var(\widetilde c_1-\widehat c_1)}\\
&\ge\sqrt{\frac{c_1}{{b}}}\Bigl(\sqrt{\underline v}-\sqrt{\frac{K_0}{{g}^2c_1}}\Bigr)
\ge\frac{\sqrt{\underline v}}2\sqrt{\frac{c_1}{{b}}}.
\end{align*}
If ${g}^2c_1<4K_0/\underline v$, then
\[
\frac{c_1+{g}^{-2}}{b}\le\Bigl(1+\frac{4K_0}{\underline v}\Bigr)\frac1{{b}{g}^2},
\]
so the $({b}{g}^2)^{-1}$ lower bound suffices. Together with Lemma~\ref{lem:coefficient-concentration}(b), these bounds complete the proof of Lemma~\ref{lem:coefficient-variance}.
\end{proof}

We next state a local risk approximation that applies to all three group size estimators.

\begin{appendixlemma}
\label{lem:local-excess-risk}
For $c_1,c_3>0$, let $a_1,a_3>0$ satisfy
\begin{equation}
\Bigl|\frac{a_1}{c_1}-1\Bigr|+\Bigl|\frac{a_3}{c_3}-1\Bigr|\le\epsilon,
\qquad 0<\epsilon\le\frac12.
\label{eq:local-risk-relative-error}
\end{equation}
Then, uniformly over such $a_1,a_3$,
\begin{equation}
\ell\!\Bigl(1+\sqrt{\frac{a_3}{a_1}}\Bigr)-\ell({{G}}^*)
=\frac{\sqrt{c_3}}{4c_1^{3/2}}
\Bigl\{a_1-c_1-\frac{c_1}{c_3}(a_3-c_3)\Bigr\}^2\{1+O(\epsilon)\}.
\label{eq:local-excess-risk}
\end{equation}
\end{appendixlemma}

\begin{proof}[Proof of Lemma~\ref{lem:local-excess-risk}.]
We first rewrite the excess loss as
\begin{align}
\ell\!\Bigl(1+\sqrt{\frac{a_3}{a_1}}\Bigr)-\ell({{G}}^*)
&=c_1\sqrt{\frac{a_3}{a_1}}+c_3\sqrt{\frac{a_1}{a_3}}-2\sqrt{c_1c_3}\notag\\
&=\frac{(\sqrt{c_3a_1}-\sqrt{c_1a_3})^2}{\sqrt{a_1a_3}}\notag\\
&=\frac{(c_3a_1-c_1a_3)^2}
{\sqrt{a_1a_3}(\sqrt{c_3a_1}+\sqrt{c_1a_3})^2}.
\label{eq:ratio-risk-proof}
\end{align}
By \eqref{eq:local-risk-relative-error}, we have
\[
1-\epsilon\le\frac{a_j}{c_j}\le1+\epsilon,\qquad j=1,3.
\]
We therefore bound the denominator in \eqref{eq:ratio-risk-proof} as follows:
\begin{align}
(1-\epsilon)^2
&\le\sqrt{\frac{a_1a_3}{c_1c_3}}
\Bigl(\frac{\sqrt{a_1/c_1}+\sqrt{a_3/c_3}}2\Bigr)^2\notag\\
&=\frac{\sqrt{a_1a_3}(\sqrt{c_3a_1}+\sqrt{c_1a_3})^2}{4(c_1c_3)^{3/2}}
\le(1+\epsilon)^2.
\label{eq:local-risk-denominator-bounds}
\end{align}
By \eqref{eq:local-risk-denominator-bounds} and $0<\epsilon\le1/2$, we obtain
\begin{equation}
\Bigl|\frac{4(c_1c_3)^{3/2}}{\sqrt{a_1a_3}(\sqrt{c_3a_1}+\sqrt{c_1a_3})^2}-1\Bigr|
\le\frac{2\epsilon+\epsilon^2}{(1-\epsilon)^2}\le10\epsilon.
\label{eq:local-risk-denominator-approximation}
\end{equation}
Finally, substituting
$
c_3a_1-c_1a_3
=c_3\bigl\{a_1-c_1-\frac{c_1}{c_3}(a_3-c_3)\bigr\},
$
and \eqref{eq:local-risk-denominator-approximation} into \eqref{eq:ratio-risk-proof} yields \eqref{eq:local-excess-risk}. This completes the proof of Lemma~\ref{lem:local-excess-risk}.
\end{proof}

\subsubsection{Theorem~\ref{thm:ratioest}\ref{part:ratioest-debiased}: Debiased estimator}
\label{app:debiased-risk-proof}

\begin{proof}[Proof of Theorem~\ref{thm:ratioest}\ref{part:ratioest-debiased}.]
By \eqref{eqn:RG2}, for any group size estimator ${\widehat {G}}$ taking values in $[2,{{G}}_{\max}]$, we have
\begin{align*}
\mathcal R({\widehat {G}})
&=\E\{\ell({\widehat {G}})-\ell({{G}}^*)\}\\
&=c_1\E({\widehat {G}}-1)+c_3\E\Bigl\{\frac1{{\widehat {G}}-1}\Bigr\}-2\sqrt{c_1c_3},
\end{align*}
where, for a fixed group size ${G}\in[2,{{G}}_{\max}]$,
\begin{equation}
\ell({G})-\ell({{G}}^*)
=c_1({G}-1)+\frac{c_3}{{G}-1}-2\sqrt{c_1c_3}
=\frac{c_1({G}-{{G}}^*)^2}{{G}-1}.
\label{eq:exact-regret}
\end{equation}
It follows that
\begin{equation}
0\le\ell({G})-\ell({{G}}^*)\le c_1{G}+c_3\le c_1{{G}}_{\max}+c_3,
\qquad 2\le {G}\le {{G}}_{\max}.
\label{eq:bounded-excess-risk}
\end{equation}
We examine the weak-signal, weak-identification and regular regimes in turn.

\smallskip
\noindent\textbf{(i) Weak signal: $c_1{g}\sqrt {b}\to0$.}
Splitting the risk according to the sign of $\widetilde c_1$, we obtain
\begin{align}
\mathcal R({\widehat {G}}_{\rm debias})
&=\{\ell({{G}}_{\max})-\ell({{G}}^*)\}\Pr(\widetilde c_1\le0)+\E\{[\ell({\widehat {G}}_{\rm debias})-\ell({{G}}^*)]\mathbf1_{\{\widetilde c_1>0\}}\}.
\label{eq:debiased-weak-risk-decomposition}
\end{align}
By $c_1{{G}}_{\max}\to\infty$ and $c_3\asymp1$, we have
\begin{equation}
\ell({{G}}_{\max})-\ell({{G}}^*)
=c_1({{G}}_{\max}-1)+\frac{c_3}{{{G}}_{\max}-1}-2\sqrt{c_1c_3}
\sim c_1{{G}}_{\max},
\label{eq:upper-boundary-risk}
\end{equation}
Then, by \eqref{eq:conditional-sign-normal-approximation} and an argument similar to that used to derive \eqref{eq:conditional-lower-integrated}, we obtain
\begin{align}
\Pr(\widetilde c_1\le0)
&=\E[\Pr(\widetilde c_1\le0\mid X_{1:{b}})\mathbf1_{\{\sigma_{b}\ge\nu\}}]
+\E[\Pr(\widetilde c_1\le0\mid X_{1:{b}})\mathbf1_{\{\sigma_{b}<\nu\}}]\notag\\
&\le\E\Bigl[\Phi\Bigl(-\frac{{g}\sqrt {b}c_{1,{b}}}{\sigma_{b}}\Bigr)
\mathbf1_{\{\sigma_{b}\ge\nu\}}\Bigr]
+C_{\rm N}\bigl[{b}^{-1/3}+\E\min\{1,({g}c_{1,{b}})^{3/2}\}\bigr]+\Pr(\sigma_{b}<\nu)\notag\\
&\le\E\Bigl[\Phi\Bigl(-\frac{{g}\sqrt {b}c_{1,{b}}}{\sigma_{b}}\Bigr)
\mathbf1_{\{\sigma_{b}\ge\nu\}}\Bigr]
+C_{\rm N}({b}^{-1/3}+{g}c_1)+\Pr(\sigma_{b}<\nu)\notag\\
&\le\frac12\Pr(\sigma_{b}\ge\nu)+C_{\rm N}({b}^{-1/3}+{g}c_1)+\Pr(\sigma_{b}<\nu)\notag\\
&=\frac12+O({b}^{-1/3}+{g}c_1+c_1),\qquad {g}\ge {g}_0,
\label{eq:conditional-integrated-normal}
\end{align}
where the last equality follows from {Lemma~\ref{lem:conditional-decomposition-variance}(b)}. By Assumption~\ref{assump:asym}, ${b},{g}\to\infty$, \eqref{eq:appendix-half-probability} and \eqref{eq:conditional-integrated-normal}, we have
\begin{equation}
\frac12-o(1)\le\Pr(\widetilde c_1\le0)\le\frac12+o(1).
\label{eq:debiased-sign-probability-limit}
\end{equation}
For the second term in \eqref{eq:debiased-weak-risk-decomposition}, splitting $\{\widetilde c_1>0\}$ at any $t>0$ and applying \eqref{eq:bounded-excess-risk}, we obtain
\begin{align}
\E\{[\ell({\widehat {G}}_{\rm debias})-\ell({{G}}^*)]\mathbf1_{\{\widetilde c_1>0\}}\}&=\E\{[\ell({\widehat {G}}_{\rm debias})-\ell({{G}}^*)]
(\mathbf1_{\{0<\widetilde c_1\le t\}}+\mathbf1_{\{\widetilde c_1>t\}})\}\notag\\
&\le(c_1{{G}}_{\max}+c_3)\Pr(0<\widetilde c_1\le t)
+c_1(2+\sqrt{C_0/t})+c_3,
\label{eq:positive-denominator-split}
\end{align}
where $C_0>0$ is a fixed constant satisfying $|\widetilde c_3|\le C_0$ by Assumption~\ref{assump:upper}, so that ${\widehat {G}}_{\rm debias}\le2+\sqrt{C_0/t}$ on $\{\widetilde c_1>t\}$. By {Lemma~\ref{lem:conditional-decomposition-variance}(b)} and \eqref{eq:conditional-completion-clt},
\begin{align}
\Pr(0<\widetilde c_1\le t)
&\le\E\biggl[\Bigl\{\Phi\Bigl(\frac{{g}\sqrt {b}(t-c_{1,{b}})}{\sigma_{b}}\Bigr)
-\Phi\Bigl(-\frac{{g}\sqrt {b}c_{1,{b}}}{\sigma_{b}}\Bigr)\Bigr\}
\mathbf1_{\{\sigma_{b}\ge\nu\}}\biggr]\notag\\
&\quad+O\bigl({b}^{-1/3}+\E\min\{1,({g}c_{1,{b}})^{3/2}\}\bigr)
+\Pr(\sigma_{b}<\nu)
\notag\\
&\le\frac{t{g}\sqrt {b}}{\sqrt{2\pi}\nu}
+O({b}^{-1/3}+{g}c_1+c_1),
\label{eq:positive-denominator-small-ball}
\end{align}
where the last inequality uses $\sup_u\Phi'(u)=1/\sqrt{2\pi}$.

Taking $t=({{G}}_{\max}^2{g}\sqrt {b})^{-1/2}$ and substituting \eqref{eq:positive-denominator-small-ball} into \eqref{eq:positive-denominator-split}, by Assumption~\ref{assump:asym}, we obtain
\begin{align}
\frac{\E\{[\ell({\widehat {G}}_{\rm debias})-\ell({{G}}^*)]\mathbf1_{\{\widetilde c_1>0\}}\}}{c_1{{G}}_{\max}}
&=O\Bigl(t{g}\sqrt {b}+\frac1{{{G}}_{\max}\sqrt t}
+{b}^{-1/3}+{g}c_1+\frac1{c_1{{G}}_{\max}}\Bigr)\notag\\
&=O\Bigl(\frac{{g}^{1/2}{b}^{1/4}}{{{G}}_{\max}}
+\frac{{g}^{1/4}{b}^{1/8}}{{{G}}_{\max}^{1/2}}
+{b}^{-1/3}+{g}c_1+\frac1{c_1{{G}}_{\max}}\Bigr)\longrightarrow0.
\label{eq:positive-denominator-risk}
\end{align}
By \eqref{eq:upper-boundary-risk}, \eqref{eq:debiased-sign-probability-limit} and \eqref{eq:positive-denominator-risk}, the decomposition \eqref{eq:debiased-weak-risk-decomposition} yields
\begin{equation*}
\mathcal R({\widehat {G}}_{\rm debias})
=\{\ell({{G}}_{\max})-\ell({{G}}^*)\}\Pr(\widetilde c_1\le0)+o(c_1{{G}}_{\max})
\sim\frac{c_1{{G}}_{\max}}2.
\end{equation*}

\smallskip
\noindent\textbf{(ii) Weak identification: ${b}c_1=O(1)$ and $c_1{g}\sqrt {b}\to\infty$.}
First, the upper bound follows directly from \eqref{eq:bounded-excess-risk}:
\[
\mathcal R({\widehat {G}}_{\rm debias})\le c_1{{G}}_{\max}+c_3=O(c_1{{G}}_{\max}).
\]
For the lower bound, by \eqref{eq:exact-regret}, we have
\[
\ell({\widehat {G}}_{\rm debias})-\ell({{G}}^*)
=\sqrt{c_1c_3}\Bigl\{\frac{{\widehat {G}}_{\rm debias}-1}{{{G}}^*-1}
+\frac{{{G}}^*-1}{{\widehat {G}}_{\rm debias}-1}-2\Bigr\}.
\]
Since $c_3\asymp1$, it suffices to show that $({\widehat {G}}_{\rm debias}-1)/({{G}}^*-1)$ exceeds one by a fixed amount with probability bounded away from zero. When $\widetilde c_1>0$,
\[
\frac{{\widehat {G}}_{\rm debias}-1}{{{G}}^*-1}
=\sqrt{\frac{\widetilde c_3/c_3}{\widetilde c_1/c_1}}.
\]
We therefore seek an event on which $\widetilde c_1$ underestimates $c_1$ by a fixed proportion while $\widetilde c_3$ remains close to $c_3$. In what follows,  we will obtain this underestimate from the prompt fluctuation $c_{1,{b}}-c_1$, and show that $\widetilde c_1-c_{1,{b}}$ is $o_p(c_1)$. Accordingly, we write
\[
\widetilde c_1-c_1=(c_{1,{b}}-c_1)+(\widetilde c_1-c_{1,{b}}).
\]
Fix $0<\delta<1/2$, to be chosen below, and define the event
\begin{equation}
E_{{n}}=\Bigl\{c_{1,{b}}\le(1-\delta)c_1,\ |\widetilde c_1-c_{1,{b}}|\le\frac{\delta c_1}{2},\
|\widetilde c_3/c_3-1|\le\frac\delta4\Bigr\},
\label{eq:weak-identification-event}
\end{equation}
on which
\[
\widetilde c_1\le c_{1,{b}}+|\widetilde c_1-c_{1,{b}}|
\le(1-\delta/2)c_1,\qquad
\widetilde c_3\ge(1-\delta/4)c_3.
\]
Consequently, on $E_{{n}}\cap\{\widetilde c_1>0\}$,
\[
\frac{c_3\widetilde c_1}{c_1\widetilde c_3}
\le\frac{1-\delta/2}{1-\delta/4}<1.
\]
On $E_{{n}}\cap\{\widetilde c_1\le0\}$, ${\widehat {G}}_{\rm debias}={{G}}_{\max}$. Therefore, on $E_{{n}}$, for sufficiently large ${{n}}$, we have
\[
\frac{{\widehat {G}}_{\rm debias}-1}{{{G}}^*-1}
\ge\min\Bigl\{\sqrt{\frac{1-\delta/4}{1-\delta/2}},\frac{{{G}}_{\max}-1}{{{G}}^*-1}\Bigr\}
=\sqrt{\frac{1-\delta/4}{1-\delta/2}}>1,
\]
where the equality follows from $({{G}}_{\max}-1)/({{G}}^*-1)\to\infty$. Consequently,
\begin{equation}
\ell({\widehat {G}}_{\rm debias})-\ell({{G}}^*)
\ge\sqrt{c_1c_3}\Bigl\{\sqrt{\frac{1-\delta/4}{1-\delta/2}}
+\sqrt{\frac{1-\delta/2}{1-\delta/4}}-2\Bigr\}
\gtrsim\sqrt{c_1},
\label{eq:weak-identification-risk-lower-bound}
\end{equation}
which implies $\mathcal R({\widehat {G}}_{\rm debias})\gtrsim\sqrt{c_1}\Pr(E_{{n}}),$
and it remains to bound $\Pr(E_{{n}})$ away from zero. By Markov's inequality,
\begin{align}
\Pr(E_{{n}})
&\ge\Pr\{c_{1,{b}}\le(1-\delta)c_1\}
-\Pr(|\widetilde c_1-c_{1,{b}}|>\delta c_1/2)
-\Pr(|\widetilde c_3-c_3|>\delta c_3/4)\notag\\
&\ge\Pr\{c_{1,{b}}\le(1-\delta)c_1\}
-\frac{4\E(\widetilde c_1-c_{1,{b}})^2}{\delta^2c_1^2}
-\frac{16\E(\widetilde c_3-c_3)^2}{\delta^2c_3^2},
\label{eq:weak-identification-probability-reduction}
\end{align}
where last term is $o(1)$ by Lemma~\ref{lem:coefficient-concentration}(b).

We then show that $\E(\widetilde c_1-c_{1,{b}})^2=o(c_1^2)$.
Taking expectations in \eqref{eq:conditional-variance-decomposition} and applying {Lemma~\ref{lem:conditional-decomposition-variance}(a)--(b)}, we obtain
\begin{equation}
\E(\widetilde c_1-c_{1,{b}})^2
=\frac{\E\sigma_{b}^2}{{b}{g}^2}+O({b}^{-2}{g}^{-2})
=O\Bigl(\frac{c_1}{{b}{g}}+\frac1{{b}{g}^2}\Bigr).
\label{eq:weak-identification-response-error}
\end{equation}
In particular,
\[
\frac{\E(\widetilde c_1-c_{1,{b}})^2}{c_1^2}
=O\Bigl\{\frac1{{b}{g}c_1}+\frac1{(c_1{g}\sqrt {b})^2}\Bigr\}\to0.
\]

It remains to prove that $\liminf_{{{n}}\to\infty}\Pr\{c_{1,{b}}\le(1-\delta)c_1\}>0$ for a suitable fixed $\delta>0$. By \eqref{eq:weak-identification-response-error}, we have
\[
\frac{\E(\widetilde c_1-c_{1,{b}})^2}{c_1/{b}}
=O\Bigl(\frac1{g}+\frac1{{g}^2c_1}\Bigr)\longrightarrow0,
\]
where ${g}^2c_1=(c_1{g}\sqrt {b})^2/({b}c_1)\to\infty$. Hence, by Lemma~\ref{lem:coefficient-variance} and the law of total variance, we obtain
\begin{equation}
\Var(c_{1,{b}})
={\Var(\widetilde c_1)}-\E(\widetilde c_1-c_{1,{b}})^2
\asymp\frac{c_1}{{b}}.
\label{eq:prompt-variance-scale}
\end{equation}
We next bound the third absolute moment to establish the required lower-tail probability.
For $q=2,3$, we have
\begin{align}
\E\Bigl|\frac12\|{\bm g}(X_1;\theta)-{\bm g}(X_2;\theta)\|^2-c_1\Bigr|^q&\le2^{q-1}\Bigl[\E\Bigl\{\frac12\|{\bm g}(X_1;\theta)-{\bm g}(X_2;\theta)\|^2\Bigr\}^q+c_1^q\Bigr]\notag\\
&\le C_0\Bigl\{\frac12\E\|{\bm g}(X_1;\theta)-{\bm g}(X_2;\theta)\|^2+c_1\Bigr\}
=2C_0c_1,
\label{eq:prompt-pair-centered-moments}
\end{align}
where $C_0>0$ is a constant independent of ${b}$ and ${g}$, the first inequality uses $|a-b|^q\le2^{q-1}(|a|^q+|b|^q)$, and the second follows from Assumption~\ref{assump:upper}  and $\E\|{\bm g}(X_1;\theta)-{\bm g}(X_2;\theta)\|^2=2c_1$.

We have
\begin{align}
\E|c_{1,{b}}-c_1|^3
&\le\E\Bigl|\frac1{\lfloor {b}/2\rfloor}\sum_{i=1}^{\lfloor {b}/2\rfloor}
\Bigl\{\frac12\|{\bm g}(X_{2i-1};\theta)-{\bm g}(X_{2i};\theta)\|^2-c_1\Bigr\}\Bigr|^3\notag\\
&\le\frac{C_1}{\lfloor {b}/2\rfloor^3}\bigl\{\lfloor {b}/2\rfloor c_1+(\lfloor {b}/2\rfloor c_1)^{3/2}\bigr\}=O\Bigl\{\frac{c_1}{{b}^2}+\bigl(\frac{c_1}{{b}}\bigr)^{3/2}\Bigr\}
=O(c_1/{b}^2).
\label{eq:prompt-third-moment}
\end{align}
Here, the derivation is similar to that of \eqref{eq:biased-denominator-moments}, where the first inequality follows from the paired representation and Jensen's inequality. For the second inequality, we apply Lemma~\ref{lem:conditional-vector-rosenthal} with $p=3$ and use \eqref{eq:prompt-pair-centered-moments}. The constant $C_1>0$ is independent of ${b}$ and ${g}$, and the last equality follows from ${b}c_1=O(1)$.

For any $0<\delta<1$, we have
\begin{align*}
\E|c_{1,{b}}-c_1|
&=2\E(c_1-c_{1,{b}})_+\\
&=2\E\bigl[(c_1-c_{1,{b}})_+\mathbf1_{\{c_{1,{b}}>(1-\delta)c_1\}}\bigr]
+2\E\bigl[(c_1-c_{1,{b}})_+\mathbf1_{\{c_{1,{b}}\le(1-\delta)c_1\}}\bigr]\\
&\le2\delta c_1+2c_1\Pr\{c_{1,{b}}\le(1-\delta)c_1\},
\end{align*}
which implies
\begin{align}
\Pr\{c_{1,{b}}\le(1-\delta)c_1\}
&\ge\frac{\E|c_{1,{b}}-c_1|}{2c_1}-\delta\notag\\
&\ge\frac{\bigl\{\E[|c_{1,{b}}-c_1|^{1/2}|c_{1,{b}}-c_1|^{3/2}]\bigr\}^2}
{2c_1\E|c_{1,{b}}-c_1|^3}-\delta\notag\\
&=\frac{\{\Var(c_{1,{b}})\}^2}{2c_1\E|c_{1,{b}}-c_1|^3}-\delta
\ge a-\delta,
\label{eq:weak-identification-deficit}
\end{align}
where $a>0$ is a fixed constant, the second inequality is the Cauchy--Schwarz inequality, and the last follows from \eqref{eq:prompt-variance-scale} and \eqref{eq:prompt-third-moment}. Choosing $0<\delta<\min(1/2,a/2)$ and setting $p=a/2$, we obtain $\Pr\{c_{1,{b}}\le(1-\delta)c_1\}\ge p>0$.

Substituting \eqref{eq:weak-identification-response-error} and \eqref{eq:weak-identification-deficit} into \eqref{eq:weak-identification-probability-reduction}, we obtain $\Pr(E_{{n}})\ge p-o(1)$. Hence \eqref{eq:weak-identification-risk-lower-bound} implies
\[
\mathcal R({\widehat {G}}_{\rm debias})\gtrsim\sqrt{c_1}\{p-o(1)\}.
\]

\smallskip
\noindent\textbf{(iii) Regular regime: $c_1{g}\sqrt {b}\to\infty$ and ${b}c_1\to\infty$.}
For a sequence $\epsilon_{{n}}\to0$, define the event 
\begin{equation*}
E_{{n}}=\{|\widetilde c_1/c_1-1|+|\widetilde c_3/c_3-1|\le\epsilon_{{n}}\},
\end{equation*}
on which $\widetilde c_1$ and $\widetilde c_3$ are relatively close to $c_1$ and $c_3$, allowing us to apply the local risk approximation in Lemma~\ref{lem:local-excess-risk}. In particular,
\[
\sqrt{\frac{1-\epsilon_{{n}}}{1+\epsilon_{{n}}}}
\le\frac{\sqrt{\widetilde c_3/\widetilde c_1}}{\sqrt{c_3/c_1}}
\le\sqrt{\frac{1+\epsilon_{{n}}}{1-\epsilon_{{n}}}},
\qquad
1\ll\sqrt{c_3/c_1}\ll {{G}}_{\max}.
\]
Thus, for sufficiently large ${{n}}$, ${\widehat {G}}_{\rm debias}\in(2,{{G}}_{\max})$ on $E_{{n}}$. By Lemma~\ref{lem:local-excess-risk} with $(a_1,a_3)=(\widetilde c_1,\widetilde c_3)$, we obtain
\begin{align}
\mathcal R({\widehat {G}}_{\rm debias})
&=\frac{\sqrt{c_3}}{4c_1^{3/2}}
\E\biggl[\Bigl\{\widetilde c_1-c_1-\frac{c_1}{c_3}(\widetilde c_3-c_3)\Bigr\}^2\mathbf1_{E_{{n}}}\biggr]
\{1+O(\epsilon_{{n}})\}\notag\\
&\quad+\E\{[\ell({\widehat {G}}_{\rm debias})-\ell({{G}}^*)]\mathbf1_{E_{{n}}^c}\}.
\label{eq:debiased-regular-risk-decomposition}
\end{align}
We first upper bound the probability
\begin{align}
\Pr(E_{{n}}^c)
&=\Pr\Bigl\{\frac{|\widetilde c_1-c_1|}{c_1}+\frac{|\widetilde c_3-c_3|}{c_3}>\epsilon_{{n}}\Bigr\}\notag\\
&\le\Pr(|\widetilde c_1-c_1|>\epsilon_{{n}}c_1/2)
+\Pr(|\widetilde c_3-c_3|>\epsilon_{{n}}c_3/2).
\label{eq:debiased-localization-split}
\end{align}
To apply Lemma~\ref{lem:coefficient-concentration}(a) to both terms in \eqref{eq:debiased-localization-split}, it suffices to choose $\epsilon_{{n}}$ and $u\ge1$ such that
\begin{equation*}
K\Bigl[\sqrt{u\Var(\widetilde c_j)}+\frac u{b}\Bigr]
\le\frac{\epsilon_{{n}}c_j}{2},\qquad j=1,3,
\end{equation*}
where $K$ is the constant in Lemma~\ref{lem:coefficient-concentration}(a).
By Lemma~\ref{lem:coefficient-variance} and Assumption~\ref{assump:asym},
\begin{align*}
\frac{{\Var(\widetilde c_1)}}{c_1^2}
&\asymp\frac1{{b}c_1}+\frac1{(c_1{g}\sqrt {b})^2}\asymp {{n}}^{-(\kappa_2-\kappa_1)}+{{n}}^{-(\kappa_2+2\kappa_3-2\kappa_1)}.
\end{align*}
Since $\kappa_2-\kappa_1>0$ and $\kappa_2+2\kappa_3-2\kappa_1>0$ in this regime, choose $\epsilon_{{n}}={{n}}^{-\kappa_\epsilon}$ with
\[
0<\kappa_\epsilon<\frac12\min\{\kappa_2-\kappa_1,\ \kappa_2+2\kappa_3-2\kappa_1\}.
\]
For any fixed $0<\delta<\min\{\kappa_2-\kappa_1,\ \kappa_2+2\kappa_3-2\kappa_1\}-2\kappa_\epsilon$ and sufficiently large ${{n}}$, we have
\begin{equation}
\min\Bigl\{\frac{\epsilon_{{n}}^2c_1^2}{{\Var(\widetilde c_1)}},
{b}\epsilon_{{n}}c_1,\ {b}\epsilon_{{n}}^2\Bigr\}\ge {{n}}^\delta.
\label{eq:debiased-localization-sequence}
\end{equation}
By Lemma~\ref{lem:coefficient-concentration}(b), we have ${b}\Var(\widetilde c_3)\le C_{\rm num}$ for a fixed constant $C_{\rm num}>0$. Taking $u=k{{n}}^\delta$, we obtain from \eqref{eq:debiased-localization-sequence} that
\begin{align}
K\Bigl[\sqrt{u{\Var(\widetilde c_1)}}+\frac u{b}\Bigr]
&\le K(\sqrt k+k)\epsilon_{{n}}c_1\le\frac{\epsilon_{{n}}c_1}{2},\notag\\
K\Bigl[\sqrt{u\Var(\widetilde c_3)}+\frac u{b}\Bigr]
&\le K(\sqrt{kC_{\rm num}}+k\epsilon_{{n}})\epsilon_{{n}}\le\frac{\epsilon_{{n}}c_3}{2},
\label{eq:debiased-localization-thresholds}
\end{align}
where $k$ is chosen so that $K(\sqrt k+k)\le1/2$ and $K(\sqrt{kC_{\rm num}}+k)\le\underline c_3/2$. The last inequality in each line then follows from $\epsilon_{{n}}\le1$ and $c_3\ge\underline c_3>0$ for sufficiently large ${{n}}$.

We obtain
\begin{align}
\Pr(E_{{n}}^c)
&\le\sum_{j\in\{1,3\}}\Pr\biggl\{|\widetilde c_j-c_j|>K\Bigl[\sqrt{u\Var(\widetilde c_j)}+\frac u{b}\Bigr]\biggr\}\le2e^{-u}+2e^{-u}=4e^{-k{{n}}^\delta},
\label{eq:debiased-localization-tail}
\end{align}
where the first inequality follows from \eqref{eq:debiased-localization-split} and \eqref{eq:debiased-localization-thresholds}, and the second follows from Lemma~\ref{lem:coefficient-concentration}(a).
Combining \eqref{eq:bounded-excess-risk} and \eqref{eq:debiased-localization-tail}, we obtain
\begin{align}
\frac{\E\{[\ell({\widehat {G}}_{\rm debias})-\ell({{G}}^*)]\mathbf1_{E_{{n}}^c}\}}
{{{\tau_1}^2}/c_1^{3/2}}
&\le\frac{(c_1{{G}}_{\max}+c_3)\Pr(E_{{n}}^c)}{{{\tau_1}^2}/c_1^{3/2}}\notag\\
&\le\frac{4c_1^{3/2}(c_1{{G}}_{\max}+c_3)}{{{\tau_1}^2}}e^{-k{{n}}^\delta}\notag\\
&\lesssim {b}\sqrt{c_1}(c_1{{G}}_{\max}+c_3)e^{-k{{n}}^\delta}\notag\\
&=O\Bigl[\bigl\{{{n}}^{\kappa_2+\kappa_4-3\kappa_1/2}
+{{n}}^{\kappa_2-\kappa_1/2}\bigr\}e^{-k{{n}}^\delta}\Bigr]\to0,
\label{eq:debiased-exceptional-risk}
\end{align}
where the third inequality follows from Lemma~\ref{lem:coefficient-variance}, and the last equality follows from Assumption~\ref{assump:asym}.

We then consider the first term on the RHS of \eqref{eq:debiased-regular-risk-decomposition}. We have
\begin{align}
&\E\biggl[\Bigl\{\widetilde c_1-c_1-\frac{c_1}{c_3}(\widetilde c_3-c_3)\Bigr\}^2\mathbf1_{E_{{n}}}\biggr]\notag\\
&\quad=\E\Bigl\{\widetilde c_1-c_1-\frac{c_1}{c_3}(\widetilde c_3-c_3)\Bigr\}^2
-\E\biggl[\Bigl\{\widetilde c_1-c_1-\frac{c_1}{c_3}(\widetilde c_3-c_3)\Bigr\}^2\mathbf1_{E_{{n}}^c}\biggr]\notag\\
&\quad={\Var(\widetilde c_1)}
-\frac{2c_1}{c_3}\Cov(\widetilde c_1,\widetilde c_3)
+\frac{c_1^2}{c_3^2}\Var(\widetilde c_3)+O(e^{-k{{n}}^\delta})\notag\\
&\quad={\Var(\widetilde c_1)}
+O\Bigl\{c_1\sqrt{\frac{{\Var(\widetilde c_1)}}{b}}+\frac{c_1^2}{b}\Bigr\}
+O(e^{-k{{n}}^\delta})\notag\\
&\quad={\Var(\widetilde c_1)}\{1+O(\sqrt{c_1})\}+O(e^{-k{{n}}^\delta})\sim{\Var(\widetilde c_1)},
\label{eq:debiased-risk-moment}
\end{align}
where the second equality follows from Assumption~\ref{assump:upper} and \eqref{eq:debiased-localization-tail}, the third equality follows from the Cauchy--Schwarz inequality and $\Var(\widetilde c_3)=O({b}^{-1})$ in Lemma~\ref{lem:coefficient-concentration}(b), and the fourth follows from Lemma~\ref{lem:coefficient-variance}. The asymptotic equivalence follows since
\begin{align*}
\frac{e^{-k{{n}}^\delta}}{{{\tau_1}^2}}
&\lesssim\frac{{b}}{c_1+{g}^{-2}}e^{-k{{n}}^\delta}
\le\frac{{b}}{c_1}e^{-k{{n}}^\delta}=O\bigl({{n}}^{\kappa_1+\kappa_2}e^{-k{{n}}^\delta}\bigr)\to0.
\end{align*}
Substituting \eqref{eq:debiased-exceptional-risk} and \eqref{eq:debiased-risk-moment} into \eqref{eq:debiased-regular-risk-decomposition} and using $\epsilon_{{n}}\to0$, we obtain
\begin{equation*}
\mathcal R({\widehat {G}}_{\rm debias})
\sim\frac{\sqrt{c_3}{{\tau_1}^2}}{4c_1^{3/2}}.
\end{equation*}
This completes the proof of Theorem~\ref{thm:ratioest}\ref{part:ratioest-debiased}.
\end{proof}

\subsubsection{Theorem~\ref{thm:ratioest}\ref{part:ratioest-biased}: Biased estimator}
\label{app:raw-risk-proof}

\begin{proof}[Proof of Theorem~\ref{thm:ratioest}\ref{part:ratioest-biased}.]
By Assumptions~\ref{assump:upper}--\ref{assump:lower},
\[
\E\widehat c_1=c_1+\eta_0,\qquad
\eta_0=\frac{c_2}{{g}}\Bigl\{1+\frac{c_3}{c_2({g}-1)}\Bigr\}\sim\frac{c_2}{{g}}.
\]
First, we prove $\widehat c_3\ge\widehat c_1$. For ${b}\ge3$ and ${g}\ge2$, we have
\begin{align}
\widehat c_3
&=\frac1{b}\sum_{i=1}^{b}\biggl[
\Bigl\{\frac1{g}\sum_{j=1}^{g}\|S_\theta(X_i,Y_{i,j})\|^2\Bigr\}
\Bigl\{\frac1{{g}-1}\sum_{j=1}^{g}(Z_{i,j}-\bar Z_i)^2\Bigr\}
+\|{\widehat{\bm g}}_{g}(X_i;\theta)\|^2\biggr]\notag\\
&\ge\frac1{b}\sum_{i=1}^{b}\biggl[
\frac1{{g}({g}-1)}\Bigl\|\sum_{j=1}^{g} S_\theta(X_i,Y_{i,j})(Z_{i,j}-\bar Z_i)\Bigr\|^2
+\|{\widehat{\bm g}}_{g}(X_i;\theta)\|^2\biggr]\notag\\
&=\frac{2-{g}^{-1}}{b}\sum_{i=1}^{b}\|{\widehat{\bm g}}_{g}(X_i;\theta)\|^2
\ge\frac1{{b}-1}\sum_{i=1}^{b}\|{\widehat{\bm g}}_{g}(X_i;\theta)\|^2\notag\\
&\ge\frac1{{b}-1}\Bigl\{\sum_{i=1}^{b}\|{\widehat{\bm g}}_{g}(X_i;\theta)\|^2
-{b}\|{\widehat{\bm g}}_{{b},{g}}(\theta)\|^2\Bigr\}=\widehat c_1,
\label{eq:biased-ratio-lower-boundary}
\end{align}
where the first inequality follows from the Cauchy--Schwarz inequality, and the second inequality follows from $2-{g}^{-1}\ge3/2\ge {b}/({b}-1)$.
Thus ${\widehat {G}}_{\rm bias}\in[2,{{G}}_{\max}]$ whenever $\widehat c_1>0$. When $\widehat c_1=0$, by \eqref{eq:raw-denominator-lower-tail} and \eqref{eq:bounded-excess-risk}, we obtain the following bound for any ${\widehat {G}}$ taking values in $[2,{{G}}_{\max}]$:
\[
\E\{[\ell({\widehat {G}})-\ell({{G}}^*)]\mathbf1_{\{\widehat c_1=0\}}\}
\le(c_1{{G}}_{\max}+c_3)\Pr(\widehat c_1=0)
=O\{(c_1{{G}}_{\max}+c_3)e^{-c{b}}\}.
\]
This bound decreases faster than every power of ${{n}}^{-1}$ by Assumption~\ref{assump:asym}.

\smallskip
\noindent\textbf{(i) Large prompt and completion groups: ${b}c_1,{g}c_1\to\infty$.}
For a sequence $\epsilon_{{n}}\to0$, define
\[
E_{{n}}=\{|\widehat c_1/c_1-1|+|\widehat c_3/c_3-1|\le\epsilon_{{n}}\}.
\]
On $E_{{n}}$, ${\widehat {G}}_{\rm bias}=1+\sqrt{\widehat c_3/\widehat c_1}\in(2,{{G}}_{\max})$ for sufficiently large ${{n}}$. We use the decomposition in \eqref{eq:debiased-regular-risk-decomposition} with $(\widetilde c_1,\widetilde c_3,{\widehat {G}}_{\rm debias})$ replaced by $(\widehat c_1,\widehat c_3,{\widehat {G}}_{\rm bias})$:
\begin{align}
\mathcal R({\widehat {G}}_{\rm bias})
&=\frac{\sqrt{c_3}}{4c_1^{3/2}}
\E\biggl[\Bigl\{\widehat c_1-c_1-\frac{c_1}{c_3}(\widehat c_3-c_3)\Bigr\}^2\mathbf1_{E_{{n}}}\biggr]
\{1+O(\epsilon_{{n}})\}\notag\\
&\quad+\E\{[\ell({\widehat {G}}_{\rm bias})-\ell({{G}}^*)]\mathbf1_{E_{{n}}^c}\}.
\label{eq:biased-regular-risk-decomposition}
\end{align}
We first control the second term in \eqref{eq:biased-regular-risk-decomposition}. As in the derivation of \eqref{eq:debiased-localization-sequence}, choose $\epsilon_{{n}}={{n}}^{-\kappa_\epsilon}$, now with $0<\kappa_\epsilon<\min\{(\kappa_2-\kappa_1)/2,\ \kappa_3-\kappa_1\}$. By \eqref{eq:varianceest-moment-map},
\begin{equation}
|\E\widehat c_1-c_1|=\eta_0=o(\epsilon_{{n}}c_1),\qquad
|\E\widehat c_3-c_3|=O({g}^{-1})=o(\epsilon_{{n}}c_3).
\label{eq:biased-localization-bias}
\end{equation}
Following the derivation of \eqref{eq:debiased-localization-tail}, we obtain, for sufficiently large ${{n}}$,
\begin{align}
\Pr(E_{{n}}^c)
&\le\sum_{j\in\{1,3\}}\Pr\{|\widehat c_j-\E\widehat c_j|>\epsilon_{{n}}c_j/4\}
\le4e^{-k{{n}}^\delta},
\label{eq:biased-localization-tail}
\end{align}
where $k,\delta>0$ are fixed constants. The first inequality uses \eqref{eq:biased-localization-bias}, and the second follows from Lemma~\ref{lem:coefficient-concentration}(a).
By Lemma~\ref{lem:coefficient-variance} and $\eta_0\sim c_2/{g}$, we have
\begin{equation}
\E(\widehat c_1-c_1)^2={\Var(\widehat c_1)}+\eta_0^2
\asymp\frac{c_1}{b}+\frac1{{g}^2}.
\label{eq:biased-denominator-mse}
\end{equation}
The derivation of \eqref{eq:debiased-exceptional-risk} then applies with this mean squared error in place of ${{\tau_1}^2}$, yielding
\begin{align}
\frac{\E\{[\ell({\widehat {G}}_{\rm bias})-\ell({{G}}^*)]\mathbf1_{E_{{n}}^c}\}}
{\E(\widehat c_1-c_1)^2/c_1^{3/2}}
&\lesssim {b}\sqrt{c_1}(c_1{{G}}_{\max}+c_3)e^{-k{{n}}^\delta}\to0.
\label{eq:biased-exceptional-risk}
\end{align}

We next consider the first term in \eqref{eq:biased-regular-risk-decomposition}. By Lemma~\ref{lem:coefficient-concentration}(b) and \eqref{eq:biased-denominator-mse},
\begin{equation}
\frac{c_1^2\E(\widehat c_3-c_3)^2}{c_3^2\E(\widehat c_1-c_1)^2}
\lesssim\frac{c_1^2/{b}+c_1^2/{g}^2}{c_1/{b}+{g}^{-2}}
\le c_1+c_1^2.
\label{eq:biased-numerator-negligible}
\end{equation}
By similar derivations used in \eqref{eq:debiased-risk-moment}, we obtain
\begin{align}
&\E\biggl[\Bigl\{\widehat c_1-c_1-\frac{c_1}{c_3}(\widehat c_3-c_3)\Bigr\}^2\mathbf1_{E_{{n}}}\biggr]\notag\\
&\quad=\E(\widehat c_1-c_1)^2
-\frac{2c_1}{c_3}\E[(\widehat c_1-c_1)(\widehat c_3-c_3)]
+\frac{c_1^2}{c_3^2}\E(\widehat c_3-c_3)^2+O(e^{-k{{n}}^\delta})\notag\\
&\quad=\E(\widehat c_1-c_1)^2\{1+O(\sqrt{c_1})\}+O(e^{-k{{n}}^\delta})
\sim{\Var(\widehat c_1)}+\eta_0^2,
\label{eq:biased-risk-moment}
\end{align}
where the first equality follows from Assumption~\ref{assump:upper} and \eqref{eq:biased-localization-tail}, and the second equality follows from \eqref{eq:biased-numerator-negligible}, and the asymptotic equivalence follows from \eqref{eq:biased-denominator-mse} and Assumption~\ref{assump:asym}.
Substituting \eqref{eq:biased-exceptional-risk} and \eqref{eq:biased-risk-moment} into \eqref{eq:biased-regular-risk-decomposition} and using $\epsilon_{{n}}\to0$, we obtain
\begin{align*}
\mathcal R({\widehat {G}}_{\rm bias})
&\sim\frac{\sqrt{c_3}}{4c_1^{3/2}}\{{{\sigma_1}^2}+\eta_0^2\}\sim\frac{\sqrt{c_3}}{4c_1^{3/2}}
\Bigl\{{{\sigma_1}^2}+\frac{c_2^2}{{g}^2}\Bigr\}.
\end{align*}

\smallskip
\noindent\textbf{(ii) Small completion groups: ${g}c_1\to0$.}
By \eqref{eq:raw-denominator-lower-tail} and ${g}c_1\to0$, we have $\widehat c_1/c_1\xrightarrow{p}\infty$, so the condition in Lemma~\ref{lem:local-excess-risk} fails with probability tending to one. We therefore analyse the exact excess MSE using \eqref{eq:exact-regret}. To control ${\widehat {G}}_{\rm bias}-1$ and its reciprocal, we first consider the event on which $\widehat c_1$ is not too small and $\widehat c_3$ remains close to $c_3$. Specifically, let
\begin{equation*}
F_{{n}}=\{\widehat c_1\ge\bar c/{g},\ c_3/2\le\widehat c_3\le2c_3\},
\end{equation*}
where $\bar c>0$ is the constant in \eqref{eq:raw-denominator-lower-tail}, and Lemma~\ref{lem:coefficient-concentration}(c) then yields 
\begin{equation}
\Pr(F_{{n}}^c)=O(e^{-c{b}})
\label{eq:biased-good-event-c-prob}
\end{equation}
On the event $F_{{n}}$, by \eqref{eq:biased-ratio-lower-boundary} and Assumption~\ref{assump:asym}, we have
\[
1\le\sqrt{\widehat c_3/\widehat c_1}
\le\sqrt{2c_3{g}/\bar c}=o({{G}}_{\max}),
\]
so ${\widehat {G}}_{\rm bias}=1+\sqrt{\widehat c_3/\widehat c_1}$. 
By \eqref{eq:exact-regret}, the risk satisfies
\begin{align}
\mathcal R({\widehat {G}}_{\rm bias})
&=\E\biggl[\Bigl\{c_1\sqrt{\frac{\widehat c_3}{\widehat c_1}}
+c_3\sqrt{\frac{\widehat c_1}{\widehat c_3}}-2\sqrt{c_1c_3}\Bigr\}\mathbf1_{F_{{n}}}\biggr]+\E\{[\ell({\widehat {G}}_{\rm bias})-\ell({{G}}^*)]\mathbf1_{F_{{n}}^c}\}\notag\\
&=c_1\sqrt {g}\,\E\Bigl[\sqrt{\frac{\widehat c_3}{{g}\widehat c_1}}\mathbf1_{F_{{n}}}\Bigr]
+\frac{c_3}{\sqrt {g}}\E\Bigl[\sqrt{\frac{{g}\widehat c_1}{\widehat c_3}}\mathbf1_{F_{{n}}}\Bigr]-2\sqrt{c_1c_3}\Pr(F_{{n}})
+O\{(c_1{{G}}_{\max}+c_3)\Pr(F_{{n}}^c)\}\notag\\
&=\frac{c_3}{\sqrt {g}}\E\Bigl[\sqrt{\frac{{g}\widehat c_1}{\widehat c_3}}\mathbf1_{F_{{n}}}\Bigr]+{g}^{-1/2}O\Bigl\{{g}c_1\sqrt{\frac{2c_3}{\bar c}}+2\sqrt{{g}c_1c_3}
+\sqrt {g}(c_1{{G}}_{\max}+c_3)e^{-c{b}}\Bigr\}\notag\\
&=\frac{c_3}{\sqrt {g}}\E\Bigl[\sqrt{\frac{{g}\widehat c_1}{\widehat c_3}}\mathbf1_{F_{{n}}}\Bigr]
+o({g}^{-1/2}),
\label{eq:biased-small-group-risk-decomposition}
\end{align}
where the second equality follows from \eqref{eq:bounded-excess-risk} and the third equality follows from \eqref{eq:biased-good-event-c-prob} and the fact that $\widehat c_3/({g}\widehat c_1)\le2c_3/\bar c$ on $F_{{n}}$. Here, the last equality follows from ${g}c_1\to0$ and Assumption~\ref{assump:asym}, under which $\sqrt {g}(c_1{{G}}_{\max}+c_3)e^{-c{b}}\to0$.
It remains to show that the expectation in \eqref{eq:biased-small-group-risk-decomposition} converges to $\sqrt{c_2/c_3}$. We have
\begin{align}
\E\biggl[\Bigl|\sqrt{\frac{{g}\widehat c_1}{\widehat c_3}}
-\sqrt{\frac{c_2}{c_3}}\Bigr|\mathbf1_{F_{{n}}}\biggr]&\le\biggl\{\E\Bigl[\Bigl|
\frac{{g}\widehat c_1}{\widehat c_3}-\frac{c_2}{c_3}\Bigr|\mathbf1_{F_{{n}}}\Bigr]\Pr(F_{{n}})\biggr\}^{1/2}\notag\\
&\le\biggl\{\E\Bigl[\Bigl(\frac{|{g}\widehat c_1-c_2|}{\widehat c_3}
+\frac{c_2|\widehat c_3-c_3|}{c_3\widehat c_3}\Bigr)\mathbf1_{F_{{n}}}\Bigr]\biggr\}^{1/2}\notag\\
&\le\Bigl\{\frac{2}{c_3}\E|{g}\widehat c_1-c_2|
+\frac{2c_2}{c_3^2}\E|\widehat c_3-c_3|\Bigr\}^{1/2},
\label{eq:biased-small-group-ratio-error}
\end{align}
where the first inequality follows from $|\sqrt a-\sqrt b|\le\sqrt{|a-b|}$ and the Cauchy--Schwarz inequality, and the third inequality follows from the fact that $\widehat c_3\ge c_3/2$ on $F_{{n}}$. Since $\E|\widehat c_3-c_3|\to0$ by Lemma~\ref{lem:coefficient-concentration}(b), it suffices to prove $\E|{g}\widehat c_1-c_2|\to0$.

By direct calculation, we have
\begin{align}
\widehat c_1
&=\frac1{{b}-1}\sum_{i=1}^{b}\Bigl\|{\bm g}(X_i;\theta)-{\bar{\bm g}}_{b}(\theta)
+\varepsilon_i-\frac1{b}\sum_{\ell=1}^{b}\varepsilon_\ell\Bigr\|^2\notag\\
&=c_{1,{b}}+\frac2{{b}-1}\sum_{i=1}^{b}
\{{\bm g}(X_i;\theta)-{\bar{\bm g}}_{b}(\theta)\}^\top\varepsilon_i+\frac1{{b}-1}\sum_{i=1}^{b}
\Bigl\|\varepsilon_i-\frac1{b}\sum_{\ell=1}^{b}\varepsilon_\ell\Bigr\|^2.
\label{eq:biased-sample-variance-decomposition}
\end{align}
For the middle term in \eqref{eq:biased-sample-variance-decomposition}, by the Cauchy--Schwarz inequality,
\begin{align}
&\E\Bigl|\frac2{{b}-1}\sum_{i=1}^{b}
\{{\bm g}(X_i;\theta)-{\bar{\bm g}}_{b}(\theta)\}^\top\varepsilon_i\Bigr|
\le2\E\Bigl[\sqrt{c_{1,{b}}}\sqrt{\frac1{{b}-1}\sum_{i=1}^{b}\|\varepsilon_i\|^2}\Bigr]\notag\\
&\quad\le2\biggl\{\E c_{1,{b}}\,
\E\Bigl(\frac1{{b}-1}\sum_{i=1}^{b}\|\varepsilon_i\|^2\Bigr)\biggr\}^{1/2}
=2\sqrt{\frac{{b}c_1\eta_0}{{b}-1}}
=O(\sqrt{c_1/{g}}).
\label{eq:biased-cross-term-bound}
\end{align}
where the last equality follows from $\E\|\varepsilon_i\|^2=\eta_0$. 
For the last term in \eqref{eq:biased-sample-variance-decomposition}, we show that
\[
{g}\E\biggl|\frac1{{b}-1}\sum_{i=1}^{b}
\Bigl\|\varepsilon_i-\frac1{b}\sum_{\ell=1}^{b}\varepsilon_\ell\Bigr\|^2-\eta_0\biggr|
=O({b}^{-1/2}).
\]
By direct calculation, we have
\begin{align*}
&\E\biggl[\frac1{{b}-1}\sum_{i=1}^{b}
\Bigl\|\varepsilon_i-\frac1{b}\sum_{\ell=1}^{b}\varepsilon_\ell\Bigr\|^2\biggr]\\
&\quad=\frac1{{b}-1}\Bigl\{\sum_{i=1}^{b}\E\|\varepsilon_i\|^2
-\frac1{b}\sum_{i=1}^{b}\sum_{\ell=1}^{b}\E(\varepsilon_i^\top\varepsilon_\ell)\Bigr\}
=\frac{{b}\eta_0-\eta_0}{{b}-1}=\eta_0,
\end{align*}
Let $\pi$ be a permutation of $\{1,\ldots,{b}\}$, and let $\Pi$ denote the set of all such permutations. By derivations similar to those in \eqref{eq:paired-sample-variance} and \eqref{eq:biased-denominator-moments}, with ${\widehat{\bm g}}_{g}(X_i;\theta)$ replaced by $\varepsilon_i$, we obtain
\begin{align}
\E\biggl[\Bigl\{\frac1{{b}-1}\sum_{i=1}^{b}
\Bigl\|\varepsilon_i-\frac1{b}\sum_{\ell=1}^{b}\varepsilon_\ell\Bigr\|^2-\eta_0\Bigr\}^2\biggr]
&=\E\biggl[\Bigl\{\frac1{{b}!}\sum_{\pi\in\Pi}\Bigl(
\frac1{2\lfloor {b}/2\rfloor}\sum_{i=1}^{\lfloor {b}/2\rfloor}
\|\varepsilon_{\pi(2i-1)}-\varepsilon_{\pi(2i)}\|^2-\eta_0\Bigr)\Bigr\}^2\biggr]\notag\\
&\le\frac1{{b}!}\sum_{\pi\in\Pi}\E\biggl[\Bigl\{
\frac1{2\lfloor {b}/2\rfloor}\sum_{i=1}^{\lfloor {b}/2\rfloor}
\|\varepsilon_{\pi(2i-1)}-\varepsilon_{\pi(2i)}\|^2-\eta_0\Bigr\}^2\biggr]\notag\\
&=\E\biggl[\Bigl\{\frac1{2\lfloor {b}/2\rfloor}\sum_{i=1}^{\lfloor {b}/2\rfloor}
\|\varepsilon_{2i-1}-\varepsilon_{2i}\|^2-\eta_0\Bigr\}^2\biggr]\notag\\
&=\frac1{4\lfloor {b}/2\rfloor}\Var(\|\varepsilon_1-\varepsilon_2\|^2)
\le\frac1{4\lfloor {b}/2\rfloor}\E\|\varepsilon_1-\varepsilon_2\|^4\notag\\
&\le\frac2{\lfloor {b}/2\rfloor}\{\E\|\varepsilon_1\|^4+\E\|\varepsilon_2\|^4\}
=O\{({b}{g}^2)^{-1}\},
\label{eq:biased-noise-variance}
\end{align}
where the first inequality follows from Jensen's inequality, the last inequality uses $\|u-v\|^4\le8(\|u\|^4+\|v\|^4)$, and the last equality follows from {Lemma~\ref{lem:within-prompt-moment-bounds}} with $q=2$.
By the Cauchy--Schwarz inequality and \eqref{eq:biased-noise-variance}, we obtain
\begin{align}
&{g}\E\biggl|\frac1{{b}-1}\sum_{i=1}^{b}
\Bigl\|\varepsilon_i-\frac1{b}\sum_{\ell=1}^{b}\varepsilon_\ell\Bigr\|^2-\eta_0\biggr|\notag\\
&\quad\le {g}\biggl\{\E\biggl[\Bigl\{\frac1{{b}-1}\sum_{i=1}^{b}
\Bigl\|\varepsilon_i-\frac1{b}\sum_{\ell=1}^{b}\varepsilon_\ell\Bigr\|^2-\eta_0\Bigr\}^2\biggr]\biggr\}^{1/2}
={g}\,O\{({b}{g}^2)^{-1/2}\}=O({b}^{-1/2}).
\label{eq:biased-noise-absolute-error}
\end{align}
Combining \eqref{eq:biased-sample-variance-decomposition}, \eqref{eq:biased-cross-term-bound} and \eqref{eq:biased-noise-absolute-error}, we obtain
\begin{align}
\E|{g}\widehat c_1-c_2|
&\le {g}\E c_{1,{b}}+\frac{2{g}}{{b}-1}\E\Bigl|\sum_{i=1}^{b}
\{{\bm g}(X_i;\theta)-{\bar{\bm g}}_{b}(\theta)\}^\top\varepsilon_i\Bigr|\notag\\
&\quad+{g}\E\biggl|\frac1{{b}-1}\sum_{i=1}^{b}
\Bigl\|\varepsilon_i-\frac1{b}\sum_{\ell=1}^{b}\varepsilon_\ell\Bigr\|^2-\eta_0\biggr|
+|{g}\eta_0-c_2|\notag\\
&\le {g}\E c_{1,{b}}+O(\sqrt{{g}c_1})+O({b}^{-1/2})+|{g}\eta_0-c_2|\notag\\
&={g}c_1+O(\sqrt{{g}c_1}+{b}^{-1/2})+\frac{c_3}{{g}-1}
\longrightarrow0.
\label{eq:biased-small-group-consistency}
\end{align}

Since $\Pr(F_{{n}})=1-o(1)$ by \eqref{eq:biased-good-event-c-prob}, combining \eqref{eq:biased-small-group-ratio-error} and \eqref{eq:biased-small-group-consistency}, we have
\begin{equation*}
\E\Bigl[\sqrt{\frac{{g}\widehat c_1}{\widehat c_3}}\mathbf1_{F_{{n}}}\Bigr]
=\sqrt{\frac{c_2}{c_3}}+o(1),
\end{equation*}
which, together with \eqref{eq:biased-small-group-risk-decomposition}, implies
\begin{equation*}
\mathcal R({\widehat {G}}_{\rm bias})\sim\sqrt{\frac{c_2c_3}{{g}}}.
\end{equation*}

\smallskip
\noindent\textbf{(iii) Weak identification with large completion groups: ${g}c_1\to\infty$ and ${b}c_1=O(1)$.}
For the lower bound, define $E_{{n}}$ as in \eqref{eq:weak-identification-event}, with $(\widetilde c_1,\widetilde c_3)$ replaced by $(\widehat c_1,\widehat c_3)$ and the same choices of $\delta,p$ as in \eqref{eq:weak-identification-deficit}. Specifically,
\[
E_{{n}}=\Bigl\{c_{1,{b}}\le(1-\delta)c_1,\ |\widehat c_1-c_{1,{b}}|\le\frac{\delta c_1}{2},\
|\widehat c_3-c_3|\le\frac{\delta c_3}{4}\Bigr\}.
\]
As in \eqref{eq:weak-identification-risk-lower-bound}, we have $\ell({\widehat {G}}_{\rm bias})-\ell({{G}}^*)\gtrsim\sqrt{c_1}$ on $E_{{n}}\cap F_{{n}}$, and hence
\begin{equation}
\mathcal R({\widehat {G}}_{\rm bias})\gtrsim\sqrt{c_1}\Pr(E_{{n}}\cap F_{{n}}).
\label{eq:biased-weak-identification-risk-lower-reduction}
\end{equation}
It suffices to show that $\Pr(E_{{n}})\ge p-o(1)$. Similarly to \eqref{eq:weak-identification-probability-reduction}, we obtain
\begin{align}
\Pr(E_{{n}})
&\ge\Pr\{c_{1,{b}}\le(1-\delta)c_1\}
-\frac{4\E(\widehat c_1-c_{1,{b}})^2}{\delta^2c_1^2}
-\frac{16\E(\widehat c_3-c_3)^2}{\delta^2c_3^2}\notag\\
&\ge p-\frac{4\E(\widehat c_1-c_{1,{b}})^2}{\delta^2c_1^2}
-O({b}^{-1}+{g}^{-2}),
\label{eq:biased-weak-identification-probability}
\end{align}
where the second inequality follows from \eqref{eq:weak-identification-deficit} and Lemma~\ref{lem:coefficient-concentration}(b). We then prove $\E(\widehat c_1-c_{1,{b}})^2=o(c_1^2)$. By Lemma~\ref{lem:coefficient-concentration}(b) and \eqref{eq:weak-identification-response-error},
\begin{align}
\E(\widehat c_1-c_{1,{b}})^2
&\le2\E(\widetilde c_1-c_{1,{b}})^2+2\E(\widehat c_1-\widetilde c_1)^2\notag\\
&=O\Bigl\{\frac{c_1}{{b}{g}}+\frac1{{b}{g}^2}\Bigr\}
+2\Var(\widehat c_1-\widetilde c_1)+2\eta_0^2\notag\\
&=O\Bigl\{\frac{c_1}{{b}{g}}+\frac1{{b}{g}^2}+\frac1{{g}^2}\Bigr\}
=o(c_1^2),
\label{eq:biased-weak-identification-error}
\end{align}
Substituting \eqref{eq:biased-weak-identification-error} into \eqref{eq:biased-weak-identification-probability}, we have
$
\Pr(E_{{n}}\cap F_{{n}})\ge p-o(1),
$
which, together with \eqref{eq:biased-weak-identification-risk-lower-reduction}, proves 
\[
\mathcal R({\widehat {G}}_{\rm bias})\gtrsim\sqrt{c_1}.
\]

For the upper bound, Assumption~\ref{assump:asym} implies ${{G}}_{\max}-1>\sqrt{c_3/c_1}$ for sufficiently large ${{n}}$. On $F_{{n}}\cap\{\sqrt{\widehat c_3/\widehat c_1}>{{G}}_{\max}-1\}$, we have
\begin{align*}
\ell\!\Bigl(1+\sqrt{\frac{\widehat c_3}{\widehat c_1}}\Bigr)-\ell({{G}}_{\max})
&=\Bigl(\sqrt{\frac{\widehat c_3}{\widehat c_1}}-{{G}}_{\max}+1\Bigr)
\biggl\{c_1-\frac{c_3}{({{G}}_{\max}-1)\sqrt{\widehat c_3/\widehat c_1}}\biggr\}\ge0.
\end{align*}
On $F_{{n}}\cap\{\sqrt{\widehat c_3/\widehat c_1}\le {{G}}_{\max}-1\}$, ${\widehat {G}}_{\rm bias}=1+\sqrt{\widehat c_3/\widehat c_1}$. Hence, on $F_{{n}}$,
\begin{equation}
\ell({\widehat {G}}_{\rm bias})\le\ell\!\Bigl(1+\sqrt{\frac{\widehat c_3}{\widehat c_1}}\Bigr).
\label{eq:biased-truncated-loss-comparison}
\end{equation}
Consequently, we have
\begin{align*}
\mathcal R({\widehat {G}}_{\rm bias})
&=\E\{[\ell({\widehat {G}}_{\rm bias})-\ell({{G}}^*)]\mathbf1_{F_{{n}}}\}
+\E\{[\ell({\widehat {G}}_{\rm bias})-\ell({{G}}^*)]\mathbf1_{F_{{n}}^c}\}\\
&\le\E\Bigl[\Bigl\{c_1\sqrt{\frac{\widehat c_3}{\widehat c_1}}
+c_3\sqrt{\frac{\widehat c_1}{\widehat c_3}}-2\sqrt{c_1c_3}\Bigr\}\mathbf1_{F_{{n}}}\Bigr]
+(c_1{{G}}_{\max}+c_3)\Pr(F_{{n}}^c)\\
&\lesssim c_1\E[\widehat c_1^{-1/2}\mathbf1_{F_{{n}}}]
+\E[\sqrt{\widehat c_1}\mathbf1_{F_{{n}}}]
+(c_1{{G}}_{\max}+c_3)\Pr(F_{{n}}^c),
\end{align*}
where the first inequality follows from \eqref{eq:exact-regret}, \eqref{eq:bounded-excess-risk} and \eqref{eq:biased-truncated-loss-comparison}, and the second inequality follows from the fact that $\sqrt{\widehat c_3}\le\sqrt{2c_3}$ and $c_3/\sqrt{\widehat c_3}\le\sqrt{2c_3}$ on $F_{{n}}$.
It remains to bound the two expectations. Since $\eta_0=o(c_1)$ on this regime, we have
\[
\E\widehat c_1=c_1+\eta_0=O(c_1),\qquad
\widehat c_1^{-1/2}\mathbf1_{F_{{n}}}\le\sqrt{{g}/\bar c}.
\]
which implies
\begin{align*}
\mathcal R({\widehat {G}}_{\rm bias})
&\lesssim c_1\sqrt {g}+\sqrt{\E\widehat c_1}
+(c_1{{G}}_{\max}+c_3)\Pr(F_{{n}}^c)\\
&=O(c_1\sqrt {g}+\sqrt{c_1})=O(c_1\sqrt {g}).
\end{align*}
The last equality follows from $\sqrt{c_1}/(c_1\sqrt {g})=({g}c_1)^{-1/2}\to0$. This completes the proof of Theorem~\ref{thm:ratioest}.
\end{proof}

\medskip
\begingroup
\subsection{Proof of Theorem~\ref{thm:pessi}}
\label{app:pessimistic-risk-proof}

We first restate Theorem~\ref{thm:pessi} below with the explicit asymptotic expressions corresponding to the rightmost panel of Figure~\ref{fig:estasymptotic}. To state the result, define the denominator signal-to-noise ratio $t_{{n}}:=c_1/{{\tau_1}}$. By Lemma~\ref{lem:coefficient-variance},
\[
 t_{{n}}\asymp\min\{\sqrt{{b}c_1},\,c_1{g}\sqrt {b}\},
\]
so $t_{{n}}\to\infty$ in the regular regime (${b}c_1\to\infty$ and $c_1{g}\sqrt {b}\to\infty$), whereas $t_{{n}}=O(1)$ whenever the estimation error in $\widetilde c_1$ is at least of the same order as $c_1$ itself (covering the weak-signal and weak-identification regimes as well as the transition boundary $c_1{g}\sqrt {b}\asymp1$). Moreover, in the weak-signal regime, the leading constant depends on the fraction of the denominator variance ${{\tau_1}}^2$ attributable to within-prompt completion sampling, defined by
\[
 \rho_{{n}}^2:=\frac{\E\{\Var(\widetilde c_1\mid X_{1:{b}})\}}{{{\tau_1}}^2},
 \qquad 0\le\rho_{{n}}\le1.
\]

\begin{restatedtheoremthree}[Population standard deviation and fixed multiplier]
Suppose Assumptions~\ref{assump:upper}--\ref{assump:asym} hold. Fix $k>0$ and let ${\widehat {G}}_{\rm pes}$ be any measurable minimiser of \eqref{eqn:pessiest} on $[2,{{G}}_{\max}]$.
\begin{enumerate}
\item[(i)] If $t_{{n}}=O(1)$, the expected excess MSE satisfies
\[
\mathcal R({\widehat {G}}_{\rm pes})\asymp\sqrt{c_3{{\tau_1}}}
\asymp\sqrt{c_3}\left(\frac{c_1+{g}^{-2}}{{b}}\right)^{1/4}.
\]
If $t_{{n}}\to0$, then
\begin{equation}
\mathcal R({\widehat {G}}_{\rm pes})
=\sqrt{c_3{{\tau_1}}}\left[\E\sqrt{k+\bigl(\widetilde c_1/{{\tau_1}}\bigr)_+}+o(1)\right].
\label{eq:fixed-mean-weak-leading}
\end{equation}
In the weak-signal regime $c_1{g}\sqrt {b}\to0$, with $Z_0\sim N(0,1)$, the leading constant is given by
\begin{equation}
 \frac{\mathcal R({\widehat {G}}_{\rm pes})}{\sqrt{c_3{{\tau_1}}}}
 =\E\sqrt{k+\rho_{{n}}(Z_0)_+}+o(1).
\label{eq:fixed-mean-weak-constant}
\end{equation}
If ${g}^2c_1\to0$, then $\rho_{{n}}\to1$ and the limit is $C(k):=\E\sqrt{k+(Z_0)_+}$. If ${g}^2c_1\to\infty$, then $\rho_{{n}}\to0$ and the limit is $\sqrt k$. On the boundary ${g}^2c_1\asymp1$, any subsequence with $\rho_{{n}}\to\rho$ has leading constant $\E\sqrt{k+\rho(Z_0)_+}$.
In the interior of the weak-identification regime, where ${b}c_1\to0$ and $c_1{g}\sqrt {b}\to\infty$, \eqref{eq:fixed-mean-weak-leading} reduces to
\[
\mathcal R({\widehat {G}}_{\rm pes})\sim\sqrt{k c_3{{\tau_1}}}.
\]
The constants in the two-sided order bound may depend on $k$ and an upper bound for $t_{{n}}$.
\item[(ii)] If $t_{{n}}\to\infty$, equivalently ${b}c_1\to\infty$ and $c_1{g}\sqrt {b}\to\infty$, then
\[
\mathcal R({\widehat {G}}_{\rm pes})
\sim\frac{(1+k^2)\sqrt{c_3}\,{{\tau_1}}^2}{4c_1^{3/2}}
\asymp\frac{\sqrt{c_3}(c_1+{g}^{-2})}{{b}c_1^{3/2}}.
\]
\end{enumerate}
\end{restatedtheoremthree}

\begin{proof}[Proof of Theorem~\ref{thm:pessi}.]
For the true loss, balancing its two derivatives gives ${{G}}^*-1=\sqrt{c_3/c_1}$. We seek a comparable expression for the minimiser of $Q:=Q_k$. Its limiting slope is $c_{1,U}:=(\widetilde c_1)_++k{{\tau_1}}$, which is positive on every pilot. We will show that, for an interior choice, the same square-root formula applies with denominator $c_{1,U}$ and a numerator within $3k\tau_3$ of $\widetilde c_3$, where $\tau_3:=\{\Var(\widetilde c_3)\}^{1/2}$.

To bound the derivative of the penalty, represent the covariance matrix of $(\widetilde c_1,\widetilde c_3)$ as the Gram matrix of two vectors $\bm u,\bm v\in\mathbb R^2$. Their lengths are ${{\tau_1}}$ and $\tau_3$, and their inner product is the covariance. Hence the standard deviation in \eqref{eq:loss-variance} is the length of the vector
\[
 \bm z({G}):={G}\bm u+\frac{\bm v}{{G}-1}.
\]
At a point where this length is positive, write $\bm w:=\bm z({G})/\|\bm z({G})\|$. The derivative of the length is the component of $\bm z'({G})$ in direction $\bm w$, so
\[
 \sigma'({G})=\bm w^\top\!\left(\bm u-\frac{\bm v}{({G}-1)^2}\right).
\]
It remains to compare $\bm w$ with the unit vector $\bm u/{{\tau_1}}$. For nonzero vectors $\bm z,\bm y$, the triangle inequality gives
$\|\bm z/\|\bm z\|-\bm y/\|\bm y\|\|\le2\|\bm z-\bm y\|/\|\bm y\|$.
Taking $\bm y={G}\bm u$ and $\bm z=\bm z({G})$ yields
\begin{equation}
 |\sigma'({G})-{{\tau_1}}|
 \le {{\tau_1}}\|\bm w-\bm u/{{\tau_1}}\|+\frac{\tau_3}{({G}-1)^2}
 \le \frac{2\tau_3}{{G}({G}-1)}+\frac{\tau_3}{({G}-1)^2}
 \le \frac{3\tau_3}{({G}-1)^2}.
 \label{eq:covariance-standard-deviation-derivative}
\end{equation}
Where $\sigma({G})>0$, the objective in \eqref{eqn:pessiest} has derivative $Q'({G})=c_{1,U}-\widetilde c_3/({G}-1)^2+k\{\sigma'({G})-{{\tau_1}}\}$, so \eqref{eq:covariance-standard-deviation-derivative} implies
\begin{equation}
c_{1,U}-\frac{\widetilde c_3+3k\tau_3}{({G}-1)^2}
\le Q'({G})\le
c_{1,U}-\frac{\widetilde c_3-3k\tau_3}{({G}-1)^2}.
\label{eq:covariance-objective-derivative}
\end{equation}
The right-hand bound on $Q'$ is negative when $({G}-1)^2<(\widetilde c_3-3k\tau_3)/c_{1,U}$. The left-hand bound is positive when $({G}-1)^2>(\widetilde c_3+3k\tau_3)/c_{1,U}$. These signs determine where the minimiser can lie even if $\sigma$ vanishes at a point. Indeed, $Q$ is absolutely continuous, and a zero of $\sigma$ requires ${G}({G}-1)\bm u=-\bm v$, which has at most one solution for ${G}\ge2$ since $\bm u\ne0$.

Suppose $\widetilde c_3>3k\tau_3$ and both endpoints in the following display lie in $(2,{{G}}_{\max})$. Integrating the derivative bound shows that $Q$ decreases strictly to the left of the smaller endpoint and increases strictly to the right of the larger one. Every minimiser therefore satisfies
\begin{equation}
1+\sqrt{\frac{\widetilde c_3-3k\tau_3}{c_{1,U}}}
\le{\widehat {G}}_{\rm pes}\le
1+\sqrt{\frac{\widetilde c_3+3k\tau_3}{c_{1,U}}}.
\label{eq:covariance-minimizer-bracket}
\end{equation}
Setting $a_3:=c_{1,U}({\widehat {G}}_{\rm pes}-1)^2$, we can write ${\widehat {G}}_{\rm pes}$ on such samples in the plug-in form
\begin{equation}
{\widehat {G}}_{\rm pes}=1+\sqrt{\frac{a_3}{c_{1,U}}},\qquad
|a_3-\widetilde c_3|\le 3k\tau_3.
\label{eq:covariance-effective-numerator}
\end{equation}

Consider first a bounded sequence $t_{{n}}$. To compare the selected denominator with $c_1$, measure both in units of ${{\tau_1}}$. In these units the adjusted denominator is $d_{{n}}:=c_{1,U}/{{\tau_1}}=k+(\widetilde c_1/{{\tau_1}})_+$. It is at least $k$ on every pilot. Also $(\widetilde c_1)_+\le c_1+|\widetilde c_1-c_1|$ and $\E|\widetilde c_1-c_1|\le{{\tau_1}}$, so
\begin{equation}
 \E d_{{n}}\le t_{{n}}+1+k=O(1).
 \label{eq:fixed-mean-denominator}
\end{equation}
We next verify that the two roots in \eqref{eq:covariance-minimizer-bracket} are admissible. Let $F_{{n}}$ be the event that $|\widetilde c_j-c_j|\le c_3/8$ for $j=1,3$. Since $c_3\asymp1$ and both variances are $O({b}^{-1})$, \eqref{eq:coefficient-bernstein} with $u$ a sufficiently small constant times ${b}$ gives $\Pr(F_{{n}}^c)\le4e^{-c{b}}$. On $F_{{n}}$, for all large ${{n}}$,
\[
 k{{\tau_1}}\le c_{1,U}\le c_3/4,\qquad
 c_3/2\le\widetilde c_3-3k\tau_3\le\widetilde c_3+3k\tau_3\le2c_3.
\]
The lower root therefore exceeds $2$. The upper root is $O({{\tau_1}}^{-1/2})=O(c_1^{-1/2})$ because $t_{{n}}$ is bounded, and $c_1^{-1/2}=o({{G}}_{\max})$ by Assumption~\ref{assump:asym}. It follows from \eqref{eq:covariance-effective-numerator} that $a_3/c_3\in[1/2,2]$ on $F_{{n}}$ and
\[
 \E[(a_3/c_3-1)^2\mathbf1_{F_{{n}}}]
 \lesssim \E(\widetilde c_3-c_3)^2+\tau_3^2=O({b}^{-1}).
\]

We can now evaluate the cost of using $d_{{n}}$. To isolate the effect of the denominator, first keep the numerator at $c_3$ and choose any $d>0$. Substituting ${G}-1=\sqrt{c_3/({{\tau_1}}d)}$ in the true loss and subtracting its oracle value gives, at $t=c_1/{{\tau_1}}$,
\[
\frac{\ell({G})-\ell({{G}}^*)}{\sqrt{c_3{{\tau_1}}}}
 = f_t(d),\qquad f_t(d):=\sqrt d+\frac{t}{\sqrt d}-2\sqrt t
 =\frac{(\sqrt d-\sqrt t)^2}{\sqrt d}.
\]
On $F_{{n}}$ the selected group uses numerator $a_3$ and denominator ${{\tau_1}}d_{{n}}$. Write $r:=a_3/c_3$. Its exact normalised loss is
\[
 \sqrt{d_{{n}}}\,r^{-1/2}+\frac{t_{{n}}}{\sqrt{d_{{n}}}}\,r^{1/2}-2\sqrt{t_{{n}}}.
\]
Both $r^{1/2}$ and $r^{-1/2}$ are Lipschitz functions of $r$ on $[1/2,2]$. Replacing them by $1$ costs $O({b}^{-1/2})$ in expectation on $F_{{n}}$. Indeed, $r-1$ has restricted second moment $O({b}^{-1})$, and both coefficients multiplying these functions have bounded second moments by \eqref{eq:fixed-mean-denominator} and $d_{{n}}\ge k$.

It remains to remove the restriction to $F_{{n}}$. By \eqref{eq:bounded-excess-risk}, the expected loss on $F_{{n}}^c$ is at most $(c_1{{G}}_{\max}+c_3)\Pr(F_{{n}}^c)$. The exponential probability bound makes this $o(\sqrt{c_3{{\tau_1}}})$ under the polynomial scales in Assumption~\ref{assump:asym}. Also $\E f_{t_{{n}}}(d_{{n}})^2=O(1)$ by the same coefficient bounds, so Cauchy--Schwarz permits us to remove $\mathbf1_{F_{{n}}}$ from its expectation. Thus
\begin{equation}
 \frac{\mathcal R({\widehat {G}}_{\rm pes})}{\sqrt{c_3{{\tau_1}}}}
 =\E f_{t_{{n}}}(d_{{n}})+o(1).
 \label{eq:fixed-mult-weak-expansion}
\end{equation}

The bounded second moment gives the upper bound in part~(i). For the lower bound, we need a positive probability that the penalty moves the denominator away from the oracle. Since $d_{{n}}\ge\widetilde c_1/{{\tau_1}}+k$, Cantelli's inequality gives
\[
 \Pr(d_{{n}}\ge t_{{n}}+k/2)\ge
 \Pr\{(\widetilde c_1-c_1)/{{\tau_1}}\ge-k/2\}\ge\frac{k^2}{4+k^2}.
\]
For $d\ge t$, the function $f_t(d)$ increases with $d$. If $0\le t_{{n}}\le T$, its value on this event is at least $\min_{0\le t\le T}f_t(t+k/2)$. This minimum is positive by continuity and $k>0$. Substitution in \eqref{eq:fixed-mult-weak-expansion} proves the two-sided order, including on ${b}c_1\asymp1$ and $c_1{g}\sqrt {b}\asymp1$.

If $t_{{n}}\to0$, the bound $d_{{n}}\ge k$ gives $|f_{t_{{n}}}(d_{{n}})-\sqrt{d_{{n}}}|\le t_{{n}}/\sqrt k+2\sqrt{t_{{n}}}$. This proves \eqref{eq:fixed-mean-weak-leading}.

In the interior of weak identification, $t_{{n}}\lesssim\sqrt{{b}c_1}\to0$, and we can evaluate the remaining expectation. Conditional on the observed prompts, $\widetilde c_1$ has nonnegative mean $c_{1,{b}}$, with $\E c_{1,{b}}=c_1$ by \eqref{eq:conditional-centering}. The completion error satisfies $\E(\widetilde c_1-c_{1,{b}})^2=o(c_1^2)$ by \eqref{eq:weak-identification-response-error}. Consequently,
\[
 \frac{\E|\widetilde c_1|}{{{\tau_1}}}
 \le\frac{\E c_{1,{b}}+\{\E(\widetilde c_1-c_{1,{b}})^2\}^{1/2}}{{{\tau_1}}}
 =t_{{n}}\{1+o(1)\}\longrightarrow0.
\]
Since $0\le\sqrt{k+u}-\sqrt k\le u/(2\sqrt k)$ for $u\ge0$, the expectation of $\sqrt{d_{{n}}}$ tends to $\sqrt k$.

To obtain the weak-signal constant in \eqref{eq:fixed-mean-weak-constant}, fix the prompts $X_{1:{b}}$. The random part of $\widetilde c_1$ then comes from the completions, and its conditional mean is $c_{1,{b}}$. We first approximate the conditional law of ${g}\sqrt {b}(\widetilde c_1-c_{1,{b}})$. This calculation needs ${g}c_1\to0$, a condition that holds under weak signal and on the transition $c_1{g}\sqrt {b}\asymp1$.

Recall from \eqref{eq:conditional-variance-definition} that $\sigma_{b}^2$ is the conditional variance of the leading sum in this approximation. Its leading contribution under ${g}c_1\to0$ is the prompt average $2{b}^{-1}\sum_i\tr(\Sigma_i^2)$, where $\Sigma_i$ is the conditional covariance in \eqref{eq:conditional-quadratic-leading-moments}. To see this, \eqref{eq:conditional-summand-approximation} bounds the squared error from replacing the summands in conditional $L^2$. In the variance of the replacement sum, \eqref{eq:conditional-leading-variance} bounds the remaining quadratic form by $O({g}c_{1,{b}})$. The reverse triangle inequality for the first replacement and $(\sqrt a-\sqrt b)^2\le|a-b|$ for the second give
\[
\E\left(\sigma_{b}-\sqrt{\frac2{b}\sum_{i=1}^{b}\tr(\Sigma_i^2)}\right)^2
=O(c_1+{g}^{-1})+O({g}c_1+{g}^{-1})=o(1),
\]
because $\E c_{1,{b}}=c_1$. Write $v_{{n}}:=2\E\tr(\Sigma_1^2)$. The independent prompt contributions are bounded, so their average in the preceding display has mean $v_{{n}}$ and variance $O({b}^{-1})$. Since $v_{{n}}\asymp1$ by \eqref{eq:exp-sigma-bound} and \eqref{eq:conditional-covariance-upper-bound}, it follows that $\E(\sigma_{b}-\sqrt{v_{{n}}})^2=o(1)$ and $\E\sigma_{b}^2=v_{{n}}+o(1)$.

We may now average the conditional normal approximation over the prompts. Its error in \eqref{eq:conditional-completion-clt} tends to zero in probability on $\{\sigma_{b}\ge\nu\}$, since ${g}c_{1,{b}}\to0$ in probability by Markov's inequality. The probability of the complement tends to zero by Lemma~\ref{lem:conditional-decomposition-variance}(b). The difference between two distribution functions is bounded by $1$, so its expected error tends to zero as well. Using $\sigma_{b}/\sqrt{v_{{n}}}\to1$ in probability gives
\[
\frac{{g}\sqrt {b}(\widetilde c_1-c_{1,{b}})}{\sqrt{v_{{n}}}}\xrightarrow{d} Z_0.
\]

To express this normal approximation in units of the total standard error ${{\tau_1}}$, decompose the estimate as
\[
 \frac{\widetilde c_1}{{{\tau_1}}}
 =\frac{c_{1,{b}}}{{{\tau_1}}}
  +\frac{\sqrt{v_{{n}}}}{{g}\sqrt {b}\,{{\tau_1}}}\,
     \frac{{g}\sqrt {b}(\widetilde c_1-c_{1,{b}})}{\sqrt{v_{{n}}}}.
\]
Under weak signal the first term vanishes in $L^1$, since it is nonnegative and its mean is $t_{{n}}\to0$. For the second term, \eqref{eq:conditional-variance-decomposition} gives ${b}{g}^2\E\{\Var(\widetilde c_1\mid X_{1:{b}})\}=v_{{n}}+o(1)$. As ${b}{g}^2{{\tau_1}}^2\gtrsim1$, the coefficient of the standardised completion error is $\rho_{{n}}+o(1)$. Along a subsequence with $\rho_{{n}}\to\rho$ we therefore have $\widetilde c_1/{{\tau_1}}\xrightarrow{d}\rho Z_0$. The prompt term can disappear on this scale in probability while still contributing to its variance.

It remains to take the expectation in \eqref{eq:fixed-mean-weak-leading}. Equation~\eqref{eq:fixed-mean-denominator} bounds the second moment of $\sqrt{k+(\widetilde c_1/{{\tau_1}})_+}$, so convergence in distribution suffices for convergence of its mean. The resulting limit is $\E\sqrt{k+\rho(Z_0)_+}$. Every subsequence of $\rho_{{n}}\in[0,1]$ has a convergent subsequence, and this expectation is continuous in $\rho$. This proves the stated approximation even when $\rho_{{n}}$ itself has no limit.

If ${g}^2c_1\to0$, \eqref{eq:biased-denominator-moments} applied to the prompt gradients and \eqref{eq:prompt-pair-centered-moments} give ${b}{g}^2\Var(c_{1,{b}})=O({g}^2c_1)=o(1)$. Total variance then gives $\rho_{{n}}\to1$. If ${g}^2c_1\to\infty$, Lemma~\ref{lem:coefficient-variance} gives ${b}{g}^2{{\tau_1}}^2\asymp1+{g}^2c_1\to\infty$, while $v_{{n}}=O(1)$, so $\rho_{{n}}\to0$.

In the regular regime, $t_{{n}}\to\infty$ and the error in the denominator is small relative to $c_1$. Here we use a quadratic expansion of the loss near the oracle. Let $E_{{n}}$ be the event
\[
 |\widetilde c_1/c_1-1|+|\widetilde c_3/c_3-1|\le\epsilon_{{n}},
\]
where $\epsilon_{{n}}={{n}}^{-\kappa_\epsilon}$ is chosen as in \eqref{eq:debiased-localization-sequence}. To control the expected loss outside $E_{{n}}$, use the following consequence of \eqref{eq:coefficient-bernstein} for either coefficient $T\in\{\widetilde c_1,\widetilde c_3\}$ and each fixed $p\ge2$.
\begin{equation}
\E|T-\E T|^p\le C_p\bigl[\{\Var(T)\}^{p/2}+{b}^{-p}\bigr].
\label{eq:fixed-mean-local-moments}
\end{equation}
Indeed, integrating the tail with threshold $C\{\sqrt{u\Var(T)}+u/{b}\}$ gives \eqref{eq:fixed-mean-local-moments}. {Since both $({\tau_1}+{b}^{-1})/(\epsilon_{{n}}c_1)$ and $(\tau_3+{b}^{-1})/(\epsilon_{{n}}c_3)$ tend to zero at a polynomial rate,} Markov's inequality with sufficiently large $p$ gives $\Pr(E_{{n}}^c)=O({{n}}^{-m})$ for each fixed $m>0$.

On $E_{{n}}$, $\widetilde c_1$ is positive, so $c_{1,U}=\widetilde c_1+k{{\tau_1}}$. Both $c_{1,U}/c_1$ and $(\widetilde c_3\pm3k\tau_3)/c_3$ tend to $1$ uniformly. Since $1\ll\sqrt{c_3/c_1}\ll {{G}}_{\max}$, the roots are interior and $a_3/c_3=1+o(1)$ there.

Lemma~\ref{lem:local-excess-risk} now expresses the leading loss as the square of a difference. Its first term is the error in the adjusted denominator. Its second is the numerator error expressed in the same units. Including the negligible contribution from $E_{{n}}^c$ via \eqref{eq:bounded-excess-risk}, we obtain
\begin{align}
\mathcal R({\widehat {G}}_{\rm pes})
&=\frac{\sqrt{c_3}}{4c_1^{3/2}}
\E\biggl[\Bigl\{\widetilde c_1-c_1+k{{\tau_1}}-\frac{c_1}{c_3}(a_3-c_3)\Bigr\}^2
\mathbf1_{E_{{n}}}\biggr]\{1+o(1)\}
+o\!\left(\frac{\sqrt{c_3}{{\tau_1}}^2}{c_1^{3/2}}\right).
\label{eq:pess-regular-risk-decomposition}
\end{align}

It remains to evaluate the squared error in this expansion. On $E_{{n}}$, \eqref{eq:covariance-effective-numerator} bounds $|a_3-c_3|$ by $|\widetilde c_3-c_3|+3k\tau_3$. Since $\tau_3^2=O({b}^{-1})$ and ${{\tau_1}}^2\gtrsim c_1/{b}$, this gives
\[
\E\!\left[\left\{\frac{c_1}{c_3}(a_3-c_3)\right\}^2\mathbf1_{E_{{n}}}\right]
\le \frac{2(1+9k^2)c_1^2\tau_3^2}{c_3^2}=O(c_1{{\tau_1}}^2)=o({{\tau_1}}^2).
\]
For the denominator, unbiasedness separates the variance of the pilot from the squared penalty,
\[
 \E(\widetilde c_1-c_1+k{{\tau_1}})^2={{\tau_1}}^2+k^2{{\tau_1}}^2.
\]
Restricting this expectation to $E_{{n}}$ costs $o({{\tau_1}}^2)$ by boundedness of $\widetilde c_1$ and the tail bound for $E_{{n}}^c$. Cauchy--Schwarz and the preceding numerator bound give $o({{\tau_1}}^2)$ for the cross term as well. The square in \eqref{eq:pess-regular-risk-decomposition} therefore has expectation $(1+k^2){{\tau_1}}^2+o({{\tau_1}}^2)$ on $E_{{n}}$, which proves part~(ii).

On the transition $c_1{g}\sqrt {b}\asymp1$ with ${b}c_1\to\infty$, the pilot also has a Gaussian limit, but its mean must be retained. Indeed, ${g}^2c_1=(c_1{g}\sqrt {b})^2/({b}c_1)\to0$, so the same variance calculation gives $\rho_{{n}}\to1$ and $(c_{1,{b}}-c_1)/{{\tau_1}}\to0$ in mean square. Hence $(\widetilde c_1-c_1)/{{\tau_1}}\xrightarrow{d}Z_0$. Along any subsequence on which $t_{{n}}\to t$, \eqref{eq:fixed-mult-weak-expansion} yields
\[
 \frac{\mathcal R({\widehat {G}}_{\rm pes})}{\sqrt{c_3{{\tau_1}}}}
 \longrightarrow \E f_t\bigl(k+(t+Z_0)_+\bigr).
\]
Here the bounded second moments used in \eqref{eq:fixed-mult-weak-expansion} again justify taking expectations.
\end{proof}
\endgroup

\begingroup
\subsection{Replacing the population standard deviation by the jackknife}
\label{app:covariance-pessimistic-risk}
Theorem~\ref{thm:pessi} treats the covariance matrix of $(\widetilde c_1,\widetilde c_3)^\top$ as known. In practice, as described in Section~\ref{sec:pessimistic}, we estimate this covariance matrix from the pilot data using the grouped jackknife estimator $\widehat V_K$ in \eqref{eq:grouped-jackknife-cov} (Section~\ref{app:jackknife-cov}) and minimise the empirical criterion $\widehat Q_k({G})$ in \eqref{eq:main-floored-pess}. Write $\widehat\tau_1=\{(\widehat V_K)_{11}\}^{1/2}$ and $\widehat\tau_3=\{(\widehat V_K)_{33}\}^{1/2}$ for the jackknife standard errors of $\widetilde c_1$ and $\widetilde c_3$, respectively, and let
\begin{equation}
\widehat Q_k({G})={G}(\widetilde c_1)_++\frac{\widetilde c_3}{{G}-1}
+k\bigl\{\widehat\sigma({G})+{G}(\epsilon_0-{{\widehat\tau_1}})_+\bigr\},
\qquad
D:=(\widetilde c_1)_++k\max\{{{\widehat\tau_1}},\epsilon_0\},
\label{eq:estimated-sd-floor-rule}
\end{equation}
where $D$ denotes the effective large-${G}$ linear slope of $\widehat Q_k({G})$. Here, the completion-scale floor $\epsilon_0=a/({g}\sqrt {b})$ reflects the within-prompt variation from sampling ${g}$ completions per prompt, while the enlarged floor $\epsilon_0=\max\{a/({g}\sqrt {b}),\,{\beta} w_{\rm env}/{b}\}$ (with $a,{\beta}>0$ and an upper bound $w_{\rm env}$ on $\|{\bm g}(X;\theta)-{\bm g}(\theta)\|^2$) additionally guards against between-prompt variation that may be missed when the pilot batch size ${b}$ is small.

Before analysing the integer-grid estimator under the grouped jackknife, we note that replacing $\sigma({G})$ by $\widehat\sigma({G})$ on the continuous interval $[2,{{G}}_{\max}]$ already preserves the risk bounds of Theorem~\ref{thm:pessi} whenever the estimated standard errors concentrate at their population scales. Specifically, since the jackknife covariance matrix $\widehat V_K$ is positive semidefinite and the floor term adjusts the large-${G}$ linear slope to $D$, the derivative bounds in \eqref{eq:covariance-objective-derivative} continue to hold on every pilot sample with $D$ and $\widehat\tau_3$ in place of $c_{1,U}$ and $\tau_3$ (including when ${{\widehat\tau_1}}=0$, in which case $\widehat\sigma({G})=\widehat\tau_3/({G}-1)$). Suppose there exist constants $0<d_-<d_+<\infty$ and events $J_{{n}}$ on which
\[
d_-{{\tau_1}}\le\max\{{{\widehat\tau_1}},\epsilon_0\}\le d_+{{\tau_1}},
\qquad \widehat\tau_3\le \frac{d_+}{\sqrt {b}},
\]
such that
\[
(c_1{{G}}_{\max}+c_3)\Pr(J_{{n}}^c)=o(r_{{n}}),
\qquad
r_{{n}}=
\begin{cases}
\sqrt{c_3{{\tau_1}}},&c_1/{{\tau_1}}=O(1),\\
\sqrt{c_3}{{\tau_1}}^2/c_1^{3/2},&c_1/{{\tau_1}}\to\infty.
\end{cases}
\]
By \eqref{eq:bounded-excess-risk}, the contribution outside $J_{{n}}$ is $o(r_{{n}})$ in expectation. When $t_{{n}}=O(1)$, the argument for \eqref{eq:fixed-mult-weak-expansion} applies on $J_{{n}}$ with $d_{{n}}=D/{{\tau_1}}$, where $d_{{n}}\ge kd_-$ controls its reciprocal and $\E[d_{{n}}\mathbf1_{J_{{n}}}]\le t_{{n}}+1+kd_+$ controls its mean; intersecting $J_{{n}}$ with the Cantelli event $\{(\widetilde c_1-c_1)/{{\tau_1}}\ge-kd_-/2\}$ similarly yields the matching lower bound, so that $\mathcal R({\widehat {G}}_{\rm pes})\asymp r_{{n}}$.

In the regular regime ($t_{{n}}\to\infty$), the denominator adjustment on $J_{{n}}$ is $O({{\tau_1}})$, so intersecting $J_{{n}}$ with the relative-error event in \eqref{eq:pess-regular-risk-decomposition} yields $\mathcal R({\widehat {G}}_{\rm pes})=O(r_{{n}})$. If, in addition,
\[
 \E\!\left[\left(\frac{\max\{{{\widehat\tau_1}},\epsilon_0\}}{{{\tau_1}}}-1\right)^2\mathbf1_{J_{{n}}}\right]=o(1),
\]
then the difference between the estimated penalty $k\max\{{{\widehat\tau_1}},\epsilon_0\}$ and $k{{\tau_1}}$ has $L^2$ norm $o({{\tau_1}})$ on $J_{{n}}$. By the Cauchy--Schwarz inequality and \eqref{eq:fixed-mean-local-moments} with $p=4$, the regular-regime leading constant of Theorem~\ref{thm:pessi} is also preserved.

When minimising \eqref{eq:estimated-sd-floor-rule} over the integer grid $\{2,\ldots,{{G}}_{\max}\}$ using the grouped jackknife with a fixed number of groups $K\ge2$, two additional effects arise: the relative jackknife standard error $\widehat\tau_1/\tau_1$ can retain non-vanishing randomness, and the selected group size is restricted to integer values. The following theorem accounts for both effects.

\begin{appendixtheorem}[Grouped jackknife with an enlarged floor]\label{thm:enlarged-floor}
Suppose Assumptions~\ref{assump:upper}--\ref{assump:asym} hold, $\Var\{\|{\bm g}(X;\theta)-{\bm g}(\theta)\|^2\}\asymp c_1$, and $k,a,{\beta},w_{\rm env}>0$ are fixed. Partition the ${b}$ prompts into $K\ge2$ equal groups independently of the data, keeping $K$ fixed. Use the enlarged floor $\epsilon_0:=\max\{a/({g}\sqrt {b}),{\beta}w_{\rm env}/{b}\}$ and let $\lambda_{{n}}:=\epsilon_0/{{\tau_1}}$.

Let $Z\sim N(0,1)$ and let $S_K\ge0$ be independent of $Z$, with $(K-1)S_K^2\sim\chi^2_{K-1}$. Define the two functions
\[
\begin{aligned}
 C_{k,K}(\lambda)&:=\E\sqrt{(Z)_+ + k\max\{S_K,\lambda\}},\\
 A_{k,K}(\lambda)&:=1+k^2\E\bigl[\max\{S_K,\lambda\}^2\bigr].
\end{aligned}
\]
Let ${\widehat {G}}_{\rm pes}$ be the smallest minimiser of \eqref{eq:estimated-sd-floor-rule} over $\{2,\ldots,{{G}}_{\max}\}$.
\begin{enumerate}
\item[(i)] If $c_1=O(\epsilon_0)$, then $\mathcal R({\widehat {G}}_{\rm pes})\asymp\sqrt{c_3\epsilon_0}$. This includes weak signal, weak identification and the transition $c_1{g}\sqrt {b}\asymp1$. Under weak identification ($c_1{g}\sqrt {b}\to\infty$ and ${b}c_1=O(1)$), the floor equals ${\beta}w_{\rm env}/{b}$ for all large ${{n}}$. Two cases admit the following leading terms.
\[
\mathcal R({\widehat {G}}_{\rm pes})\sim
\begin{cases}
C_{k,K}(\lambda_{{n}})\sqrt{c_3{{\tau_1}}},
   & c_1{g}\sqrt {b}\to0,\quad {g}=O(\sqrt {b}),\\[2pt]
\sqrt{k {\beta} w_{\rm env}c_3/{b}},
   & {b}c_1\to0,\quad {g}/\sqrt {b}\to\infty .
\end{cases}
\]
\item[(ii)] In the regular regime, define $w_{{n}}:=\sqrt{c_3}\,{{\tau_1}}/c_1^{3/2}$. If $\liminf_{{n}} w_{{n}}>0$, in particular if $0<\kappa_{b}\le2$, then
\[
\mathcal R({\widehat {G}}_{\rm pes})\asymp\frac{\sqrt{c_3}{{\tau_1}}^2}{c_1^{3/2}}.
\]
If $w_{{n}}\to\infty$, in particular if $\kappa_{b}<2$, then
\[
\mathcal R({\widehat {G}}_{\rm pes})\sim
 A_{k,K}(\lambda_{{n}})\frac{\sqrt{c_3}{{\tau_1}}^2}{4c_1^{3/2}} .
\]
In particular, $A_{k,K}(\lambda_{{n}})\to1+k^2$ whenever $\epsilon_0=o({{\tau_1}})$.
\end{enumerate}
\end{appendixtheorem}

\begin{proof}[Proof of Theorem~\ref{thm:enlarged-floor}.]
Let $x:={\widehat {G}}_{\rm pes}-1$ be the selected integer in the ${G}-1$ coordinate, and let $x^*:={{G}}^*-1=\sqrt{c_3/c_1}$. We first locate $x$ for a given pilot. The error in the effective numerator is bounded by
\[
 U:=|\widetilde c_3-c_3|+3k\widehat\tau_3.
\]
Suppose $U<c_3$ and both roots $\sqrt{(c_3\pm U)/D}$ lie in $[1,{{G}}_{\max}-1]$. The criterion decreases to the left of the smaller root and increases to the right of the larger one. An integer strictly below the floor of the smaller root therefore has a neighbour to its right with a smaller criterion value. An integer strictly above the ceiling of the larger root has such a neighbour to its left. Hence
\begin{equation}
 \sqrt{\frac{c_3-U}{D}}-1\ \le x\le\sqrt{\frac{c_3+U}{D}}+1 .
 \label{eq:full-jk-bracket}
\end{equation}
In \eqref{eq:grouped-jackknife-cov}, replacing the replicate mean $K^{-1}\sum_{\ell=1}^K\widetilde c_j^{(-\ell)}$ by $c_j$ can only enlarge the sum of squares, so
\[
(\widehat V_K)_{jj}
\le\frac{K-1}{K}\sum_{k=1}^K\bigl(\widetilde c_j^{(-k)}-c_j\bigr)^2,
\qquad j=1,3.
\]
Since $K\ge2$ is fixed, each replicate uses ${b}(1-1/K)\asymp {b}$ prompts, and Lemmas~\ref{lem:coefficient-concentration}(b) and~\ref{lem:coefficient-variance} give
\[
 \E{{\widehat\tau_1}}^2=O({{\tau_1}}^2),\qquad \E U^2\le 2\Var(\widetilde c_3)+18k^2\E\widehat\tau_3^2=O({b}^{-1}).
\]

Suppose $c_1=O(\epsilon_0)$. Since ${b}^{-1}=O(\epsilon_0)$ and $({g}\sqrt {b})^{-1}=O(\epsilon_0)$, Lemma~\ref{lem:coefficient-variance} gives ${{\tau_1}}=O(\epsilon_0)$. It follows that
\[
 {{\tau_1}}^2\asymp\frac{c_1}{{b}}+\frac{1}{{b}{g}^2}=O(\epsilon_0^2),\qquad
 D\ge k\epsilon_0,\qquad
 \E\bigl[(D/\epsilon_0)^2\bigr]=O(1),
\]
using $D\le c_1+|\widetilde c_1-c_1|+k{{\widehat\tau_1}}+k\epsilon_0$ and $\E{{\widehat\tau_1}}^2=O({{\tau_1}}^2)=O(\epsilon_0^2)$.

Let $F_{{n}}$ be the event that each full-sample and delete-group estimate of $c_j$ is within ${b}^{-1/4}$ of $c_j$, for $j=1,3$. Each variance is $O({b}^{-1})$. Applying Lemma~\ref{lem:coefficient-concentration}(a) with $u$ a sufficiently small multiple of $\sqrt {b}$ and summing over the $2(K+1)$ estimates gives $\Pr(F_{{n}}^c)\le C\exp(-d\sqrt {b})$.

On $F_{{n}}$, all replicate deviations vanish uniformly, so the jackknife sums of squares give $D=o(1)$ and $U=o(1)$. Thus both roots in \eqref{eq:full-jk-bracket} diverge. They also stay below the cap, since $D\ge k\epsilon_0$ and $\epsilon_0^{-1/2}=O(c_1^{-1/2})=o({{G}}_{\max})$. The interval endpoints are consequently both equivalent to $\sqrt{c_3/D}$, and shifting either endpoint by one unit has a vanishing relative effect. We have shown that
\[
x=\{1+o(1)\}\sqrt{\frac{c_3}{D}}
\quad\text{uniformly on }F_{{n}}.
\]
As in \eqref{eq:fixed-mult-weak-expansion}, substitution in the exact loss gives
\[
 \frac{\mathcal R({\widehat {G}}_{\rm pes})}{\sqrt{c_3\epsilon_0}}
 =\E f_{c_1/\epsilon_0}(D/\epsilon_0)+o(1).
\]
Indeed, the relative error in $x$ is uniformly $o(1)$ on $F_{{n}}$, and the two terms $\sqrt{D/\epsilon_0}$ and $(c_1/\epsilon_0)/\sqrt{D/\epsilon_0}$ have bounded second moments. The complement contributes $o(1)$ by \eqref{eq:bounded-excess-risk} and its exponential probability bound. For the lower bound, Cantelli's inequality gives
\[
 \Pr(D\ge c_1+k\epsilon_0/2)\ge
 \frac{k^2\epsilon_0^2}{4{{\tau_1}}^2+k^2\epsilon_0^2},
\]
which stays bounded away from zero. The upper and lower bounds on $f_t$ used after \eqref{eq:fixed-mult-weak-expansion} now give the claimed two-sided order. Under weak identification, $({g}\sqrt {b})^{-1}=o(c_1)=O({b}^{-1})$, so the enlarged floor equals ${\beta}w_{\rm env}/{b}$. Whenever $c_1=o(\epsilon_0)$, the expansion reduces to
\begin{equation}
 \frac{\mathcal R({\widehat {G}}_{\rm pes})}{\sqrt{c_3\epsilon_0}}
 =\E\sqrt{D/\epsilon_0}+o(1).
 \label{eq:full-jk-weak-mean}
\end{equation}

Consider first ${b}c_1\to0$ and ${g}/\sqrt {b}\to\infty$. The enlarged floor then equals ${\beta}w_{\rm env}/{b}$, and
\[
 \frac{c_1}{\epsilon_0}=O({b}c_1)\to0,\qquad
 \frac{{{\tau_1}}^2}{\epsilon_0^2}=O({b}c_1+{b}/{g}^2)\to0.
\]
Both $\widetilde c_1/\epsilon_0$ and ${{\widehat\tau_1}}/\epsilon_0$ vanish in mean square. The maximum in $D$ therefore contributes $k+o_{L^2}(1)$ after division by $\epsilon_0$, while the positive-part estimate contributes $o_{L^2}(1)$. Equation~\eqref{eq:full-jk-weak-mean} now gives $\mathcal R({\widehat {G}}_{\rm pes})\sim\sqrt{k {\beta} w_{\rm env}c_3/{b}}$.

In the other weak-signal case, the floor and the jackknife standard error have comparable size. We therefore need the joint law of $\widetilde c_1$ and ${{\widehat\tau_1}}$ on the scale ${{\tau_1}}$. The same law will determine the regular-regime constant. We derive it by expressing both statistics in terms of the $K$ group means.

For a subsample of $m$ prompts, relabelled $1,\ldots,m$, \eqref{eq:conditional-group-representation} and \eqref{eq:prompt-hoeffding-coefficients} give
\[
\widetilde c_{1,m}-c_1=\frac1m\sum_{i=1}^m h_i+R_m,
\]
where $h_i=m\{\E(\widetilde c_{1,m}\mid X_i,Y_{i,1:{g}})-c_1\}$ depends only on $(X_i,Y_{i,1:{g}})$ (not on $m$), and
\[
R_m=-\frac{1}{m(m-1)}\sum_{1\le i\ne\ell\le m}\{{\widehat{\bm g}}_{g}(X_i;\theta)-{\bm g}(\theta)\}^\top\{{\widehat{\bm g}}_{g}(X_\ell;\theta)-{\bm g}(\theta)\}.
\]
Every kernel in $R_m$ has conditional mean zero given either of its two prompt blocks. In the square of $R_m$, a product consequently has mean zero unless both factors involve the same unordered pair. Write $V_{g}:=\Cov\{{\widehat{\bm g}}_{g}(X_1;\theta)\}$. For a matching pair, independence of the two blocks gives second moment $\tr(V_{g}^2)$. Counting the two orders of the pair yields
\[
 \E R_m^2=\frac{2\tr(V_{g}^2)}{m(m-1)}
 \le\frac{2\{\tr(V_{g})\}^2}{m(m-1)}
 =O\!\left(\frac{(c_1+{g}^{-1})^2}{m^2}\right)=o({{\tau_1}}^2)
 \qquad (m\asymp {b}),
\]
where $\tr(V_{g})=O(c_1+{g}^{-1})$ by Lemma~\ref{prop:score-closed-forms} and ${{\tau_1}}^2\asymp(c_1+{g}^{-2})/{b}$ by Lemma~\ref{lem:coefficient-variance}. The remainder is uncorrelated with the linear sum by \eqref{eq:corr-ET}. Taking $m={b}$ therefore gives $\Var(h_i)\sim {b}{{\tau_1}}^2$.

Define the standardised group sums
\[
L_j:=\frac{1}{\sqrt{({b}/K)\Var(h_1)}}\sum_{i\in I_j} h_i,
\qquad
\bar L:=\frac1K\sum_{j=1}^K L_j.
\]
The group sums are independent conditional on the data-independent partition, and their joint law is the same for every equal-sized partition. Each sum has a standard normal limit in the following two cases.
\begin{enumerate}
\item[(a)] In the regular regime, $h_i$ is bounded under Assumption~\ref{assump:upper} and $({b}/K)\Var(h_i)\sim {b}^2{{\tau_1}}^2/K\to\infty$ since ${b}c_1\to\infty$, so the Lindeberg condition holds for each group.
\item[(b)] When $c_1{g}\sqrt {b}\to0$ and ${g}=O(\sqrt {b})$, the product $(c_1{g}\sqrt {b})({g}/\sqrt {b})={g}^2c_1$ tends to zero. The same conditions hold for a group of ${b}/K$ prompts. The weak-signal calculation in the proof of Theorem~\ref{thm:pessi} in Section~\ref{app:pessimistic-risk-proof} therefore gives a standard normal limit for the group estimate after centring and division by its own standard error. Removing $R_{{b}/K}$ costs $o_{L^2}({{\tau_1}})$, and the variance of the group estimate is asymptotic to $\Var(h_1)/({b}/K)$. This gives the asserted limit for $L_j$.
\end{enumerate}
To see what the jackknife estimates, let $\bar h_j$ denote the mean of $h_i$ in group $I_j$, and $\bar h:=K^{-1}\sum_j\bar h_j$ the full mean. Removing group $j$ gives the exact identity
\[
 \frac{K\bar h-\bar h_j}{K-1}-\bar h
 =-\frac{\bar h_j-\bar h}{K-1}.
\]
The delete-group deviations therefore measure the spread of the group means. With $s_{g}^2:=\Var(h_1)$, their contribution to the jackknife variance is
\[
 \frac{K-1}{K}\sum_{j=1}^K
 \left(\frac{\bar h_j-\bar h}{K-1}\right)^2
 =\frac{s_{g}^2}{{b}}\,\frac{1}{K-1}\sum_{j=1}^K(L_j-\bar L)^2,
\]
whereas the full mean is $\bar h=(s_{g}/\sqrt {b})\sqrt K\,\bar L$. Since $s_{g}/\sqrt {b}\sim{{\tau_1}}$ and all $K+1$ remainders are $o_{L^2}({{\tau_1}})$, restoring them changes the following standardised statistics by $o_p(1)$. We obtain
\begin{equation}
 \left(\frac{\widetilde c_1-c_1}{{{\tau_1}}},\frac{{{\widehat\tau_1}}^2}{{{\tau_1}}^2}\right)
 =\left(\sqrt K\,\bar L,\ \frac1{K-1}\sum_{j=1}^K(L_j-\bar L)^2\right)+o_p(1)
 \ \xrightarrow{d}\ (Z,S_K^2).
 \label{eq:full-jk-joint}
\end{equation}
The two limiting coordinates are the scaled mean and sample variance of $K$ independent standard normal variables. By Cochran's theorem they are independent, with $Z\sim N(0,1)$ and $(K-1)S_K^2\sim\chi^2_{K-1}$.

Return to weak signal with ${g}=O(\sqrt {b})$. Here $\epsilon_0\asymp({g}\sqrt {b})^{-1}\asymp{{\tau_1}}$, so $\lambda_{{n}}$ stays bounded above and away from zero. Also $c_1/{{\tau_1}}\to0$. Along any subsequence with $\lambda_{{n}}\to\lambda$, taking the positive part and the maximum in the definition of $D$ gives
\[
 D/{{\tau_1}}\xrightarrow{d}(Z)_++k\max\{S_K,\lambda\}.
\]
The second moments of $D/{{\tau_1}}$ are bounded, so we can take expectations of square roots. By \eqref{eq:full-jk-weak-mean} and continuity of $C_{k,K}$, the risk is asymptotic to $C_{k,K}(\lambda_{{n}})\sqrt{c_3{{\tau_1}}}$ along the full sequence.

In the regular regime, Lemma~\ref{lem:coefficient-variance} gives $\epsilon_0=O({{\tau_1}})$ and ${{\tau_1}}=o(c_1)$. Let $E_{{n}}$ be the event that both the full estimates and all $K$ replicates lie within relative error $\epsilon_{{n}}$ of their respective means $c_1,c_3$, with $\epsilon_{{n}}$ as in \eqref{eq:debiased-localization-sequence}. From \eqref{eq:fixed-mean-local-moments}, $\Pr(E_{{n}}^c)=O({{n}}^{-m})$ for each fixed $m>0$.

On $E_{{n}}$, $D/c_1=1+o(1)$ and $U/c_3=o(1)$ uniformly, so both roots are interior. We first evaluate the loss at a point inside the interval, then account for the integer spacing. Let $x_0$ be the point of the closed interval between the roots nearest to $x$. By \eqref{eq:full-jk-bracket}, $|x-x_0|\le1$. If $a_3:=Dx_0^2$, the defining inequalities for the interval give
\[
 |a_3-c_3|\le U,\qquad 1+x_0=1+\sqrt{a_3/D}.
\]
Both $x_0/x^*$ and $x/x^*$ thus tend to $1$ uniformly on $E_{{n}}$.

We first evaluate the loss at $1+x_0$. The numerator error has negligible second moment on the scale of the denominator error, since $(c_1/c_3)^2\E U^2=O(c_1^2/{b})=o({{\tau_1}}^2)$. Applying Lemma~\ref{lem:local-excess-risk} as in \eqref{eq:pess-regular-risk-decomposition} gives
\[
\E\bigl[\{\ell(1+x_0)-\ell({{G}}^*)\}\mathbf1_{E_{{n}}}\bigr]
=\frac{\sqrt{c_3}}{4c_1^{3/2}}\E\bigl[(D-c_1)^2\mathbf1_{E_{{n}}}\bigr]\{1+o(1)\}
+o\!\left(\frac{\sqrt{c_3}{{\tau_1}}^2}{c_1^{3/2}}\right).
\]
To evaluate the expectation, note that on $E_{{n}}$ the positive part of $\widetilde c_1$ equals $\widetilde c_1$. Thus the standardised denominator error is the sum of the centred pilot estimate and its estimated penalty,
\[
 \frac{D-c_1}{{{\tau_1}}}
 =\frac{\widetilde c_1-c_1}{{{\tau_1}}}
   +k\max\{{{\widehat\tau_1}}/{{\tau_1}},\lambda_{{n}}\}.
\]
Equation~\eqref{eq:fixed-mean-local-moments} with $p=4$ for the full sample and each replicate bounds the fourth moment of the left-hand side uniformly, using ${b}{{\tau_1}}\to\infty$. Hence we can take second moments in the joint limit along any convergent subsequence of $\lambda_{{n}}$. The cross term drops out because $Z$ has mean zero and is independent of $S_K$. By continuity in $\lambda$ and boundedness of $\lambda_{{n}}$, this gives
\[
 \frac{\E[(D-c_1)^2\mathbf1_{E_{{n}}}]}{{{\tau_1}}^2}
 =\E\bigl[(Z+k\max\{S_K,\lambda_{{n}}\})^2\bigr]+o(1)
 =A_{k,K}(\lambda_{{n}})+o(1),
\]
so that
\begin{equation}
 \E\bigl[\{\ell(1+x_0)-\ell({{G}}^*)\}\mathbf1_{E_{{n}}}\bigr]
 =\frac{\sqrt{c_3}{{\tau_1}}^2}{4c_1^{3/2}}\{A_{k,K}(\lambda_{{n}})+o(1)\}.
 \label{eq:enlarged-floor-unrounded-risk}
\end{equation}

It remains to determine when moving from $x_0$ to the integer $x$ affects the leading loss. Write $\|Y\|_{2,E_{{n}}}:=\{\E(Y^2\mathbf1_{E_{{n}}})\}^{1/2}$. For any $y>0$ the exact loss is $c_1(y-x^*)^2/y$. Since $x_0/x^*\to1$ uniformly on $E_{{n}}$, \eqref{eq:enlarged-floor-unrounded-risk} implies
\[
 \|x_0-x^*\|_{2,E_{{n}}}^2
 \asymp \frac{x^*}{c_1}\,\frac{\sqrt{c_3}{{\tau_1}}^2}{c_1^{3/2}}
 =\frac{c_3{{\tau_1}}^2}{c_1^3}=w_{{n}}^2.
\]
Thus $w_{{n}}$ measures the displacement due to estimation in the same units as the grid spacing. Moving to the integer changes the root mean squared displacement by at most one unit, since
\[
 \big|\|x-x^*\|_{2,E_{{n}}}-\|x_0-x^*\|_{2,E_{{n}}}\big|\le1.
\]
Both $x$ and $x_0$ are uniformly equivalent to $x^*$ on $E_{{n}}$, and $E_{{n}}^c$ contributes $o(\sqrt{c_3}{{\tau_1}}^2/c_1^{3/2})$ by \eqref{eq:bounded-excess-risk}. It follows that the integer rule has the same leading constant when $w_{{n}}\to\infty$, and the asserted upper bound when $\liminf_{{n}} w_{{n}}>0$. Lemma~\ref{lem:coefficient-variance} gives $w_{{n}}\gtrsim({b}c_1^2)^{-1/2}$, which yields the stated sufficient conditions on $\kappa_{b}$.

If $w_{{n}}$ stays bounded away from zero, a one-unit displacement can affect the leading constant. To obtain a lower bound, we exhibit pilots for which the error in the group size exceeds the rounding allowance. Fix $w_0>0$ with $w_{{n}}\ge w_0$ eventually, and choose $M\ge8/w_0$. By \eqref{eq:full-jk-joint}, the event $\{\widetilde c_1-c_1\ge M{{\tau_1}}\}$ has limiting probability $1-\Phi(M)>0$.

The numerator error is smaller in relative terms, since $\E U^2=O({b}^{-1})$ and $c_1^2/({b}{{\tau_1}}^2)=O(c_1)$. Thus intersecting the preceding event with $E_{{n}}$ and $\{U/c_3\le M{{\tau_1}}/(4c_1)\}$ leaves its limiting probability unchanged. On this intersection the denominator is at least $c_1+M{{\tau_1}}$. Substituting in the upper root and using $\sqrt{(1+u/4)/(1+u)}\le1-u/4$ for sufficiently small $u>0$ gives
\[
 x\le 1+x^*\sqrt{\frac{1+M{{\tau_1}}/(4c_1)}{1+M{{\tau_1}}/c_1}}
 \le 1+x^*\left(1-\frac{M{{\tau_1}}}{4c_1}\right).
\]
Hence $x^*-x\ge Mw_{{n}}/4-1\ge Mw_{{n}}/8$. On this event the exact loss is at least $c_1(Mw_{{n}}/8)^2/x^*$, of order $\sqrt{c_3}{{\tau_1}}^2/c_1^{3/2}$. Its probability stays positive, proving the lower bound. Finally, $\E S_K^2=1$ and $0\le\max\{S_K,\lambda_{{n}}\}^2-S_K^2\le\lambda_{{n}}^2$ give $A_{k,K}(\lambda_{{n}})\to1+k^2$ when $\lambda_{{n}}\to0$.
\end{proof}
\endgroup

\subsection{Choosing the penalty multiplier}\label{app:pilot-confidence}
While Theorem~\ref{thm:enlarged-floor} characterises the expected excess MSE across repeated pilot samples, we now show how to calibrate the penalty multiplier $k$ so that, on the observed pilot sample, the adjusted denominator $D$ in \eqref{eq:estimated-sd-floor-rule} upper-bounds $c_1$ with a prescribed asymptotic confidence level $1-\alpha$. On the same coverage event, we also bound the realised excess loss $\Delta({G}):=\ell({G})-\ell({{G}}^*)$ at the selected integer group size ${G}={\widehat {G}}_{\rm pes}$ (which equals ${{N}}$ times the realised excess gradient MSE).
To state the regularity conditions for the normal and Student-$t$ approximations, define the first-order contribution of the $i$th pilot prompt by $h_i:={b}\{\E(\widetilde c_1\mid X_i,Y_{i,1:{g}})-c_1\}$ for $i=1,\ldots,{b}$, and let $W_i:=\|{\bm g}(X_i;\theta)-{\bm g}(\theta)\|^2$, so that $\E W_i=c_1$. By \eqref{eq:conditional-group-representation}--\eqref{eq:conditional-centering}, $\E(h_i\mid X_i)=W_i-c_1$, which gives the orthogonal decomposition
\[
 h_i=(W_i-c_1)+e_i,\qquad \E(e_i\mid X_i)=0,\qquad
 s_{g}^2=\Var(h_i)=\Var(W_i)+\E e_i^2,
\]
where $e_i$ captures the within-prompt completion sampling error. To establish the central limit theorem for $\widetilde c_1$, we assume
\begin{equation}
\begin{gathered}
 \liminf_{{{n}}\to\infty}{g}^2\E e_i^2>0,\qquad c_1^2=o({b}s_{g}^2),\\
 \E(W_i-c_1)^4=o({b}s_{g}^4)\quad\text{or}\quad\Var(W_i)=o({g}^{-2}),
 \end{gathered}
 \label{eq:pilot-confidence-conditions}
\end{equation}
where the first line ensures that the within-prompt completion error has standard deviation at least of order ${g}^{-1}$ and that the cross-prompt remainder is asymptotically negligible, while the second line requires either a Lyapunov fourth-moment condition on $W_i$ or that the between-prompt variance is negligible relative to the within-prompt completion noise.
To quantify the excess loss incurred by this upper-confidence adjustment, note that when the denominator is increased from $c_1$ to $c_1+v$ ($v\ge0$) while holding the numerator at $c_3$, the continuous minimiser $1+\sqrt{c_3/(c_1+v)}$ incurs an exact excess loss of
\[
 \frac{\sqrt{c_3}\,v^2}
 {\sqrt{c_1+v}\,(\sqrt{c_1+v}+\sqrt{c_1})^2}
 \asymp \frac{\sqrt{c_3}\,v^2}{(c_1+v)^{3/2}}.
\]
Setting the effective denominator adjustment scale to $v_{{n}}:=k\max\{{{\tau_1}},\epsilon_0\}$ and accounting for both the $O_p({b}^{-1/2})$ estimation error in $\widetilde c_3$ and the unit grid spacing near the oracle group size yields the realised excess-loss scale
\begin{equation}
 R_{{{n}},k}:=\frac{\sqrt{c_3}\,v_{{n}}^2}{(c_1+v_{{n}})^{3/2}}
        +\frac{\sqrt{c_1}}{{b} c_3^{3/2}}
        +\frac{c_1^{3/2}}{\sqrt{c_3}} .
 \label{eq:pilot-confidence-scale}
\end{equation}

\begin{appendixproposition}[Fixed-confidence group size selection]\label{prop:pilot-confidence}
Suppose the pilot samples independent prompts and conditionally independent completions as in Section~\ref{sec:tradeoff}. Let Assumptions~\ref{assump:upper} and~\ref{assump:asym} hold, with $c_3\asymp1$ and \eqref{eq:pilot-confidence-conditions}. Use either floor, with fixed $a,{\beta}>0$ and a uniformly bounded prespecified envelope $w_{\rm env}$.

Fix $0<\alpha<1/2$. For deletion of one of $K\ge2$ equal groups at a time, suppose $K$ is fixed, ${b}/K\to\infty$, and the partition is independent of the data. Take $k=t_{K-1,1-\alpha}$, the $1-\alpha$ quantile of Student's $t$ law with $K-1$ degrees of freedom. For deletion of one prompt at a time ($K={b}$), take $k=z_{1-\alpha}$, the corresponding standard normal quantile. In either case,
\[
 \liminf_{{{n}}\to\infty}\Pr\{D\ge c_1,\ {\widehat {G}}_{\rm pes}<{{G}}_{\max}\}\ge1-\alpha.
\]
Moreover, for every $0<\delta<1-\alpha$, there exists a finite constant $C_{\alpha,\delta}$ such that
\begin{equation}
 \liminf_{{{n}}\to\infty}\Pr\{D\ge c_1,\ {\widehat {G}}_{\rm pes}<{{G}}_{\max},\
        \Delta({\widehat {G}}_{\rm pes})\le C_{\alpha,\delta}R_{{{n}},k}\}
 \ge 1-\alpha-\delta .
 \label{eq:pilot-confidence-joint}
\end{equation}
Consequently, for every deterministic sequence $M_{{n}}\to\infty$, the limit inferior in \eqref{eq:pilot-confidence-joint} is at least $1-\alpha$ when $M_{{n}}R_{{{n}},k}$ replaces $C_{\alpha,\delta}R_{{{n}},k}$.
\end{appendixproposition}

At the nominal $90\%$ level ($\alpha=0.10$), using $K=16$ groups yields $k=t_{15,0.90}\approx1.34$, while the leave-one-prompt-out jackknife ($K={b}$) yields $k=z_{0.90}\approx1.28$. With either choice, Proposition~\ref{prop:pilot-confidence} gives a limiting lower bound of $0.90$ for coverage and avoidance of the upper boundary. For any fixed confidence below $0.90$, the same pilot also attains excess loss at most a fixed multiple of $R_{{{n}},k}$. Allowing that multiple to grow arbitrarily slowly gives the full $0.90$ guarantee. In particular, whenever $c_1\le k\epsilon_0$---which holds eventually in the weak-signal regime under either positive floor---the floor alone ensures $D\ge c_1$ on every pilot sample.

\begin{proof}[Proof of Proposition~\ref{prop:pilot-confidence}.]
Consider first the error in $\widetilde c_1$ before studentisation. The expansion from Section~\ref{app:covariance-pessimistic-risk} separates a sum of independent prompt contributions from a remainder involving pairs of prompts. With $M_i:={\widehat{\bm g}}_{g}(X_i;\theta)$ and $u_{ij}:=-(M_i-{\bm g}(\theta))^\top(M_j-{\bm g}(\theta))$, it reads
\[
 \widetilde c_1-c_1=\frac1{b}\sum_{i=1}^{b} h_i+R_{b},\qquad
 R_{b}=\binom {b}2^{-1}\sum_{1\le i<j\le {b}}u_{ij}.
\]
The kernel $u_{ij}$ has mean zero conditional on block $i$ and also conditional on block $j$. Thus distinct unordered pairs are uncorrelated, and their sum is also uncorrelated with the sum of the $h_i$. Since ${g}^{-2}=O(s_{g}^2)$ by the first condition of \eqref{eq:pilot-confidence-conditions}, we have
\[
 \E u_{12}^2\lesssim(c_1+{g}^{-1})^2=o({b}s_{g}^2).
\]
For $m\asymp {b}$ this makes $\E R_m^2=2\E u_{12}^2/\{m(m-1)\}=o(s_{g}^2/{b})$. In particular, the sum of the $h_i$ accounts for the full variance to first order, so ${{\tau_1}}^2\sim s_{g}^2/{b}$.

We next verify the central limit theorem for that sum. The two alternatives in \eqref{eq:pilot-confidence-conditions} concern $W_i$. Both use the same fourth-moment bound for $e_i$. To obtain it, define the within-prompt gradient error $\varepsilon_i:=M_i-{\bm g}(X_i;\theta)$. Conditional on $X_i$, this is a centred second-order U-statistic. The pairing representation in \eqref{eq:paired-sample-variance} writes it as an average over pairings of the completions. Within each pairing there are $\lfloor {g}/2\rfloor$ independent, bounded, centred pair kernels. Applying Lemma~\ref{lem:conditional-vector-rosenthal} within each pairing and Jensen's inequality across pairings gives
\[
 \E(\|\varepsilon_i\|^p\mid X_i)\lesssim {g}^{-p/2},\qquad p=4,8.
\]

To pass from $\varepsilon_i$ to $e_i$, use \eqref{eq:lemma-gradient-ustat} and the within-prompt expansion \eqref{eq:conditional-within-group-expansion}--\eqref{eq:conditional-quadratic-mean}. With $a_{g}:=({g}-1)^2/\{({g}-1)^2+1\}$, the completion error has a linear and a quadratic term,
\[
 e_i=h_i-(W_i-c_1)
    =2\{a_{g} {\bm g}(X_i;\theta)-{\bm g}(\theta)\}^\top\varepsilon_i
+\frac{{g}({g}-1)}{{g}^2-2{g}+2}\{Q_i-\E(Q_i\mid X_i)\}.
\]
For the linear term, write $a_{g}{\bm g}(X_i;\theta)-{\bm g}(\theta)=a_{g}\{{\bm g}(X_i;\theta)-{\bm g}(\theta)\}-(1-a_{g}){\bm g}(\theta)$. Its squared norm is $O(W_i+{g}^{-4})$ because $1-a_{g}=O({g}^{-2})$. The conditional fourth moment of the linear term is therefore $O\{(W_i^2+{g}^{-8}){g}^{-2}\}$.

In the quadratic term, $Q_i$ is the difference between $\|\varepsilon_i\|^2$ and an average bounded by $C/{g}$. The eighth-moment bound for $\varepsilon_i$ therefore gives conditional fourth moment $O({g}^{-4})$ for the centred $Q_i$. Combining the two bounds and using $\E W_i^2=\Var(W_i)+c_1^2$ gives
\[
 \E e_i^4\lesssim {g}^{-2}\{\Var(W_i)+c_1^2\}+{g}^{-4}
       =o({b}s_{g}^4),
\]
where the last step uses ${g}^{-2}\lesssim s_{g}^2$, $\Var(W_i)\le s_{g}^2$, and $c_1^2=o({b}s_{g}^2)$.

Under the fourth-moment alternative we can now bound $\E h_i^4$ by $8\E(W_i-c_1)^4+8\E e_i^4=o({b}s_{g}^4)$. This is Lyapunov's condition for $\sum_i h_i/(\sqrt {b}\,s_{g})$. Under the small-variance alternative,
\[
 \frac{\Var(W_i)}{s_{g}^2}
 =\frac{\Var(W_i)}{{g}^{-2}}\,\frac{{g}^{-2}}{s_{g}^2}\longrightarrow0.
\]
The centred $W_i$ then make a negligible contribution in mean square to $(\sqrt {b}\,s_{g})^{-1}\sum_i h_i$, and $\Var(e_i)/s_{g}^2\to1$. Lyapunov's theorem applied to $e_i$ gives the same standard normal limit. Both arguments also apply to group sums of size ${b}/K$ for fixed $K$.

For fixed $K$, apply this central limit theorem within each group and use \eqref{eq:full-jk-joint}. The studentised estimate converges to $Z/S_K$. Since $Z$ and $S_K$ are independent and $(K-1)S_K^2\sim\chi^2_{K-1}$, the ratio has Student's $t_{K-1}$ law.

When we delete one prompt at a time, the jackknife instead uses the sample variance of all ${b}$ first-order contributions. We first check that this sample variance, divided by $s_{g}^2$, tends to $1$. Under the fourth-moment alternative, Chebyshev's inequality applies to the average of $h_i^2$ since $\E h_i^4=o({b}s_{g}^4)$. Under the other alternative, it applies to the average of $e_i^2$. The average of $(W_i-c_1)^2/s_{g}^2$ vanishes in probability by its expectation, and Cauchy--Schwarz handles the cross term. In both cases the sample mean correction vanishes since $\E\bar h^2=s_{g}^2/{b}$, with $\bar h:={b}^{-1}\sum_i h_i$.

We must still account for the pair kernel in each replicate. Let $r_i:=({b}-1)^{-1}\sum_{j\ne i}u_{ij}$ be the mean of the kernels involving prompt $i$. Deleting that prompt removes their sum $({b}-1)r_i$ from the total $\binom {b}2 R_{b}$ and leaves $\binom{{b}-1}{2}$ pairs. Since ${b}^{-1}\sum_i r_i=R_{b}$, centring the replicates gives
\[
 \widetilde c_1^{(-i)}-\frac1{b}\sum_{\ell=1}^{b}\widetilde c_1^{(-\ell)}
 =-\frac{h_i-\bar h}{{b}-1}-\frac{2(r_i-R_{b})}{{b}-2}.
\]
The first term gives the jackknife variance $\{{b}({b}-1)\}^{-1}\sum_i(h_i-\bar h)^2={{\tau_1}}^2\{1+o_p(1)\}$. For the second term, $\E r_i^2=\E u_{12}^2/({b}-1)$ by conditional centring. Its contribution to the jackknife sum of squares has expectation $O(\E u_{12}^2/{b}^2)=o({{\tau_1}}^2)$. Cauchy--Schwarz gives the same negligible order in probability for the cross term. Thus ${{\widehat\tau_1}}/{{\tau_1}}\to1$ in probability, and Slutsky's lemma gives a standard normal limit for the studentised estimate.

We now use these limits to obtain coverage. On $\{{{\widehat\tau_1}}>0\}$ the event $\widetilde c_1+k{{\widehat\tau_1}}\ge c_1$ is exactly the event that the studentised estimate is at least $-k$. The limiting relative standard error is positive almost surely in both constructions. By symmetry and continuity of the limiting law, the chosen quantile gives
\[
 \Pr(\widetilde c_1+k{{\widehat\tau_1}}\ge c_1)\longrightarrow1-\alpha.
\]
Taking the positive part and applying the floor can only raise this upper endpoint. Therefore $E_{{n}}:=\{D\ge c_1\}$ has $\liminf_{{n}}\Pr(E_{{n}})\ge1-\alpha$.

We need the size of the error on this event as well. The positive part satisfies $|(\widetilde c_1)_+-c_1|\le|\widetilde c_1-c_1|$, and the studentisation results give ${{\widehat\tau_1}}=O_p({{\tau_1}})$. Hence $D-c_1=O_p(v_{{n}})$ and $D=o_p(1)$. For the numerator, \eqref{eq:app-score-unbiased} expresses $\widetilde c_3$ as a mean of bounded prompt contributions. Its delete-one variance estimate is their sample variance divided by ${b}$. For fixed $K$, each delete-group mean has variance $O({b}^{-1})$. In either case $U:=|\widetilde c_3-c_3|+3k\widehat\tau_3=O_p({b}^{-1/2})$.

Coverage now explains why the selected integer avoids the cap. On $E_{{n}}$, except on a set of probability $o(1)$, we have $U\le c_3/2$ and the lower root $\sqrt{(c_3-U)/D}$ exceeds $1$. For the upper root, use $D\ge c_1$ to obtain
\[
 \sqrt{(c_3+U)/D}\le\sqrt{3c_3/(2c_1)}<{{G}}_{\max}-2.
\]
Even after allowing one unit for rounding and adding $1$ to return to the group size, \eqref{eq:full-jk-bracket} gives ${\widehat {G}}_{\rm pes}<{{G}}_{\max}$.

On these pilots let $x_0$ be the nearest point to ${\widehat {G}}_{\rm pes}-1$ between the roots. As in Section~\ref{app:covariance-pessimistic-risk}, define ${G}_0:=1+x_0$ and $a_3:=Dx_0^2$. Then $|a_3-c_3|\le U$ and $|{\widehat {G}}_{\rm pes}-{G}_0|\le1$. Set ${G}_0:={{G}}^*$ on all other pilots. We first bound the cost of using ${G}_0$ and then the cost of the one-unit displacement.

At ${G}_0$, the exact loss is the squared difference $(\sqrt{c_3D}-\sqrt{c_1a_3})^2$ divided by $\sqrt{Da_3}$. To see how each coefficient contributes, write
\[
 \sqrt{c_3D}-\sqrt{c_1a_3}
 =\sqrt{c_3}\,\frac{D-c_1}{\sqrt D+\sqrt{c_1}}
  -\sqrt{c_1}\,\frac{a_3-c_3}{\sqrt{a_3}+\sqrt{c_3}}.
\]
We square this identity, using $(u-v)^2\le 2u^2+2v^2$. Since $a_3\asymp c_3$ and $D\ge c_1$ on the coverage event under consideration, we obtain
\[
 \Delta({G}_0)=\frac{(\sqrt{c_3D}-\sqrt{c_1a_3})^2}{\sqrt{Da_3}}
 \ \lesssim\ \sqrt{c_3}\,\frac{(D-c_1)^2}{D^{3/2}}
       +\frac{\sqrt{c_1}}{c_3^{3/2}}\,U^2 .
\]
The numerator term is $O_p(\sqrt{c_1}/({b} c_3^{3/2}))$. For the denominator term on $E_{{n}}$, write $D=c_1+t$, where $0\le t=O_p(v_{{n}})$. The function $t^2/(c_1+t)^{3/2}$ increases for $t\ge0$, and raising its argument from $t$ to $Lt$ for $L\ge1$ multiplies the value by at most $L^2$. We may therefore replace $t$ by $v_{{n}}$ to obtain the first term of \eqref{eq:pilot-confidence-scale} up to an $O_p(1)$ factor.

For the integer choice, $|{\widehat {G}}_{\rm pes}-{G}_0|\le1$ and ${G}_0-1$ diverges in probability on the pilots under consideration. Hence the two denominators ${\widehat {G}}_{\rm pes}-1$ and ${G}_0-1$ are comparable with probability tending to one on those pilots. Bounding $({\widehat {G}}_{\rm pes}-{{G}}^*)^2$ by $2({G}_0-{{G}}^*)^2+2$ in the exact loss formula gives
\[
 \Delta({\widehat {G}}_{\rm pes})\lesssim \Delta({G}_0)+\frac{c_1}{{G}_0-1},
 \qquad
 \frac{c_1}{{G}_0-1}=O_p\!\left(c_1\sqrt{\frac{c_1+v_{{n}}}{c_3}}\right).
\]
When $v_{{n}}\le c_1$, the third term of $R_{{{n}},k}$ absorbs the rounding cost. When $v_{{n}}>c_1$, the first term has order $\sqrt{c_3v_{{n}}}$, and the rounding cost is smaller by a factor $O_p(c_1/c_3)$. Thus $\mathbf1_{E_{{n}}}\Delta({\widehat {G}}_{\rm pes})/R_{{{n}},k}$ is tight.

Finally, the joint event in \eqref{eq:pilot-confidence-joint} can fail in three ways. Coverage can fail, with limiting upper probability at most $\alpha$. On the coverage event the selected integer can reach the cap, with probability $o(1)$. The remaining possibility is that the realised loss exceeds its bound on the coverage event. Tightness of $\mathbf1_{E_{{n}}}\Delta({\widehat {G}}_{\rm pes})/R_{{{n}},k}$ bounds this last limiting upper probability by $\delta$ for a sufficiently large fixed $C_{\alpha,\delta}$. For $M_{{n}}\to\infty$ the corresponding probability tends to zero. Subtracting these probabilities gives the two stated confidence bounds.
\end{proof}

\section{Experimental design and supplementary results}
\label{app:current-experiments}

\subsection{Gradient estimation experiments}
\label{app:mse-details}
\noindent\textbf{Experimental configurations.}
The gradient-MSE evaluation in Section~\ref{sec:offline-diagnosis} is conducted across 19 policy checkpoints, covering steps 0, 50, 100, and 150 for decimal arithmetic, base conversion, and prime factorisation in Reasoning Gym \citep{stojanovski2025reasoninggym}, steps 0, 150, 300, and 450 for GSM8K \citep{cobbe2021training} (as in Figure~\ref{fig:mse-geometry}), and step 50 for equation solving, fraction arithmetic, and knights-and-knaves puzzles, where step 0 refers to the initial instruct model. At each checkpoint, the completion pool is generated at temperature $1$ and scored by the task's binary verifier. When evaluating a selected group size ${\widehat {G}}$, prompts are sampled with replacement and completions are drawn without replacement from each sampled prompt's pool. If ${{N}}=512$ is not divisible by ${\widehat {G}}$, completion counts across prompts differ by at most one so that the total budget remains 512.

To approximate the oracle optimal MSE in Figure~\ref{fig:mse-mechanisms}, for each checkpoint and sample size configuration $({b},{g})$, let $\overline{\operatorname{MSE}}_m$ denote the average estimated MSE of method $m$ across the 25 repetitions (excluding degenerate replicates in which $\widetilde c_1=\widetilde c_2=\widetilde c_3=0$). We then take
\[
 d=\min\Bigl\{\min_{{{G}}\in\{2,4,8,16,32,64\}}\widehat{\operatorname{MSE}}({{G}}),\ \min_{m}\overline{\operatorname{MSE}}_m\Bigr\},
\]
where $\widehat{\operatorname{MSE}}({{G}})$ denotes the estimated MSE at a fixed group size ${{G}}$. We adopt this approximation because computing the exact oracle MSE would require Monte Carlo gradient evaluations at every integer ${{G}}\in[2,64]$, which is computationally prohibitive. Taking the minimum over the six fixed group sizes and the evaluated methods ensures that the reported relative excess MSE is always nonnegative.

\begin{figure}[!htbp]
\centering
\includegraphics[width=\textwidth]{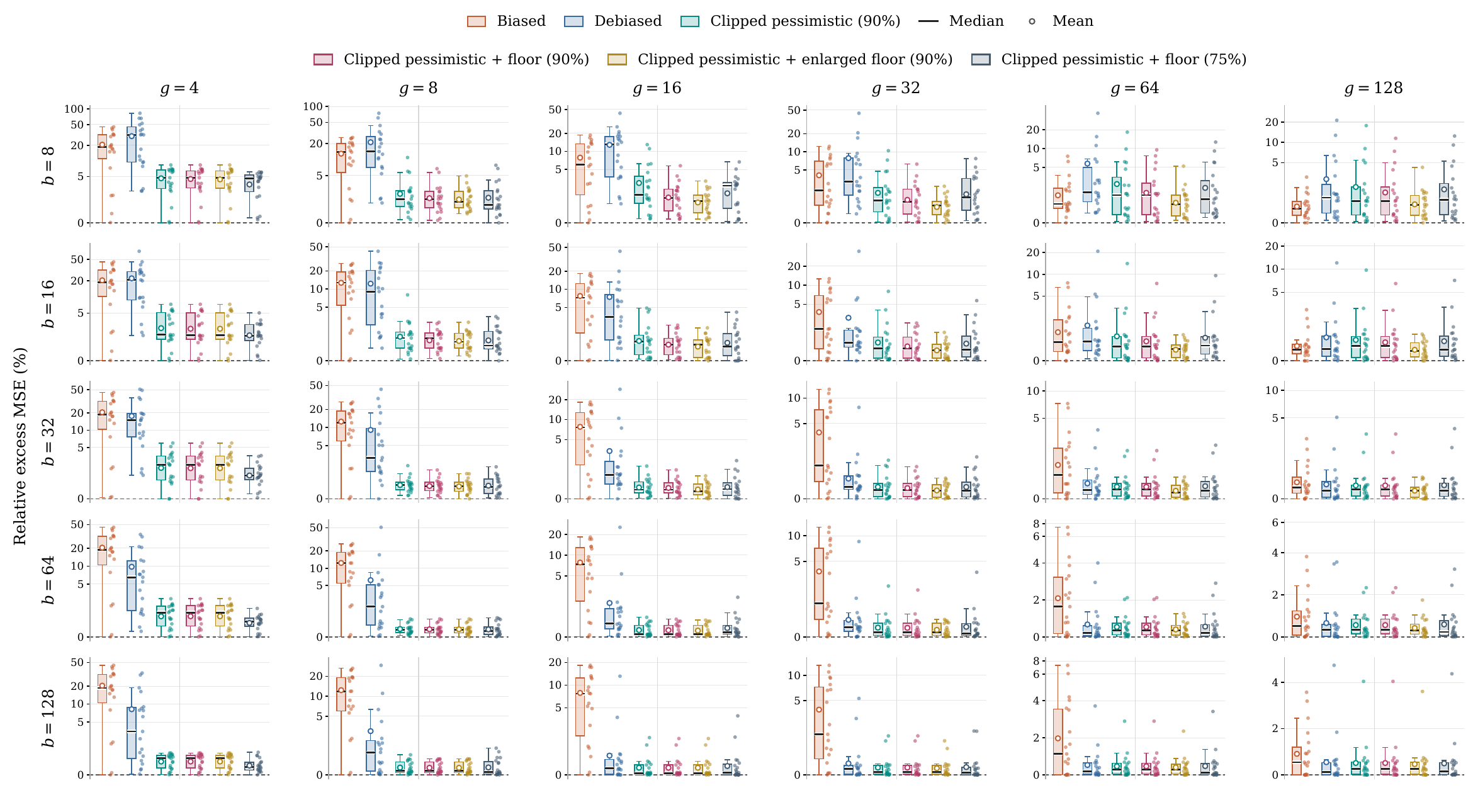}
\caption{Relative excess MSE with $a={\beta}w_{\rm env}=1000$. Black lines and white circles indicate checkpoint-level medians and means, respectively.}
\label{fig:mse-floor-1000}
\end{figure}

\begin{figure}[!htbp]
\centering
\includegraphics[width=\textwidth]{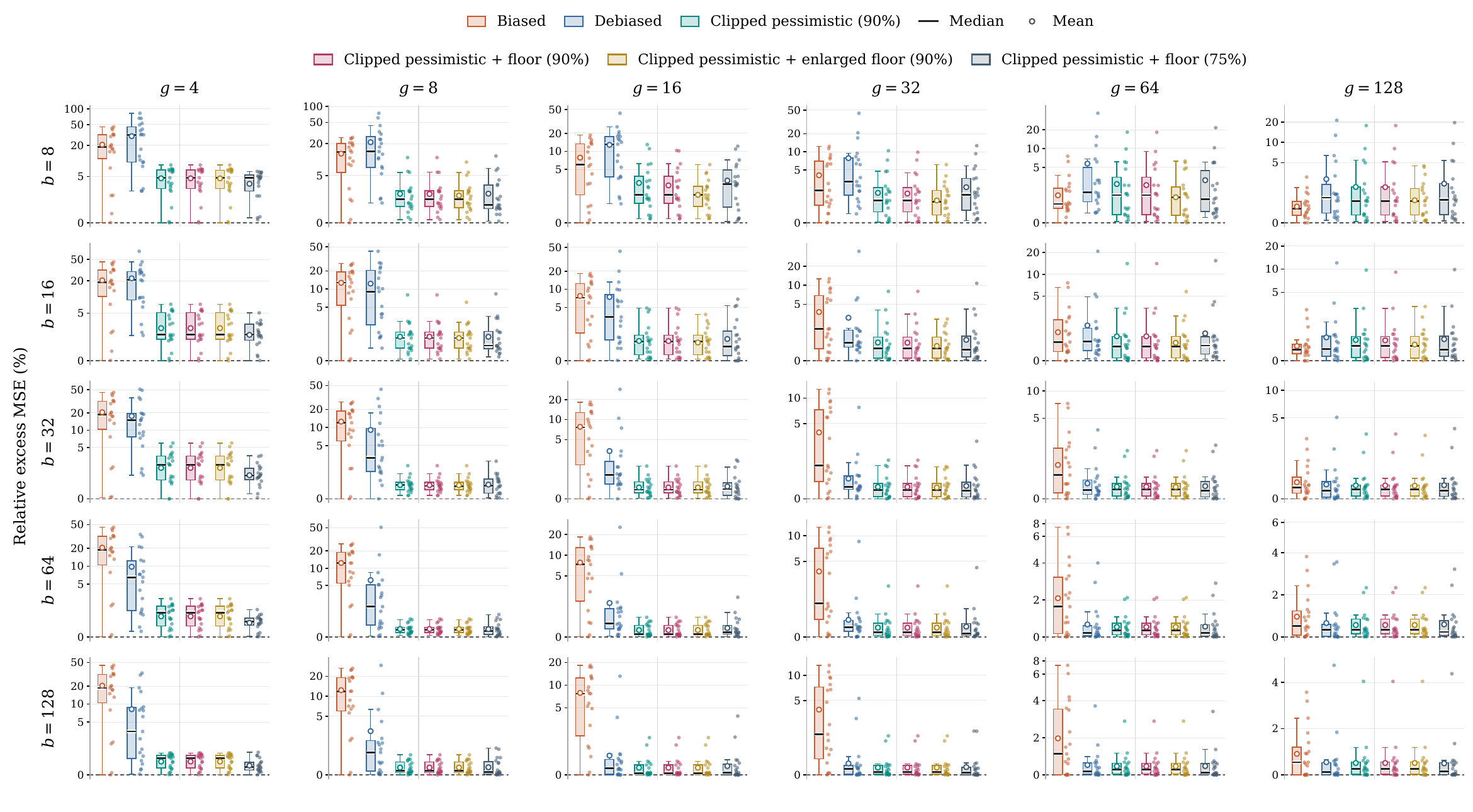}
\caption{Relative excess MSE with $a={\beta}w_{\rm env}=100$, using the same reference and axis ranges as Figure~\ref{fig:mse-floor-1000}.}
\label{fig:mse-floor-100}
\end{figure}

\begin{figure}[!htbp]
\centering
\includegraphics[width=\textwidth]{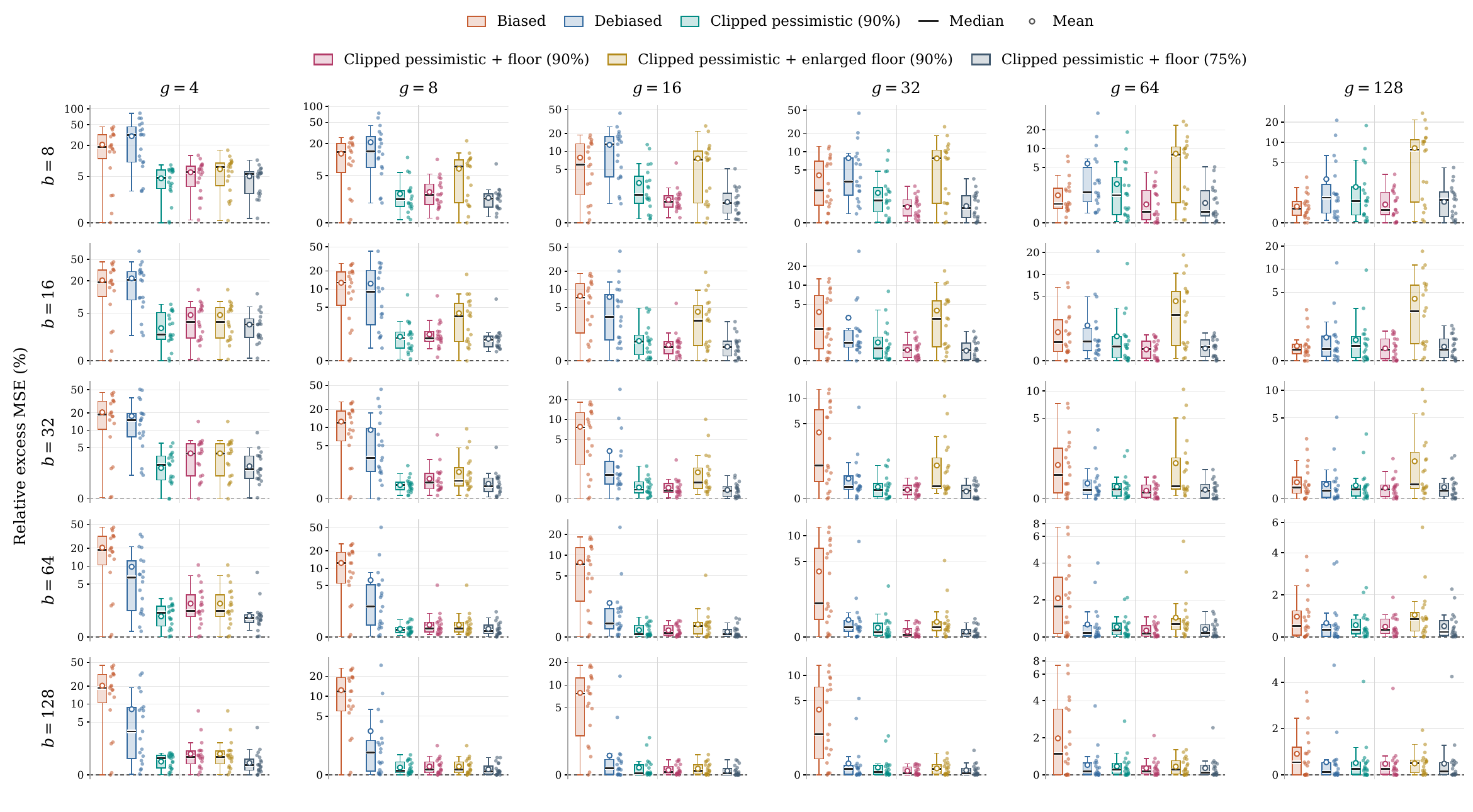}
\caption{Relative excess MSE with $a={\beta}w_{\rm env}=10000$, using the same reference and axis ranges as Figure~\ref{fig:mse-floor-1000}.}
\label{fig:mse-floor-10000}
\end{figure}

\smallskip
\noindent\textbf{Sensitivity results with different floors and confidence levels.}
We next evaluate the sensitivity of the pessimistic estimator in \eqref{eq:main-floored-pess} to the standard-error floor $\epsilon_0$ and the nominal confidence level $1-\alpha$. In addition to the estimators in Figure~\ref{fig:mse-mechanisms}, we consider the enlarged floor $\epsilon_0=\max\{a/({g}\sqrt {b}),{\beta}w_{\rm env}/{b}\}$ from Theorem~\ref{thm:enlarged-floor} at the $90\%$ confidence level ($k=z_{0.90}$), together with the completion-scale floor $\epsilon_0=a/({g}\sqrt {b})$ at the $75\%$ confidence level ($k=z_{0.75}$). Figure~\ref{fig:mse-floor-1000} presents the relative excess MSEs at $a={\beta}w_{\rm env}=1000$, while Figures~\ref{fig:mse-floor-100} and~\ref{fig:mse-floor-10000} examine $a={\beta}w_{\rm env}=100$ and $10000$, respectively. The reference MSE $d$ is taken as the minimum across all evaluated methods and settings in Figures~\ref{fig:mse-floor-1000}--\ref{fig:mse-floor-10000} so that the three figures share a common nonnegative scale.

\begin{figure}[!htbp]
\centering
\includegraphics[width=\textwidth]{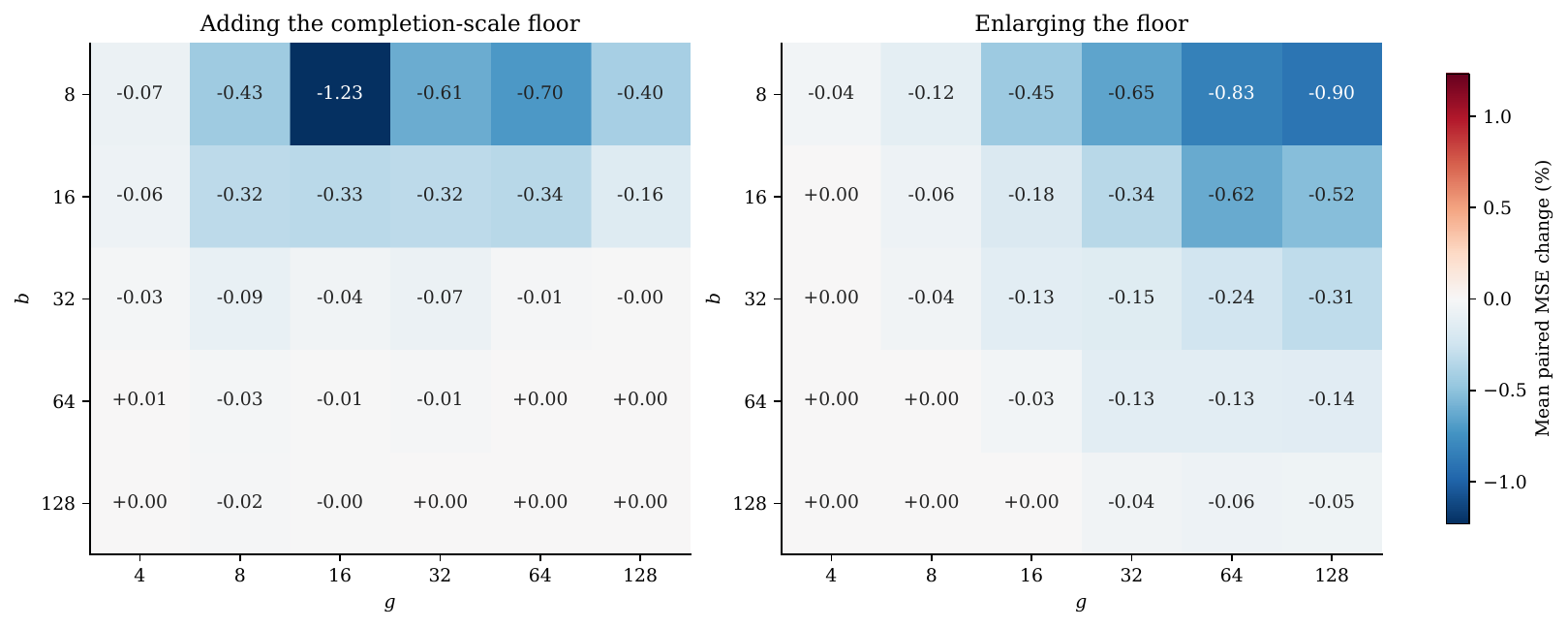}
\caption{Paired MSE changes (\%) from adding the completion-scale floor (left) and enlarging it (right), at $a={\beta}w_{\rm env}=1000$. Negative values indicate lower MSE.}
\label{fig:mse-floor-gains}
\end{figure}

In addition, Figure~\ref{fig:mse-floor-gains} displays the average percentage changes in gradient MSE when sequentially adding the completion-scale floor and enlarging it at $a={\beta}w_{\rm env}=1000$ and $k=z_{0.90}$, while Table~\ref{tab:mse-floor-selection} reports how often each modification changes the selected group size ${\widehat {G}}$ and the resulting impact on the gradient MSE. We summarise the main findings as follows:

\noindent (i) \textbf{Effects of the completion-scale and enlarged floors:} At $a={\beta}w_{\rm env}=1000$, introducing the completion-scale floor reduces the average MSE relative to the clipped pessimistic estimator in \textbf{23} out of \textbf{30} sample size configurations, and subsequently enlarging the floor with the prompt-scale term ${\beta}w_{\rm env}/{b}$ yields a further reduction in \textbf{23} out of \textbf{30} configurations. As shown in Figure~\ref{fig:mse-floor-gains} and Table~\ref{tab:mse-floor-selection}, the completion-scale floor intervenes mainly when both ${b}$ and ${g}$ are small and generally lowers the MSE when it modifies ${\widehat {G}}$, whereas the enlarged floor provides additional protection primarily when ${b}$ is small and ${g}$ is large, guarding against underestimated between-prompt variation under small batch sizes.

\noindent (ii) \textbf{Sensitivity to the floor magnitude:} Across $a={\beta}w_{\rm env}\in\{100,1000,10000\}$, the pessimistic estimators consistently outperform the biased and debiased baselines in Figures~\ref{fig:mse-floor-1000}--\ref{fig:mse-floor-10000}. As summarised in Table~\ref{tab:mse-floor-selection}, the intermediate choice $a={\beta}w_{\rm env}=1000$ achieves the largest overall MSE reductions for both floor specifications. By contrast, a smaller floor with $a={\beta}w_{\rm env}=100$ rarely alters the selected group size and performs very similarly to the clipped estimator, whereas an overly large floor with $a={\beta}w_{\rm env}=10000$ over-regularises the denominator under the enlarged rule.

\noindent (iii) \textbf{Sensitivity to the confidence level:} Under the completion-scale floor, lowering the confidence level from $90\%$ to $75\%$ frequently shifts the selected group size between adjacent integers, yet the resulting average MSE remains virtually unchanged across sample size configurations, demonstrating that our method is robust to moderate variations in the quantile multiplier $k$.

\begin{table}[!htbp]
\centering\small
\renewcommand{\arraystretch}{1.15}
\setlength{\tabcolsep}{6pt}
\caption{Changes in group size selection and gradient MSE across three floor settings, where the $75\%$ vs.\ $90\%$ comparison uses the completion-scale floor.}
\label{tab:mse-floor-selection}
\begin{tabular}{@{}r@{\hspace{12pt}}lrrrr@{}}
\toprule
$a={\beta}w_{\rm env}$ & Modification & \shortstack{Changed ${\widehat {G}}$\\(\%)} & \shortstack{Lower\\MSE} & \shortstack{Higher\\MSE} & \shortstack{Mean MSE\\change (\%)}\\
\midrule
100 & Add floor & 25 (0.18\%) & 25 & 0 & $-0.015$\\
& Enlarge floor & 214 (1.52\%) & 196 & 18 & $-0.161$\\
& $75\%$ vs.\ $90\%$ & 6,234 (44.20\%) & 3,454 & 2,780 & $+0.007$\\
\addlinespace[5pt]
1,000 & Add floor & 254 (1.80\%) & 223 & 31 & $-0.176$\\
& Enlarge floor & 1,935 (13.72\%) & 1,120 & 815 & $-0.205$\\
& $75\%$ vs.\ $90\%$ & 6,324 (44.84\%) & 3,488 & 2,836 & $+0.005$\\
\addlinespace[5pt]
10,000 & Add floor & 3,673 (26.04\%) & 1,556 & 2,117 & $-0.080$\\
& Enlarge floor & 6,733 (47.74\%) & 1,424 & 5,309 & $+1.804$\\
& $75\%$ vs.\ $90\%$ & 6,851 (48.58\%) & 4,257 & 2,594 & $-0.236$\\
\botrule
\end{tabular}
\end{table}

\subsection{Policy optimisation on Reasoning Gym and GSM8K}
\label{app:policy-details}
\label{app:main-experiment-settings}

For the experiments in Figure~\ref{fig:policy-current} and Table~\ref{tab:policy-current}, all methods use a sampling budget of ${{N}}=512$ completions per update, a maximum completion length of 512 tokens, and a learning rate of $10^{-6}$. Models are trained for 200 updates on the Reasoning Gym tasks \citep{stojanovski2025reasoninggym} (evaluated on 2,000 test problems per task) and for 500 updates on GSM8K \citep{cobbe2021training} (evaluated on its 1,319 test problems), with all test evaluations performed using greedy decoding.

\begin{table}[!htbp]
\centering\small
\renewcommand{\arraystretch}{1.04}
\setlength{\tabcolsep}{4pt}
\caption{Final test accuracy (\%) of \texttt{Qwen2.5-1.5B-Instruct} on Reasoning Gym and GSM8K, averaged over five random seeds (with standard deviations in parentheses). Avg.\ denotes the mean accuracy across tasks, and the best and second-best results in each column are shown in bold and underlined, respectively.}
\label{tab:policy-current}
\begin{tabular}{@{}lrrrr@{\hspace{9pt}}r@{}}
\toprule
Method & Equations & Baseconv. & Decimal & GSM8K & \textbf{Avg.}\\
\midrule
${{G}}=2$ & $54.66\,(0.50)$ & $68.81\,(0.70)$ & $48.09\,(4.13)$ & $78.68\,(0.85)$ & 62.56\\
${{G}}=4$ & $\underline{55.26}\,(1.45)$ & $\underline{70.18}\,(1.04)$ & $50.13\,(0.57)$ & $\underline{79.33}\,(0.32)$ & \underline{63.73}\\
${{G}}=8$ & $54.81\,(0.93)$ & $68.92\,(1.04)$ & $\underline{50.77}\,(1.78)$ & $79.06\,(0.70)$ & 63.39\\
${{G}}=16$ & $51.77\,(1.46)$ & $67.22\,(0.72)$ & $50.52\,(2.88)$ & $79.06\,(0.82)$ & 62.14\\
${{G}}=32$ & $49.35\,(2.20)$ & $62.68\,(2.39)$ & $48.47\,(2.14)$ & $79.04\,(0.72)$ & 59.89\\
${{G}}=64$ & $44.22\,(2.84)$ & $61.02\,(2.09)$ & $47.76\,(2.10)$ & $78.79\,(0.38)$ & 57.95\\
\midrule
\textbf{Pessimistic} & $\mathbf{55.71\,(0.12)}$ & $\mathbf{70.58\,(0.52)}$ & $\mathbf{50.98\,(0.98)}$ & $\mathbf{80.24\,(1.15)}$ & \textbf{64.38}\\
\botrule
\end{tabular}
\end{table}

\subsection{Policy optimisation on mathematical reasoning}
\label{app:math-details}

For the mathematical-reasoning study in Table~\ref{tab:qwen3-current}, we first warm-start \texttt{Qwen3-8B} with one epoch of supervised fine-tuning on 496 mathematical solutions to encourage shorter, less repetitive reasoning traces and reduce rollout cost during subsequent reinforcement learning. Starting from this checkpoint, we post-train the model on a subset of 800 problems from DAPO-Math-17k \citep{yu2025dapo} for 300 updates with a maximum completion length of 4,096 tokens, using a sampling budget of ${{N}}=256$ completions per update, a learning rate of $10^{-6}$, and sample sizes $({b},{g})=(32,32)$ for group size selection in Algorithm~\ref{alg:adaptive-group}. We evaluate the trained models across ten external mathematics benchmarks: MATH500 \citep{lightman2024verify} (500 problems), Minerva \citep{lewkowycz2022solving} (272 problems), OlympiadBench \citep{he2024olympiadbench} (675 problems), AIME 2024 and 2025 (30 problems each), AMC 2023 (40 problems), HMMT February 2024 and 2025 (30 problems each), BRUMO 2025 (30 problems), and CMIMC 2025 (40 problems), where the HMMT, BRUMO, and CMIMC datasets are taken from MathArena \citep{balunovic2025matharena}. Following standard practice, we report greedy accuracy (one deterministic completion per problem) on the three larger benchmarks---MATH500, Minerva, and OlympiadBench---and Average@16 (over 16 completions per problem sampled at temperature $0.6$) on the seven competition benchmarks with smaller test sets.

\subsection{Policy optimisation on code generation}
\label{app:coding-details}
For the code-generation study in Table~\ref{tab:coding-current}, \texttt{DeepSeek-Coder-1.3B-Base} \citep{guo2024deepseekcoder} is post-trained on AceCode \citep{zeng2025acecoder} for 500 updates with a maximum completion length of 1,024 tokens. We use the 21,880 problems obtained with the original AceCode hard-example selection procedure. We randomly sample 5,000 problems without replacement to form the test split and use the remaining 16,880 problems for training. All methods use the same training and test splits. We evaluate the trained models on HumanEval+ (164 problems), MBPP+ (378 problems) \citep{liu2023evalplus}, and the AceCode test split (5,000 problems), with the eight completions per problem for Average@8 and Pass@8 sampled at temperature $0.6$.

\begin{table}[!htbp]
\centering
\small
\renewcommand{\arraystretch}{1.04}
\setlength{\tabcolsep}{2.5pt}
\caption{Accuracy (\%) of \texttt{DeepSeek-Coder-1.3B-Base} on HumanEval+ (HE+), MBPP+ and AceCode after 500 training steps, with Avg. denoting the equal-weight average across the three datasets. The best and second-best results in each column are shown in bold and underlined, respectively.}
\label{tab:coding-current}
\begin{tabular}{@{}lrrrr@{\hspace{8pt}}rrrr@{\hspace{8pt}}rrrr@{}}
\toprule
& \multicolumn{4}{c}{Greedy} & \multicolumn{4}{c}{Average@8} & \multicolumn{4}{c}{Pass@8}\\
\cline{2-5}\cline{6-9}\cline{10-13}\noalign{\vskip2pt}
Method & HE+ & MBPP+ & AceCode & Avg. & HE+ & MBPP+ & AceCode & Avg. & HE+ & MBPP+ & AceCode & Avg.\\
\midrule
${{G}}=4$ & \underline{36.59} & 51.06 & 26.34 & 37.99 & 34.07 & 51.16 & 25.74 & 36.99 & 53.66 & 65.34 & 40.12 & 53.04\\
${{G}}=8$ & 35.98 & 51.32 & 26.36 & 37.89 & 33.92 & 50.93 & 26.02 & 36.95 & 53.05 & 65.61 & 40.20 & 52.95\\
${{G}}=16$ & 35.98 & 53.17 & 25.98 & 38.38 & \underline{34.60} & 51.85 & 26.20 & \underline{37.55} & \textbf{56.71} & 64.81 & 39.76 & 53.76\\
${{G}}=32$ & \textbf{37.80} & \underline{53.44} & \underline{26.80} & \textbf{39.35} & 33.99 & 51.39 & \underline{26.32} & 37.23 & 54.27 & 65.87 & 40.52 & 53.55\\
${{G}}=64$ & 33.54 & 52.91 & 26.34 & 37.60 & 33.92 & \underline{52.18} & 25.95 & 37.35 & 52.44 & \underline{67.20} & \textbf{41.74} & \underline{53.79}\\
\midrule
\textbf{Pessimistic} & \underline{36.59} & \textbf{54.23} & \textbf{27.00} & \underline{39.27} & \textbf{34.98} & \textbf{52.45} & \textbf{26.74} & \textbf{38.06} & \underline{56.10} & \textbf{67.72} & \underline{40.96} & \textbf{54.93}\\
\botrule
\end{tabular}
\end{table}

\subsection{Training time and estimation cost}
\label{app:estimation-overhead}

All experiments are conducted on an NVIDIA GH200 GPU cluster. Table~\ref{tab:policy-estimation-overhead} reports the end-to-end training time of our adaptive method alongside the time spent on the pilot study for group size selection, averaged across runs. Here, training time measures the total wall-clock runtime of each run (including model initialisation, policy updates, group size selection, checkpointing, and in-training evaluations), while estimation time covers the pilot study, namely completion generation, reward evaluation, score computation, and variance estimation across all iterations $t\in\mathcal T$ in Algorithm~\ref{alg:adaptive-group}. Because all methods use the same per-update sampling budget ${{N}}$, the training time of the fixed-${{G}}$ baselines is comparable to that of our adaptive method excluding the estimation time. When weighted by the number of GPUs per run, group size selection accounts for only about $8.1\%$ of total GPU-hours across all benchmarks.

\begin{table}[!htbp]
\centering\small
\renewcommand{\arraystretch}{1.04}
\setlength{\tabcolsep}{5pt}
\caption{Mean wall-clock training and estimation times per run on NVIDIA GH200 GPUs, where GPUs denotes the number of GPUs per run and Time share is the percentage of training time spent on the pilot study for group size selection.}
\label{tab:policy-estimation-overhead}
\begin{tabular}{@{}lrrrrrr@{}}
\toprule
Study & Runs & GPUs & Steps & \shortstack{Training time\\(min)} & \shortstack{Estimation time\\(min)} & \shortstack{Time share\\(\%)}\\
\midrule
Equations & 5 & 1 & 200 & 72.18 & 5.84 & 8.08\\
Base conversion & 5 & 1 & 200 & 95.68 & 6.73 & 7.03\\
Decimal arithmetic & 5 & 1 & 200 & 80.58 & 6.01 & 7.46\\
GSM8K & 5 & 1 & 500 & 238.29 & 24.62 & 10.33\\
DAPO (\texttt{Qwen3-8B}) & 1 & 4 & 300 & 809.70 & 58.26 & 7.20\\
AceCode (\texttt{DeepSeek-Coder-1.3B}) & 1 & 1 & 500 & 459.80 & 45.20 & 9.83\\
\botrule
\end{tabular}
\end{table}

\section{Implementation details for variance estimation}
\label{app:implementation}
\label{app:streaming-variance}

By the definitions of $\widehat{\bm c}=(\widehat c_0,\widehat c_1,\widehat c_2,\widehat c_3)^\top$ and $\widetilde{\bm c}=\bm A^{-1}\widehat{\bm c}$ in Section~\ref{sec:tradeoffvarest}, computing the variance estimators from the pilot data $\{X_i,(Y_{i,j},Z_{i,j})_{j=1}^{g}\}_{i=1}^{b}$ amounts to evaluating the four sample sums
\begin{equation}
\begin{gathered}
\sum_{i=1}^{b}{\widehat{\bm g}}_{g}(X_i;\theta),\qquad
\sum_{i=1}^{b}\|{\widehat{\bm g}}_{g}(X_i;\theta)\|^2,\qquad
\sum_{i=1}^{b}\sum_{j=1}^{g}(Z_{i,j}-\bar Z_{i,-j})^2\|S_\theta(X_i,Y_{i,j})\|^2,\\
\sum_{i=1}^{b}
\Bigl\{\frac1{g}\sum_{j=1}^{g}\|S_\theta(X_i,Y_{i,j})\|^2\Bigr\}
\Bigl\{\frac1{{g}-1}\sum_{j=1}^{g}(Z_{i,j}-\bar Z_i)^2\Bigr\},
\end{gathered}
\label{eq:streaming-sums}
\end{equation}
where $S_\theta(X_i,Y_{i,j})=\nabla_\theta\log\pi_\theta(Y_{i,j}\mid X_i)$ and ${\widehat{\bm g}}_{g}(X_i;\theta)={g}^{-1}\sum_{j=1}^{g}(Z_{i,j}-\bar Z_{i,-j})S_\theta(X_i,Y_{i,j})$. For each prompt $X_i$, its contribution to \eqref{eq:streaming-sums} is therefore determined by the prompt-level gradient ${\widehat{\bm g}}_{g}(X_i;\theta)$ and the per-completion score norms $\|S_\theta(X_i,Y_{i,j})\|^2$.

In practice, computing these quantities involves two main difficulties. First, standard backpropagation over a batch returns only the summed gradient across completions, while running a separate backward pass for each of the ${b}{g}$ completions to obtain $\|S_\theta(X_i,Y_{i,j})\|^2$ would be computationally expensive. To avoid per-completion backward passes, we compute $\|S_\theta(X_i,Y_{i,j})\|^2$ and ${\widehat{\bm g}}_{g}(X_i;\theta)$ layer by layer within a single backward pass over a microbatch of completions (Section~\ref{app:per-completion}). Second, even with microbatch backpropagation, processing all ${b}{g}$ completions in one pass or storing the $d$-dimensional gradient vectors across all prompts would exceed GPU memory. We therefore process the prompts in small batches and update the four sums in \eqref{eq:streaming-sums} in a streaming fashion using $O(d)$ GPU memory (Section~\ref{app:streaming-accum}). This streaming procedure also yields the jackknife covariance estimate without additional forward or backward passes (Section~\ref{app:jackknife-cov}).

\subsection{Computing per-completion score norms}
\label{app:per-completion}
Consider a linear layer $\ell$ with input $h_{i,j,t}$ and pre-activation output $u_{i,j,t}=Wh_{i,j,t}+{\bm b_{\rm NN}}$ at token position $t$. By the chain rule, the gradient of $\log\pi_\theta(Y_{i,j}\mid X_i)$ with respect to $W$ is
\begin{equation}
\nabla_W\log\pi_\theta(Y_{i,j}\mid X_i)
=\sum_t\frac{\partial\log\pi_\theta(Y_{i,j}\mid X_i)}{\partial u_{i,j,t}}h_{i,j,t}^{\mathsf T},
\label{eq:streaming-score-recovery}
\end{equation}
and the gradient with respect to ${\bm b_{\rm NN}}$ is $\sum_t\partial\log\pi_\theta(Y_{i,j}\mid X_i)/\partial u_{i,j,t}$. In the forward pass over a microbatch of completions for prompt $X_i$, we save the layer inputs $h_{i,j,t}$, and a single backward pass on $\sum_{(i,j)}\log\pi_\theta(Y_{i,j}\mid X_i)$ gives the output derivatives $\partial\log\pi_\theta(Y_{i,j}\mid X_i)/\partial u_{i,j,t}$ for each completion. Evaluating \eqref{eq:streaming-score-recovery} at layer $\ell$ then gives the corresponding parameter block $S_{\theta,\ell}(X_i,Y_{i,j})$ of $S_\theta(X_i,Y_{i,j})$. Since
\[
\|S_\theta(X_i,Y_{i,j})\|^2=\sum_\ell\|S_{\theta,\ell}(X_i,Y_{i,j})\|^2
\quad\text{and}\quad
{\widehat{\bm g}}_{{g},\ell}(X_i;\theta)=\frac1{g}\sum_{j=1}^{g}(Z_{i,j}-\bar Z_{i,-j})S_{\theta,\ell}(X_i,Y_{i,j}),
\]
we add $\|S_{\theta,\ell}(X_i,Y_{i,j})\|^2$ to $\|S_\theta(X_i,Y_{i,j})\|^2$ and ${g}^{-1}(Z_{i,j}-\bar Z_{i,-j})S_{\theta,\ell}(X_i,Y_{i,j})$ to the $\ell$th block of ${\widehat{\bm g}}_{g}(X_i;\theta)$, and then discard $S_{\theta,\ell}(X_i,Y_{i,j})$ before moving to the next layer. After all completions for prompt $X_i$ are processed, we obtain ${\widehat{\bm g}}_{g}(X_i;\theta)$, $\|{\widehat{\bm g}}_{g}(X_i;\theta)\|^2$, and $\|S_\theta(X_i,Y_{i,j})\|^2$ for $j=1,\ldots,{g}$, which give prompt $i$'s contributions to the four sums in \eqref{eq:streaming-sums}.

\subsection{Streaming accumulation}
\label{app:streaming-accum}
Of the four sums in \eqref{eq:streaming-sums}, only the first is a $d$-dimensional vector while the other three are scalars. Since processing all ${b}$ prompts and ${b}{g}$ completions at once would exceed GPU memory, we evaluate the prompts sequentially in small batches. We partition the ${b}$ prompts into $K$ equal-sized subsets $I_1,\ldots,I_K$ (with $K=8$ for ${b}=8$ and $K=16$ for ${b}\ge 16$, so each subset contains ${b}/K$ prompts). As each prompt $X_i$ is processed as in Section~\ref{app:per-completion}, we add its contributions to the four sums in \eqref{eq:streaming-sums} for its subset and then discard ${\widehat{\bm g}}_{g}(X_i;\theta)$. Once all ${b}/K$ prompts in a subset $I_k$ have been processed, the resulting sums are moved to CPU memory. Summing over $k=1,\ldots,K$ gives the full-sample sums in \eqref{eq:streaming-sums} and hence $\widehat{\bm c}$ and $\widetilde{\bm c}=\bm A^{-1}\widehat{\bm c}$, reducing the GPU memory requirement from $O({b}{g}d)$ to $O(d)$.

\subsection{Jackknife covariance estimation}
\label{app:jackknife-cov}
The sums over the subsets $I_1,\ldots,I_K$ saved in Section~\ref{app:streaming-accum} can be directly reused to estimate the covariance by jackknife. Each replicate removes one subset of prompts together with all its completions. Subtracting the $k$th subset's sums from the full-sample sums,
\begin{equation}
\sum_{i\notin I_k}{\widehat{\bm g}}_{g}(X_i;\theta)
=\sum_{i=1}^{b}{\widehat{\bm g}}_{g}(X_i;\theta)
-\sum_{i\in I_k}{\widehat{\bm g}}_{g}(X_i;\theta)
\label{eq:streaming-delete-group}
\end{equation}
(and similarly for the three scalar sums in \eqref{eq:streaming-sums}), gives the jackknife estimates $\widetilde{\bm c}^{(-k)}$ after replacing ${b}$ with ${b}-{b}/K$ in the definitions of $\widehat{\bm c}$ and $\bm A$ in Section~\ref{sec:tradeoffvarest}, without any additional forward or backward passes. The jackknife covariance estimator for coordinates $r,s\in\{1,3\}$ is then
\begin{equation}
(\widehat V_K)_{rs}
=\frac{K-1}{K}\sum_{k=1}^K
\Bigl(\widetilde c_r^{(-k)}-\frac1K\sum_{\ell=1}^K\widetilde c_r^{(-\ell)}\Bigr)
\Bigl(\widetilde c_s^{(-k)}-\frac1K\sum_{\ell=1}^K\widetilde c_s^{(-\ell)}\Bigr),
\label{eq:grouped-jackknife-cov}
\end{equation}
with $K={b}$ corresponding to the standard leave-one-prompt-out jackknife. Substituting $(\widehat V_K)_{11}$, $(\widehat V_K)_{13}$, and $(\widehat V_K)_{33}$ for $\Var(\widetilde c_1)$, $\Cov(\widetilde c_1,\widetilde c_3)$, and $\Var(\widetilde c_3)$ in \eqref{eq:loss-variance} then gives the estimated standard errors $\widehat\sigma({G})$ and $\widehat\tau_1$ used in \eqref{eq:main-floored-pess} of Section~\ref{sec:pessimistic}, namely,
\[
\widehat\sigma({G})=\left\{{G}^2(\widehat V_K)_{11}
+\frac{2{G}}{{G}-1}(\widehat V_K)_{13}
+\frac{(\widehat V_K)_{33}}{({G}-1)^2}\right\}^{1/2}
\quad\text{and}\quad
\widehat\tau_1=\{(\widehat V_K)_{11}\}^{1/2}.
\]
\end{appendices}

\end{document}